\documentclass{article}

\PassOptionsToPackage{numbers, compress}{natbib}

\usepackage{setspace} 
\usepackage[final]{neurips_2024}
\usepackage{xcolor}
\usepackage[hidelinks]{hyperref}
\definecolor{linkcolor}{RGB}{0, 0, 128}
\definecolor{forestgreen}{rgb}{0.13, 0.55, 0.13}
\hypersetup{
colorlinks   = true,
citecolor    = linkcolor,
linkcolor    = linkcolor,
urlcolor     = linkcolor,
}
\usepackage{xspace}
\usepackage{amsmath}
\usepackage{mathtools}
\usepackage{cleveref}
\usepackage{fdsymbol}
\usepackage[normalem]{ulem}
\usepackage{booktabs}
\usepackage{adjustbox}
\usepackage{colortbl}
\usepackage[inline]{enumitem}
\usepackage{wrapfig}
\usepackage{multirow}
\usepackage{makecell}
\usepackage{placeins}
\usepackage{xcolor}
\usepackage[most]{tcolorbox}

\newcommand{\emoji}[1]{\includegraphics[height=1em]{emojis/#1}}
\newcommand{\unicode}[1]{\includegraphics[height=.9em]{unicode/#1}}
\newcommand{\data}{\textbf{\textsc{OLMoE-Mix}}}
\newcommand{\model}{\textbf{\textsc{OLMoE}}}

\newcommand{\modelsmall}{\textbf{\textsc{OLMoE-1B-7B}}}
\newcommand{\modelsmallnew}{\textbf{\textsc{OLMoE-1B-7B-0125}}}
\newcommand{\modelsmallsft}{\textbf{\textsc{OLMoE-1B-7B-SFT}}}
\newcommand{\modelsmalldpo}{\textbf{\textsc{OLMoE-1B-7B-Instruct}}}
\newcommand{\modelsmalldate}{\textbf{\textsc{OLMoE-1B-7B-0924}}}

\newcommand{\huggingface}{\raisebox{-1.5pt}{\includegraphics[height=1.05em]{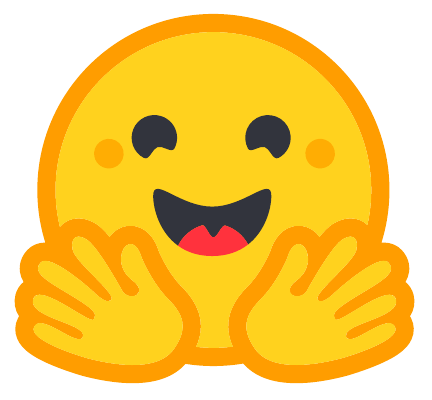}}\xspace}
\newcommand{\github}{\raisebox{-1.5pt}{\includegraphics[height=1.05em]{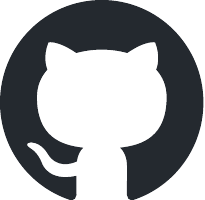}}\xspace}
\newcommand{\wandb}{\raisebox{-1.5pt}{\includegraphics[height=1.05em]{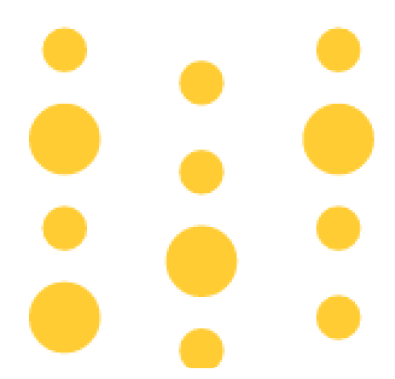}}\xspace}
\newcommand{\olmoeLogoWithText}{\raisebox{-.3em}{\rlap{{\textcolor{white}{ OLMoE}}}\includegraphics[height=1.5em]{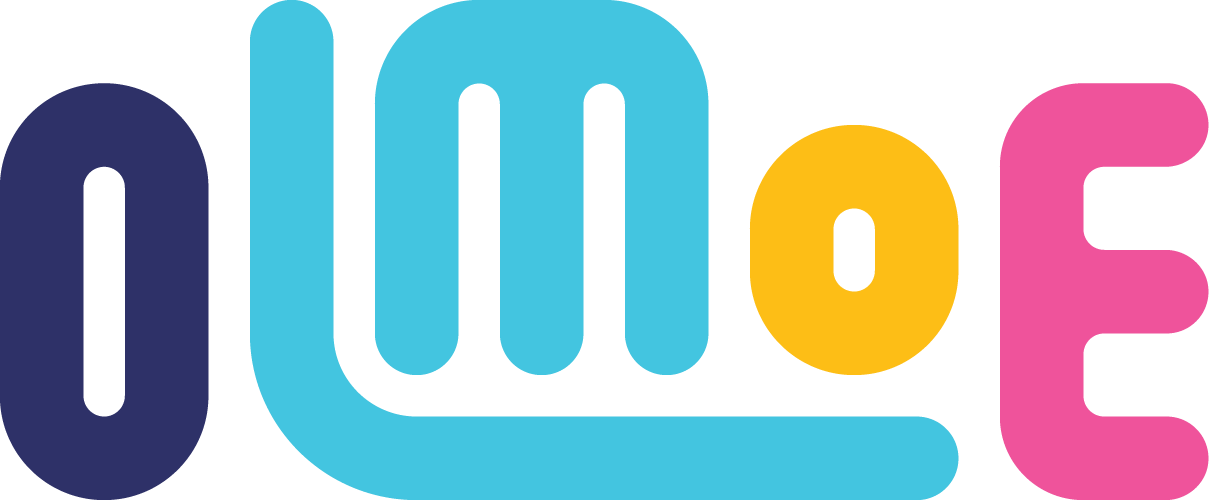}}\xspace}

\definecolor{olmoDarkBlue}{HTML}{012e59}
\definecolor{olmoBlue}{HTML}{265ed4}
\definecolor{olmoLightBlue}{HTML}{012e59}
\definecolor{olmoTeal}{HTML}{00d5ff}
\definecolor{olmoYellow}{HTML}{ffbb00}
\definecolor{olmoOrange}{HTML}{ff9100}
\definecolor{olmoePink}{HTML}{f0539b}
\definecolor{olmoeYellow}{HTML}{fdbe15}
\definecolor{olmoeDarkYellow}{HTML}{fdac15}
\definecolor{olmoeBlue}{HTML}{2E3168}
\definecolor{olmoeLightBlue}{HTML}{43c5e0}
\colorlet{lightOlmoePink}{olmoePink!50}
\colorlet{lightOlmoeYellow}{olmoeYellow!50}
\colorlet{lightOlmoeLightBlue}{olmoeLightBlue!50}

\newcommand{\pa}{{\color{olmoePink}\boldsymbol{a}}}

\newcommand{\yc}{{\color{olmoeDarkYellow}\boldsymbol{c}}}
\newcommand{\bp}{{\color{olmoeBlue}\boldsymbol{p}}}
\newcommand{\bw}{{\color{olmoeLightBlue}\boldsymbol{w}}}

\makeatletter
\AtBeginDocument{%
\renewcommand{\sectionautorefname}{\S\@gobble}

\renewcommand{\subsectionautorefname}{\S\@gobble} 
\renewcommand{\subsubsectionautorefname}{\S\@gobble} 
}
\makeatother

\title{
\olmoeLogoWithText{}: Open Mixture-of-Experts Language Models
}

\author{
\hspace{-.8em}
{\bf
Niklas Muennighoff$\hspace{.1em}^{\yc\pa}$
}
\hspace{.2em}
{\bf
Luca Soldaini$\hspace{.1em}^{\pa}$
}
\hspace{.2em}
{\bf
Dirk Groeneveld$\hspace{.1em}^{\pa}$
}
\hspace{.2em}
{\bf
Kyle Lo$\hspace{.1em}^{\pa}$
}
\hspace{.2em}
{\bf
Jacob Morrison$\hspace{.1em}^{\pa}$
}
\\
\hspace{-.8em}
{\bf
Sewon Min$\hspace{.1em}^{\pa}$
}
\hspace{.75em}
{\bf
Weijia Shi$\hspace{.1em}^{\bw}$
}
\hspace{.75em}
{\bf
Pete Walsh$\hspace{.1em}^{\pa}$
}
\hspace{.75em}
{\bf
Oyvind Tafjord$\hspace{.1em}^{\pa}$
}
\hspace{.75em}
{\bf
Nathan Lambert$\hspace{.1em}^{\pa}$
}
\\
\hspace{-.8em}
{\bf
Yuling Gu$\hspace{.1em}^{\pa}$
}
\hspace{.35em}
{\bf
Shane Arora$\hspace{.1em}^{\pa}$
}
\hspace{.35em}
{\bf
Akshita Bhagia$\hspace{.1em}^{\pa}$
}
\hspace{.35em}
{\bf
Dustin Schwenk$\hspace{.1em}^{\pa}$
}
\hspace{.35em}
{\bf
David Wadden$\hspace{.1em}^{\pa}$
}
\\
\hspace{-.8em}
{\bf
Alexander Wettig$\hspace{.1em}^{\pa\bp}$
}
\hspace{.3em}
{\bf
Binyuan Hui
}
\hspace{.3em}
{\bf
Tim Dettmers$\hspace{.1em}^{\pa}$
}
\hspace{.3em}
{\bf
Douwe Kiela$\hspace{.1em}^{\yc}$
}
\hspace{.3em}
{\bf
Ali Farhadi$\hspace{.1em}^{\pa\bw}$
}
\\
\hspace{-.8em}
{\bf
Noah A. Smith$\hspace{.1em}^{\pa\bw}$
}
\hspace{.5em}
{\bf
Pang Wei Koh$\hspace{.1em}^{\pa\bw}$
}
\hspace{.5em}
{\bf
Amanpreet Singh$\hspace{.1em}^{\yc}$
}
\hspace{.5em}
{\bf
Hannaneh Hajishirzi$\hspace{.1em}^{\pa\bw}$
}
\\
\hspace{-1em}
$^{\pa}\hspace{.1em}$Allen Institute for AI
\hspace{.05em}
$^{\yc}\hspace{.1em}$Contextual AI
\hspace{.05em}
$^{\bw}\hspace{.1em}$University of Washington
\hspace{.05em}
$^{\bp}\hspace{.1em}$Princeton University
\\
\hspace{-1em}
{\tt \href{mailto:n.muennighoff@gmail.com}{n.muennighoff@gmail.com}}
\hspace{.5em}
{\tt \href{mailto:hannah@allenai.org}{hannah@allenai.org}}
}

\date{\today}

\begin{document}

\maketitle

\begin{abstract}

We introduce \model{},\footnote{This paper describes the first \model{} from 09/2024. See \autoref{sec:olmoe0125} for an overview of a newer version.} a fully open, state-of-the-art language model leveraging sparse Mixture-of-Experts (MoE). \modelsmall{} has 7 billion (B) parameters but uses only 1B per input token. We pretrain it on 5 trillion tokens and further adapt it to create \modelsmalldpo{}. Our models outperform all available models with similar active parameters, even surpassing larger ones like Llama2-13B-Chat and DeepSeekMoE-16B. We present various experiments on MoE training, analyze routing in our model showing high specialization, and open-source all aspects of our work: model weights, training data, code, and logs.

\begin{center}
\begin{tabular}{rcl}
\multirow{2}{*}{\huggingface} & \textbf{Model} & \href{https://hf.co/allenai/OLMoE-1B-7B-0924}{\path{hf.co/allenai/OLMoE-1B-7B-0924}}\\[0.2em]
& \textbf{Data} & \href{https://hf.co/datasets/allenai/OLMoE-mix-0924}{\path{hf.co/datasets/allenai/OLMoE-mix-0924}}\\[0.2em]
\multirow{1}{*}{\github} & \textbf{Code} & \href{https://github.com/allenai/OLMoE}{\path{github.com/allenai/OLMoE}}\\[0.2em]
\multirow{2}{*}{\wandb} & \multirow{2}{*}{\textbf{Logs}} & \href{https://wandb.ai/ai2-llm/olmoe/reports/OLMoE-1B-7B-0924--Vmlldzo4OTcyMjU3}{\path{wandb.ai/ai2-llm/olmoe/reports/}}\\
& & \href{https://wandb.ai/ai2-llm/olmoe/reports/OLMoE-1B-7B-0924--Vmlldzo4OTcyMjU3}{\path{OLMoE-1B-7B-0924--Vmlldzo4OTcyMjU3}}\\
\end{tabular}
\end{center}
\end{abstract}

\begin{figure*}[htbp]
\centering
\begin{center}
\includegraphics[width=0.9\textwidth]{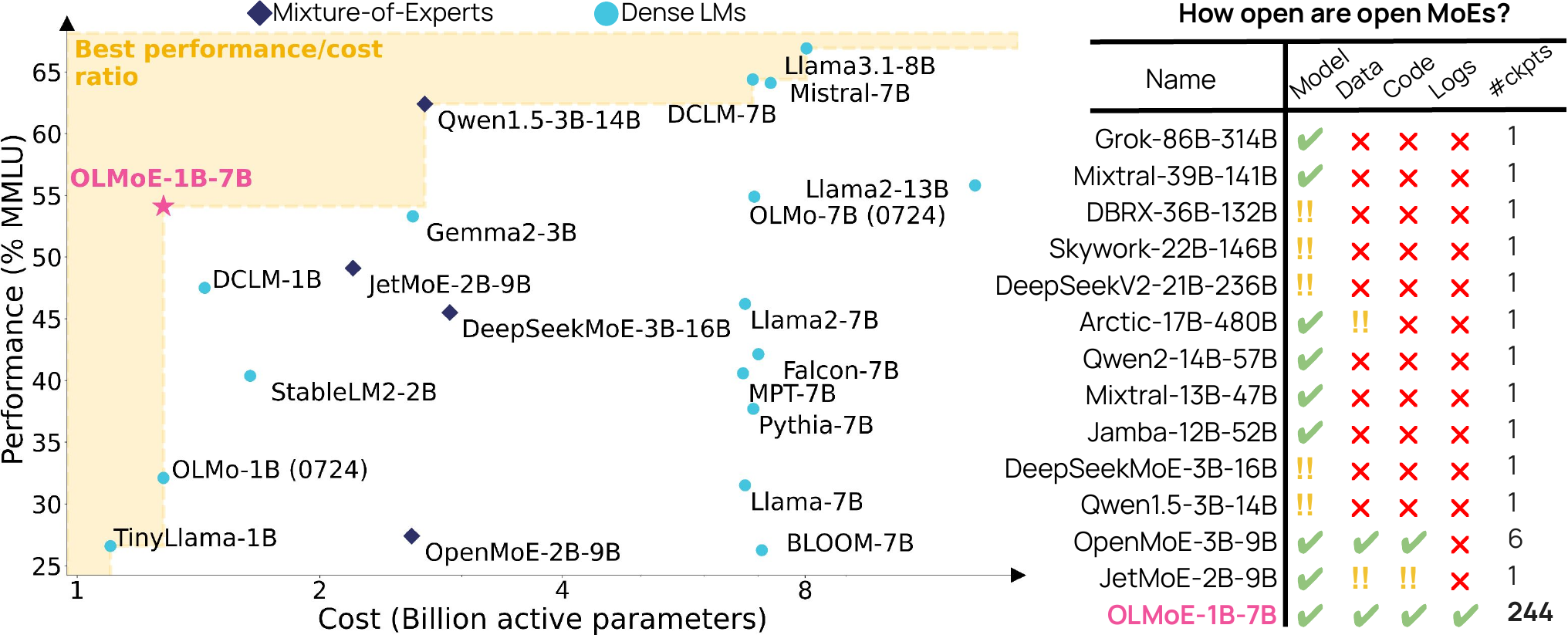}
\caption{\textbf{Performance, cost, and degree of openness of open MoE and dense LMs.} Model names contain rounded parameter counts: \texttt{model-active-total} for MoEs and \texttt{model-total} for dense LMs. \texttt{\#ckpts} is the number of intermediate checkpoints available. 
We highlight MMLU as a summary of overall performance; see \autoref{sec:results} for more results. \modelsmall{} performs best among models with similar active parameter counts and is the most open MoE.}
\label{fig:overview}
\end{center}
\end{figure*}

\newpage
\begin{spacing}{0}
\tableofcontents
\end{spacing}
\newpage

\section{Introduction}
\label{sec:intro}

Despite significant advances in Large Language Models (LMs) on various tasks, there remains a clear trade-off between performance and cost in both training and inference. High-performing LMs are inaccessible for many academics and open-source developers as they are prohibitively expensive to build and deploy.\footnote{For example, even with 16 H100 GPUs and several optimizations, Llama 3 405B only achieves a decoding throughput of around 100 tokens per second~\citep{dubey2024llama3herdmodels}.} One approach to improve the cost-performance trade-off lies in using sparsely-activated Mixture-of-Experts (MoEs)~\citep{shazeer2017outrageously}. MoEs have several experts in each layer, only a subset of which is activated at a time (see \autoref{fig:olmoe}). This makes MoEs significantly more efficient than dense models with a similar number of total parameters, which activate all parameters for every input~\citep{yun2024inferenceoptimalmixtureofexpertlargelanguage}. For this reason, industry frontier models use MoEs including Gemini-1.5~\citep{geminiteam2024gemini} and reportedly GPT-4~\citep{gpt4moe}.

Most MoE models, however, are closed-source: While some have publicly released model weights~\citep{deepseekai2024deepseekv2,jiang2024mixtral,shen2024jetmoe,jambateam2024jamba15hybridtransformermambamodels,qwen_moe}, they offer limited to no information about their training data, code, or recipes (see Figure~\ref{fig:overview}). While there have been prior efforts to make language modeling research fully accessible~\citep{biderman2023pythiasuiteanalyzinglarge,groeneveld2024olmo,li2024datacomplm,liu2023llm360,workshop2023bloom,zhang2024mapneohighlycapabletransparent}, they have been largely limited to dense LMs. This comes despite MoEs requiring \textit{more} openness as they add complex new design questions to LMs, such as how many total versus active parameters to use, whether to use many small or few large experts, if experts should be shared, and what routing algorithm to use. The lack of open resources and findings about these details prevents the field from building cost-efficient open MoEs that approach the capabilities of closed-source frontier models.

To address these issues, we introduce \model{}, a fully open Mixture-of-Experts language model with state-of-the-art performance among similarly-sized models. In particular, we pretrain \modelsmall{} for 5.1 trillion tokens with 6.9B total parameters, of which only 1.3B are activated for each input token. This leads to a similar inference cost as using dense models with around 1B parameters, such as OLMo 1B~\citep{groeneveld2024olmo} or TinyLlama 1B~\citep{zhang2024tinyllamaopensourcesmalllanguage}, but requires more GPU memory to store its 7B total parameters. Our experiments show that MoEs train $\sim$2$\times$ faster than dense LMs with equivalent active parameters. In \autoref{fig:overview}, we show that \modelsmall{} significantly outperforms all open 1B models and displays competitive performance to dense models with significantly higher inference costs and memory storage (e.g., similar MMLU scores to Llama2-13B, which is $\sim$10$\times$ more costly). Via instruction- and preference tuning, we create \modelsmalldpo{}, which we find exceeds various larger instruct models including Llama2-13B-Chat~\citep{touvron2023llama2openfoundation}, OLMo-7B-Instruct (0724), and DeepSeekMoE-16B~\citep{deepseekai2024deepseek} on common benchmarks (MMLU, GSM8k, HumanEval, etc.).

Our comprehensive set of controlled experiments highlights key design choices for MoEs (see \autoref{tab:moedesign}) and LMs in general. One critical design decision for making MoEs performant is the use of fine-grained routing with granular experts~\citep{deepseekai2024deepseek}: we employ 64 small experts in each layer with 8 being activated. The choice of routing algorithm is also important: we find dropless~\citep{gale2022megablocksefficientsparsetraining} token-based routing~\citep{shazeer2017outrageously} outperforms expert-based routing~\citep{zhou2022mixtureofexperts}. Our findings also include those that challenge prior work, such as the ineffectiveness of shared experts~\citep{deepseekai2024deepseek} and the limited benefits of sparsely upcycling a pretrained dense LM into an MoE~\citep{komatsuzaki2023sparse} unless under small compute budgets. Finally, we analyze the routing behavior in \modelsmall{}, finding that routing saturates early in pretraining, experts are rarely co-activated, and experts exhibit domain and vocabulary specialization. 

We hope our fully open MoE facilitates more research and analysis to improve our understanding of these models. We release training code, intermediate checkpoints (every 5000 steps), training logs, and training data under open-source licenses (Apache 2.0~{\url{http://www.apache.org/licenses/LICENSE-2.0}} or ODC-By 1.0~{\url{https://opendatacommons.org/licenses/by/1-0/}}).

\section{Pretraining and Adaptation}
\label{sec:pretraining}

\begin{figure}[htbp]
\centering
\begin{center}
\includegraphics[width=\textwidth]{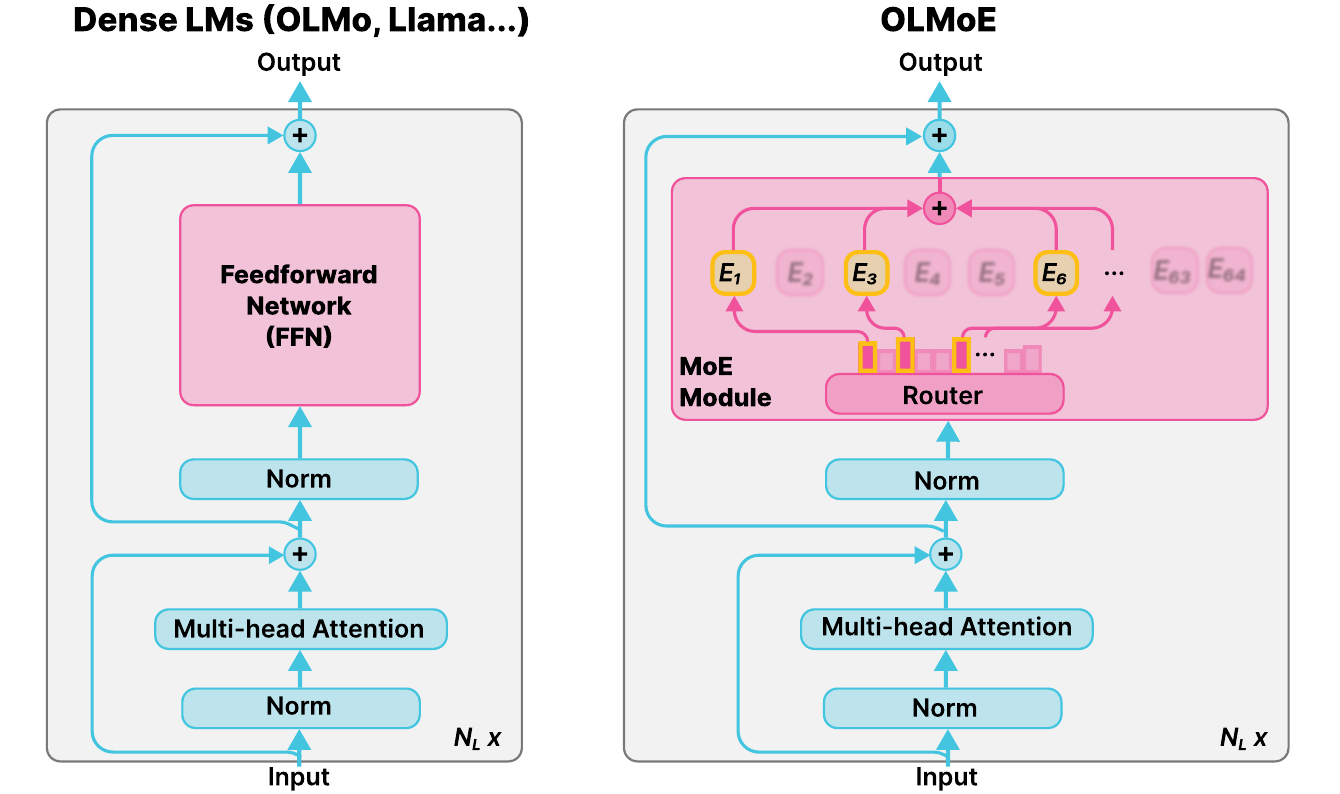}
\caption{\textbf{Comparison of the architecture of dense LMs and MoE models like \model{}.} The figure excludes some details, e.g., \modelsmall{} also uses QK-Norm (\autoref{sec:qknorm}).}
\label{fig:olmoe}
\end{center}
\end{figure}

\begin{table}[htbp]
\centering
\setlength{\tabcolsep}{4pt}
\begin{tabular}{>{\raggedright}p{2.7cm} p{6.4cm} p{1.2cm} p{2.6cm}}
\toprule
\textbf{Design choice} & \textbf{Description} & \textbf{Exper-iment} & \modelsmall{} \\
\midrule
Active params & \# active parameters per input token  & \autoref{sec:moevsdense} & 1.3B active\\
Total params & Total \# of parameters in the model & \autoref{sec:moevsdense} & 6.9B total \\\addlinespace
Expert granularity & Using fine-grained small experts vs.~a few large experts~\citep{dai2024deepseekmoeultimateexpertspecialization} & \autoref{sec:granularity} & 64 small experts with 8  activated \\\addlinespace
Expert sharing & Whether or not to include a shared expert~\citep{dai2024deepseekmoeultimateexpertspecialization} & \autoref{sec:shared} & No shared expert \\\addlinespace
Routing algorithm & How inputs are assigned to experts, e.g., assignment on a per token basis (e.g., 2 experts per token) or per expert basis (e.g., 2 tokens per expert), and whether or not all tokens get assigned or some get dropped~\citep{gale2022megablocksefficientsparsetraining,zhou2022mixtureofexperts} & \autoref{sec:expertchoice} & Dropless~\citep{gale2022megablocksefficientsparsetraining} MoE with token choice \\\addlinespace
Sparse upcycling & Whether to start from a dense model~\citep{komatsuzaki2023sparse,zhang2024bamjustlikethat} & \autoref{sec:upcycling} & Not used \\\addlinespace
Load balancing loss & Auxiliary loss to penalize unequal assignment to experts that may harm performance~\citep{shazeer2017outrageously} & \autoref{sec:lbloss} & Used with weight 0.01 \\\addlinespace
Router z-loss & Auxiliary loss to penalize large logits in the router that may cause instabilities~\citep{zoph2022stmoe} & \autoref{sec:zloss} & Used with weight 0.001 \\
\bottomrule
\end{tabular}
\vspace{.2em}
\caption{\textbf{Key MoE design choices and our setup for \modelsmall{} based on our experiments.} Full configuration for \modelsmall{} is in \autoref{sec:config}.}
\label{tab:moedesign}
\end{table}

\begin{table}[htbp]
\centering 
\begin{tabular}{llrrrr}
\toprule
\multirow{3}{*}{\textbf{Source}} & \multirow{3}{*}{\textbf{Doc Type}} & \textbf{GPT-NeoX} & \multirow{3}{*}{\shortstack[l]{\textbf{~~~Words}\\\textit{(billions)}}} & \textbf{UTF-8} & \multirow{3}{*}{\shortstack[l]{\textbf{Documents}\\\textit{~~~~(millions)}}} \\
& & \textbf{tokens} &  & \textbf{bytes} & \\
& & \textit{(billions)} &  & \textit{(GB)} & \\
\midrule
DCLM-Baseline~\citep{li2024datacomplm} & web pages & 3,860 & 3,380 & 16,700 & 2,950 \\
StarCoder~\citep{li2023starcoder,kocetkov2022stack3tbpermissively} & code & 101 & 63.9 & 325 & 78.7 \\
peS2o~\citep{peS2o,soldaini2024dolma} & STEM papers & 57.2 & 51.3 & 268 & 38.8 \\
arXiv~\citep{together2023redpajama} & STEM papers & 21.1 & 23.5 & 88.8 & 1.55 \\
OpenWebMath~\citep{paster2023openwebmath} & math web pages & 12.7 & 10.2 & 42.4 & 2.91 \\
Algebraic Stack~\citep{azerbayev2023llemma} & math proofs code & 12.6 & 9.6 & 39.3 & 2.83 \\
\begin{tabular}{@{}l@{}}
English Wikipedia \\
\hspace{.5em}~\&  Wikibooks~\citep{soldaini2024dolma} 
\end{tabular}
& encyclopedic & 3.69 & 3.16 & 16.2 & 6.17 \\
\midrule
\multicolumn{2}{c}{\textbf{Total}} & \textbf{4,060} & \textbf{3,530} & \textbf{17,400} & \textbf{3,080} \\
\bottomrule
\end{tabular}
\vspace{.5em}
\caption{\textbf{Composition of the pretraining data for \modelsmall{}}. StarCoder, peS2o, and Wikipedia parts come from Dolma 1.7~\citep{soldaini2024dolma}. Links to our data are in \autoref{sec:artifacts}.}
\label{tab:data}
\end{table}

\paragraph{Pretraining architecture} \model{} is a decoder-only LM consisting of $N_{L}$ transformer~\citep{vaswani2023attention} layers. The feedforward network (FFN) in dense models like OLMo~\citep{groeneveld2024olmo}, is replaced with an MoE module consisting of $N_{E}$ smaller FFN modules called experts, of which a subset of $k$ experts is activated for each processed input token $x$ (also see \autoref{fig:olmoe}):
\begin{equation}
\label{eq:moe}
\text{MoE module}(x)=\sum_{i \in \text{Top}-k(r(x))} \mathrm{softmax} \left( r(x) \right)_i {E_i(x)}
\end{equation}
where $r$, called the router, is a learned linear layer mapping from the input logits to the chosen $k$ experts. A softmax is applied to the router outputs to compute routing probabilities for all $N_E$ experts. Each selected expert $E_i$ processes the input $x$, the output of which is then multiplied with its respective routing probability. The results are then summed across all chosen Top-$k$ experts to constitute the output of the MoE module for a single layer of the model out of its $N_L$ total layers. Key decisions in designing an MoE model include determining the number of activated and total parameters, the design of the experts (e.g., granularity, whether or not to include shared experts), and the choice of the routing algorithm. Moreover, training an MoE model can involve initializing from a dense model (sparse upcycling) and changing the training objective, such as including auxiliary load balancing and router z-losses. Experiments related to these design choices are in \autoref{sec:moespecific}; \autoref{tab:moedesign} shows our final decisions. 

In summary, we use 1.3B active parameters out of a total of 6.9B, with 8 activated experts out of 64 per layer. We use dropless token choice routing~\citep{gale2022megablocksefficientsparsetraining}: For each input token, the learned router network determines 8 experts to process it. We train \modelsmall{} from scratch with two auxiliary losses: load balancing loss ($\mathcal{L}_{LB}$)~\citep{shazeer2017outrageously} and router z-loss ($\mathcal{L}_{RZ}$)~\citep{zoph2022stmoe}, which we define and experiment with in \autoref{sec:lbloss} and \autoref{sec:zloss}, respectively. We multiply them with respective loss weights, $\alpha$ and $\beta$, and sum them linearly with the cross entropy loss ($\mathcal{L}_{\textit{CE}}$) to arrive at our final training loss:
\begin{equation}
\label{eq:loss}
\mathcal{L} = \mathcal{L}_{\textit{CE}} + \alpha \mathcal{L}_{\textit{LB}} + \beta \mathcal{L}_{\textit{RZ}}
\end{equation}
Our full pretraining configuration for \modelsmall{} is in \autoref{sec:config}.

\begin{table}[t]
\centering
\begin{tabular}{l|cr}
\toprule
\textbf{Source} & \textbf{Domain} & \textbf{Samples} \\
\midrule
\multicolumn{3}{c}{\textit{Instruction Tuning}} \\
\midrule
Tulu 2 SFT Mix~\citep{ivison2023camels} & Various & 326,154 \\
No Robots~\citep{no_robots} & Various & 9,500 \\
CodeFeedback-Filtered-Instruction~\citep{zheng2024opencodeinterpreter} & Coding & 156,526 \\
MetaMathQA~\citep{yu2024metamathbootstrapmathematicalquestions} & Math & 98,750 \\
Advanced (non-chat) subset of Daring Anteater~\citep{wang2024helpsteer2opensourcedatasettraining} & Various & 17,082 \\
\midrule
\multicolumn{3}{c}{\textit{Preference Tuning (DPO~\citep{rafailov2023direct})}} \\
\midrule
UltraFeedback~\citep{cui2023ultrafeedback} binarized and filtered for TruthfulQA~\citep{lin2022truthfulqa} contamination & Various & 60,800 \\
\bottomrule
\end{tabular}
\vspace{.5em}
\caption{\textbf{Adaptation training data for \modelsmall{}.} Links to our data are in \autoref{sec:artifacts}.}
\label{tab:adaptds}
\end{table}
\paragraph{Pretraining data} We mix data from DCLM~\citep{li2024datacomplm} and Dolma 1.7~\citep{soldaini2024dolma}, which includes: (1) a quality-filtered subset of Common Crawl, referred to as DCLM-Baseline, (2) StarCoder, Algebraic Stack and arXiv, used in both DCLM and Dolma 1.7, and (3) peS2o and Wikipedia from Dolma 1.7. We refer to our pretraining dataset as \data{}.

To all sources above, we apply a filter that removes all documents with a sequence of 32 or more repeated n-grams, where an n-gram is any span of 1 to 13 tokens. 
For the StarCoder subset, we also remove any document from a repository with fewer than 2 stars on GitHub, whose most frequent word constitutes over 30\% of the document, or whose top-2 most frequent words constitute over 50\% of the document.

We shuffle all samples randomly at the beginning of each epoch and train for a total of 5.133T tokens (1.3 epochs following \citet{muennighoff2023scaling}). During our annealing phase (final 100B tokens) we first reshuffle the entire dataset and then linearly decay the learning rate to 0, following prior work~\citep{groeneveld2024olmo,li2024datacomplm}. Our pretraining data statistics are in \autoref{tab:data}.

\paragraph{Adaptation} We create \modelsmalldpo{} by following a standard adaptation recipe split into \textbf{instruction tuning}~\citep{mishra2022crosstask,wei2022finetuned, sanh2022multitask,shen2023mixtureofexperts,zadouri2023pushingmixtureexpertslimit} followed by \textbf{preference tuning}~\citep{christiano2023deep,bai2022constitutionalaiharmlessnessai,rafailov2023direct,ethayarajh2024kto} building on prior open models~\citep{tunstall2023zephyr,ivison2023camels,wang2023far}. In our instruction tuning dataset, we add more code and math data to boost performance on downstream coding and math applications. Other models, such as GPT-4~\citep{openai2023gpt4} and Llama 3~\citep{dubey2024llama3herdmodels} similarly include samples from math datasets like GSM8k~\citep{cobbe2021training} or MATH~\citep{hendrycks2021measuringmathematicalproblemsolving} during pretraining. We also include No Robots and a subset of Daring Anteater as they are of high quality and add diversity, two key factors for successful adaptation~\citep{wang2023far,zhou2023lima,longpre2023flancollectiondesigningdata,muennighoff2023octopack}. We describe our adaptation datasets in \autoref{tab:adaptds} and hyperparameters in \autoref{sec:config}.

\section{Results}
\label{sec:results}

Our evaluation procedure consists of three parts: \textbf{During pretraining}, \textbf{After pretraining}, and \textbf{After adaptation}. We detail the setup for each in \autoref{sec:evalsetup}.

\begin{figure}[htbp]
\centering
\begin{center}
\includegraphics[width=\textwidth]{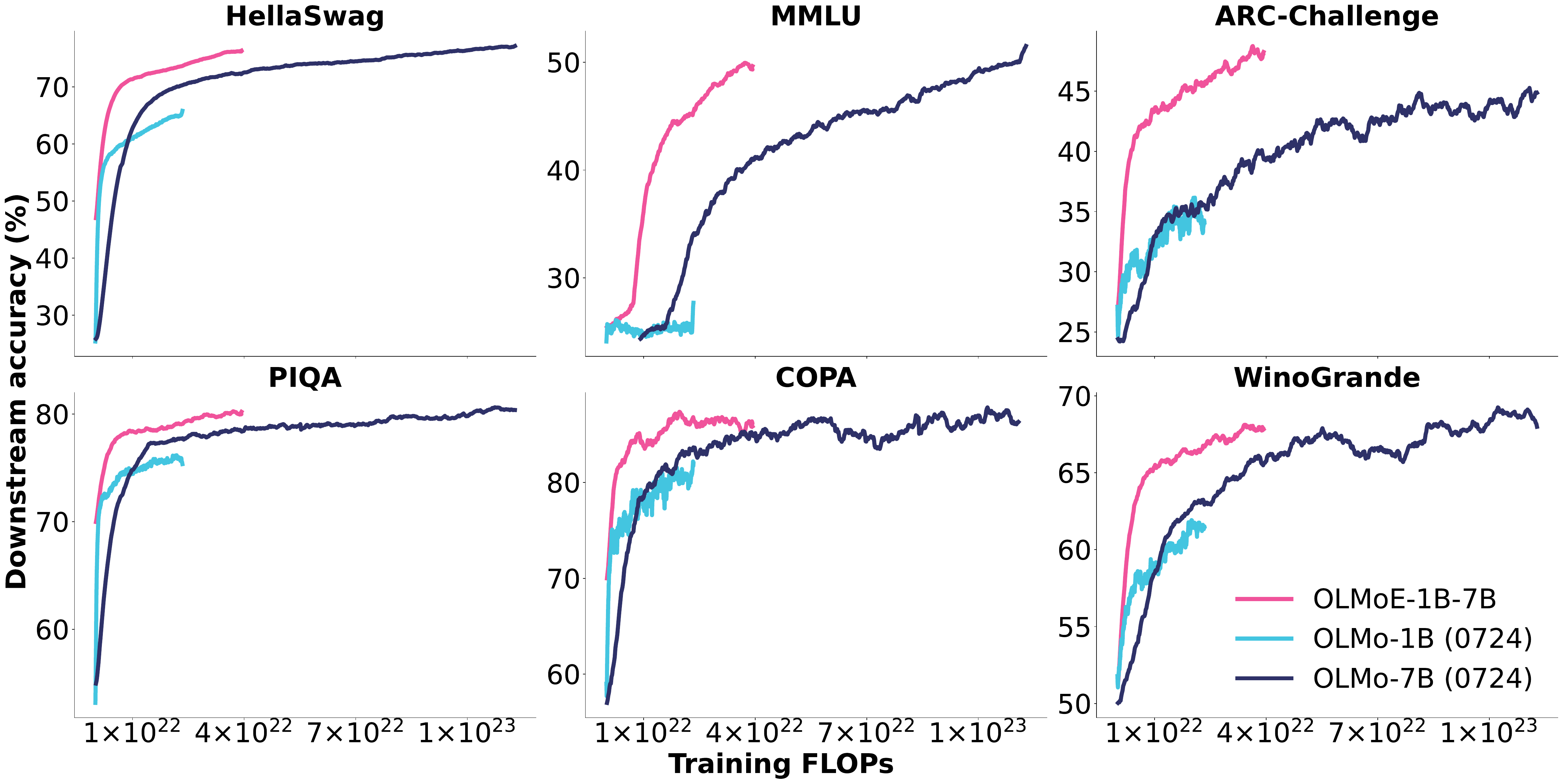}
\caption{\textbf{Evaluation of \modelsmall{} and the current best OLMo models during pretraining.} \modelsmall{} differs from the OLMo models in its MoE architecture, several training hyperparameters, and its training dataset, see \autoref{sec:pretraining}. A version of this plot with tokens as the x-axis and markers where annealing starts is in \autoref{sec:addeval}. More results, logs, and configurations: \url{https://wandb.ai/ai2-llm/olmoe/reports/Plot-OLMoE-1B-7B-vs-OLMo-7B-vs-OLMo-1B--Vmlldzo4OTcyMjEz}}
\label{fig:trainingevalflops}
\end{center}
\end{figure}

\paragraph{During pretraining} In \autoref{fig:trainingevalflops} we benchmark the performance of \modelsmall{} during pretraining with the current best OLMo models~\citep{groeneveld2024olmo} on commonly used downstream tasks. We find that across all tasks \modelsmall{} reaches better performance with less compute (FLOPs) than the dense OLMo models. \modelsmall{} matches or outperforms OLMo-7B at the end of training despite \modelsmall{} having used less than half as many FLOPs for training and using only 1B active parameters. This is likely a result of the dataset and modeling changes we make to the OLMo setup including MoE-related changes, stability, and performance improvements, outlined in \autoref{sec:config}. \autoref{sec:addeval} contains training and validation loss plots showing very smooth loss curves without major loss spikes during the 5T tokens of our pretraining. 

\begin{table}[htbp]
\centering
{\footnotesize
\begin{tabular}{l|cc|cccccc}
\toprule
& \textbf{Active} & \textbf{Open} & \multirow{2}{*}{\textbf{MMLU}} & \textbf{Hella-} & \textbf{ARC-} & \textbf{ARC-} & \multirow{2}{*}{\textbf{PIQA}} & \textbf{Wino-} \\
& \textbf{params} & \textbf{Data} & & \textbf{Swag} & \textbf{Chall.} & \textbf{Easy} & & \textbf{Grande} \\
\midrule
\multicolumn{9}{c}{LMs with $\sim$7-9B active parameters} \\
\midrule    
Llama2-7B~\citep{touvron2023llama2openfoundation} & 6.7B & \emoji{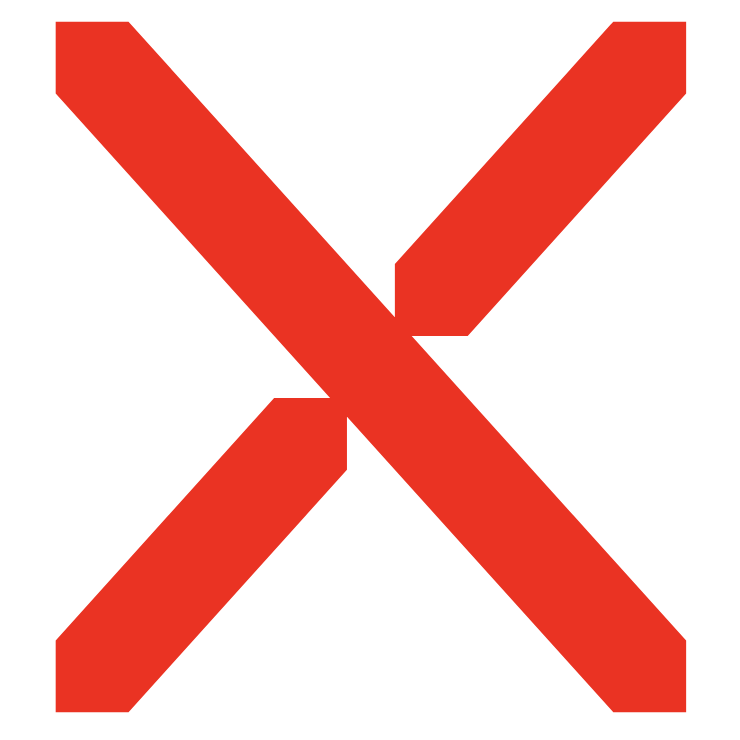} & 46.2 & 78.9 & 54.2 & 84.0 & 77.5 & 71.7 \\
OLMo-7B (0724)~\citep{groeneveld2024olmo} & 6.9B & \emoji{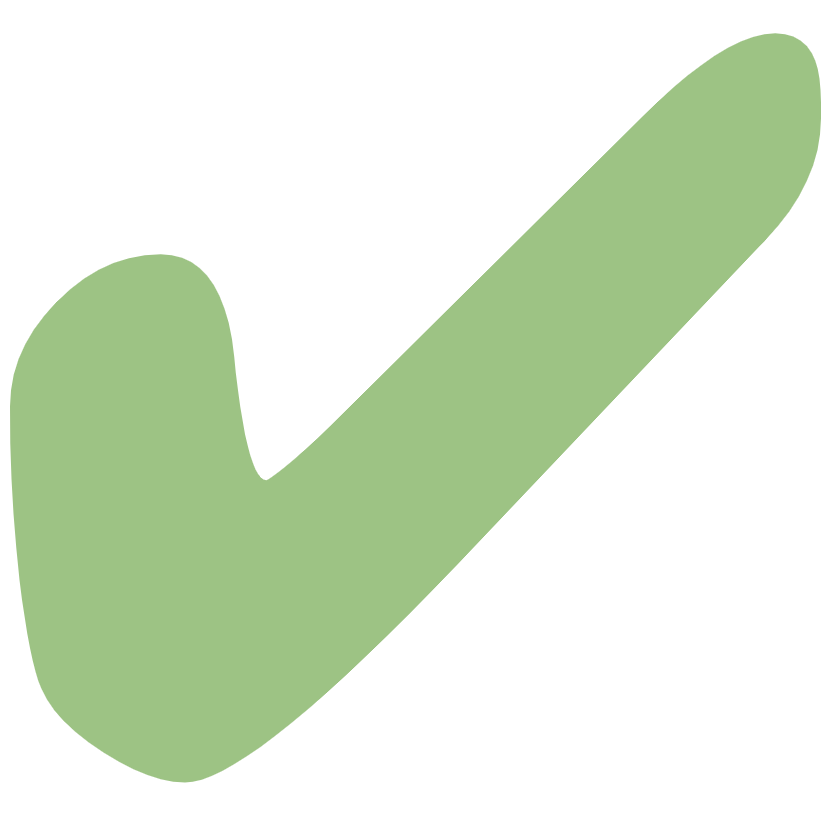} & 54.9 & 80.5 & 68.0 & 85.7 & 79.3 & 73.2 \\
Mistral-7B~\citep{jiang2023mistral} & 7.3B & \emoji{olmoe_cross} & 64.0 & 83.0 & 78.6 & 90.8 & 82.8 & 77.9 \\
DCLM-7B~\citep{li2024datacomplm} & 6.9B & \emoji{olmoe_checkmark} & 64.4 & 82.3 & 79.8 & 92.3 & 80.1 & 77.3 \\
Llama3.1-8B~\citep{dubey2024llama3herdmodels} & 8.0B & \emoji{olmoe_cross} & 66.9 & 81.6 & 79.5 & 91.7 & 81.1 & 76.6 \\
Gemma2-9B~\citep{gemmateam2024gemma2improvingopen} & 9.2B & \emoji{olmoe_cross} & \textbf{70.6} & \textbf{87.3} & \textbf{89.5} & \textbf{95.5} & \textbf{86.1} & \textbf{78.8} \\
\midrule
\multicolumn{9}{c}{LMs with $\sim$2-3B active parameters} \\
\midrule
OpenMoE-3B-9B~\citep{xue2024openmoe}& 2.6B & \emoji{olmoe_checkmark} & 27.4 & 44.4 & 29.3 & 50.6 & 63.3 & 51.9 \\
StableLM-2B~\citep{bellagente2024stablelm216b} & 1.6B & \emoji{olmoe_cross} & 40.4 & 70.3 & 50.6 & 75.3 & 75.6 & 65.8 \\
DeepSeek-3B-16B~\citep{dai2024deepseekmoeultimateexpertspecialization} & 2.9B & \emoji{olmoe_cross} & 45.5 & 80.4 & 53.4 & 82.7 & 80.1 & \textbf{73.2} \\
JetMoE-2B-9B~\citep{shen2024jetmoe} & 2.2B & \emoji{olmoe_cross} & 49.1 & \textbf{81.7} & 61.4 & 81.9 & 80.3 & 70.7 \\
Gemma2-3B~\citep{gemmateam2024gemma2improvingopen} & 2.6B & \emoji{olmoe_cross} & 53.3 & 74.6 & 67.5 & 84.3 & 78.5 & 71.8 \\ 
Qwen1.5-3B-14B~\citep{qwen_moe} & 2.7B & \emoji{olmoe_cross} & \textbf{62.4} & 80.0 & \textbf{77.4} & \textbf{91.6} & \textbf{81.0} & 72.3 \\
\midrule
\multicolumn{9}{c}{LMs with $\sim$1B active parameters} \\
\midrule
Pythia-1B~\citep{biderman2023pythiasuiteanalyzinglarge} & 1.1B & \emoji{olmoe_checkmark} & 31.1 & 48.0 & 31.4 & 63.4 & 68.9 & 52.7 \\
OLMo-1B (0724)~\citep{groeneveld2024olmo} & 1.3B & \emoji{olmoe_checkmark} & 32.1 & 67.5 & 36.4 & 53.5 & 74.0 & 62.9 \\
TinyLlama-1B~\citep{zhang2024tinyllamaopensourcesmalllanguage} & 1.1B & \emoji{olmoe_checkmark} & 33.6 & 60.8 & 38.1 & 69.5 & 71.7 & 60.1 \\
Llama3.2-1B~\citep{dubey2024llama3herdmodels} & 1.2B & \emoji{olmoe_cross} & 38.2 & 67.3 & 43.5 & 71.6 & 73.7 & 62.5 \\
DCLM-1B~\citep{li2024datacomplm} & 1.4B & \emoji{olmoe_checkmark} & 48.5 & 75.1 & 57.6 & 79.5 & 76.6 & 68.1 \\
\rowcolor{lightOlmoeYellow}
\modelsmall{} & 1.3B & \emoji{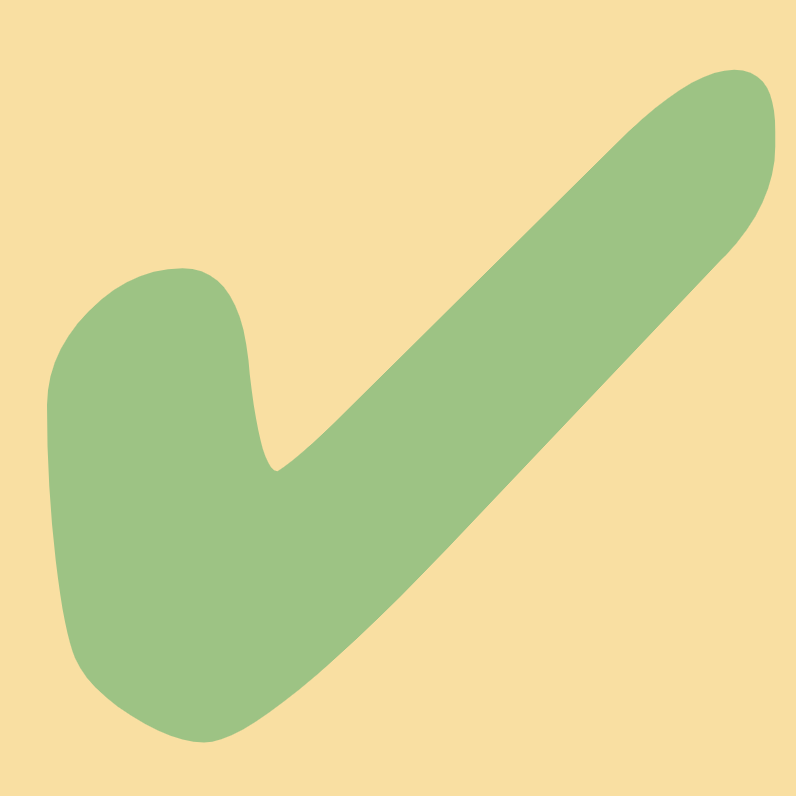} & \textbf{54.1} & \textbf{80.0} & \textbf{62.1} & \textbf{84.2} & \textbf{79.8} & \textbf{70.2} \\
\bottomrule
\end{tabular}
\vspace{.5em}
}
\caption{\textbf{\modelsmall{} after pretraining versus larger MoEs and dense LMs.} We compare with dense LMs close to \modelsmall{} either in active parameters (1B, approximates speed and cost) or total parameters (7B, approximates memory requirements). Model names contain rounded parameter counts: \texttt{model-active-total} for MoEs and \texttt{model-total} for dense LMs (this leads to some differences to official names, e.g., while called ``Gemma2-2B'' it actually has 2.6B active and total parameters~\citep{gemmateam2024gemma2improvingopen}). Chall.~= Challenge. We run all evaluations ourselves with 5 few-shots, see \autoref{sec:evalsetup} for details.
}
\label{tab:eval}
\end{table}

\begin{table}[htbp]
{\footnotesize
\begin{tabular}{l|ccccccc|c}
\toprule
&  &  &  & \textbf{Human-} & \textbf{Alpaca-} & & \\
\textbf{Task ($\rightarrow$)} & \textbf{MMLU} & \textbf{GSM8k} & \textbf{BBH} & \textbf{Eval} & \textbf{Eval 1.0} & \textbf{XSTest} & \textbf{IFEval} & \textbf{Avg} \\
\textbf{Setup ($\rightarrow$)} & \scriptsize{0-shot} & \scriptsize{8-shot CoT} & \scriptsize{3-shot} & \scriptsize{0-shot} & \scriptsize{0-shot} &\scriptsize{0-shot} & \scriptsize{0-shot} & \scriptsize{} \\
\textbf{Metric ($\rightarrow$)} & \scriptsize{EM} & \scriptsize{EM} & \scriptsize{EM} & \scriptsize{Pass@10} & \scriptsize{\%win} & \scriptsize{F1} & \scriptsize{Loose Acc} \\
\midrule
OLMo-1B (0724) & 25.0 & 7.0 & 22.5 & 16.0 & - & 67.6 & 20.5 & - \\
+SFT & 36.0 & 12.5 & 27.2 & 21.2 & 41.5 & 81.9 & 26.1 & 35.9 \\
+DPO & 36.7 & 12.5 & 30.6 & 22.0 & 50.9 & 79.8 & 24.2 & 37.4\\
\midrule
OLMo-7B (0724) & 50.8 & 32.5 & 36.9 & 32.3 & - & 80.8 & 19.6 & - \\
+SFT & 54.2 & 25.0 & 35.7 & 38.5 & 70.9 & 86.1 & 39.7 & 49.3 \\
+DPO & 52.8 & 9.0 & 16.6 & 35.0 & 83.5 & \textbf{87.5} & 37.9 & 49.1 \\
\midrule
JetMoE-2B-9B & 45.6 & 43.0 & 37.2 & 54.6 & - & 68.2 & 20.0 & - \\
+SFT & 46.1 & 53.5 & 35.6 & 64.8 & 69.3 & 55.6 & 30.5 & 50.4 \\
\midrule
DeepSeek-3B-16B & 37.7 & 18.5 & 39.4 & 48.3 & - & 65.9 & 13.5 & - \\
+Chat & 48.5 & 46.5 & \textbf{40.8} & \textbf{70.1} & 74.8 & 85.6 & 32.3 & 57.0 \\
\midrule
Qwen1.5-3B-14B & \textbf{60.4} & 13.5 & 27.2 & 60.2 & - & 73.4 & 20.9 & - \\
+Chat & 58.9 & \textbf{55.5} & 21.3 & 59.7 & 83.9 & 85.6 & 36.2 & 57.3 \\
\midrule
{\modelsmall{}} & 49.8 & 3.0 & 33.6 & 22.4 & - & 59.7 & 16.6 & - \\
\textbf{+SFT} & 51.4 & 40.5 & 38.0 & 51.6 & 69.2 & 84.1 & 43.3 & 54.0 \\
\rowcolor{lightOlmoeYellow}
\textbf{+DPO} & 51.9 & 45.5 & 37.0 & 54.8 & \textbf{84.0} & 82.6 & \textbf{48.1} & \textbf{57.7} \\
\bottomrule
\end{tabular}
}
\vspace{.5em}
\caption{\textbf{\modelsmall{} after adaptation versus other models.} We find the JetMoE chat model (\url{https://hf.co/jetmoe/jetmoe-8b-chat}) has random scores thus we exclude it. Model names contain rounded parameter counts: \texttt{model-active-total} for MoEs and \texttt{model-total} for dense LMs. We run all evaluations ourselves (\autoref{sec:evalsetup}). Models use different mixes for adaptation, e.g., \model{} is trained on an improved version of the pipeline used for OLMo models.}
\label{tab:adaptresults}
\end{table}

\paragraph{After pretraining} In \autoref{tab:eval} we benchmark \modelsmall{} on common downstream tasks. We find that \modelsmall{} performs best among models that use less than 2B active parameters, making it the most economical option for many use cases of LMs. For larger budgets, Qwen1.5-3B-14B has stronger performance but has more than double the active and total parameters than \modelsmall{}. We find that despite requiring $\sim$6--7$\times$ less compute per forward pass, \modelsmall{} outperforms some dense LMs with 7B parameters such as Llama2-7B~\citep{touvron2023llama2openfoundation}, but falls short of others like Llama3.1-8B~\citep{dubey2024llama3herdmodels}. \autoref{fig:overview} compares MMLU performance with active parameters, a proxy for the value of a model given its cost, of \modelsmall{} and other LMs. \modelsmall{} is the state of the art in its cost regime.

\paragraph{After adaptation} In \autoref{tab:adaptresults}, we benchmark our instruction (SFT) and preference (DPO) tuning of \modelsmall{}. SFT improves our model on all tasks measured. We observe a $>$10$\times$ gain on GSM8k, likely due to our inclusion of additional math data to account for the relatively small amounts of math data during pretraining (\autoref{sec:pretraining}). DPO helps on most tasks, especially AlpacaEval which aligns with findings from prior work~\citep{wang2023far,ivison2023camels,muennighoff2024generativerepresentationalinstructiontuning}. Our DPO model, which we refer to as \modelsmalldpo{}, has the highest average among all models benchmarked. We find it to outperform the chat version of Qwen1.5-3B-14B despite Qwen having $>$2$\times$ more parameters and its pretrained model outperforming \modelsmall{} in \autoref{tab:eval}. The 84\% score on AlpacaEval also outperforms much larger dense models on the leaderboard,\footnote{\url{https://tatsu-lab.github.io/alpaca_eval/}} such as Llama2-13B-Chat~\citep{touvron2023llama2openfoundation}.

\FloatBarrier

\section{Experimenting with Alternative Design Choices}
\label{sec:ablations}

In this section, we present pretraining and adaptation experiments that have led to \modelsmall{}. We group them into experiments on settings specific to Mixture-of-Experts (\autoref{sec:moespecific}), experiments on settings applicable to both dense LMs and MoEs (\autoref{sec:notmoespecific}), and adaptation experiments (\autoref{sec:adapt}). In pretraining experiments, we often use MMLU Var, a version of MMLU~\citep{hendrycks2021measuringmassivemultitasklanguage} with varying few-shots and a different format that provides signal earlier during training. We describe our full evaluation setup in \autoref{sec:evalsetup} and provide additional experiments in \autoref{sec:addablations}. Each experiment links to a Weights \& Biases report with more validation and downstream results, and the full configurations of the runs. To isolate the impact of changes and minimize confounders, we vary only one hyperparameter for each experiment. Nevertheless, due to the large number of hyperparameters, some results may change under different configurations and we cannot guarantee the correctness of each of our hyperparameter choices. Models are not comparable across different experiments, as we vary the base model to incorporate successful findings.

\subsection{MoE-specific Pretraining Settings}
\label{sec:moespecific}

\subsubsection{Mixture-of-Experts vs.~Dense}
\label{sec:moevsdense}

\begin{figure*}[htbp]
\centering
\begin{center}
\includegraphics[width=\textwidth]{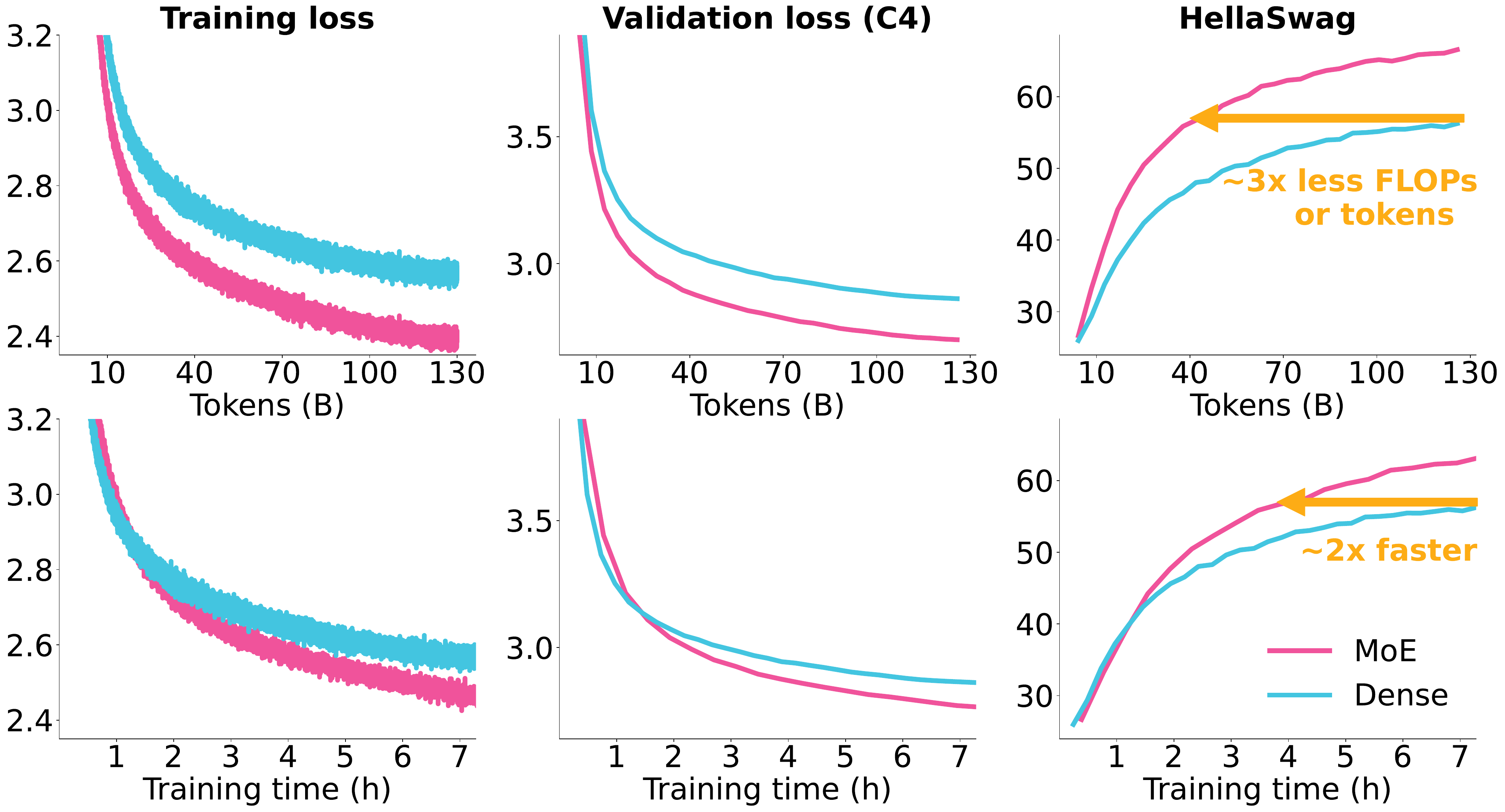}
\caption{\textbf{MoE vs.~Dense.} We train a 1.3B parameter dense model and a 1.3B active, 6.9B total parameter MoE model, each on 128 H100 GPUs. Apart from MoE-related changes, we train both with the same configuration for 130B tokens. The MoE contains 64 experts out of which 8 are activated with an FFN dimension of 1,024, while the dense model has an FFN dimension of 8,192. Thus both have the same number of active parameters. \textbf{Top:} The MoE reaches the final dense performance with $\sim$3$\times$ fewer tokens (or FLOPs, as both have the same active parameters ignoring the trivial router parameters). \textbf{Bottom:} Due to some memory overhead, this equates to $\sim$2$\times$ faster training. More results, logs, and configurations: \url{https://wandb.ai/ai2-llm/olmoe/reports/Plot-MoE-vs-Dense--Vmlldzo4OTM0Mjkx}}
\label{fig:moevsdense}
\end{center}
\end{figure*}

Prior work reports various speed-ups of MoEs over dense models: \citet{artetxe2022efficientlargescalelanguage} report that MoEs require 2--4$\times$ less compute to match dense models, MoMa~\citep{lin2024momaefficientearlyfusionpretraining} exhibits 2.6$\times$ FLOP savings for language tasks, Arctic~\citep{arcticcookbook} yields 4$\times$ FLOP savings but for very different dense and MoE configurations, and Switch Transformers~\citep{fedus2022switch} train 2-7$\times$ faster with MoEs but for encoder-decoder models while the other works study decoder-only LMs~\citep{radford2019language}.

In \autoref{fig:moevsdense}, we compare MoEs and dense models in a controlled setup. We find that our MoE reaches the performance of the dense model with $\sim$3$\times$ fewer tokens equivalent to $\sim$3$\times$ less compute measured in FLOPs. However, due to the additional memory overhead of training the MoE with its 7B total parameters, it processes fewer tokens per second than the dense model (23,600 tokens per second per GPU for the MoE vs. 37,500 for dense). Thus, in terms of training time, it reaches the performance of the dense model only $\sim$2$\times$ faster. There are likely optimizations possible that would bring the speed-up closer to the 3$\times$ token speed-up, which we leave to future work. Based on these results, \colorbox{lightOlmoeYellow}{we select an MoE configuration with 6.9B total and 1.3B active parameters} matching OLMo-7B in total and OLMo-1B in active parameter count, respectively.

\subsubsection{Expert Granularity}
\label{sec:granularity}

\begin{figure*}[htbp]
\centering
\begin{center}
\includegraphics[width=\textwidth]{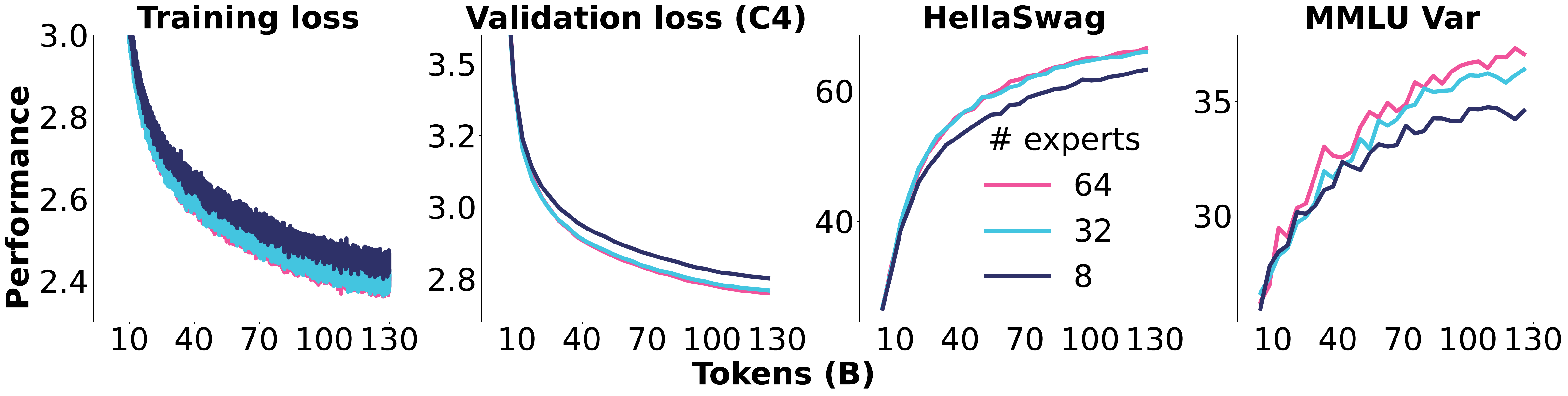}
\caption{\textbf{Expert granularity.} We vary the number of experts in tandem with the FFN dimension to ensure that active and total parameters and thus compute cost remain the same. For example, for 64 experts, the FFN dimension is 1,024 and 8 experts are activated, while for 32 experts it is 2,048 with 4 activated experts. More results, logs, and configurations: \url{https://wandb.ai/ai2-llm/olmoe/reports/Plot-Granularity--Vmlldzo4OTIxOTE4}}
\label{fig:granularity}
\end{center}
\end{figure*}

\citet{dai2024deepseekmoeultimateexpertspecialization} propose to use small fine-grained experts to allow more combinations of experts and thus make the model more flexible. For example, the Mixtral model~\citep{jiang2024mixtral} uses the common configuration of 8 experts per layer, 2 of which are activated. This allows for $\binom{8}{2}=28$ combinations per layer. By halving the size of each expert and therefore doubling the number of experts to maintain the same compute and parameter budget, we can increase the possible combinations to $\binom{16}{4}=1,820$. \citet{krajewski2024scaling} investigate compute-optimal granularity configurations finding that higher compute budgets warrant more granular experts.

In \autoref{fig:granularity}, we observe that more granular experts improve training loss, validation loss, and downstream performance. The 8-expert configuration uses 1 active expert, which yields $\binom{8}{1}=8$ combinations. By quartering the size of each expert but increasing the number to 32 with 4 active ones ($\binom{32}{4}=35,960$ combinations), we observe an improvement of around 10\% on HellaSwag and MMLU at around 130 billion tokens. However, we find that there are diminishing returns to granularity. The additional increase to 64 experts with 8 active ones ($\binom{64}{8}=4,426,165,368$ combinations) improves downstream metrics by a smaller amount of 1--2\%. For our \modelsmall{} compute budget\footnote{Approximated via $6*N*D$~\citep{kaplan2020scalinglawsneurallanguage}, where $N$ are active parameters (1B) and $D$ are training tokens (5T).} of $3\times10^{22}$, \citet{krajewski2024scaling} predict an optimal number of experts of 256 ($G=32$ in their paper). However, their predictions are for compute-optimal models~\citep{hoffmann2022trainingcomputeoptimallargelanguage,clark2022unifiedscalinglawsrouted}, while we train for 5T tokens, which is orders of magnitude beyond what would be conventionally considered optimal for our model size. Thus, their predictions may not extend to our setup, and \colorbox{lightOlmoeYellow}{we stick with 64 experts for \modelsmall{}}, also due to the diminishing returns in \autoref{fig:granularity}.

\subsubsection{Shared Experts}
\label{sec:shared}

\begin{figure*}[htbp]
\centering
\begin{center}
\includegraphics[width=\textwidth]{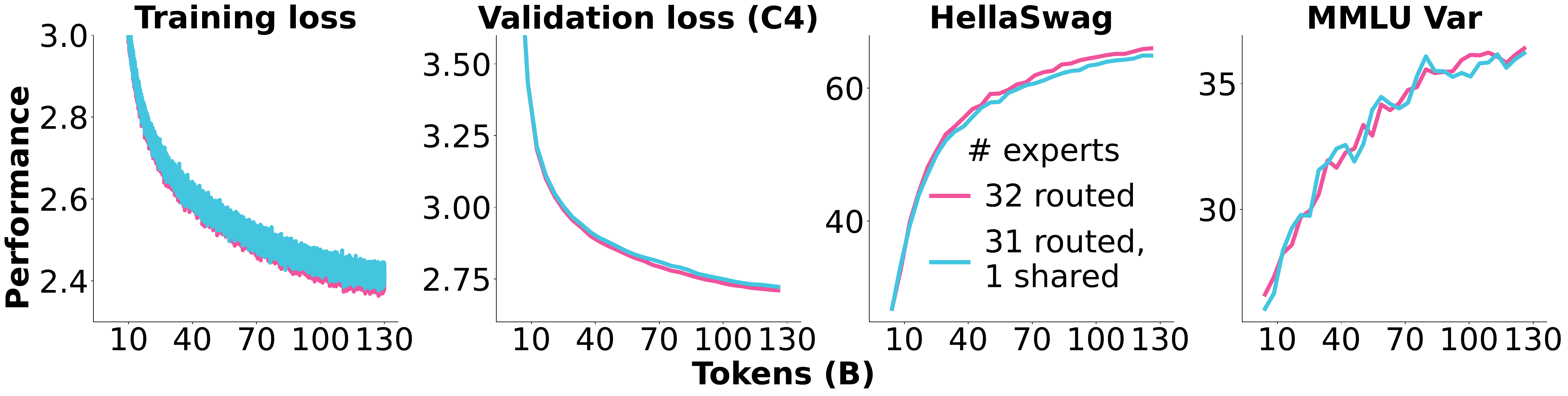}
\caption{\textbf{Shared experts.} Both setups have the same number of active and total parameters and use the same number of FLOPs. 4 of the 32 routed experts are activated, while it is 3 for the 31 routed experts of the other model, as it has 1 always-active shared expert. More results, logs, and configurations: \url{https://wandb.ai/ai2-llm/olmoe/reports/Plot-Expert-sharing--Vmlldzo4OTIyMjQz}}
\label{fig:shared}
\end{center}
\end{figure*}

\citet{dai2024deepseekmoeultimateexpertspecialization} propose training with a shared/fixed expert that is always used in addition to the routed experts. The intuition is to encourage the shared expert to learn common information and allow the other routed experts to learn more specialized knowledge. This should reduce redundancy among experts and thus lead to a better model as it can store more total information.

In \autoref{fig:shared}, we benchmark having a single shared and a single routed expert versus two routed experts. While both settings lead to similar performance, sharing an expert performs slightly worse. Sharing an expert removes flexibility from the model and thus goes against the findings in \autoref{sec:granularity} suggesting that allowing for more expert combinations improves performance. Specifically, the two models in \autoref{fig:shared} have $\binom{32}{4}=35,960$ and $\binom{31}{3}=4,495$ possible combinations per layer. Thus, removing one of the routed experts and turning it into a shared one eliminates almost 90\% of possible combinations. This likely acts as a counterforce to the potential benefits of isolating common knowledge in a shared expert. Based on these results, \colorbox{lightOlmoeYellow}{we do not use shared experts in \modelsmall{}}, but we do think that there is merit to the idea of experts that are activated more often or even always. However, rather than enforcing this behavior via a shared expert, we believe that it should be learned by the model. This is difficult with current setups due to the necessity of a load balancing loss (\autoref{sec:lbloss}) penalizing the model if tokens are not distributed equally among experts. Potential future work can explore removing the load balancing loss to allow for more flexible usage of experts.

\subsubsection{Expert Choice vs.~Token Choice}
\label{sec:expertchoice}

\begin{figure*}[htbp]
\centering
\begin{center}
\includegraphics[width=\textwidth]{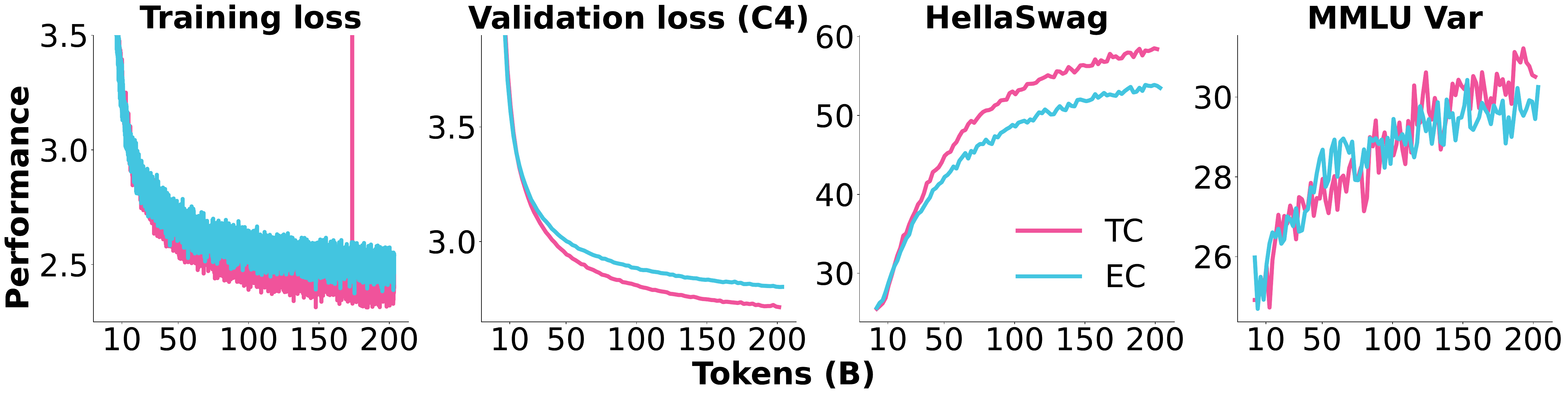}
\caption{\textbf{Expert choice (EC) vs.~token choice (TC).} Both models have an 8-expert MoE in every 2nd layer. For TC, 2 experts are activated per token, while for EC the capacity factor is 2. Thus, both models use the same number of active parameters. More results, logs, and configurations: \url{https://wandb.ai/ai2-llm/olmoe/reports/Plot-EC-vs-TC--Vmlldzo4MzkzMDM3}}
\label{fig:ectc}
\end{center}
\end{figure*}

The MoE router determines which experts process each input token (\autoref{sec:pretraining}). There are two common types~\citep{liu2024routersvisionmixtureexperts}: \textbf{expert choice (EC)}~\citep{zhou2022mixtureofexperts} and \textbf{token choice (TC)}~\citep{shazeer2017outrageously}. For EC, each expert selects a fixed number of tokens from the incoming sequence. By design, this leads to each expert processing the same number of tokens. This is the main benefit of EC as it ensures perfect load balance, which improves training throughput and removes the need for a load balancing loss. The main downside of EC is that it is not easily usable for autoregressive generation where a single token is processed at each step rather than the entire sequence in one~\citep{raposo2024mixtureofdepthsdynamicallyallocatingcompute}. Another potential downside is that EC can lead to token dropping, where some tokens are not selected by any expert, which can hurt performance~\citep{gale2022megablocksefficientsparsetraining}. At the same time, it can lead to some tokens being processed by multiple experts, which could also be beneficial as it allows the model to allocate more compute to some tokens~\citep{zhou2022mixtureofexperts}. For TC, each token selects a fixed number of experts. This can lead to many tokens choosing the same expert, hurting training efficiency. Therefore it is common to use TC with a load balancing loss~\citep{shazeer2017outrageously} to encourage equal distribution.

In \autoref{fig:ectc}, we benchmark EC and TC. We find that TC outperforms EC for the same token budget for all tasks depicted as well as other tasks like PIQA, SciQ, etc. which we report at \url{https://wandb.ai/ai2-llm/olmoe/reports/Plot-EC-vs-TC--Vmlldzo4MzkzMDM3}. While \citet{zhou2022mixtureofexperts} find EC to be better, our configuration slightly differs in that we use dropless MoEs~\citep{gale2022megablocksefficientsparsetraining} with a load balancing loss. Thus, our TC variant is expected to perform better than the TC variant in \citet{zhou2022mixtureofexperts}. We confirm findings that EC runs around 20\% faster at 29,400 tokens per second per device versus 24,400 for TC~\citep{zhou2022mixtureofexperts}. EC may be more beneficial in a multimodal setup~\citep{lin2024momaefficientearlyfusionpretraining} as dropping noisy image tokens is likely less harmful than text tokens. Thus, while \colorbox{lightOlmoeYellow}{we stick with TC for this release of \model{}}, we may revisit EC for future multimodal models.

\subsubsection{Sparse Upcycling}
\label{sec:upcycling}

\begin{figure*}[htbp]
\centering
\begin{center}
\includegraphics[width=\textwidth]{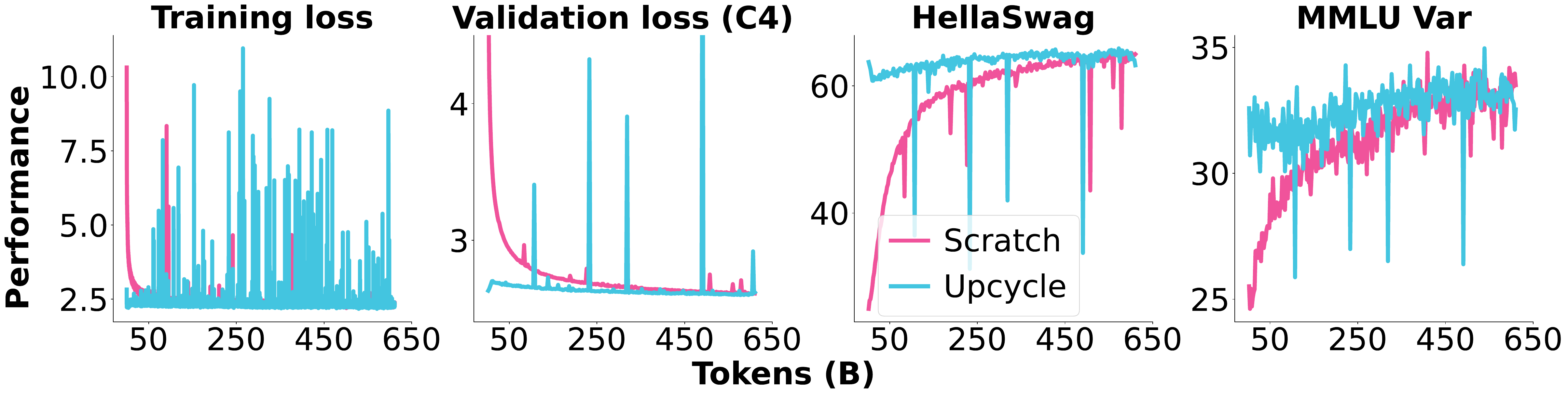}
\caption{\textbf{Sparse upcycling.} We upcycle OLMo-1B (0724) at 2T tokens into an MoE with 8 total experts of which 2 are activated and train it for an additional 610 billion tokens. We compare it to a model trained from scratch for 610 billion tokens. Except for this difference, both models use the same config, which includes some suboptimal settings that contribute to the instability, such as no QK-Norm (\autoref{sec:qknorm}) and no truncated normal init (\autoref{sec:init}). More results, logs, and configurations: \url{https://wandb.ai/ai2-llm/olmoe/reports/Plot-Scratch-vs-Upcycle--Vmlldzo4NDIyOTc4}}
\label{fig:upcycle}
\end{center}
\end{figure*}

\citet{komatsuzaki2023sparse} propose turning a dense model into a Mixture-of-Experts model via sparse upcycling: (1) The dense MLP is cloned for each desired expert to constitute MoE layers. (2) A newly initialized router is added in front of each MoE layer. (3) Pretraining continues with the new model so that the cloned MLPs can gradually specialize in different things and the router can be learned. They find that the upcycling approach maintains a performance advantage over a language model trained from scratch for up to 120\% of the compute budget of the original dense checkpoint that the sparse model was upcycled from. For example, if sparsely upcycling a 1.3B parameter model at 2 trillion tokens then only at 2.4 trillion tokens should an MoE trained from scratch catch up with the upcycled model. That is, the sparsely upcycled model would have been trained for another 400 billion tokens, thereby saving the equivalent of up to 2T tokens of compute. Other works such as MiniCPM~\citep{hu2024minicpm}, Qwen2~\citep{yang2024qwen2technicalreport} and reportedly Mixtral~\citep{mixtralupcycle,jiang2024mixtral} have adopted sparse upcycling but only share limited information about their configuration.

In \autoref{fig:upcycle}, we compare sparse upcycling OLMo-1B (0724)~\citep{groeneveld2024olmo} with training an MoE from scratch. We find that after 500B tokens, an otherwise equivalent MoE trained from scratch already catches up with the upcycled model, both on the metrics in \autoref{fig:upcycle} and our additional metrics at \url{https://wandb.ai/ai2-llm/olmoe/reports/Plot-Scratch-vs-Upcycle--Vmlldzo4NDIyOTc4}. At around 600B tokens, the MoE from scratch starts outperforming the upcycled MoE. Thus, it only requires 25\% of the compute budget of the original dense model to catch up as opposed to the 120\% reported in \citet{komatsuzaki2023sparse}. However, they use expert choice routing and study encoder-decoder models~\citep{raffel2023exploring}. Meanwhile, we use token choice routing (\autoref{sec:expertchoice}) and decoder-only models (\autoref{sec:pretraining}). Further, we upcycle a model that has already been significantly overtrained~\citep{gadre2024language}, i.e., a 1B model trained for 2T tokens. Its parameters are likely already in a very optimal range for a dense model, which may limit the amount of additional exploration possible after upcycling. This motivates us to experiment with adding noise to the upcycled weights outlined in \autoref{sec:addablations}, but we do not find it to lead to better performance. A large disadvantage of upcycling is that the upcycled MoE is constrained by some hyperparameters of the dense model. Specifically, OLMo-1B (0724) was trained without QK-Norm and normal initialization, both of which hurt stability in our experiments (\autoref{sec:qknorm}, \autoref{sec:init}). While it may be possible to simply add new QK-Norms and train them from scratch similar to the new router layer trained from scratch, it is impossible to change the initialization of the original dense model when upcycling it. Thus, as we want to change these hyperparameters and also train \modelsmall{} for around 250\% of the compute budget of the dense model (5T vs.~2T tokens), \colorbox{lightOlmoeYellow}{we do not use upcycling.}

\subsubsection{Load Balancing Loss}
\label{sec:lbloss}

\begin{figure*}[htbp]
\centering
\begin{center}
\includegraphics[width=\textwidth]{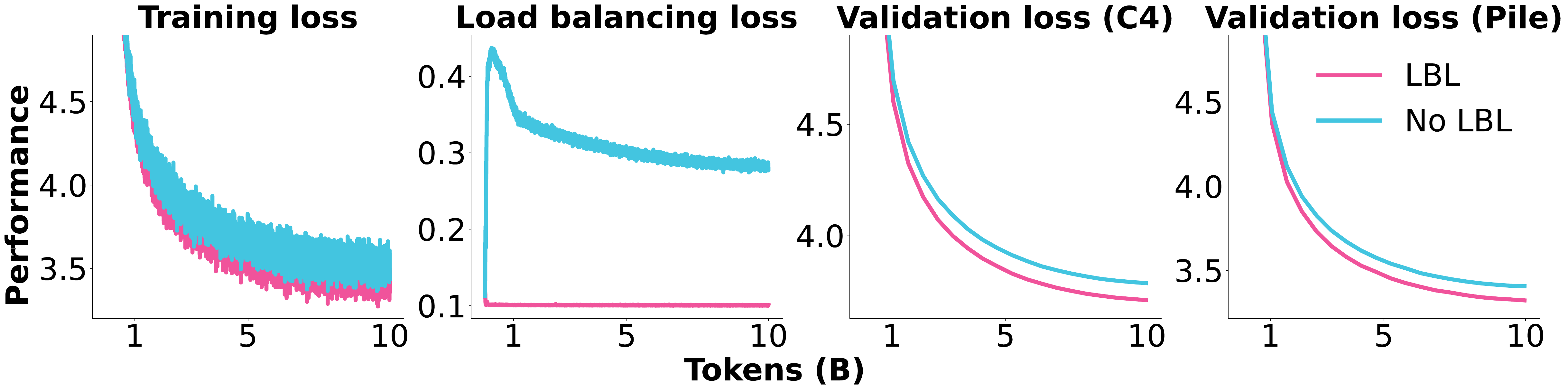}
\caption{\textbf{Impact of applying a load balancing loss (LBL).} The training loss plot excludes the load balancing loss for both models. More results, logs, and configurations: \url{https://wandb.ai/ai2-llm/olmoe/reports/Plot-LBL-vs-No-LBL--Vmlldzo4OTkyNDg4}}
\label{fig:lbloss}
\end{center}
\end{figure*}

\begin{figure*}[htbp]
\centering
\begin{center}
\includegraphics[width=0.9\textwidth]{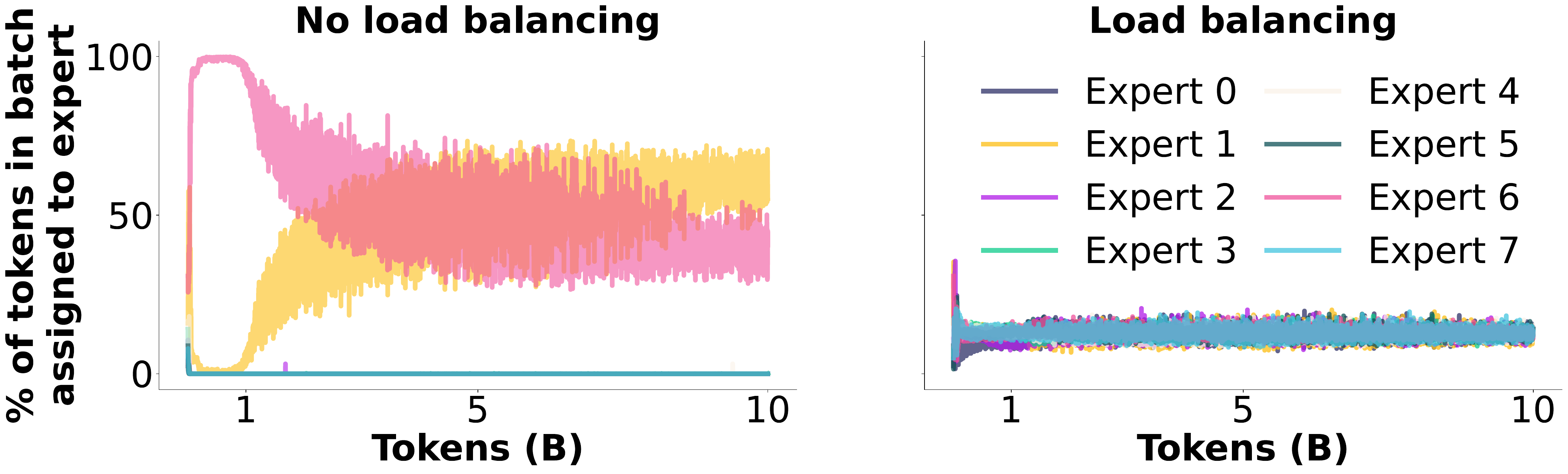}
\caption{\textbf{Expert assignment during training when using or not using a load balancing loss for the first MoE layer.} More results, logs, and configurations: \url{https://wandb.ai/ai2-llm/olmoe/reports/Plot-LBL-vs-No-LBL--Vmlldzo4OTkyNDg4}}
\label{fig:lbltoks}
\end{center}
\end{figure*}

\citet{shazeer2017outrageously} propose the load balancing loss to penalize the model if it is unbalanced, i.e., if it routes all tokens to only a few experts. This is based on the observation that without such penalty, models tend to update only a select few experts in each layer~\citep{eigen2014learningfactoredrepresentationsdeep,bengio2016conditionalcomputationneuralnetworks}. To compute the load balancing loss ($\mathcal{L}_{\textit{LB}}$) we multiply the fraction of tokens $f_i$ routed to one expert $E_i$ with the total routing probability $P_i$ allocated to $E_i$ for one batch and sum it across the number of experts $N_{E}$:
\begin{equation}  
\label{eq:lbl}
\mathcal{L}_{\textit{LB}} = N_{E} \cdot \sum_{i=1}^{N_{E}} f_i \cdot P_i
\end{equation}
The loss is further scaled by $N_{E}$ and a loss weight $\alpha$ (see \autoref{eq:loss}), which is an optional weight to determine the magnitude of the loss commonly set to 0.01~\citep{zoph2022stmoe,xue2024openmoe}. We do not experiment with changing the weight of 0.01.

In \autoref{fig:lbloss} we investigate the performance impact of using the auxiliary load balancing loss. We find that across training loss and validation losses, using the load balancing loss leads to better performance even after only a few billion tokens. We still measure the load balancing loss even when it is not used (``No LBL'') and find that while it spikes initially, it slowly decreases over the next few billion tokens. This behavior is also visible in \autoref{fig:lbltoks} (left), where initially all tokens in the first layer are assigned to the 6th expert (pink). Eventually, the model also starts assigning some tokens to the 1st expert (yellow). However, all other experts remain largely flat and are thus ``dead weights'' that take up GPU memory but are not used. Given these results, \colorbox{lightOlmoeYellow}{we use the auxiliary load balancing loss with a weight of 0.01} following prior work~\citep{shazeer2017outrageously,shen2024jetmoe}. However, getting rid of the load balancing loss is an important direction for future research as it constrains the flexibility of the model by forcing it to use all experts approximately equally. This could prevent the experts from specializing in certain data domains and may be a reason prior work has failed to find strong evidence of expert specialization~\citep{jiang2024mixtral,zoph2022stmoe}.

\subsubsection{Router Z-loss}
\label{sec:zloss}

\begin{figure*}[htbp]
\centering
\begin{center}
\includegraphics[width=\textwidth]{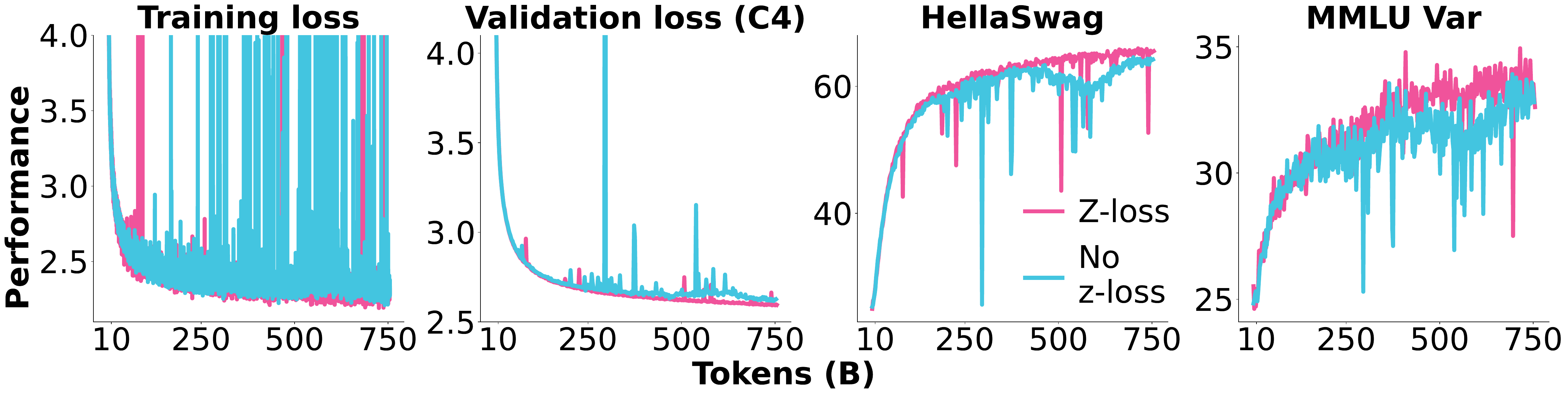}
\caption{\textbf{Router z-loss.} We compare adding router z-loss with a loss weight of 0.001 versus no additional z-loss. More results, logs, and configurations: \url{https://wandb.ai/ai2-llm/olmoe/reports/Plot-Zloss-vs-none--Vmlldzo4NDM4NjUz}}
\label{fig:zloss}
\end{center}
\end{figure*}

\citet{zoph2022stmoe} propose the router z-loss to improve both the stability and quality of MoE models. This auxiliary loss penalizes large logits coming into the gating network. Such large logits can lead to numeric overflows in the large matrix multiplications happening in the MoE layer. It is computed by exponentiating the logits $x_j$ right before the router layer summed across the number of experts $N_E$ and averaged across the batch $B$, thereby making larger logits lead to a larger loss:
\begin{equation}
\label{eq:rzl}
\mathcal{L}_{\textit{RZ}}(x) = \frac{1}{B} \cdot \sum_{i=1}^B \left(\log \sum_{j=1}^{N_{E}} \exp({x_j^{(i)}}) \right)^2
\end{equation}
The loss is further multiplied with an optional loss weight, $\beta$ (see \autoref{eq:loss}), to determine the magnitude of the loss commonly set to 0.001~\citep{zoph2022stmoe,shen2024jetmoe}. We do not experiment with changing the weight of 0.001.

In \autoref{fig:zloss}, we confirm that across training loss, validation loss, and downstream performance adding the router z-loss improves stability (less spikes) and quality (lower loss and higher downstream performance). Thus, despite it reducing throughput by $\sim$2\% \colorbox{lightOlmoeYellow}{we use the router z-loss for \modelsmall{} with a weight of 0.001} as in \citet{zoph2022stmoe}. 

\subsection{General Pretraining Settings}
\label{sec:notmoespecific}

\subsubsection{Dataset Experiments}
\label{sec:data}

\begin{figure*}[htbp]
\centering
\begin{center}
\includegraphics[width=\textwidth]{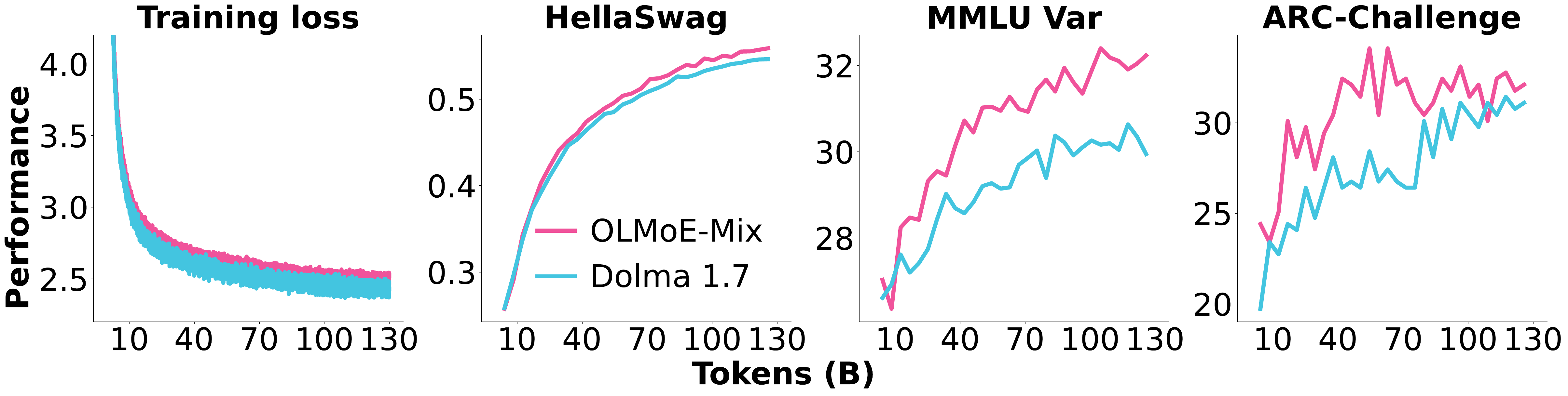}
\caption{\textbf{\data{} vs.~Dolma 1.7.} We compare our data mix described in \autoref{sec:pretraining} with Dolma 1.7 used to train prior OLMo models. Lower training loss does not mean that one dataset is better, but rather suggests which dataset is easier for the model to learn. More results, logs, and configurations: \url{https://wandb.ai/ai2-llm/olmoe/reports/Plot-Dolma-1-7-vs-Dolma-OLMoE--Vmlldzo4OTIxNTg5}}
\label{fig:dataset}
\end{center}
\end{figure*}

\citet{li2024datacomplm} release the DCLM-Baseline dataset and establish that it leads to better language models than Dolma 1.7 and other datasets as measured on common benchmarks like MMLU~\citep{hendrycks2021measuringmassivemultitasklanguage}. This motivates us to mix their DCLM dataset with some components from Dolma 1.7 that we deem to be high-quality; see \autoref{sec:pretraining}. In \autoref{fig:dataset}, we compare our mix, \data{}, with Dolma 1.7 in a controlled setup. We find that \data{} leads to clear gains on all three downstream metrics, especially MMLU. DCLM-Baseline has been created through a series of dataset ablations targeting MMLU and other downstream metrics, which explains these results. We also compare adding Reddit and FLAN to our mix as detailed in \autoref{sec:addablations}, but do not find consistent performance gains. We do not have a strong intuition for why adding these datasets does not help and a more automatic approach to dataset mixing may be desirable for future iterations~\citep{liu2024regmixdatamixtureregression,albalak2024surveydataselectionlanguage}. \colorbox{lightOlmoeYellow}{We pretrain using our mix of DCLM-Baseline and Dolma 1.7 dubbed \data{}.}

\subsubsection{Initialization}
\label{sec:init}

\begin{figure*}[htbp]
\centering
\begin{center}
\includegraphics[width=\textwidth]{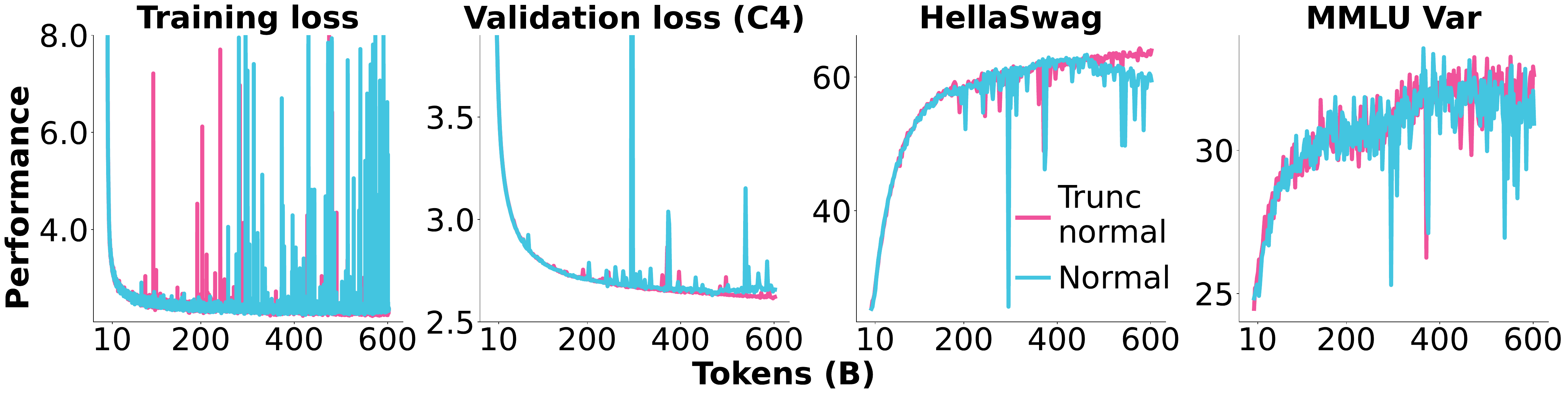}
\caption{\textbf{Initialization.} We compare a normal initialization with a standard deviation (std) of 0.02 with a truncated normal initialization with a maximum (minimum) cut-off of 0.06 (--0.06) corresponding to three stds~(3$\times$0.02). More results, logs, and configurations: \url{https://wandb.ai/ai2-llm/olmoe/reports/Plot-Init--Vmlldzo4NDIzMzM5}}
\label{fig:init}
\end{center}
\end{figure*}

Few prior works on Mixture-of-Experts share their initialization strategy. Even the most open MoEs prior to this work, JetMoE~\citep{shen2024jetmoe} and OpenMoE~\citep{xue2024openmoe}, do not mention their initialization scheme. For DeepSeekMoE~\citep{dai2024deepseekmoeultimateexpertspecialization} and DeepSeekV2~\citep{deepseekai2024deepseekv2}, the authors share that they use a normal initialization with a standard deviation (std) of 0.006. For dense language models, a normal initialization with an std of 0.02 has been commonly used as popularized by \citet{shoeybi2020megatronlmtrainingmultibillionparameter}.

In \autoref{fig:init}, we find a truncated normal initialization leads to more stable training and better performance than a regular normal initialization. The difference between the two initializations only becomes clear at around 450 billion tokens, where the model with the normal initialization starts to diverge. This is despite both models using the same configuration except for the difference in weight initialization. Having to train for hundreds of billions of tokens until an experiment provides a clear signal is one of the key challenges of pretraining ablations. \colorbox{lightOlmoeYellow}{We use the truncated normal initialization for \modelsmall{}.}

\subsubsection{RMSNorm}
\label{sec:ln}

\begin{figure*}[t]
\centering
\begin{center}
\includegraphics[width=\textwidth]{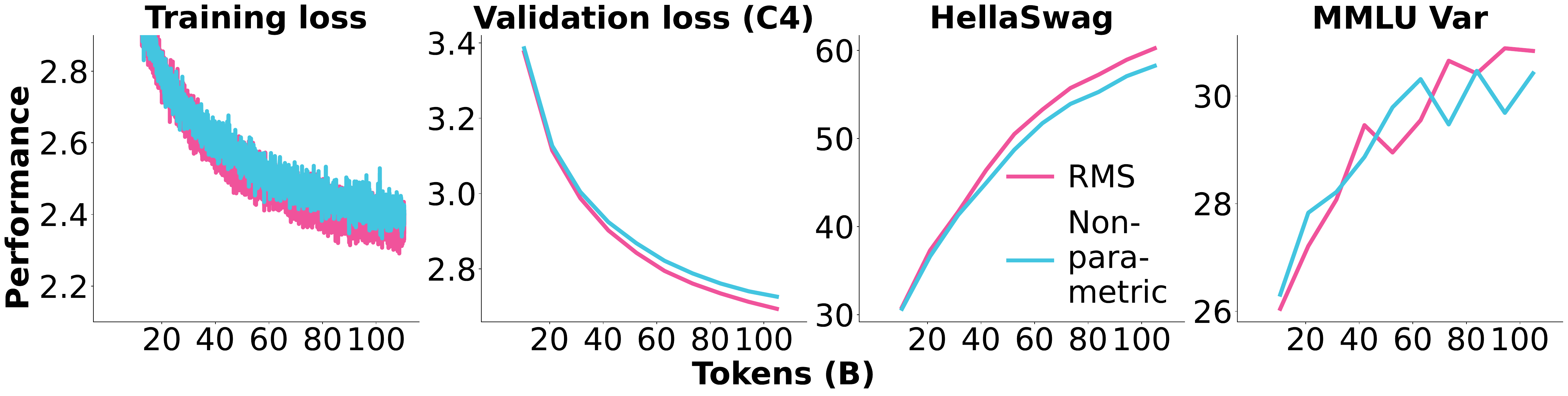}
\caption{\textbf{Non-parametric layer normalization vs. RMSNorm.} More results, logs, and configurations: \url{https://wandb.ai/ai2-llm/olmoe/reports/Plot-LN--Vmlldzo4NDQyMTAz}}
\label{fig:ln}
\end{center}
\end{figure*}

\begin{figure*}[t]
\centering
\begin{center}
\includegraphics[width=\textwidth]{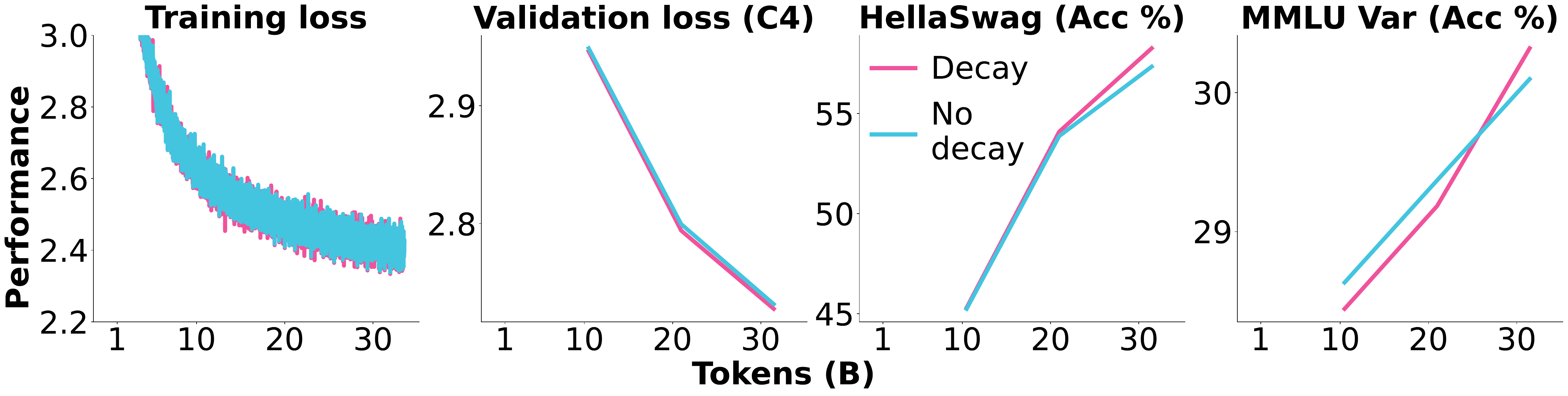}
\caption{\textbf{Decaying the RMSNorm parameters.} More results, logs, and configurations: \url{https://wandb.ai/ai2-llm/olmoe/reports/Plot-Decay-LN--Vmlldzo4NDQ1NDYy}}
\label{fig:normdecay}
\end{center}
\end{figure*}

\begin{wrapfigure}{r}{0.34\textwidth}
\vspace{-5em}
\centering
\begin{center}
\includegraphics[width=\linewidth]{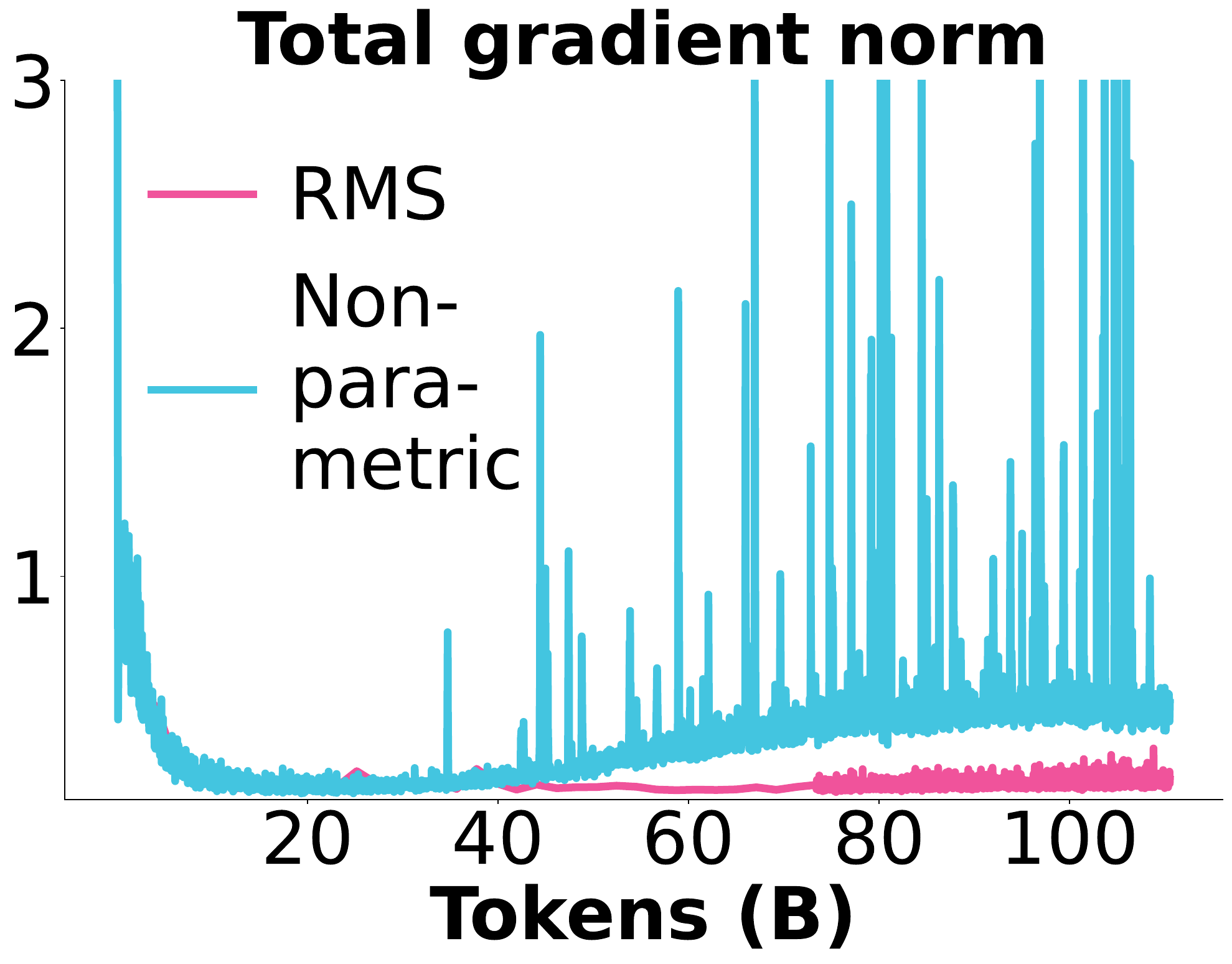}
\caption{\textbf{Total norm of the gradients when training with RMS or non-parametric normalization.} We increase the logging interval of the RMS run at 75B tokens, hence its change in thickness.}
\label{fig:lngradnorm}
\end{center}
\end{wrapfigure}

OLMo~\citep{groeneveld2024olmo} uses non-parametric layer normalization~\citep{ba2016layer}, mainly as it is significantly faster than the commonly used RMSNorm~\citep{zhang2019rootmeansquarelayer,mehta2024openelm}. This is an unusual choice as most LMs use RMSNorm, such as the Llama~\citep{touvron2023llama,touvron2023llama2openfoundation,dubey2024llama3herdmodels}, Gemma~\citep{gemmateam2024gemmaopenmodelsbased,gemmateam2024gemma2improvingopen}, and Qwen~\citep{bai2023qwen,yang2024qwen2technicalreport} model families.

In \autoref{fig:ln}, we observe that replacing the non-parametric layer normalization in OLMo with a parametric RMSNorm leads to better performance. This is likely because the non-parametric layer normalization leads to a large number of spikes in the gradients as seen in \autoref{fig:lngradnorm}. We clip gradients at 1.0, which prevents these spikes from leading to very large and potentially disruptive parameter updates. However, the clipped gradients may still harm the performance of the model as they are no longer the true gradients. Thus, despite RMSNorm lowering our training throughput by 15\%, \colorbox{lightOlmoeYellow}{we train our final model with RMSNorm.} We include the RMSNorm parameters in weight decay as we find that it performs slightly better (\autoref{fig:normdecay}) even though it is common practice to exclude them.\footnote{\url{https://github.com/karpathy/minGPT/pull/24\#issuecomment-679316025}}

\subsubsection{Decaying Embedding Parameters}
\label{sec:embdecay}

\begin{figure*}[t]
\centering
\begin{center}
\includegraphics[width=\textwidth]{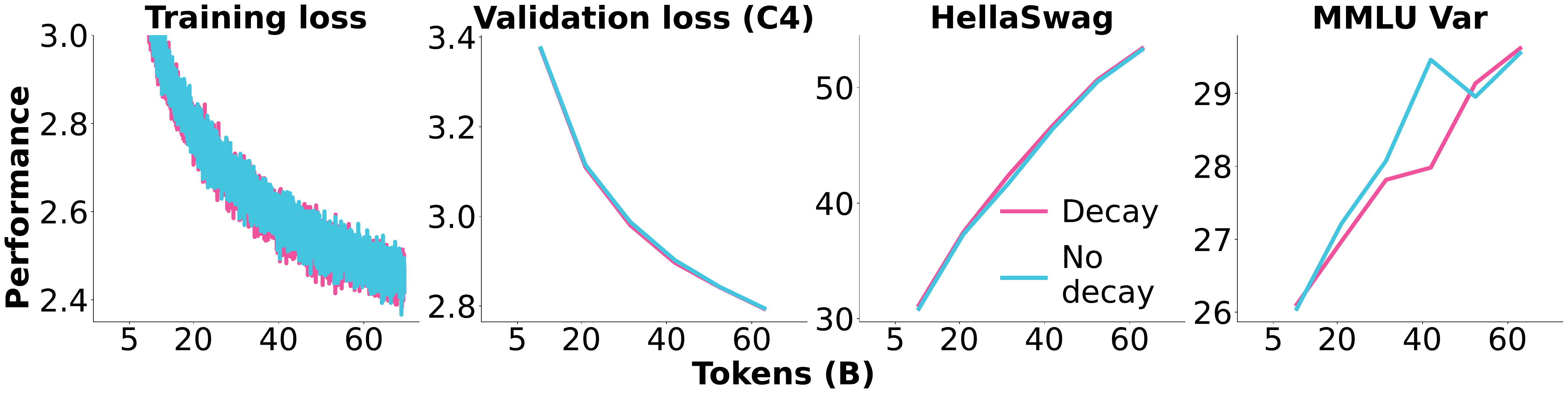}
\caption{\textbf{Decaying the embedding parameters.} More results, logs, and configurations: \url{https://api.wandb.ai/links/ai2-llm/3h22onp5}}
\label{fig:embdecay}
\end{center}
\end{figure*}

Similar to the RMSNorm parameters (\autoref{sec:ln}), embedding parameters are commonly excluded from weight decay.\footnote{\url{https://github.com/karpathy/minGPT/pull/24\#issuecomment-679316025}} In \autoref{fig:embdecay} we find that whether or not they are decayed has only a minor impact on performance, with decaying being slightly better. Thus for simplicity, \colorbox{lightOlmoeYellow}{we weight decay all parameters in \modelsmall{} including embedding and RMSNorm.}

\subsubsection{QK-Norm}
\label{sec:qknorm}

\begin{figure*}[t]
\centering
\begin{center}
\includegraphics[width=\textwidth]{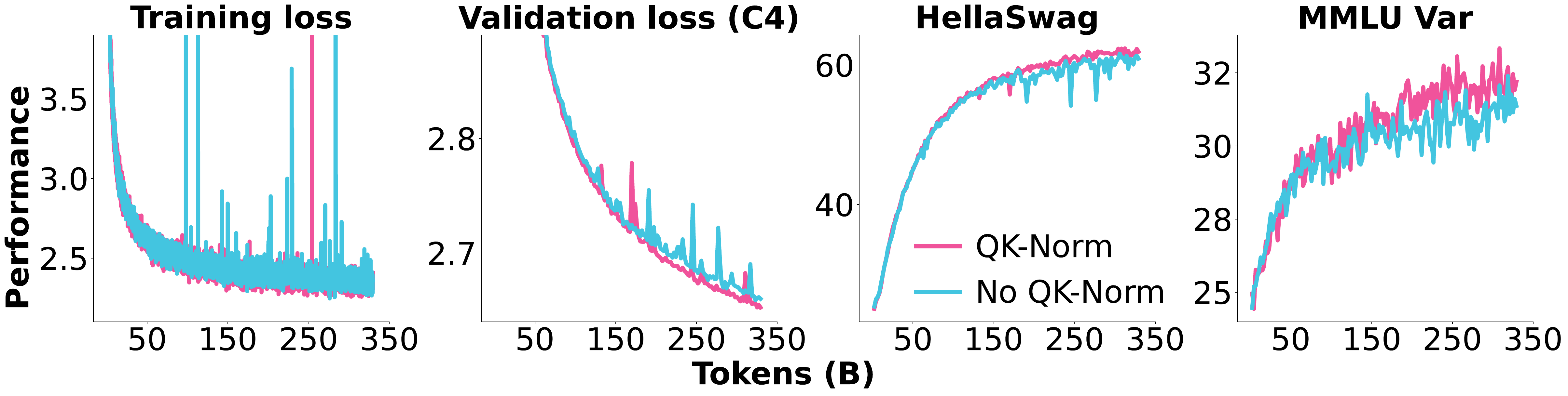}
\caption{\textbf{Query-Key layer normalization (QK-Norm).} Both models use non-parametric layer normalization. QK-Norm corresponds to additional layer normalization of the query and key projections. More results, logs, and configurations: \url{https://wandb.ai/ai2-llm/olmoe/reports/Plot-QKNorm-vs-none--Vmlldzo4NDIzMzE2}}
\label{fig:qknorm}
\end{center}
\end{figure*}

Some works have reported stability improvements from adding layer normalization after the query and key projections (``QK-Norm'')~\citep{chameleonteam2024chameleon,mehta2024openelm,dehghani2023scaling}. QK-Norm can prevent the subsequent attention operation from leading to very large logits that may lead to numeric overflows and destabilize the network, especially when training in low precision. Like layer normalization at other places in the model, the QK-Norm could be non-parametric or use the parametric RMSNorm (\autoref{sec:ln}).

In \autoref{fig:qknorm}, we compare using QK-Norm with no normalization after the query and key projections. We find that QK-Norm leads to some stability and performance improvements. We perform this experiment with non-parametric layer normalization as used in OLMo~\citep{groeneveld2024olmo}, while we used parametric RMS layer normalization~\citep{zhang2019rootmeansquarelayer} for \modelsmall{} (\autoref{sec:ln}). To ensure the benefit of QK-Norm is not an artifact of comparing with non-parametric layer normalization, we run another experiment with RMS layer normalization and still find QK-Norm to lead to slightly better training loss and to prevent a large grad norm spike.\footnote{\url{https://wandb.ai/ai2-llm/olmoe/reports/Plot-QKNorm-revisited--Vmlldzo4NTc2NTIz}} Thus, \colorbox{lightOlmoeYellow}{we use QK-Norm for \modelsmall{}} despite it reducing throughput by almost 10\%.

\subsubsection{AdamW Epsilon}
\label{sec:eps}

\begin{figure*}[htbp]
\centering
\begin{center}
\includegraphics[width=\textwidth]{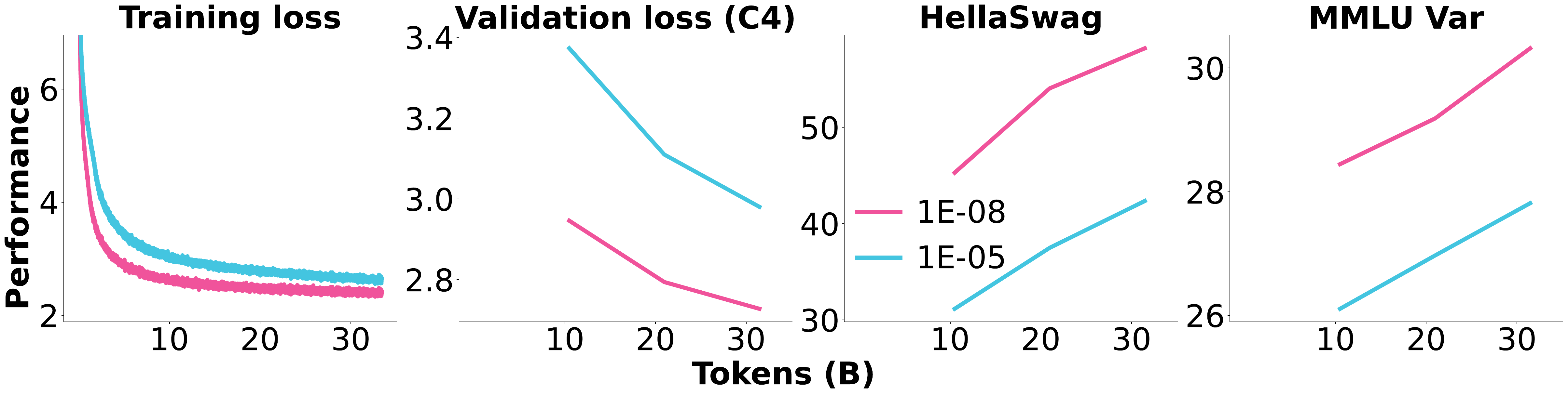}
\caption{\textbf{AdamW epsilon.} More results, logs, and configurations: \url{https://wandb.ai/ai2-llm/olmoe/reports/Plot-AdamW-eps--Vmlldzo4NDc5MDg0}}
\label{fig:adamweps}
\end{center}
\end{figure*}

\citet{groeneveld2024olmo} use an epsilon (``eps'') value of 1E-05 in the AdamW optimizer for training OLMo. A larger eps value leads to smaller steps of the optimizer but can be more stable~\citep{kingma2017adam}.

In \autoref{fig:adamweps}, we find that decreasing eps to the recommended default of 1E-08~\citep{kingma2017adam} significantly improves performance while the run remains stable. Thus, \colorbox{lightOlmoeYellow}{we set eps to 1E-08} for our final run.


\subsection{Adaptation Settings}
\label{sec:adapt}

\begin{wraptable}{r}{7cm}
\begin{tabular}{l|ccc}
\toprule
\multirow{2}{*}{Data ($\downarrow$)}& \multicolumn{2}{c}{\modelsmall{}} \\
& After pretraining & After SFT \\
\midrule
SFT data & 12.22 & 12.16 \\
Github & 13.85 & 14.85 \\
Wikipedia & 14.48 & 14.24 \\
C4 & 9.09 & 9.13 \\
\bottomrule
\end{tabular}
\caption{\textbf{Load balancing loss (\autoref{eq:lbl}) over a subset of the respective corpora prior to scaling with the load balancing loss weight $\alpha$}. While we use load balancing loss during pretraining, we do not use it during SFT.}
\label{tab:adaptlbl}
\end{wraptable}

We experiment with small design choices for adaptation using our evaluation setup described in \autoref{sec:evalsetup}. \textbf{(1) Auxiliary losses:} \citet{zoph2022stmoe} find that using the auxiliary load balancing loss (\autoref{sec:lbloss}) during regular finetuning leads to small performance gains. For instruction tuning, however, \citet{shen2023mixtureofexperts} do not find conclusive evidence in favor of using the load balancing or router z-loss with only small differences in performance, both in support of and against the auxiliary losses. In \autoref{tab:adaptationablation} we display experiments with the load balancing loss during adaptation and find that not using it leads to better performance (54.0 vs. 52.8 after instruction tuning (SFT) and 57.7 vs. 57.1 after preference tuning (DPO)). One potential problem of deactivating the load balancing loss is that it may harm balance among experts and turn some into dead weights as observed during pretraining in \autoref{sec:lbloss}. However, when measuring the load balancing loss in \autoref{tab:adaptlbl} on our SFT data (\autoref{sec:pretraining}), we find that the loss actually decreases slightly during SFT (12.16 vs. 12.22).
This is likely because which experts certain tokens get routed to is determined early during pretraining, as we find later in the analysis section (\autoref{sec:router}). We also visualize the activation patterns of experts of the model after pretraining, and the models after SFT and DPO trained without load balancing in \autoref{sec:addanalysis} (\autoref{fig:routing_tulu}) finding that the distribution remains around the same. Thus, as our models adapted without load balancing perform better and we find it not to impact routing substantially, \colorbox{lightOlmoeYellow}{we do not use load balancing during adaptation}. \textbf{(2) Annealing checkpoint:} We also experiment with using the checkpoint pre-annealing (\autoref{sec:pretraining}) for adaptation and find the checkpoint post-annealing leads to better performance (53.8 vs. 54.0 after SFT and 56.3 vs 57.7 after DPO), thus \colorbox{lightOlmoeYellow}{we use the post-annealing checkpoint.} \textbf{(3) Preference algorithm:} Since the release of DPO (Direct Preference Optimization)~\citep{rafailov2023direct}, a variety of preference algorithms have been proposed~\citep{ethayarajh2024kto,hong2024orpomonolithicpreferenceoptimization,meng2024simposimplepreferenceoptimization}. We experiment with KTO~\citep{ethayarajh2024kto} and find that it matches DPO in \autoref{tab:adaptationablation} for our setup (\autoref{sec:config}). While we release both models, \colorbox{lightOlmoeYellow}{we use DPO} for our final \modelsmalldpo{} model, as it scores higher on AlpacaEval, which has a smaller chance of data contamination than our other benchmarks~\citep{xu2024benchmarkdatacontaminationlarge}.

\begin{table*}[htbp]
\setlength{\tabcolsep}{3.9pt}
\begin{tabular}{l|ccccccc|c}
\toprule
&  &  &  & \textbf{Human-} & \textbf{Alpaca-} & & \\
\textbf{Task ($\rightarrow$)} & \textbf{MMLU} & \textbf{GSM8k} & \textbf{BBH} & \textbf{Eval} & \textbf{Eval 1.0} & \textbf{XSTest} & \textbf{IFEval} & \textbf{Avg} \\
\textbf{Setup ($\rightarrow$)} & \scriptsize{0-shot} & \scriptsize{8-shot CoT} & \scriptsize{0-shot} & \scriptsize{0-shot} & \scriptsize{0-shot} &\scriptsize{0-shot} & \scriptsize{0-shot} & \scriptsize{0-shot} \\
\textbf{Metric ($\rightarrow$)} & \scriptsize{EM} & \scriptsize{EM} & \scriptsize{EM} & \scriptsize{Pass@10} & \scriptsize{\%win} & \scriptsize{F1} & \scriptsize{Loose Acc} & \\
\midrule
\multicolumn{1}{l|}{\begin{tabular}[c]{@{}l@{}}\modelsmall\\ w/o annealing\end{tabular}} & 49.0 & 2.0 & 31.5 & 18.9 & - & 62.1 & 18.5 & - \\
\multicolumn{1}{l|}{+SFT} & 50.2 & 43.0 & 35.6 & 55.5 & 68.9 & 83.8 & 39.7 & 53.8 \\
\multicolumn{1}{l|}{+DPO} & 50.9 & 36.0 & 35.8 & \textbf{58.8} & 81.7 & 83.2 & 47.9 & 56.3 \\
\midrule
\modelsmall & 49.8 & 3.0 & 33.6 & 22.4 & - & 59.7 & 16.6 & - \\
\multicolumn{1}{l|}{\begin{tabular}[c]{@{}l@{}}+SFT\end{tabular}} & 51.4 & 40.5 & 38.0 & 51.6 & 69.2 & 84.1 & 43.3 & 54.0 \\
\rowcolor{lightOlmoeYellow}
\multicolumn{1}{l|}{\begin{tabular}[c]{@{}l@{}}+DPO\end{tabular}} & \textbf{51.9} & \textbf{45.5} & 37.0 & 54.8 & \textbf{84.0} & 82.6 & \textbf{48.1} & \textbf{57.7} \\
\multicolumn{1}{l|}{\begin{tabular}[c]{@{}l@{}}+KTO\end{tabular}} & 51.2 & \textbf{45.5} & 34.1 & 57.1 & 81.6 & \textbf{86.6} & 47.5 & \textbf{57.7} \\
\midrule
\multicolumn{1}{l|}{\begin{tabular}[c]{@{}l@{}}+SFT \\ (load balancing)\end{tabular}} & 50.9 & 36.5 & 35.7 & 52.4 & 66.9 & 84.8 & 42.3 & 52.8 \\
\multicolumn{1}{l|}{\begin{tabular}[c]{@{}l@{}}+DPO \\ (load balancing)\end{tabular}} & 51.1 & 42.5 & \textbf{39.3} & 55.6 & 82.9 & 82.1 & 46.0 & 57.1 \\

\bottomrule
\end{tabular}
\caption{\textbf{Adaptation experiments of \modelsmall{}.} We compare using the pretrained checkpoint prior to annealing for adaptation, using the checkpoint after the additional 100B tokens of annealing, and using the checkpoint after the additional 100B tokens of annealing and with load balancing loss (\autoref{sec:lbloss}) during adaptation. We apply DPO/KTO to the respective SFT model.}
\label{tab:adaptationablation}
\end{table*}


\section{MoE Analysis}

By advancing open and cost-efficient models (\autoref{sec:intro}), \modelsmall{} enables new research into LMs and MoEs. Making use of our released intermediate checkpoints, data, and code, we define and analyze four properties specific to MoEs: \textbf{Router saturation} (\autoref{sec:router}), \textbf{Expert co-activation} (\autoref{sec:expert}), \textbf{Domain specialization} (\autoref{sec:domain}), and \textbf{Vocabulary specialization} (\autoref{sec:token}).

\subsection{Router Saturation}
\label{sec:router}

\begin{figure*}[htbp]
\centering
\includegraphics[width=\textwidth]{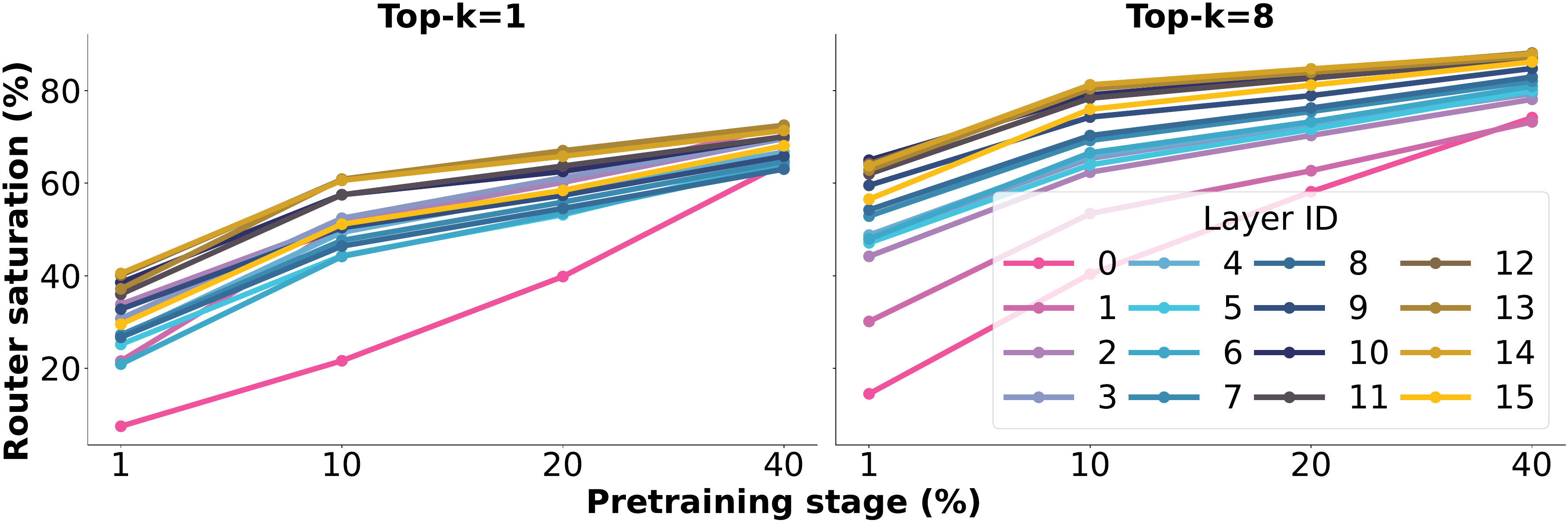}
\caption{\textbf{Router saturation during pretraining measured on a random 0.5\% of the C4 validation data.} We compute saturation by comparing the routing to the top-$k$ experts at four intermediate checkpoints (1, 10, 20, and 40\% of pretraining) to the final pretraining checkpoint (\autoref{eq:routersaturation}).}
\label{fig:saturation}
\end{figure*}

We define router saturation as the proportion of expert activations at some intermediary checkpoint at time $t$ that matches the expert IDs activated at some final checkpoint over the same dataset:
\begin{equation}\label{eq:routersaturation}
\text{Router Saturation}(t) = \frac{1}{N} \sum_{i=1}^{N} \frac{|\mathcal{E}_{i}^{(t)} \cap \mathcal{E}_{i}^{(T)}|}{k},
\end{equation}
where:
\begin{itemize}
\item $N$: The total number of tokens in the dataset.
\item $k$: The number of top-$k$ experts activated per input token. While we train with $k=8$ (\autoref{sec:pretraining}), we also analyze $k=1$ by only looking at the expert with the highest routing probability.
\item $\mathcal{E}_{i}^{(t)}$: The set of $k$ experts activated for the $i$th token at the $t$th checkpoint.
\item $\mathcal{E}_{i}^{(T)}$: The set of $k$ experts activated for the $i$th token at the final checkpoint $T$.
\item $|\mathcal{E}_{i}^{(t)} \cap \mathcal{E}_{i}^{(T)}|$: The number of common experts activated for the $i$th token between the $t$th and final checkpoints.
\end{itemize}
Router saturation thus corresponds to whether the router weights are still learning which expert will process certain data. A value of 100\% indicates that the router at the intermediate checkpoint will route to the same experts as the final checkpoint router. However, even at 100\% saturation the router weight can still change and adapt the exact router probability for each expert. These probabilities are used to scale the output of the respective expert in the model. For \modelsmall{} with its 64 experts, random routing equals a saturation of $1/64=1.6\%$ for $k=1$ and $8/64=12.5\%$ for $k=8$. 

In \autoref{fig:saturation} we find that after 1\% of pretraining (5000 steps or 20B tokens), up to $\sim$60\% of routing to the top-8 activated experts has already saturated (right). Thus the model already uses the same 8 experts for given input data as it will at the end of pretraining. This early saturation aligns with prior work~\citep{xue2024openmoe}. At 40\% of pretraining, saturation reaches up to $\sim$80\%. However, which top-1 expert has the highest routing probability saturates slower (left). We find that routing in later layers saturates earlier during pretraining. Layer 0 is an outlier saturating significantly more slowly than other layers. \citet{dai2024deepseekmoeultimateexpertspecialization} do not use an MoE in the first layer as they find that load balancing converges more slowly for the first layer. This is likely linked to our findings on saturation. Because routing in the first layer saturates slower, the experts that certain input data get routed to frequently change. These changes may lead to one expert suddenly getting significantly more data than others thereby impairing load balancing. We are excited about future work further investigating what happens in the first layer by building on our open release.

\subsection{Expert Co-activation}
\label{sec:expert}

\begin{figure*}[htbp]
\centering
\includegraphics[width=0.31\textwidth,trim={1cm .34cm 4cm .5cm},clip]{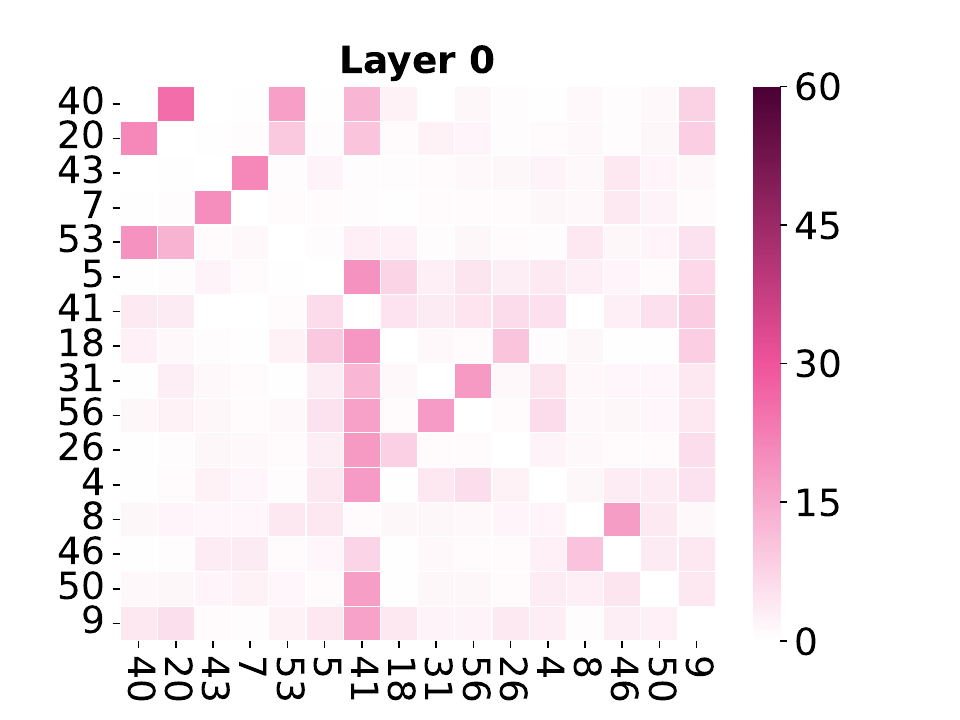}
\includegraphics[width=0.31\textwidth,trim={1cm .34cm 4cm .5cm},clip]{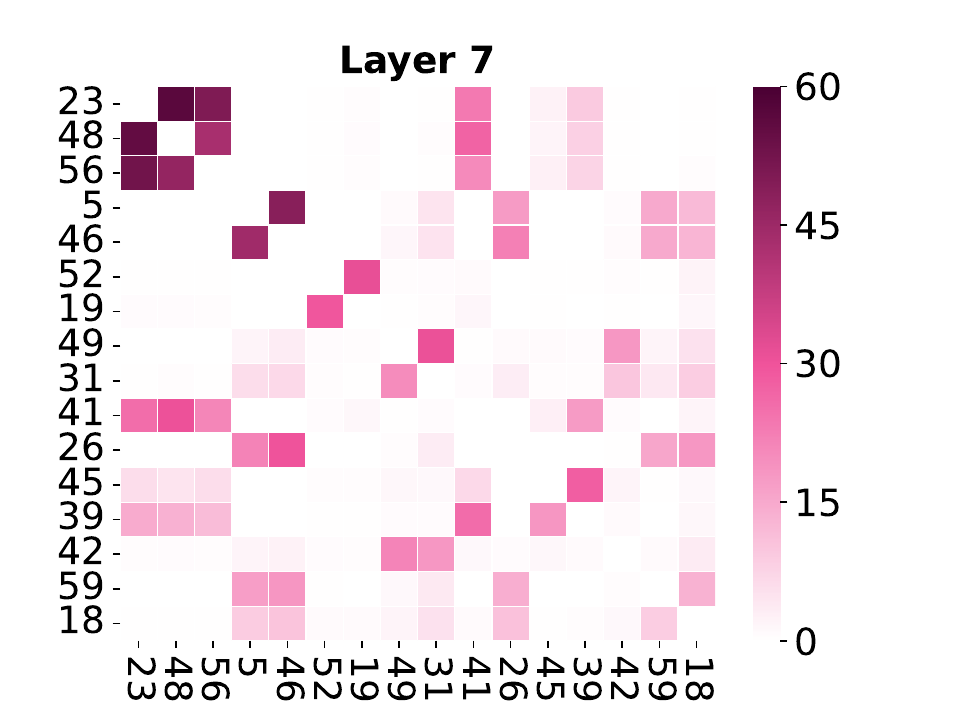}
\includegraphics[width=0.36\textwidth,trim={1cm .34cm 2cm .5cm},clip]{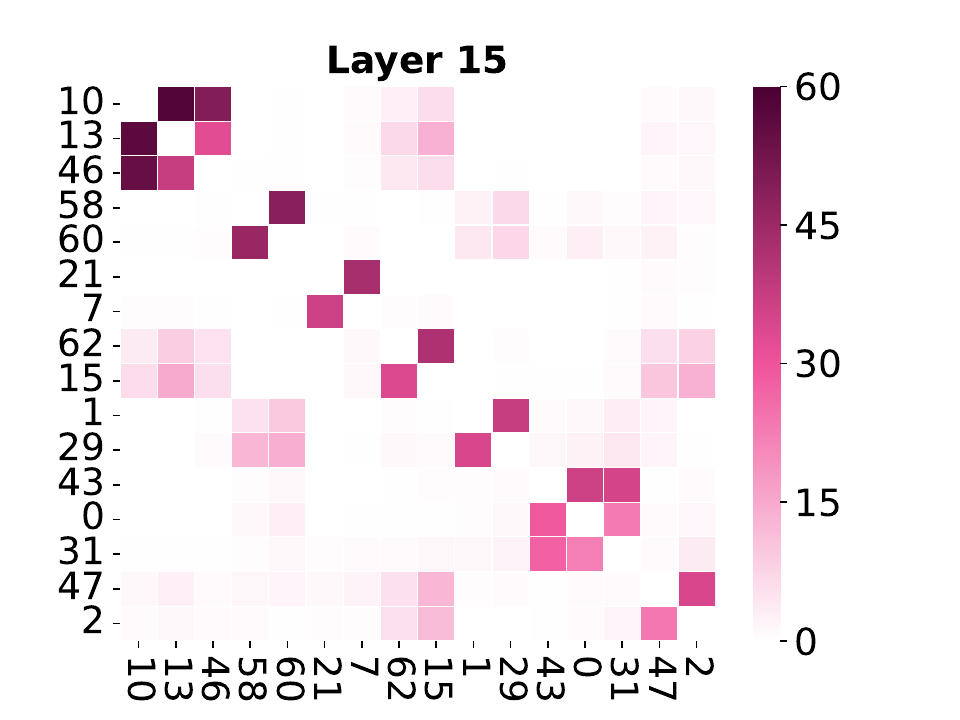}
\caption{\textbf{Co-activation among experts of \modelsmall{} on a random 0.5\% of the C4 validation data.} We display the 32 experts with the highest maximum co-activation score via their expert IDs on the x- and y-axis.}
\label{fig:coactivation}
\end{figure*}

We define expert co-activation as the proportion of times two specific experts, $E_i$ and $E_j$, are simultaneously activated out of the total number of activations of one of those experts:
\begin{equation}
\text{Expert co-activation}(E_i, E_j) = \frac{N_{E_i, E_j}}{N_{E_i}},
\end{equation}
where:
\begin{itemize}
\item $E_i$: The first expert.
\item $E_j$: The second expert.
\item $N_{E_i, E_j}$: The number of times experts $E_i$ and $E_j$ are activated together.
\item $N_{E_i}$: The total number of times expert $E_i$ is activated.
\end{itemize}
A co-activation of 100\% indicates that if $E_i$ is activated, $E_j$ is also always activated. A value of 0\% indicates that the experts never co-occur. If multiple expert pairs have high co-activation, it may suggest that these experts could be merged, benefiting less from keeping them separate. In a distributed setup, we could place highly co-activated experts on the same device to reduce communication costs during model inference. 

In \autoref{fig:coactivation}, we find that there is no strong co-activation among experts in one layer, with only few exceptions. This may indicate that there is little redundancy across different experts. Overall, layers 7 and 15 show similar co-activation patterns with several groups of 3 or 2 experts that tend to get activated together. We investigate tokens that activate these experts in \autoref{sec:token}. Further, in \autoref{sec:addanalysis} (\autoref{fig:cross_layer_sankey}), we investigate whether experts across layers, rather than within one layer, tend to process tokens together.

\subsection{Domain Specialization}
\label{sec:domain}

\begin{figure*}[htbp]
\centering
\includegraphics[width=\textwidth]{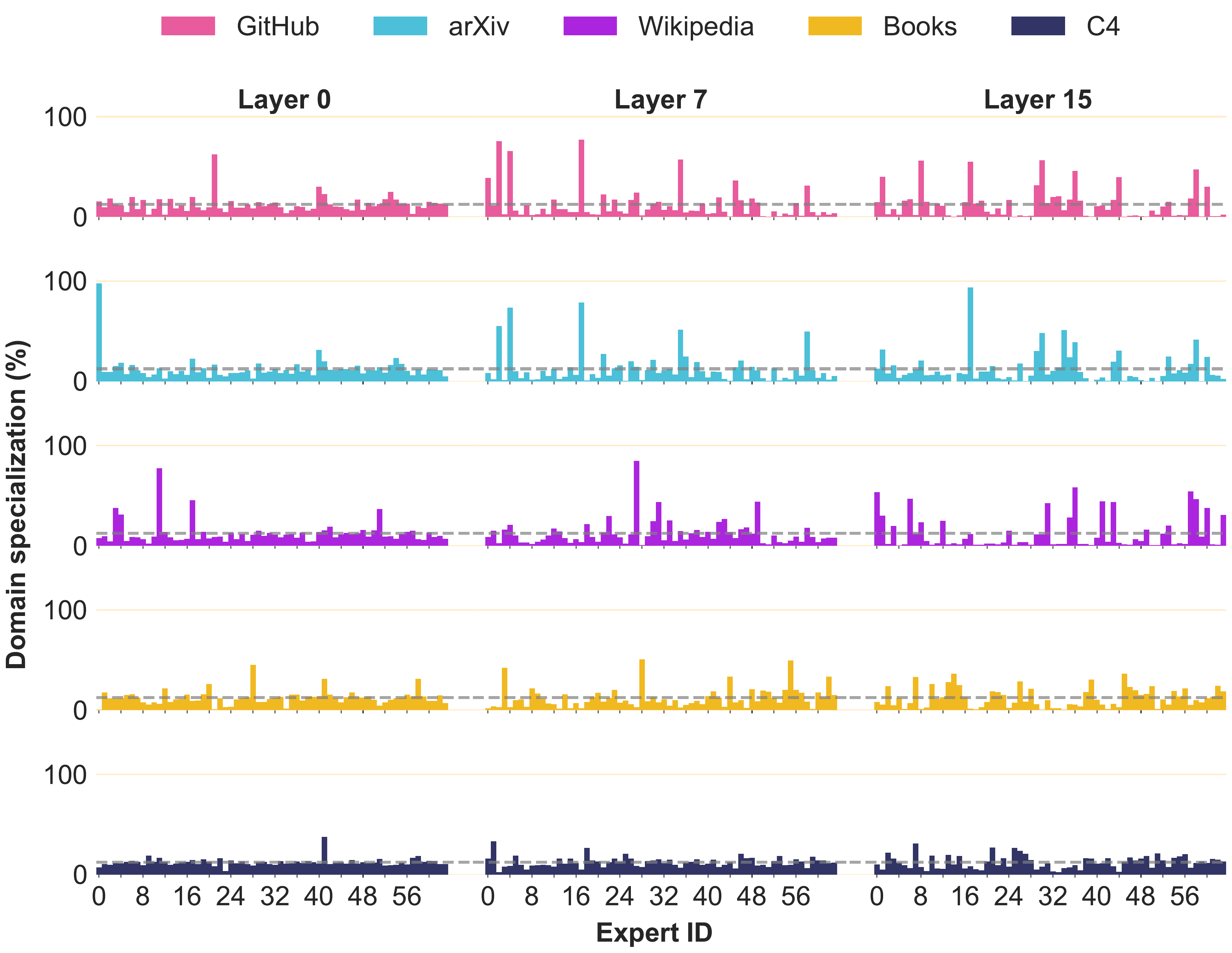}
\vskip 20pt
\includegraphics[width=\textwidth]{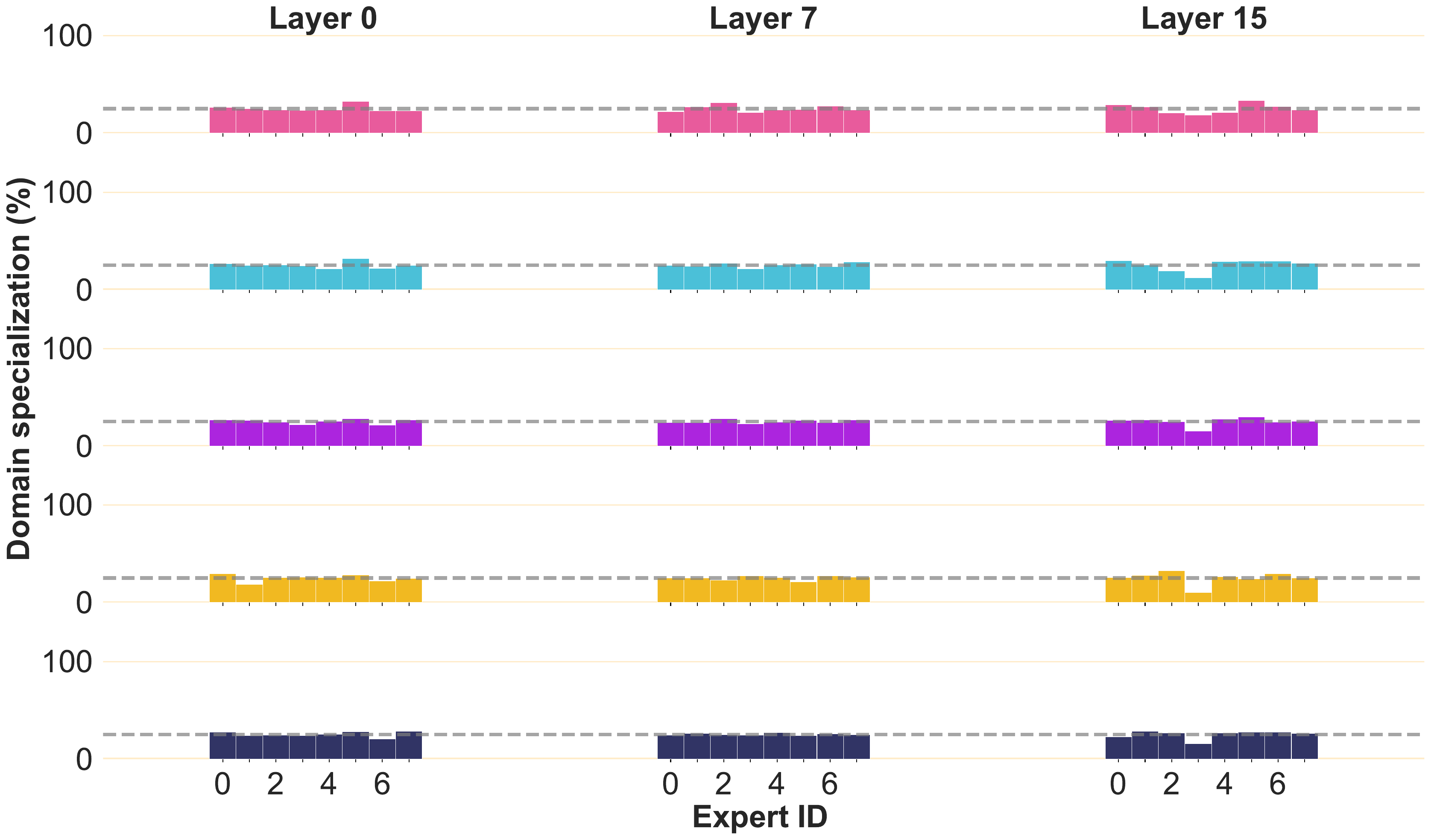}
\caption{\textbf{Domain specialization of \modelsmall{} (top) vs. Mixtral-8x7B (bottom).} We visualize how often tokens from different domains get routed to the 64 (\model{}) or 8 (Mixtral) experts at the end of pretraining. We consider tokens routed to any of the $k=8$ (\model{}) or $k=2$ (Mixtral) active experts (\autoref{eq:domainspec}). Horizontal gray lines correspond to random chance or uniform routing (8/64=12.5\% per expert for \modelsmall{} with 8 active out of 64 total experts per layer and 2/8=25\% for Mixtral with 2 active out of 8 total experts per layer). See \autoref{fig:domainspec_top1} for $k=1$ results.}
\label{fig:domainspec}
\end{figure*}

We define domain specialization as the proportion of tokens from a particular domain $D$ that get routed to a particular expert $E_i$:
\begin{equation}
\label{eq:domainspec}
\text{Domain specialization}(E_i, D) = \frac{N_{E_i, D}^{(k)}}{N_D},
\end{equation}
where:
\begin{itemize}
\item $E_i$: The $i$th expert in the model.
\item $D$: The domain from which the data originates.
\item $k$: The number of experts considered (e.g., $k = 8$ means considering the top 8 experts with the highest routing probabilities).
\item $N_{E_i,D}^{(k)}$: The number of tokens from domain $D$ for which $E_i$ is among the top-$k$ selected experts.
\item $N_D$: The total number of tokens from domain $D$ processed by the MoE.
\end{itemize}

Domain specialization thus refers to the specialization of expert $E_i$ to domain $D$. A value of 100\% indicates that all data from that domain is routed to $E_i$, whereas 0\% indicates the expert is never used for that domain and can be removed from the model without affecting performance in that domain.

In \autoref{fig:domainspec} (top) we find many examples of experts that are activated significantly above or below random chance for \textit{specific domains}. E.g., for arXiv, which has a very specific distribution with lots of scientific text, the first expert in layer 0 is nearly 100\% specialized. This suggests that there is little redundancy in the knowledge of the experts in \modelsmall{}, as they specialize in different kinds of data. GitHub and arXiv are often activated together in layer 7, which we explore further in \autoref{sec:token}. For \textit{generic domains}, such as C4~\citep{raffel2023exploring}, which is a web crawl containing various kinds of data, expert activations in \modelsmall{} are much more balanced. This highlights that the load balancing (\autoref{sec:lbloss}) works as intended and the model makes proper use of all experts for generic data. Mixtral-8x7B~\citep{jiang2024mixtral} in \autoref{fig:domainspec} (bottom), however, exhibits little domain specialization across both \textit{unique} and \textit{generic} \textit{domains}. Experts are activated close to the uniform routing baseline for all layers and domains. Thus, there may be more redundancy across experts in Mixtral, as they likely contain similar knowledge. We hypothesize that this is due to Mixtral being upcycled from Mistral~\citep{mixtralupcycle}. The initialization from a dense model may limit the amount of possible specialization in the experts as they all start from the same local optimum. This is likely why training from scratch eventually outperforms upcycling in our pretraining experiments (\autoref{sec:upcycling}).

\subsection{Vocabulary Specialization}
\label{sec:token}

\begin{figure*}[htbp]
\centering
\includegraphics[width=\textwidth]{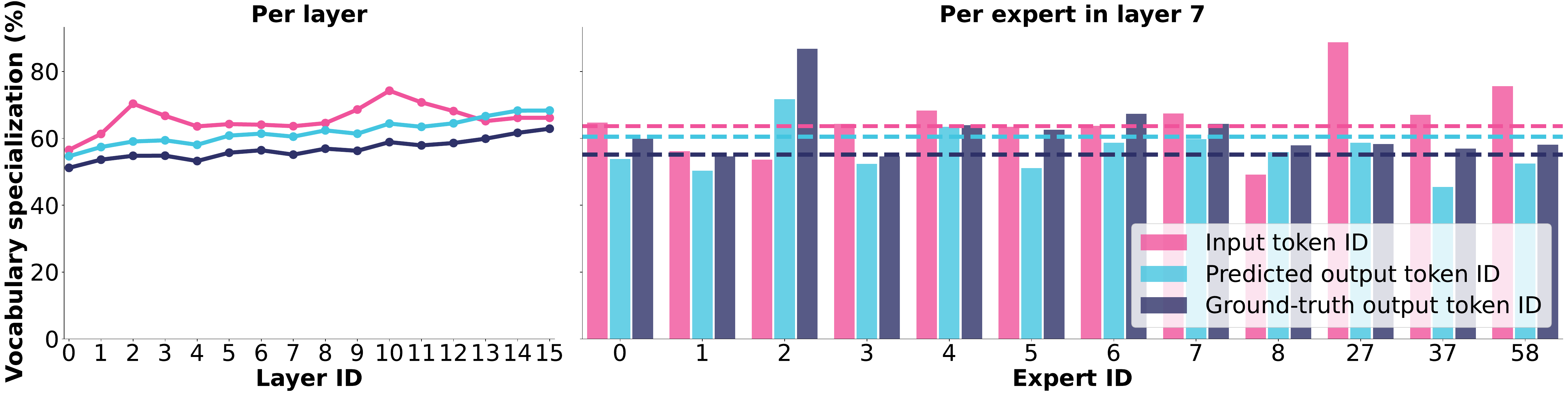}
\caption{\textbf{Vocabulary specialization of \modelsmall{} across layers and experts.} To compute vocabulary specialization per layer (left) we average the specialization of each expert in that layer. Dashed lines (right) correspond to the average of layer 7 as depicted left. We display the first 32 experts out of 64. This plot is for $k=1$ (\autoref{eq:token}) and we provide $k=8$ and a comparison with Mixtral-8x7B in \autoref{sec:addanalysis}.}
\label{fig:tokenspec}
\end{figure*}

\begin{table}[htbp]
\centering 
\footnotesize
\setlength{\tabcolsep}{3pt}
\begin{tabular}{p{1.4cm} p{6cm} p{6cm}}
\toprule
\textbf{Expert ID} & \textbf{Input token IDs} & \textbf{Predicted output token IDs} \\
\midrule
27 & \colorbox{lightOlmoeLightBlue}{\unicode{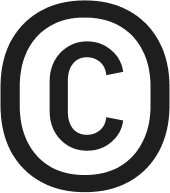}} (100\%) \colorbox{lightOlmoeLightBlue}{\unicode{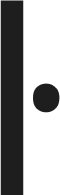}} (100\%) \colorbox{lightOlmoeLightBlue}{$^{3}$} (100\%) \colorbox{lightOlmoeLightBlue}{\unicode{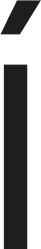}} (100\%) \colorbox{lightOlmoeLightBlue}{\unicode{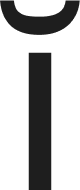}} (100\%) \colorbox{lightOlmoeLightBlue}{\unicode{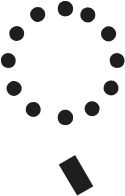}} (100\%) \colorbox{lightOlmoeLightBlue}{\unicode{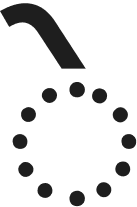}} (100\%) \colorbox{lightOlmoeLightBlue}{\unicode{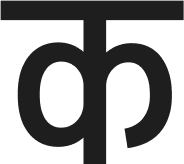}} (100\%) \colorbox{lightOlmoeLightBlue}{\unicode{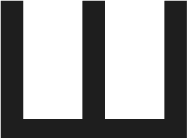}} (100\%) \colorbox{lightOlmoeLightBlue}{\unicode{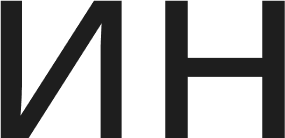}} (100\%) \colorbox{lightOlmoeLightBlue}{ \unicode{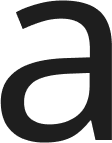}} (100\%) & \colorbox{lightOlmoeLightBlue}{\unicode{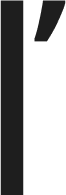}} (100\%) § (100\%) \colorbox{lightOlmoeLightBlue}{\unicode{copyright}} (100\%) \unicode{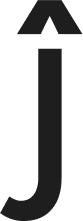} (100\%) \colorbox{lightOlmoeLightBlue}{\unicode{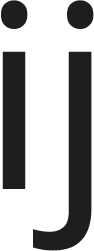}} (100\%) \colorbox{lightOlmoeLightBlue}{\unicode{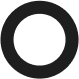}} (100\%) \colorbox{lightOlmoeLightBlue}{\unicode{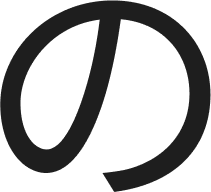}} (100\%) \colorbox{lightOlmoeLightBlue}{\unicode{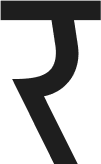}} (100\%) \colorbox{lightOlmoeLightBlue}{\unicode{devanagari_ka}} (100\%) \colorbox{lightOlmoeLightBlue}{\unicode{devanagari_virama}} (100\%) \colorbox{lightOlmoeLightBlue}{\unicode{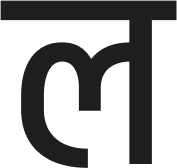}} (100\%) \\
\midrule
58 & \colorbox{lightOlmoeLightBlue}{ (“} (100\%) \colorbox{lightOlmoeLightBlue}{ ("} (100\%) \colorbox{lightOlmoeLightBlue}{ ‘} (94\%) \colorbox{lightOlmoeLightBlue}{ ’} (92\%) \colorbox{lightOlmoeLightBlue}{ “} (92\%) \colorbox{lightOlmoeLightBlue}{ (} (92\%) \colorbox{lightOlmoeLightBlue}{ "} (90\%) \colorbox{lightOlmoeLightBlue}{ '} (89\%) \colorbox{lightOlmoeLightBlue}{“} (88\%) \colorbox{lightOlmoeLightBlue}{ \$} (87\%) \colorbox{lightOlmoeLightBlue}{ [} (87\%) \colorbox{lightOlmoeLightBlue}{ £} (86\%) & \colorbox{lightOlmoeLightBlue}{such} (100\%) \colorbox{lightOlmoeLightBlue}{486} (100\%) \colorbox{lightOlmoeLightBlue}{see} (95\%) \colorbox{lightOlmoeLightBlue}{which} (91\%) \colorbox{lightOlmoeLightBlue}{driving} (91\%) \colorbox{lightOlmoeLightBlue}{UK} (90\%) \colorbox{lightOlmoeLightBlue}{who} (88\%) \colorbox{lightOlmoeLightBlue}{including} (88\%) \colorbox{lightOlmoeLightBlue}{normal} (88\%)
\\
\midrule
7 & \colorbox{lightOlmoeLightBlue}{ Him} (100\%) \colorbox{lightOlmoeLightBlue}{inde} (100\%) \colorbox{lightOlmoeLightBlue}{ Jesus} (98\%) \colorbox{lightOlmoeLightBlue}{God} (90\%) \colorbox{lightOlmoeLightBlue}{ pray} (81\%) \colorbox{lightOlmoeLightBlue}{ Holy} (80\%) \colorbox{lightOlmoeLightBlue}{ Quran} (80\%) \colorbox{lightOlmoeLightBlue}{ God} (77\%) \colorbox{lightOlmoeLightBlue}{ Lord} (76\%) \colorbox{lightOlmoeLightBlue}{ glory} (75\%) \colorbox{lightOlmoeLightBlue}{ Spirit} (66\%) \colorbox{lightOlmoeLightBlue}{ Christ} (65\%) & \colorbox{lightOlmoeLightBlue}{rella} (100\%) \colorbox{lightOlmoeLightBlue}{ Him} (94\%) \colorbox{lightOlmoeLightBlue}{ sin} (90\%) \colorbox{lightOlmoeLightBlue}{ prince} (80\%) \colorbox{lightOlmoeLightBlue}{ glory} (72\%) \colorbox{lightOlmoeLightBlue}{ Jesus} (69\%) \colorbox{lightOlmoeLightBlue}{ Lord} (68\%) \colorbox{lightOlmoeLightBlue}{ Christ} (65\%) \colorbox{lightOlmoeLightBlue}{ Spirit} (55\%) \colorbox{lightOlmoeLightBlue}{ Holy} (53\%) \colorbox{lightOlmoeLightBlue}{ God} (50\%) \colorbox{lightOlmoeLightBlue}{ Prayer} (50\%) \\
\midrule
37 & \colorbox{lightOlmoeLightBlue}{ Sunday} (100\%) \colorbox{lightOlmoeLightBlue}{ Tuesday} (100\%) \colorbox{lightOlmoeLightBlue}{ Thursday} (100\%) \colorbox{lightOlmoeLightBlue}{ Olympic} (100\%) \colorbox{lightOlmoeLightBlue}{ Christmas} (100\%) \colorbox{lightOlmoeLightBlue}{ rugby} (100\%) \colorbox{lightOlmoeLightBlue}{ Championship} (100\%) \colorbox{lightOlmoeLightBlue}{ weekends} (100\%) & \colorbox{lightOlmoeLightBlue}{days} (91\%) \colorbox{lightOlmoeLightBlue}{ anniversary} (90\%) \colorbox{lightOlmoeLightBlue}{ month} (88\%) \colorbox{lightOlmoeLightBlue}{ week} (84\%) \colorbox{lightOlmoeLightBlue}{mpi} (83\%) \colorbox{lightOlmoeLightBlue}{ semester} (81\%) \colorbox{lightOlmoeLightBlue}{mand} (80\%) \colorbox{lightOlmoeLightBlue}{ Olympics} (78\%) \colorbox{lightOlmoeLightBlue}{ cent} (76\%) \colorbox{lightOlmoeLightBlue}{ season} (76\%) \colorbox{lightOlmoeLightBlue}{ perm} (75\%) \\
\midrule
43 & \colorbox{lightOlmoeLightBlue}{ Armenian} (100\%) \colorbox{lightOlmoeLightBlue}{ijan} (100\%) \colorbox{lightOlmoeLightBlue}{enia} (96\%) \colorbox{lightOlmoeLightBlue}{ Iraq} (95\%) \colorbox{lightOlmoeLightBlue}{ Iranian} (92\%) \colorbox{lightOlmoeLightBlue}{ Iran} (92\%) \colorbox{lightOlmoeLightBlue}{ Saudi} (90\%) \colorbox{lightOlmoeLightBlue}{ northern} (90\%) \colorbox{lightOlmoeLightBlue}{ Lebanon} (90\%) \colorbox{lightOlmoeLightBlue}{ Singapore} (88\%) \colorbox{lightOlmoeLightBlue}{ Turkey} (88\%) \colorbox{lightOlmoeLightBlue}{ Asia} (87\%) \colorbox{lightOlmoeLightBlue}{ Egypt} (86\%) \colorbox{lightOlmoeLightBlue}{ western} (86\%) &
\colorbox{lightOlmoeLightBlue}{enia} (90\%) \colorbox{lightOlmoeLightBlue}{ invasion} (80\%) \colorbox{lightOlmoeLightBlue}{ Arabia} (76\%) \colorbox{lightOlmoeLightBlue}{ irregular} (66\%) \colorbox{lightOlmoeLightBlue}{ regions} (64\%) \colorbox{lightOlmoeLightBlue}{ border} (63\%) \colorbox{lightOlmoeLightBlue}{ Kong} (61\%) \colorbox{lightOlmoeLightBlue}{ians} (61\%) \colorbox{lightOlmoeLightBlue}{ bases} (60\%) \colorbox{lightOlmoeLightBlue}{ Republic} (59\%) \colorbox{lightOlmoeLightBlue}{ Ireland} (58\%) \colorbox{lightOlmoeLightBlue}{ Korea} (58\%) \colorbox{lightOlmoeLightBlue}{ War} (55\%) \colorbox{lightOlmoeLightBlue}{ Carolina} (52\%) \\
\midrule
4 & \colorbox{lightOlmoeLightBlue}{ sq} (89\%) \colorbox{lightOlmoeLightBlue}{Main} (70\%) \colorbox{lightOlmoeLightBlue}{ reversal} (69\%) \colorbox{lightOlmoeLightBlue}{YR} (63\%) \colorbox{lightOlmoeLightBlue}{GC} (56\%) \colorbox{lightOlmoeLightBlue}{Overall} (50\%) \colorbox{lightOlmoeLightBlue}{79} (50\%) \colorbox{lightOlmoeLightBlue}{main} (50\%) \colorbox{lightOlmoeLightBlue}{ RE} (46\%) \colorbox{lightOlmoeLightBlue}{ PCR} (46\%) \colorbox{lightOlmoeLightBlue}{ tomb} (45\%) \colorbox{lightOlmoeLightBlue}{normal} (43\%) \colorbox{lightOlmoeLightBlue}{ intensity} (41\%) \colorbox{lightOlmoeLightBlue}{ Overall} (41\%) \colorbox{lightOlmoeLightBlue}{ median} (41\%) & \colorbox{lightOlmoeLightBlue}{YR} (90\%) \colorbox{lightOlmoeLightBlue}{ Character} (88\%) \colorbox{lightOlmoeLightBlue}{ sq} (77\%) \colorbox{lightOlmoeLightBlue}{Os} (76\%) \colorbox{lightOlmoeLightBlue}{ GHz} (71\%) \colorbox{lightOlmoeLightBlue}{fluence} (60\%) \colorbox{lightOlmoeLightBlue}{amycin} (60\%) \colorbox{lightOlmoeLightBlue}{ pixels} (56\%) \colorbox{lightOlmoeLightBlue}{=} (53\%) \colorbox{lightOlmoeLightBlue}{arc} (52\%) \colorbox{lightOlmoeLightBlue}{ Story} (52\%) \colorbox{lightOlmoeLightBlue}{ =} (51\%) \colorbox{lightOlmoeLightBlue}{anth} (50\%) \colorbox{lightOlmoeLightBlue}{GHz} (50\%) \colorbox{lightOlmoeLightBlue}{ cm} (46\%) \\
\midrule
0 & \colorbox{lightOlmoeLightBlue}{ESM} (100\%) \colorbox{lightOlmoeLightBlue}{icillin} (100\%) \colorbox{lightOlmoeLightBlue}{agra} (98\%) \colorbox{lightOlmoeLightBlue}{ aust} (96\%) \colorbox{lightOlmoeLightBlue}{asa} (93\%) \colorbox{lightOlmoeLightBlue}{ pills} (92\%) \colorbox{lightOlmoeLightBlue}{ mg} (85\%) \colorbox{lightOlmoeLightBlue}{ uk} (82\%) \colorbox{lightOlmoeLightBlue}{ login} (82\%) \colorbox{lightOlmoeLightBlue}{doc} (81\%) \colorbox{lightOlmoeLightBlue}{ generic} (81\%) \colorbox{lightOlmoeLightBlue}{cd} (81\%) \colorbox{lightOlmoeLightBlue}{ Essay} (81\%) \colorbox{lightOlmoeLightBlue}{ password} (81\%) \colorbox{lightOlmoeLightBlue}{ Content} (80\%) &  \colorbox{lightOlmoeLightBlue}{*,} (100\%) \colorbox{lightOlmoeLightBlue}{ sil} (96\%) \colorbox{lightOlmoeLightBlue}{ pills} (91\%) \colorbox{lightOlmoeLightBlue}{ vi} (90\%) \colorbox{lightOlmoeLightBlue}{ xen} (87\%) \colorbox{lightOlmoeLightBlue}{ pharmacy} (87\%) \colorbox{lightOlmoeLightBlue}{ gener} (85\%) \colorbox{lightOlmoeLightBlue}{ aust} (82\%) \colorbox{lightOlmoeLightBlue}{ mg} (75\%) \colorbox{lightOlmoeLightBlue}{ Content} (75\%) \colorbox{lightOlmoeLightBlue}{ uk} (73\%) \colorbox{lightOlmoeLightBlue}{ THAT} (73\%) \colorbox{lightOlmoeLightBlue}{ dispens} (68\%) \colorbox{lightOlmoeLightBlue}{icillin} (68\%) \colorbox{lightOlmoeLightBlue}{ generic} (66\%) \\
\midrule
3 & \colorbox{lightOlmoeLightBlue}{ grandmother} (92\%) \colorbox{lightOlmoeLightBlue}{ brother} (91\%) \colorbox{lightOlmoeLightBlue}{ Daisy} (83\%) \colorbox{lightOlmoeLightBlue}{ daughter} (78\%) \colorbox{lightOlmoeLightBlue}{ mum} (75\%) \colorbox{lightOlmoeLightBlue}{ father} (72\%) \colorbox{lightOlmoeLightBlue}{ wife} (70\%) \colorbox{lightOlmoeLightBlue}{ husband} (70\%) \colorbox{lightOlmoeLightBlue}{ lady} (63\%) \colorbox{lightOlmoeLightBlue}{ dad} (62\%) \colorbox{lightOlmoeLightBlue}{ boy} (61\%) &  \colorbox{lightOlmoeLightBlue}{hood} (36\%) \colorbox{lightOlmoeLightBlue}{mother} (35\%) \colorbox{lightOlmoeLightBlue}{inde} (31\%) \colorbox{lightOlmoeLightBlue}{ boy} (29\%) \colorbox{lightOlmoeLightBlue}{ girl} (28\%) \colorbox{lightOlmoeLightBlue}{ married} (27\%) \colorbox{lightOlmoeLightBlue}{ tri} (21\%) \colorbox{lightOlmoeLightBlue}{ Gab} (20\%) \colorbox{lightOlmoeLightBlue}{ died} (18\%) \colorbox{lightOlmoeLightBlue}{ taught} (14\%) \colorbox{lightOlmoeLightBlue}{ lived} (13\%) \colorbox{lightOlmoeLightBlue}{ knew} (10\%) \\
\midrule
48 & \colorbox{lightOlmoeLightBlue}{ compared} (42\%) \colorbox{lightOlmoeLightBlue}{!)} (41\%) \colorbox{lightOlmoeLightBlue}{Then} (41\%) \colorbox{lightOlmoeLightBlue}{’,} (40\%) \colorbox{lightOlmoeLightBlue}{),} (35\%) \colorbox{lightOlmoeLightBlue}{",} (35\%) \colorbox{lightOlmoeLightBlue}{ instead} (33\%) &
\colorbox{lightOlmoeLightBlue}{ except} (60\%) \colorbox{lightOlmoeLightBlue}{ tennis} (41\%) \colorbox{lightOlmoeLightBlue}{ Marks} (40\%) \colorbox{lightOlmoeLightBlue}{ Dunn} (33\%) \colorbox{lightOlmoeLightBlue}{ tears} (30\%) \colorbox{lightOlmoeLightBlue}{ Arizona} (30\%) \\
\midrule
23 & \colorbox{lightOlmoeLightBlue}{....} (58\%) \colorbox{lightOlmoeLightBlue}{ Therefore} (55\%) \colorbox{lightOlmoeLightBlue}{So} (46\%) \colorbox{lightOlmoeLightBlue}{!!!} (46\%) \colorbox{lightOlmoeLightBlue}{And} (44\%) \colorbox{lightOlmoeLightBlue}{ According} (41\%) \colorbox{lightOlmoeLightBlue}{."} (41\%) \colorbox{lightOlmoeLightBlue}{!!} (40\%) \colorbox{lightOlmoeLightBlue}{?"} (38\%) \colorbox{lightOlmoeLightBlue}{But} (38\%) & 
\colorbox{lightOlmoeLightBlue}{ \unicode{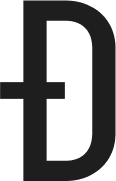}} (53\%) \colorbox{lightOlmoeLightBlue}{ Republican} (50\%) \colorbox{lightOlmoeLightBlue}{Jack} (47\%) \colorbox{lightOlmoeLightBlue}{ THIS} (40\%) \colorbox{lightOlmoeLightBlue}{ Democratic} (40\%) \colorbox{lightOlmoeLightBlue}{ according} (39\%) \colorbox{lightOlmoeLightBlue}{ So} (38\%) \colorbox{lightOlmoeLightBlue}{Step} (33\%) \\
\bottomrule
\end{tabular}
\vspace{.25em}
\caption{\textbf{Vocabulary specialization in the 7th layer of \modelsmall{}.} We use $k=1$ (\autoref{eq:token}) and a random 0.5\% of the C4 validation data excluding token IDs with $<$10 appearances.}
\label{tab:tokens}
\end{table}

We define vocabulary specialization as the proportion of tokens with a token ID $x$ (also called vocabulary element) that are routed to one particular expert $E_i$ out of all experts in that layer:
\begin{equation}
\label{eq:token}
\text{Vocabulary specialization}(E_i, x) = \frac{N_{x, E_i}^{(k)}}{N_x},
\end{equation}
where:
\begin{itemize}
\item $E_i$: The $i$th expert in the model.
\item $x$: The token ID being analyzed.
\item $k$: The number of experts considered (e.g., $k=8$ means considering the top 8 experts with the highest routing probabilities).    
\item $N_{x, E_i}$: The number of times input data is routed to $E_i$ for $x$.
\item $N_x$: The total number of times input data is routed across all experts for  $x$.
\end{itemize}
Vocabulary specialization thus refers to how specialized a particular expert is on some vocabulary item. We distinguish input and output variants of this specialization, where $x$ is either the input token ID or the next output token ID (either the ground-truth next token ID or the token ID predicted by the model). A value of 100\% indicates that for all occurrences of that vocabulary element, input data is routed to $E_i$, whereas 0\% indicates an expert that is fully irrelevant for that vocabulary element and can be effectively removed from the model without affecting performance whenever the token ID appears.

In \autoref{fig:tokenspec} we find that vocabulary specialization is higher in later layers, similar to how later layers saturate earlier (\autoref{sec:router}). Later layers also specialize more on predicted output token IDs rather than input token IDs, i.e., the routing is decided more by the token the model is about to predict rather than the original input token. This is intuitive as in earlier layers there is more uncertainty about which token the model will predict. At $\sim$90\%, expert 27 specializes the most, which we find in \autoref{tab:tokens} to activate for many non-alphabetic tokens, such as Cyrillic and Devanagari letters. Expert 43 shows specialization on geographic terms in both input and output tokens. Experts 48 and 23 both focus on connector words, such as \colorbox{lightOlmoeLightBlue}{ Then} and \colorbox{lightOlmoeLightBlue}{ Therefore}. This is likely because they commonly process tokens together with a high co-activation of 60\% in \autoref{fig:coactivation} (middle). Based on our findings in \autoref{sec:domain} that for GitHub and arXiv often the same experts in layer 7 activate, we display one such expert (expert ID 4) in \autoref{tab:tokens}. It seems to specialize in measurements, such as \colorbox{lightOlmoeLightBlue}{ sq}, \colorbox{lightOlmoeLightBlue}{YR} (year), and \colorbox{lightOlmoeLightBlue}{ GHz}. These are common terms in scientific papers corresponding to the arXiv domain and likely also in GitHub code for computations related to measurements. They are less likely to appear in books, which explains the low activation of expert ID 4 in layer 7 for book data in \autoref{fig:domainspec}. Expert 3 is among the three most active experts of layer 7 for book data in \autoref{fig:domainspec} (fourth yellow bar for layer 7). This resonates when looking at its specialization on family terms in \autoref{tab:tokens}, which are far more common in books than scientific papers or code. Overall, domain specialization and vocabulary specialization are closely linked to one another, as domains are usually characterized by their distinct word distribution. In \autoref{sec:addanalysis} (\autoref{fig:tokendomainspec}), we link them more closely by comparing the extent of vocabulary specialization across domains and expert IDs. In \autoref{sec:addanalysis} (\autoref{fig:tokenspecolmoe8}, \autoref{fig:tokenspecmixtral2}) we also find that \modelsmall{} exhibits stronger vocabulary specialization than Mixtral-8x7B.

\section{Related Work}

\paragraph{Advances in MoEs} Current LMs still largely follow the transformer architecture~\citep{vaswani2023attention} with only few architectural changes that have been widely adopted, such as decoder-only training~\citep{radford2019language}, SwiGLU activations~\citep{shazeer2020glu,dauphin2017languagemodelinggatedconvolutional}, RoPE~\citep{su2023roformerenhancedtransformerrotary}, MQA/GQA~\citep{shazeer2019fasttransformerdecodingwritehead,ainslie2023gqatraininggeneralizedmultiquery} and RMSNorm~\citep{zhang2019rootmeansquarelayer}. Model sparsity via Mixture-of-Experts is one modification still under active exploration with some early adoption but most LMs, including Llama 3~\citep{dubey2024llama3herdmodels}, still rely on a dense architecture. There has been a lot of progress in improving the sparsely-gated MoE layer since its introduction~\citep{shazeer2017outrageously}: New routing techniques~\citep{lewis2021baselayerssimplifyingtraining,roller2021hashlayerslargesparse,zuo2022tamingsparselyactivatedtransformer,gross2017hardmixturesexpertslarge,jaszczur2021sparsescalingtransformers,dua2021trickstrainingsparsetranslation,zhong2024lory,wu2024yuan20m32mixtureexperts,muqeeth2024softmergingexpertsadaptive}, fine-grained expert segmentation~\citep{dai2024deepseekmoeultimateexpertspecialization,he2024mixturemillionexperts}, stability~\citep{zoph2022stmoe} and efficiency~\citep{lepikhin2020gshardscalinggiantmodels,rajbhandari2022deepspeedmoeadvancingmixtureofexpertsinference,du2022glam,zhou2024brainformerstradingsimplicityefficiency,li2022branchtrainmergeembarrassinglyparalleltraining,sukhbaatar2024branchtrainmixmixingexpertllms,pan2024densetrainingsparseinference,ren2023pangusigmatrillionparameterlanguage} improvements. In this work, we perform many experiments to provide insights into training Mixture-of-Experts LMs. Subsequently, we train \modelsmall{} for 5T tokens. No prior MoE has been overtrained~\citep{gadre2024language} to this extent to our knowledge making \modelsmall{} the best testbed to research performance saturation of MoEs vs. dense models. With \model{} we hope to facilitate such and other research to help the field uncover whether MoEs should make it into all future LMs and with what precise configuration.

\paragraph{Open LMs} A variety of model families have been proposed under varying degrees of openness commonly categorized based on whether model weights are available. \textbf{Closed-weight} models include GPT~\citep{brown2020language,openai2023gpt4}, Gemini~\citep{geminiteam2023gemini,geminiteam2024gemini}, PaLM~\citep{chowdhery2022palm,anil2023palm2technicalreport}, Reka~\citep{rekateam2024rekacoreflashedge}, and \textbf{open-weight} ones include Llama~\citep{touvron2023llama,touvron2023llama2openfoundation,dubey2024llama3herdmodels}, Mistral~\citep{jiang2023mistral,jiang2024mixtral}, Gemma~\citep{gemmateam2024gemmaopenmodelsbased,gemmateam2024gemma2improvingopen}, Falcon~\citep{almazrouei2023falcon,penedo2023refinedwebdatasetfalconllm}, MPT~\citep{MosaicML2023Introducing}, Qwen~\citep{bai2023qwen,yang2024qwen2technicalreport}, GLM~\citep{glm2024chatglm}, Yi~\citep{ai2024yiopenfoundationmodels}, DeepSeek~\citep{deepseekai2024deepseek,deepseekai2024deepseekv2,dai2024deepseekmoeultimateexpertspecialization}, Nemotron~\citep{parmar2024nemotron415btechnicalreport,nvidia2024nemotron4340btechnicalreport}, Zamba~\citep{glorioso2024zambacompact7bssm}, InternLM~\citep{cai2024internlm2}, Baichuan~\citep{yang2023baichuan2openlargescale}, Phi~\citep{gunasekar2023textbooksneed,li2023textbooksneediiphi15,abdin2024phi3technicalreporthighly}, StableLM~\citep{bellagente2024stablelm216b}, OPT~\citep{zhang2022optopenpretrainedtransformer}. However, besides model weights, training data and code are key to enabling scientific research of these models~\citep{longpre2023dataprovenanceinitiativelarge,longpre2024consentcrisisrapiddecline} and distributing their benefits broadly~\citep{bommasani2023foundationmodeltransparencyindex}. There have been few releases also including data and code in addition to model weights which we refer to as \textbf{``fully open-source''}: BLOOM~\citep{workshop2023bloom,scao2022language,muennighoff2023crosslingual,yong2023bloom1addinglanguagesupport}, GPT-NeoX~\citep{black2022gptneox20b,gpt-neo,wang2021gpt}, StarCoder~\citep{li2023starcoder,lozhkov2024starcoder2stackv2,allal2023santacoder,muennighoff2023octopack,zhuo2024astraios}, Pythia~\citep{biderman2023pythiasuiteanalyzinglarge}, OLMo~\citep{groeneveld2024olmo}, LLM360~\citep{liu2023llm360}, Cerebras-GPT~\citep{dey2023cerebrasgptopencomputeoptimallanguage}, DCLM~\citep{li2024datacomplm}, MAP-Neo~\citep{zhang2024mapneohighlycapabletransparent}, RWKV~\citep{peng2023rwkv,peng2024eagle}, and SmolLM~\citep{allal2024SmolLM}. For Mixture-of-Experts only OpenMoE~\citep{xue2024openmoe} aims to be fully open-source, however, its poor performance limits its usefulness. We release \modelsmall{} as the first state-of-the-art Mixture-of-Experts LM that is fully open-source: model weights, data, code, and logs.

\section{Conclusion}

We open-source \modelsmall{} and \modelsmalldpo{} including model, data, code, and logs. At 1B active and 7B total parameters, our models yield state-of-the-art performance among models with a similar amount of active parameters even outperforming larger models including DeepSeekMoE-16B and Llama2-13B-Chat. We share various training experiments and define and analyze router saturation, expert co-activation, domain and vocabulary specialization of our model. Through our fully open release, we seek to help the field build better MoEs. We are excited about more iterations of \model{} to close the gap between frontier models and fully open models.

\section*{Author Contributions}
\label{sec:contribute}

\textbf{Niklas Muennighoff} proposed and led the project. He ran the pretraining experiments, pretrained the model, helped run adaptation and analysis, and wrote most of the paper.\\
\textbf{Luca Soldaini} created the pretraining dataset and advised on pretraining.\\
\textbf{Dirk Groeneveld} advised on pretraining, especially stability and throughput improvements.\\
\textbf{Kyle Lo} helped with pretraining dataset creation, analyzed data experiments, and advised on data and framing, and helped edit the paper.\\
\textbf{Jacob Morrison} co-created the adaptation dataset, ran most adaptation experiments, and helped edit the paper.\\
\textbf{Sewon Min} analyzed router saturation, expert correlation, and vocabulary specialization, and helped frame and edit the paper.\\
\textbf{Weijia Shi} analyzed domain and vocabulary specialization, advised at various project stages, and helped edit the paper.\\
\textbf{Pete Walsh} advised on pretraining, especially stability and throughput improvements.\\
\textbf{Oyvind Tafjord} ran OLMES evaluations.\\
\textbf{Nathan Lambert} co-created the adaptation dataset, advised on adaptation, and helped edit the paper.\\
\textbf{Yuling Gu} ran OLMES evaluations and helped edit the paper.\\
\textbf{Shane Arora} uploaded the models, helped with code review and framework integration.\\
\textbf{Akshita Bhagia} supported stability investigations and helped with DCLM evaluations.\\
\textbf{Dustin Schwenk} supported stability investigations.\\
\textbf{David Wadden} ran DCLM evaluations and helped with Weights \& Biases reports.\\
\textbf{Alexander Wettig} advised on pretraining, analyzed load balancing, routing, and domain specialization, and helped edit the paper.\\
\textbf{Binyuan Hui} advised on pretraining and helped with plotting and framework integration.\\
\textbf{Tim Dettmers} advised on analysis and inference experiments.\\
\textbf{Douwe Kiela} advised on framing.\\
\textbf{Ali Farhadi} advised on pretraining and framing.\\
\textbf{Noah A. Smith} advised on pretraining, and helped frame and edit the paper.\\
\textbf{Pang Wei Koh} advised on analysis, and helped frame and edit the paper.\\
\textbf{Amanpreet Singh} advised on pretraining, framing and helped edit the paper.\\
\textbf{Hannaneh Hajishirzi} was responsible for direction and advising of the overall effort and helped frame and edit the paper.

\section*{Acknowledgements}
\label{sec:ack}

\model{} would not be possible without the support of many individuals and institutions. We thank our teammates at the Allen Institute for AI, Contextual AI, and the University of Washington for their support, especially Aditya Kusupati, Ananya Harsh Jha, Caitlin Wittlif, Carissa Schoenick, Costa Huang, Crystal Nam, David Atkinson, Emma Strubell, Faeze Brahman, Hamish Ivison, Karel D'Oosterlinck, Matt Latzke, Ian Magnusson, Jack Merullo, Jay Chen, Jennifer Dumas, Jiacheng Liu, Johann Dahm, Luke Zettlemoyer, Michael Schmitz, Michael Wilson, Pradeep Dasigi, Sahil Verma, Sam Skjonsberg, Sophie Lebrecht, Stas Bekman, Taira Anderson, Valentina Pyatkin, Yanai Elazar, Yizhong Wang, and Yoganand Chandrasekhar. We also thank Armen Aghajanyan, Akshat Shrivastava, Colin Raffel, Haokun Liu, Ludwig Schmidt, Mengzhou Xia, Shayne Longpre, Sheng Shen, and Zexuan Zhong. PWK is supported by the Singapore National Research Foundation and the National AI Group in the Singapore Ministry of Digital Development and Innovation under the AI Visiting Professorship Programme (award number AIVP-2024-001).

\newpage

\bibliography{references}

\begin{thebibliography}{223}
\expandafter\ifx\csname natexlab\endcsname\relax\def\natexlab#1{#1}\fi

\bibitem[{Abdin et~al.(2024)Abdin, Jacobs, Awan, Aneja, Awadallah, Awadalla, Bach, Bahree, Bakhtiari, Bao, Behl, Benhaim, Bilenko, Bjorck, Bubeck, Cai, Cai, Mendes, Chen, Chaudhary, Chen, Chen, Chen, Chen, Chopra, Dai, Giorno, de~Rosa, Dixon, Eldan, Fragoso, Iter, Gao, Gao, Gao, Garg, Goswami, Gunasekar, Haider, Hao, Hewett, Huynh, Javaheripi, Jin, Kauffmann, Karampatziakis, Kim, Khademi, Kurilenko, Lee, Lee, Li, Li, Liang, Liden, Liu, Liu, Liu, Lin, Lin, Luo, Madan, Mazzola, Mitra, Modi, Nguyen, Norick, Patra, Perez-Becker, Portet, Pryzant, Qin, Radmilac, Rosset, Roy, Ruwase, Saarikivi, Saied, Salim, Santacroce, Shah, Shang, Sharma, Shukla, Song, Tanaka, Tupini, Wang, Wang, Wang, Wang, Ward, Wang, Witte, Wu, Wyatt, Xiao, Xu, Xu, Xu, Yadav, Yang, Yang, Yang, Yang, Yu, Yuan, Zhang, Zhang, Zhang, Zhang, Zhang, Zhang, Zhang, and Zhou}]{abdin2024phi3technicalreporthighly}
Marah Abdin, Sam~Ade Jacobs, Ammar~Ahmad Awan, Jyoti Aneja, Ahmed Awadallah, Hany Awadalla, Nguyen Bach, Amit Bahree, Arash Bakhtiari, Jianmin Bao, Harkirat Behl, Alon Benhaim, Misha Bilenko, Johan Bjorck, Sébastien Bubeck, Qin Cai, Martin Cai, Caio César~Teodoro Mendes, Weizhu Chen, Vishrav Chaudhary, Dong Chen, Dongdong Chen, Yen-Chun Chen, Yi-Ling Chen, Parul Chopra, Xiyang Dai, Allie~Del Giorno, Gustavo de~Rosa, Matthew Dixon, Ronen Eldan, Victor Fragoso, Dan Iter, Mei Gao, Min Gao, Jianfeng Gao, Amit Garg, Abhishek Goswami, Suriya Gunasekar, Emman Haider, Junheng Hao, Russell~J. Hewett, Jamie Huynh, Mojan Javaheripi, Xin Jin, Piero Kauffmann, Nikos Karampatziakis, Dongwoo Kim, Mahoud Khademi, Lev Kurilenko, James~R. Lee, Yin~Tat Lee, Yuanzhi Li, Yunsheng Li, Chen Liang, Lars Liden, Ce~Liu, Mengchen Liu, Weishung Liu, Eric Lin, Zeqi Lin, Chong Luo, Piyush Madan, Matt Mazzola, Arindam Mitra, Hardik Modi, Anh Nguyen, Brandon Norick, Barun Patra, Daniel Perez-Becker, Thomas Portet, Reid Pryzant, Heyang
  Qin, Marko Radmilac, Corby Rosset, Sambudha Roy, Olatunji Ruwase, Olli Saarikivi, Amin Saied, Adil Salim, Michael Santacroce, Shital Shah, Ning Shang, Hiteshi Sharma, Swadheen Shukla, Xia Song, Masahiro Tanaka, Andrea Tupini, Xin Wang, Lijuan Wang, Chunyu Wang, Yu~Wang, Rachel Ward, Guanhua Wang, Philipp Witte, Haiping Wu, Michael Wyatt, Bin Xiao, Can Xu, Jiahang Xu, Weijian Xu, Sonali Yadav, Fan Yang, Jianwei Yang, Ziyi Yang, Yifan Yang, Donghan Yu, Lu~Yuan, Chengruidong Zhang, Cyril Zhang, Jianwen Zhang, Li~Lyna Zhang, Yi~Zhang, Yue Zhang, Yunan Zhang, and Xiren Zhou. 2024.
\newblock \href {http://arxiv.org/abs/2404.14219} {Phi-3 Technical Report: A Highly Capable Language Model Locally on Your Phone}.

\bibitem[{AI et~al.(2024)AI, :, Young, Chen, Li, Huang, Zhang, Zhang, Li, Zhu, Chen, Chang, Yu, Liu, Liu, Yue, Yang, Yang, Yu, Xie, Huang, Hu, Ren, Niu, Nie, Xu, Liu, Wang, Cai, Gu, Liu, and Dai}]{ai2024yiopenfoundationmodels}
01. AI, :, Alex Young, Bei Chen, Chao Li, Chengen Huang, Ge~Zhang, Guanwei Zhang, Heng Li, Jiangcheng Zhu, Jianqun Chen, Jing Chang, Kaidong Yu, Peng Liu, Qiang Liu, Shawn Yue, Senbin Yang, Shiming Yang, Tao Yu, Wen Xie, Wenhao Huang, Xiaohui Hu, Xiaoyi Ren, Xinyao Niu, Pengcheng Nie, Yuchi Xu, Yudong Liu, Yue Wang, Yuxuan Cai, Zhenyu Gu, Zhiyuan Liu, and Zonghong Dai. 2024.
\newblock \href {http://arxiv.org/abs/2403.04652} {Yi: Open Foundation Models by 01.AI}.

\bibitem[{Ainslie et~al.(2023)Ainslie, Lee-Thorp, de~Jong, Zemlyanskiy, Lebrón, and Sanghai}]{ainslie2023gqatraininggeneralizedmultiquery}
Joshua Ainslie, James Lee-Thorp, Michiel de~Jong, Yury Zemlyanskiy, Federico Lebrón, and Sumit Sanghai. 2023.
\newblock \href {http://arxiv.org/abs/2305.13245} {GQA: Training Generalized Multi-Query Transformer Models from Multi-Head Checkpoints}.

\bibitem[{Albalak et~al.(2024)Albalak, Elazar, Xie, Longpre, Lambert, Wang, Muennighoff, Hou, Pan, Jeong, Raffel, Chang, Hashimoto, and Wang}]{albalak2024surveydataselectionlanguage}
Alon Albalak, Yanai Elazar, Sang~Michael Xie, Shayne Longpre, Nathan Lambert, Xinyi Wang, Niklas Muennighoff, Bairu Hou, Liangming Pan, Haewon Jeong, Colin Raffel, Shiyu Chang, Tatsunori Hashimoto, and William~Yang Wang. 2024.
\newblock \href {http://arxiv.org/abs/2402.16827} {A Survey on Data Selection for Language Models}.

\bibitem[{Allal et~al.(2023)Allal, Li, Kocetkov, Mou, Akiki, Ferrandis, Muennighoff, Mishra, Gu, Dey et~al.}]{allal2023santacoder}
Loubna~Ben Allal, Raymond Li, Denis Kocetkov, Chenghao Mou, Christopher Akiki, Carlos~Munoz Ferrandis, Niklas Muennighoff, Mayank Mishra, Alex Gu, Manan Dey, et~al. 2023.
\newblock \href {http://arxiv.org/abs/2301.03988} {SantaCoder: don't reach for the stars!}

\bibitem[{Allal et~al.(2024)Allal, Lozhkov, Bakouch, von Werra, and Wolf}]{allal2024SmolLM}
Loubna~Ben Allal, Anton Lozhkov, Elie Bakouch, Leandro von Werra, and Thomas Wolf. 2024.
\newblock \href {https://huggingface.co/blog/smollm} {SmolLM - blazingly fast and remarkably powerful}.

\bibitem[{Allen-Zhu and Li(2024)}]{allenzhu2024physics}
Zeyuan Allen-Zhu and Yuanzhi Li. 2024.
\newblock \href {http://arxiv.org/abs/2404.05405} {Physics of Language Models: Part 3.3, Knowledge Capacity Scaling Laws}.

\bibitem[{Almazrouei et~al.(2023)Almazrouei, Alobeidli, Alshamsi, Cappelli, Cojocaru, Debbah, Étienne Goffinet, Hesslow, Launay, Malartic, Mazzotta, Noune, Pannier, and Penedo}]{almazrouei2023falcon}
Ebtesam Almazrouei, Hamza Alobeidli, Abdulaziz Alshamsi, Alessandro Cappelli, Ruxandra Cojocaru, Mérouane Debbah, Étienne Goffinet, Daniel Hesslow, Julien Launay, Quentin Malartic, Daniele Mazzotta, Badreddine Noune, Baptiste Pannier, and Guilherme Penedo. 2023.
\newblock \href {http://arxiv.org/abs/2311.16867} {The Falcon Series of Open Language Models}.

\bibitem[{Anil et~al.(2023)Anil, Dai, Firat, Johnson, Lepikhin, Passos, Shakeri, Taropa, Bailey, Chen, Chu, Clark, Shafey, Huang, Meier-Hellstern, Mishra, Moreira, Omernick, Robinson, Ruder, Tay, Xiao, Xu, Zhang, Abrego, Ahn, Austin, Barham, Botha, Bradbury, Brahma, Brooks, Catasta, Cheng, Cherry, Choquette-Choo, Chowdhery, Crepy, Dave, Dehghani, Dev, Devlin, Díaz, Du, Dyer, Feinberg, Feng, Fienber, Freitag, Garcia, Gehrmann, Gonzalez, Gur-Ari, Hand, Hashemi, Hou, Howland, Hu, Hui, Hurwitz, Isard, Ittycheriah, Jagielski, Jia, Kenealy, Krikun, Kudugunta, Lan, Lee, Lee, Li, Li, Li, Li, Li, Lim, Lin, Liu, Liu, Maggioni, Mahendru, Maynez, Misra, Moussalem, Nado, Nham, Ni, Nystrom, Parrish, Pellat, Polacek, Polozov, Pope, Qiao, Reif, Richter, Riley, Ros, Roy, Saeta, Samuel, Shelby, Slone, Smilkov, So, Sohn, Tokumine, Valter, Vasudevan, Vodrahalli, Wang, Wang, Wang, Wang, Wieting, Wu, Xu, Xu, Xue, Yin, Yu, Zhang, Zheng, Zheng, Zhou, Zhou, Petrov, and Wu}]{anil2023palm2technicalreport}
Rohan Anil, Andrew~M. Dai, Orhan Firat, Melvin Johnson, Dmitry Lepikhin, Alexandre Passos, Siamak Shakeri, Emanuel Taropa, Paige Bailey, Zhifeng Chen, Eric Chu, Jonathan~H. Clark, Laurent~El Shafey, Yanping Huang, Kathy Meier-Hellstern, Gaurav Mishra, Erica Moreira, Mark Omernick, Kevin Robinson, Sebastian Ruder, Yi~Tay, Kefan Xiao, Yuanzhong Xu, Yujing Zhang, Gustavo~Hernandez Abrego, Junwhan Ahn, Jacob Austin, Paul Barham, Jan Botha, James Bradbury, Siddhartha Brahma, Kevin Brooks, Michele Catasta, Yong Cheng, Colin Cherry, Christopher~A. Choquette-Choo, Aakanksha Chowdhery, Clément Crepy, Shachi Dave, Mostafa Dehghani, Sunipa Dev, Jacob Devlin, Mark Díaz, Nan Du, Ethan Dyer, Vlad Feinberg, Fangxiaoyu Feng, Vlad Fienber, Markus Freitag, Xavier Garcia, Sebastian Gehrmann, Lucas Gonzalez, Guy Gur-Ari, Steven Hand, Hadi Hashemi, Le~Hou, Joshua Howland, Andrea Hu, Jeffrey Hui, Jeremy Hurwitz, Michael Isard, Abe Ittycheriah, Matthew Jagielski, Wenhao Jia, Kathleen Kenealy, Maxim Krikun, Sneha Kudugunta, Chang
  Lan, Katherine Lee, Benjamin Lee, Eric Li, Music Li, Wei Li, YaGuang Li, Jian Li, Hyeontaek Lim, Hanzhao Lin, Zhongtao Liu, Frederick Liu, Marcello Maggioni, Aroma Mahendru, Joshua Maynez, Vedant Misra, Maysam Moussalem, Zachary Nado, John Nham, Eric Ni, Andrew Nystrom, Alicia Parrish, Marie Pellat, Martin Polacek, Alex Polozov, Reiner Pope, Siyuan Qiao, Emily Reif, Bryan Richter, Parker Riley, Alex~Castro Ros, Aurko Roy, Brennan Saeta, Rajkumar Samuel, Renee Shelby, Ambrose Slone, Daniel Smilkov, David~R. So, Daniel Sohn, Simon Tokumine, Dasha Valter, Vijay Vasudevan, Kiran Vodrahalli, Xuezhi Wang, Pidong Wang, Zirui Wang, Tao Wang, John Wieting, Yuhuai Wu, Kelvin Xu, Yunhan Xu, Linting Xue, Pengcheng Yin, Jiahui Yu, Qiao Zhang, Steven Zheng, Ce~Zheng, Weikang Zhou, Denny Zhou, Slav Petrov, and Yonghui Wu. 2023.
\newblock \href {http://arxiv.org/abs/2305.10403} {PaLM 2 Technical Report}.

\bibitem[{Artetxe et~al.(2022)Artetxe, Bhosale, Goyal, Mihaylov, Ott, Shleifer, Lin, Du, Iyer, Pasunuru, Anantharaman, Li, Chen, Akin, Baines, Martin, Zhou, Koura, O'Horo, Wang, Zettlemoyer, Diab, Kozareva, and Stoyanov}]{artetxe2022efficientlargescalelanguage}
Mikel Artetxe, Shruti Bhosale, Naman Goyal, Todor Mihaylov, Myle Ott, Sam Shleifer, Xi~Victoria Lin, Jingfei Du, Srinivasan Iyer, Ramakanth Pasunuru, Giri Anantharaman, Xian Li, Shuohui Chen, Halil Akin, Mandeep Baines, Louis Martin, Xing Zhou, Punit~Singh Koura, Brian O'Horo, Jeff Wang, Luke Zettlemoyer, Mona Diab, Zornitsa Kozareva, and Ves Stoyanov. 2022.
\newblock \href {http://arxiv.org/abs/2112.10684} {Efficient Large Scale Language Modeling with Mixtures of Experts}.

\bibitem[{Azerbayev et~al.(2023)Azerbayev, Schoelkopf, Paster, Santos, McAleer, Jiang, Deng, Biderman, and Welleck}]{azerbayev2023llemma}
Zhangir Azerbayev, Hailey Schoelkopf, Keiran Paster, Marco~Dos Santos, Stephen McAleer, Albert~Q. Jiang, Jia Deng, Stella Biderman, and Sean Welleck. 2023.
\newblock \href {http://arxiv.org/abs/2310.10631} {Llemma: An Open Language Model For Mathematics}.

\bibitem[{Ba et~al.(2016)Ba, Kiros, and Hinton}]{ba2016layer}
Jimmy~Lei Ba, Jamie~Ryan Kiros, and Geoffrey~E. Hinton. 2016.
\newblock \href {http://arxiv.org/abs/1607.06450} {Layer Normalization}.

\bibitem[{Bai et~al.(2023{\natexlab{a}})Bai, Bai, Chu, Cui, Dang, Deng, Fan, Ge, Han, Huang, Hui, Ji, Li, Lin, Lin, Liu, Liu, Lu, Lu, Ma, Men, Ren, Ren, Tan, Tan, Tu, Wang, Wang, Wang, Wu, Xu, Xu, Yang, Yang, Yang, Yang, Yao, Yu, Yuan, Yuan, Zhang, Zhang, Zhang, Zhang, Zhou, Zhou, Zhou, and Zhu}]{bai2023qwen}
Jinze Bai, Shuai Bai, Yunfei Chu, Zeyu Cui, Kai Dang, Xiaodong Deng, Yang Fan, Wenbin Ge, Yu~Han, Fei Huang, Binyuan Hui, Luo Ji, Mei Li, Junyang Lin, Runji Lin, Dayiheng Liu, Gao Liu, Chengqiang Lu, Keming Lu, Jianxin Ma, Rui Men, Xingzhang Ren, Xuancheng Ren, Chuanqi Tan, Sinan Tan, Jianhong Tu, Peng Wang, Shijie Wang, Wei Wang, Shengguang Wu, Benfeng Xu, Jin Xu, An~Yang, Hao Yang, Jian Yang, Shusheng Yang, Yang Yao, Bowen Yu, Hongyi Yuan, Zheng Yuan, Jianwei Zhang, Xingxuan Zhang, Yichang Zhang, Zhenru Zhang, Chang Zhou, Jingren Zhou, Xiaohuan Zhou, and Tianhang Zhu. 2023{\natexlab{a}}.
\newblock \href {http://arxiv.org/abs/2309.16609} {Qwen Technical Report}.

\bibitem[{Bai et~al.(2023{\natexlab{b}})Bai, Bai, Yang, Wang, Tan, Wang, Lin, Zhou, and Zhou}]{bai2023qwenvlversatilevisionlanguagemodel}
Jinze Bai, Shuai Bai, Shusheng Yang, Shijie Wang, Sinan Tan, Peng Wang, Junyang Lin, Chang Zhou, and Jingren Zhou. 2023{\natexlab{b}}.
\newblock \href {http://arxiv.org/abs/2308.12966} {Qwen-VL: A Versatile Vision-Language Model for Understanding, Localization, Text Reading, and Beyond}.

\bibitem[{Bai et~al.(2022)Bai, Kadavath, Kundu, Askell, Kernion, Jones, Chen, Goldie, Mirhoseini, McKinnon, Chen, Olsson, Olah, Hernandez, Drain, Ganguli, Li, Tran-Johnson, Perez, Kerr, Mueller, Ladish, Landau, Ndousse, Lukosuite, Lovitt, Sellitto, Elhage, Schiefer, Mercado, DasSarma, Lasenby, Larson, Ringer, Johnston, Kravec, Showk, Fort, Lanham, Telleen-Lawton, Conerly, Henighan, Hume, Bowman, Hatfield-Dodds, Mann, Amodei, Joseph, McCandlish, Brown, and Kaplan}]{bai2022constitutionalaiharmlessnessai}
Yuntao Bai, Saurav Kadavath, Sandipan Kundu, Amanda Askell, Jackson Kernion, Andy Jones, Anna Chen, Anna Goldie, Azalia Mirhoseini, Cameron McKinnon, Carol Chen, Catherine Olsson, Christopher Olah, Danny Hernandez, Dawn Drain, Deep Ganguli, Dustin Li, Eli Tran-Johnson, Ethan Perez, Jamie Kerr, Jared Mueller, Jeffrey Ladish, Joshua Landau, Kamal Ndousse, Kamile Lukosuite, Liane Lovitt, Michael Sellitto, Nelson Elhage, Nicholas Schiefer, Noemi Mercado, Nova DasSarma, Robert Lasenby, Robin Larson, Sam Ringer, Scott Johnston, Shauna Kravec, Sheer~El Showk, Stanislav Fort, Tamera Lanham, Timothy Telleen-Lawton, Tom Conerly, Tom Henighan, Tristan Hume, Samuel~R. Bowman, Zac Hatfield-Dodds, Ben Mann, Dario Amodei, Nicholas Joseph, Sam McCandlish, Tom Brown, and Jared Kaplan. 2022.
\newblock \href {http://arxiv.org/abs/2212.08073} {Constitutional AI: Harmlessness from AI Feedback}.

\bibitem[{Bellagente et~al.(2024)Bellagente, Tow, Mahan, Phung, Zhuravinskyi, Adithyan, Baicoianu, Brooks, Cooper, Datta, Lee, Mostaque, Pieler, Pinnaparju, Rocha, Saini, Teufel, Zanichelli, and Riquelme}]{bellagente2024stablelm216b}
Marco Bellagente, Jonathan Tow, Dakota Mahan, Duy Phung, Maksym Zhuravinskyi, Reshinth Adithyan, James Baicoianu, Ben Brooks, Nathan Cooper, Ashish Datta, Meng Lee, Emad Mostaque, Michael Pieler, Nikhil Pinnaparju, Paulo Rocha, Harry Saini, Hannah Teufel, Niccolo Zanichelli, and Carlos Riquelme. 2024.
\newblock \href {http://arxiv.org/abs/2402.17834} {Stable LM 2 1.6B Technical Report}.

\bibitem[{Bengio et~al.(2016)Bengio, Bacon, Pineau, and Precup}]{bengio2016conditionalcomputationneuralnetworks}
Emmanuel Bengio, Pierre-Luc Bacon, Joelle Pineau, and Doina Precup. 2016.
\newblock \href {http://arxiv.org/abs/1511.06297} {Conditional Computation in Neural Networks for faster models}.

\bibitem[{Biderman et~al.(2023)Biderman, Schoelkopf, Anthony, Bradley, O'Brien, Hallahan, Khan, Purohit, Prashanth, Raff, Skowron, Sutawika, and van~der Wal}]{biderman2023pythiasuiteanalyzinglarge}
Stella Biderman, Hailey Schoelkopf, Quentin Anthony, Herbie Bradley, Kyle O'Brien, Eric Hallahan, Mohammad~Aflah Khan, Shivanshu Purohit, USVSN~Sai Prashanth, Edward Raff, Aviya Skowron, Lintang Sutawika, and Oskar van~der Wal. 2023.
\newblock \href {http://arxiv.org/abs/2304.01373} {Pythia: A Suite for Analyzing Large Language Models Across Training and Scaling}.

\bibitem[{Biderman et~al.(2024)Biderman, Schoelkopf, Sutawika, Gao, Tow, Abbasi, Aji, Ammanamanchi, Black, Clive, DiPofi, Etxaniz, Fattori, Forde, Foster, Hsu, Jaiswal, Lee, Li, Lovering, Muennighoff, Pavlick, Phang, Skowron, Tan, Tang, Wang, Winata, Yvon, and Zou}]{biderman2024lessons}
Stella Biderman, Hailey Schoelkopf, Lintang Sutawika, Leo Gao, Jonathan Tow, Baber Abbasi, Alham~Fikri Aji, Pawan~Sasanka Ammanamanchi, Sidney Black, Jordan Clive, Anthony DiPofi, Julen Etxaniz, Benjamin Fattori, Jessica~Zosa Forde, Charles Foster, Jeffrey Hsu, Mimansa Jaiswal, Wilson~Y. Lee, Haonan Li, Charles Lovering, Niklas Muennighoff, Ellie Pavlick, Jason Phang, Aviya Skowron, Samson Tan, Xiangru Tang, Kevin~A. Wang, Genta~Indra Winata, François Yvon, and Andy Zou. 2024.
\newblock \href {http://arxiv.org/abs/2405.14782} {Lessons from the Trenches on Reproducible Evaluation of Language Models}.

\bibitem[{Bisk et~al.(2019)Bisk, Zellers, Bras, Gao, and Choi}]{bisk2019piqareasoningphysicalcommonsense}
Yonatan Bisk, Rowan Zellers, Ronan~Le Bras, Jianfeng Gao, and Yejin Choi. 2019.
\newblock \href {http://arxiv.org/abs/1911.11641} {PIQA: Reasoning about Physical Commonsense in Natural Language}.

\bibitem[{Black et~al.(2022)Black, Biderman, Hallahan, Anthony, Gao, Golding, He, Leahy, McDonell, Phang, Pieler, Prashanth, Purohit, Reynolds, Tow, Wang, and Weinbach}]{black2022gptneox20b}
Sid Black, Stella Biderman, Eric Hallahan, Quentin Anthony, Leo Gao, Laurence Golding, Horace He, Connor Leahy, Kyle McDonell, Jason Phang, Michael Pieler, USVSN~Sai Prashanth, Shivanshu Purohit, Laria Reynolds, Jonathan Tow, Ben Wang, and Samuel Weinbach. 2022.
\newblock \href {http://arxiv.org/abs/2204.06745} {GPT-NeoX-20B: An Open-Source Autoregressive Language Model}.

\bibitem[{Black et~al.(2021)Black, Gao, Wang, Leahy, and Biderman}]{gpt-neo}
Sid Black, Leo Gao, Phil Wang, Connor Leahy, and Stella Biderman. 2021.
\newblock \href {https://doi.org/10.5281/zenodo.5297715} {GPT-Neo: Large Scale Autoregressive Language Modeling with Mesh-Tensorflow}.

\bibitem[{Bommasani et~al.(2023)Bommasani, Klyman, Longpre, Kapoor, Maslej, Xiong, Zhang, and Liang}]{bommasani2023foundationmodeltransparencyindex}
Rishi Bommasani, Kevin Klyman, Shayne Longpre, Sayash Kapoor, Nestor Maslej, Betty Xiong, Daniel Zhang, and Percy Liang. 2023.
\newblock \href {http://arxiv.org/abs/2310.12941} {The Foundation Model Transparency Index}.

\bibitem[{Brown et~al.(2020)Brown, Mann, Ryder, Subbiah, Kaplan, Dhariwal, Neelakantan, Shyam, Sastry, Askell et~al.}]{brown2020language}
Tom~B. Brown, Benjamin Mann, Nick Ryder, Melanie Subbiah, Jared Kaplan, Prafulla Dhariwal, Arvind Neelakantan, Pranav Shyam, Girish Sastry, Amanda Askell, et~al. 2020.
\newblock \href {http://arxiv.org/abs/2005.14165} {Language Models are Few-Shot Learners}.

\bibitem[{Cai(2023)}]{mixtralupcycle}
Tianle Cai. 2023.
\newblock \href {https://x.com/tianle_cai/status/1734188749117153684} {Mixtral from Mistral}.

\bibitem[{Cai et~al.(2024)Cai, Cao, Chen, Chen, Chen, Chen, Chen, Chen, Chen, Chu, Dong, Duan, Fan, Fei, Gao, Ge, Gu, Gu, Gui, Guo, Guo, He, Hu, Huang, Jiang, Jiao, Jin, Lei, Li, Li, Li, Li, Li, Li, Liu, Liu, Hong, Liu, Liu, Liu, Lv, Lv, Lv, Ma, Ma, Ma, Ning, Ouyang, Qiu, Qu, Shang, Shao, Song, Song, Sui, Sun, Sun, Tang, Wang, Wang, Wang, Wang, Wang, Wang, Wang, Wei, Weng, Wu, Xiong, Xu, Xu, Yan, Yan, Yang, Ye, Ying, Yu, Yu, Zang, Zhang, Zhang, Zhang, Zhang, Zhang, Zhang, Zhang, Zhang, Zhang, Zhang, Zhang, Zhao, Zhao, Zhao, Zhou, Zhou, Zhuo, Zou, Qiu, Qiao, and Lin}]{cai2024internlm2}
Zheng Cai, Maosong Cao, Haojiong Chen, Kai Chen, Keyu Chen, Xin Chen, Xun Chen, Zehui Chen, Zhi Chen, Pei Chu, Xiaoyi Dong, Haodong Duan, Qi~Fan, Zhaoye Fei, Yang Gao, Jiaye Ge, Chenya Gu, Yuzhe Gu, Tao Gui, Aijia Guo, Qipeng Guo, Conghui He, Yingfan Hu, Ting Huang, Tao Jiang, Penglong Jiao, Zhenjiang Jin, Zhikai Lei, Jiaxing Li, Jingwen Li, Linyang Li, Shuaibin Li, Wei Li, Yining Li, Hongwei Liu, Jiangning Liu, Jiawei Hong, Kaiwen Liu, Kuikun Liu, Xiaoran Liu, Chengqi Lv, Haijun Lv, Kai Lv, Li~Ma, Runyuan Ma, Zerun Ma, Wenchang Ning, Linke Ouyang, Jiantao Qiu, Yuan Qu, Fukai Shang, Yunfan Shao, Demin Song, Zifan Song, Zhihao Sui, Peng Sun, Yu~Sun, Huanze Tang, Bin Wang, Guoteng Wang, Jiaqi Wang, Jiayu Wang, Rui Wang, Yudong Wang, Ziyi Wang, Xingjian Wei, Qizhen Weng, Fan Wu, Yingtong Xiong, Chao Xu, Ruiliang Xu, Hang Yan, Yirong Yan, Xiaogui Yang, Haochen Ye, Huaiyuan Ying, Jia Yu, Jing Yu, Yuhang Zang, Chuyu Zhang, Li~Zhang, Pan Zhang, Peng Zhang, Ruijie Zhang, Shuo Zhang, Songyang Zhang, Wenjian Zhang,
  Wenwei Zhang, Xingcheng Zhang, Xinyue Zhang, Hui Zhao, Qian Zhao, Xiaomeng Zhao, Fengzhe Zhou, Zaida Zhou, Jingming Zhuo, Yicheng Zou, Xipeng Qiu, Yu~Qiao, and Dahua Lin. 2024.
\newblock \href {http://arxiv.org/abs/2403.17297} {InternLM2 Technical Report}.

\bibitem[{Chen et~al.(2020)Chen, Radford, Child, Wu, Jun, Luan, and Sutskever}]{chen2020generative}
Mark Chen, Alec Radford, Rewon Child, Jeffrey Wu, Heewoo Jun, David Luan, and Ilya Sutskever. 2020.
\newblock \href {https://cdn.openai.com/papers/Generative_Pretraining_from_Pixels_V2.pdf} {Generative pretraining from pixels}.

\bibitem[{Chen et~al.(2021)Chen, Tworek, Jun, Yuan, de~Oliveira~Pinto, Kaplan, Edwards, Burda, Joseph, Brockman, Ray, Puri, Krueger, Petrov, Khlaaf, Sastry, Mishkin, Chan, Gray, Ryder, Pavlov, Power, Kaiser, Bavarian, Winter, Tillet, Such, Cummings, Plappert, Chantzis, Barnes, Herbert-Voss, Guss, Nichol, Paino, Tezak, Tang, Babuschkin, Balaji, Jain, Saunders, Hesse, Carr, Leike, Achiam, Misra, Morikawa, Radford, Knight, Brundage, Murati, Mayer, Welinder, McGrew, Amodei, McCandlish, Sutskever, and Zaremba}]{chen2021evaluating}
Mark Chen, Jerry Tworek, Heewoo Jun, Qiming Yuan, Henrique~Ponde de~Oliveira~Pinto, Jared Kaplan, Harri Edwards, Yuri Burda, Nicholas Joseph, Greg Brockman, Alex Ray, Raul Puri, Gretchen Krueger, Michael Petrov, Heidy Khlaaf, Girish Sastry, Pamela Mishkin, Brooke Chan, Scott Gray, Nick Ryder, Mikhail Pavlov, Alethea Power, Lukasz Kaiser, Mohammad Bavarian, Clemens Winter, Philippe Tillet, Felipe~Petroski Such, Dave Cummings, Matthias Plappert, Fotios Chantzis, Elizabeth Barnes, Ariel Herbert-Voss, William~Hebgen Guss, Alex Nichol, Alex Paino, Nikolas Tezak, Jie Tang, Igor Babuschkin, Suchir Balaji, Shantanu Jain, William Saunders, Christopher Hesse, Andrew~N. Carr, Jan Leike, Josh Achiam, Vedant Misra, Evan Morikawa, Alec Radford, Matthew Knight, Miles Brundage, Mira Murati, Katie Mayer, Peter Welinder, Bob McGrew, Dario Amodei, Sam McCandlish, Ilya Sutskever, and Wojciech Zaremba. 2021.
\newblock \href {http://arxiv.org/abs/2107.03374} {Evaluating Large Language Models Trained on Code}.

\bibitem[{Chintala(2024)}]{gpt4moe}
Soumith Chintala. 2024.
\newblock \href {https://x.com/soumithchintala/status/1671267150101721090} {GPT-4 MoE}.

\bibitem[{Chowdhery et~al.(2022)Chowdhery, Narang, Devlin, Bosma, Mishra, Roberts, Barham, Chung, Sutton, Gehrmann et~al.}]{chowdhery2022palm}
Aakanksha Chowdhery, Sharan Narang, Jacob Devlin, Maarten Bosma, Gaurav Mishra, Adam Roberts, Paul Barham, Hyung~Won Chung, Charles Sutton, Sebastian Gehrmann, et~al. 2022.
\newblock \href {http://arxiv.org/abs/2204.02311} {PaLM: Scaling Language Modeling with Pathways}.

\bibitem[{Christiano et~al.(2023)Christiano, Leike, Brown, Martic, Legg, and Amodei}]{christiano2023deep}
Paul Christiano, Jan Leike, Tom~B. Brown, Miljan Martic, Shane Legg, and Dario Amodei. 2023.
\newblock \href {http://arxiv.org/abs/1706.03741} {Deep reinforcement learning from human preferences}.

\bibitem[{Clark et~al.(2022)Clark, de~las Casas, Guy, Mensch, Paganini, Hoffmann, Damoc, Hechtman, Cai, Borgeaud, van~den Driessche, Rutherford, Hennigan, Johnson, Millican, Cassirer, Jones, Buchatskaya, Budden, Sifre, Osindero, Vinyals, Rae, Elsen, Kavukcuoglu, and Simonyan}]{clark2022unifiedscalinglawsrouted}
Aidan Clark, Diego de~las Casas, Aurelia Guy, Arthur Mensch, Michela Paganini, Jordan Hoffmann, Bogdan Damoc, Blake Hechtman, Trevor Cai, Sebastian Borgeaud, George van~den Driessche, Eliza Rutherford, Tom Hennigan, Matthew Johnson, Katie Millican, Albin Cassirer, Chris Jones, Elena Buchatskaya, David Budden, Laurent Sifre, Simon Osindero, Oriol Vinyals, Jack Rae, Erich Elsen, Koray Kavukcuoglu, and Karen Simonyan. 2022.
\newblock \href {http://arxiv.org/abs/2202.01169} {Unified Scaling Laws for Routed Language Models}.

\bibitem[{Clark et~al.(2019)Clark, Lee, Chang, Kwiatkowski, Collins, and Toutanova}]{clark2019boolqexploringsurprisingdifficulty}
Christopher Clark, Kenton Lee, Ming-Wei Chang, Tom Kwiatkowski, Michael Collins, and Kristina Toutanova. 2019.
\newblock \href {http://arxiv.org/abs/1905.10044} {BoolQ: Exploring the Surprising Difficulty of Natural Yes/No Questions}.

\bibitem[{Clark et~al.(2018)Clark, Cowhey, Etzioni, Khot, Sabharwal, Schoenick, and Tafjord}]{clark2018thinksolvedquestionanswering}
Peter Clark, Isaac Cowhey, Oren Etzioni, Tushar Khot, Ashish Sabharwal, Carissa Schoenick, and Oyvind Tafjord. 2018.
\newblock \href {http://arxiv.org/abs/1803.05457} {Think you have Solved Question Answering? Try ARC, the AI2 Reasoning Challenge}.

\bibitem[{Cobbe et~al.(2021)Cobbe, Kosaraju, Bavarian, Chen, Jun, Kaiser, Plappert, Tworek, Hilton, Nakano, Hesse, and Schulman}]{cobbe2021training}
Karl Cobbe, Vineet Kosaraju, Mohammad Bavarian, Mark Chen, Heewoo Jun, Lukasz Kaiser, Matthias Plappert, Jerry Tworek, Jacob Hilton, Reiichiro Nakano, Christopher Hesse, and John Schulman. 2021.
\newblock \href {http://arxiv.org/abs/2110.14168} {Training Verifiers to Solve Math Word Problems}.

\bibitem[{Computer(2023)}]{together2023redpajama}
Together Computer. 2023.
\newblock \href {https://github.com/togethercomputer/RedPajama-Data} {RedPajama: An Open Source Recipe to Reproduce LLaMA training dataset}.

\bibitem[{Csordás et~al.(2024)Csordás, Irie, Schmidhuber, Potts, and Manning}]{csordás2024moeut}
Róbert Csordás, Kazuki Irie, Jürgen Schmidhuber, Christopher Potts, and Christopher~D. Manning. 2024.
\newblock \href {http://arxiv.org/abs/2405.16039} {MoEUT: Mixture-of-Experts Universal Transformers}.

\bibitem[{Cui et~al.(2023)Cui, Yuan, Ding, Yao, Zhu, Ni, Xie, Liu, and Sun}]{cui2023ultrafeedback}
Ganqu Cui, Lifan Yuan, Ning Ding, Guanming Yao, Wei Zhu, Yuan Ni, Guotong Xie, Zhiyuan Liu, and Maosong Sun. 2023.
\newblock \href {http://arxiv.org/abs/2310.01377} {UltraFeedback: Boosting Language Models with High-quality Feedback}.

\bibitem[{Dai et~al.(2024)Dai, Deng, Zhao, Xu, Gao, Chen, Li, Zeng, Yu, Wu, Xie, Li, Huang, Luo, Ruan, Sui, and Liang}]{dai2024deepseekmoeultimateexpertspecialization}
Damai Dai, Chengqi Deng, Chenggang Zhao, R.~X. Xu, Huazuo Gao, Deli Chen, Jiashi Li, Wangding Zeng, Xingkai Yu, Y.~Wu, Zhenda Xie, Y.~K. Li, Panpan Huang, Fuli Luo, Chong Ruan, Zhifang Sui, and Wenfeng Liang. 2024.
\newblock \href {http://arxiv.org/abs/2401.06066} {DeepSeekMoE: Towards Ultimate Expert Specialization in Mixture-of-Experts Language Models}.

\bibitem[{Databricks(2024)}]{dbrx}
Databricks. 2024.
\newblock \href {https://www.databricks.com/blog/introducing-dbrx-new-state-art-open-llm} {DBRX}.

\bibitem[{Dauphin et~al.(2017)Dauphin, Fan, Auli, and Grangier}]{dauphin2017languagemodelinggatedconvolutional}
Yann~N. Dauphin, Angela Fan, Michael Auli, and David Grangier. 2017.
\newblock \href {http://arxiv.org/abs/1612.08083} {Language Modeling with Gated Convolutional Networks}.

\bibitem[{DeepSeek-AI et~al.(2024{\natexlab{a}})DeepSeek-AI, :, Bi, Chen, Chen, Chen, Dai, Deng, Ding, Dong, Du, Fu, Gao, Gao, Gao, Ge, Guan, Guo, Guo, Hao, Hao, He, Hu, Huang, Li, Li, Li, Li, Li, Liang, Lin, Liu, Liu, Liu, Liu, Liu, Liu, Lu, Lu, Luo, Ma, Nie, Pei, Piao, Qiu, Qu, Ren, Ren, Ruan, Sha, Shao, Song, Su, Sun, Sun, Tang, Wang, Wang, Wang, Wang, Wang, Wu, Wu, Xie, Xie, Xie, Xiong, Xu, Xu, Xu, Yang, You, Yu, Yu, Zhang, Zhang, Zhang, Zhang, Zhang, Zhang, Zhang, Zhang, Zhao, Zhao, Zhou, Zhou, Zhu, and Zou}]{deepseekai2024deepseek}
DeepSeek-AI, :, Xiao Bi, Deli Chen, Guanting Chen, Shanhuang Chen, Damai Dai, Chengqi Deng, Honghui Ding, Kai Dong, Qiushi Du, Zhe Fu, Huazuo Gao, Kaige Gao, Wenjun Gao, Ruiqi Ge, Kang Guan, Daya Guo, Jianzhong Guo, Guangbo Hao, Zhewen Hao, Ying He, Wenjie Hu, Panpan Huang, Erhang Li, Guowei Li, Jiashi Li, Yao Li, Y.~K. Li, Wenfeng Liang, Fangyun Lin, A.~X. Liu, Bo~Liu, Wen Liu, Xiaodong Liu, Xin Liu, Yiyuan Liu, Haoyu Lu, Shanghao Lu, Fuli Luo, Shirong Ma, Xiaotao Nie, Tian Pei, Yishi Piao, Junjie Qiu, Hui Qu, Tongzheng Ren, Zehui Ren, Chong Ruan, Zhangli Sha, Zhihong Shao, Junxiao Song, Xuecheng Su, Jingxiang Sun, Yaofeng Sun, Minghui Tang, Bingxuan Wang, Peiyi Wang, Shiyu Wang, Yaohui Wang, Yongji Wang, Tong Wu, Y.~Wu, Xin Xie, Zhenda Xie, Ziwei Xie, Yiliang Xiong, Hanwei Xu, R.~X. Xu, Yanhong Xu, Dejian Yang, Yuxiang You, Shuiping Yu, Xingkai Yu, B.~Zhang, Haowei Zhang, Lecong Zhang, Liyue Zhang, Mingchuan Zhang, Minghua Zhang, Wentao Zhang, Yichao Zhang, Chenggang Zhao, Yao Zhao, Shangyan Zhou, Shunfeng
  Zhou, Qihao Zhu, and Yuheng Zou. 2024{\natexlab{a}}.
\newblock \href {http://arxiv.org/abs/2401.02954} {DeepSeek LLM: Scaling Open-Source Language Models with Longtermism}.

\bibitem[{DeepSeek-AI et~al.(2024{\natexlab{b}})DeepSeek-AI, Liu, Feng, Wang, Wang, Liu, Zhao, Dengr, Ruan, Dai, Guo, Yang, Chen, Ji, Li, Lin, Luo, Hao, Chen, Li, Zhang, Xu, Yang, Zhang, Ding, Xin, Gao, Li, Qu, Cai, Liang, Guo, Ni, Li, Chen, Yuan, Qiu, Song, Dong, Gao, Guan, Wang, Zhang, Xu, Xia, Zhao, Zhang, Li, Wang, Zhang, Zhang, Tang, Li, Tian, Huang, Wang, Zhang, Zhu, Chen, Du, Chen, Jin, Ge, Pan, Xu, Chen, Li, Lu, Zhou, Chen, Wu, Ye, Ma, Wang, Zhou, Yu, Zhou, Zheng, Wang, Pei, Yuan, Sun, Xiao, Zeng, An, Liu, Liang, Gao, Zhang, Li, Jin, Wang, Bi, Liu, Wang, Shen, Chen, Chen, Nie, Sun, Wang, Liu, Xie, Yu, Song, Zhou, Yang, Lu, Su, Wu, Li, Wei, Zhu, Xu, Huang, Li, Zhao, Sun, Li, Wang, Zheng, Zhang, Xiong, Zhao, He, Tang, Piao, Dong, Tan, Liu, Wang, Guo, Zhu, Wang, Zou, Zha, Ma, Yan, You, Liu, Ren, Ren, Sha, Fu, Huang, Zhang, Xie, Hao, Shao, Wen, Xu, Zhang, Li, Wang, Gu, Li, and Xie}]{deepseekai2024deepseekv2}
DeepSeek-AI, Aixin Liu, Bei Feng, Bin Wang, Bingxuan Wang, Bo~Liu, Chenggang Zhao, Chengqi Dengr, Chong Ruan, Damai Dai, Daya Guo, Dejian Yang, Deli Chen, Dongjie Ji, Erhang Li, Fangyun Lin, Fuli Luo, Guangbo Hao, Guanting Chen, Guowei Li, H.~Zhang, Hanwei Xu, Hao Yang, Haowei Zhang, Honghui Ding, Huajian Xin, Huazuo Gao, Hui Li, Hui Qu, J.~L. Cai, Jian Liang, Jianzhong Guo, Jiaqi Ni, Jiashi Li, Jin Chen, Jingyang Yuan, Junjie Qiu, Junxiao Song, Kai Dong, Kaige Gao, Kang Guan, Lean Wang, Lecong Zhang, Lei Xu, Leyi Xia, Liang Zhao, Liyue Zhang, Meng Li, Miaojun Wang, Mingchuan Zhang, Minghua Zhang, Minghui Tang, Mingming Li, Ning Tian, Panpan Huang, Peiyi Wang, Peng Zhang, Qihao Zhu, Qinyu Chen, Qiushi Du, R.~J. Chen, R.~L. Jin, Ruiqi Ge, Ruizhe Pan, Runxin Xu, Ruyi Chen, S.~S. Li, Shanghao Lu, Shangyan Zhou, Shanhuang Chen, Shaoqing Wu, Shengfeng Ye, Shirong Ma, Shiyu Wang, Shuang Zhou, Shuiping Yu, Shunfeng Zhou, Size Zheng, T.~Wang, Tian Pei, Tian Yuan, Tianyu Sun, W.~L. Xiao, Wangding Zeng, Wei An, Wen
  Liu, Wenfeng Liang, Wenjun Gao, Wentao Zhang, X.~Q. Li, Xiangyue Jin, Xianzu Wang, Xiao Bi, Xiaodong Liu, Xiaohan Wang, Xiaojin Shen, Xiaokang Chen, Xiaosha Chen, Xiaotao Nie, Xiaowen Sun, Xiaoxiang Wang, Xin Liu, Xin Xie, Xingkai Yu, Xinnan Song, Xinyi Zhou, Xinyu Yang, Xuan Lu, Xuecheng Su, Y.~Wu, Y.~K. Li, Y.~X. Wei, Y.~X. Zhu, Yanhong Xu, Yanping Huang, Yao Li, Yao Zhao, Yaofeng Sun, Yaohui Li, Yaohui Wang, Yi~Zheng, Yichao Zhang, Yiliang Xiong, Yilong Zhao, Ying He, Ying Tang, Yishi Piao, Yixin Dong, Yixuan Tan, Yiyuan Liu, Yongji Wang, Yongqiang Guo, Yuchen Zhu, Yuduan Wang, Yuheng Zou, Yukun Zha, Yunxian Ma, Yuting Yan, Yuxiang You, Yuxuan Liu, Z.~Z. Ren, Zehui Ren, Zhangli Sha, Zhe Fu, Zhen Huang, Zhen Zhang, Zhenda Xie, Zhewen Hao, Zhihong Shao, Zhiniu Wen, Zhipeng Xu, Zhongyu Zhang, Zhuoshu Li, Zihan Wang, Zihui Gu, Zilin Li, and Ziwei Xie. 2024{\natexlab{b}}.
\newblock \href {http://arxiv.org/abs/2405.04434} {DeepSeek-V2: A Strong, Economical, and Efficient Mixture-of-Experts Language Model}.

\bibitem[{Dehghani et~al.(2023)Dehghani, Djolonga, Mustafa, Padlewski, Heek, Gilmer, Steiner, Caron, Geirhos, Alabdulmohsin, Jenatton, Beyer, Tschannen, Arnab, Wang, Riquelme, Minderer, Puigcerver, Evci, Kumar, van Steenkiste, Elsayed, Mahendran, Yu, Oliver, Huot, Bastings, Collier, Gritsenko, Birodkar, Vasconcelos, Tay, Mensink, Kolesnikov, Pavetić, Tran, Kipf, Lučić, Zhai, Keysers, Harmsen, and Houlsby}]{dehghani2023scaling}
Mostafa Dehghani, Josip Djolonga, Basil Mustafa, Piotr Padlewski, Jonathan Heek, Justin Gilmer, Andreas Steiner, Mathilde Caron, Robert Geirhos, Ibrahim Alabdulmohsin, Rodolphe Jenatton, Lucas Beyer, Michael Tschannen, Anurag Arnab, Xiao Wang, Carlos Riquelme, Matthias Minderer, Joan Puigcerver, Utku Evci, Manoj Kumar, Sjoerd van Steenkiste, Gamaleldin~F. Elsayed, Aravindh Mahendran, Fisher Yu, Avital Oliver, Fantine Huot, Jasmijn Bastings, Mark~Patrick Collier, Alexey Gritsenko, Vighnesh Birodkar, Cristina Vasconcelos, Yi~Tay, Thomas Mensink, Alexander Kolesnikov, Filip Pavetić, Dustin Tran, Thomas Kipf, Mario Lučić, Xiaohua Zhai, Daniel Keysers, Jeremiah Harmsen, and Neil Houlsby. 2023.
\newblock \href {http://arxiv.org/abs/2302.05442} {Scaling Vision Transformers to 22 Billion Parameters}.

\bibitem[{Dehghani et~al.(2019)Dehghani, Gouws, Vinyals, Uszkoreit, and Łukasz Kaiser}]{dehghani2019universaltransformers}
Mostafa Dehghani, Stephan Gouws, Oriol Vinyals, Jakob Uszkoreit, and Łukasz Kaiser. 2019.
\newblock \href {http://arxiv.org/abs/1807.03819} {Universal Transformers}.

\bibitem[{Dey et~al.(2023)Dey, Gosal, Zhiming, Chen, Khachane, Marshall, Pathria, Tom, and Hestness}]{dey2023cerebrasgptopencomputeoptimallanguage}
Nolan Dey, Gurpreet Gosal, Zhiming, Chen, Hemant Khachane, William Marshall, Ribhu Pathria, Marvin Tom, and Joel Hestness. 2023.
\newblock \href {http://arxiv.org/abs/2304.03208} {Cerebras-GPT: Open Compute-Optimal Language Models Trained on the Cerebras Wafer-Scale Cluster}.

\bibitem[{Driess et~al.(2023)Driess, Xia, Sajjadi, Lynch, Chowdhery, Ichter, Wahid, Tompson, Vuong, Yu, Huang, Chebotar, Sermanet, Duckworth, Levine, Vanhoucke, Hausman, Toussaint, Greff, Zeng, Mordatch, and Florence}]{driess2023palmeembodiedmultimodallanguage}
Danny Driess, Fei Xia, Mehdi S.~M. Sajjadi, Corey Lynch, Aakanksha Chowdhery, Brian Ichter, Ayzaan Wahid, Jonathan Tompson, Quan Vuong, Tianhe Yu, Wenlong Huang, Yevgen Chebotar, Pierre Sermanet, Daniel Duckworth, Sergey Levine, Vincent Vanhoucke, Karol Hausman, Marc Toussaint, Klaus Greff, Andy Zeng, Igor Mordatch, and Pete Florence. 2023.
\newblock \href {http://arxiv.org/abs/2303.03378} {PaLM-E: An Embodied Multimodal Language Model}.

\bibitem[{Du et~al.(2022)Du, Huang, Dai, Tong, Lepikhin, Xu, Krikun, Zhou, Yu, Firat, Zoph, Fedus, Bosma, Zhou, Wang, Wang, Webster, Pellat, Robinson, Meier-Hellstern, Duke, Dixon, Zhang, Le, Wu, Chen, and Cui}]{du2022glam}
Nan Du, Yanping Huang, Andrew~M. Dai, Simon Tong, Dmitry Lepikhin, Yuanzhong Xu, Maxim Krikun, Yanqi Zhou, Adams~Wei Yu, Orhan Firat, Barret Zoph, Liam Fedus, Maarten Bosma, Zongwei Zhou, Tao Wang, Yu~Emma Wang, Kellie Webster, Marie Pellat, Kevin Robinson, Kathleen Meier-Hellstern, Toju Duke, Lucas Dixon, Kun Zhang, Quoc~V Le, Yonghui Wu, Zhifeng Chen, and Claire Cui. 2022.
\newblock \href {http://arxiv.org/abs/2112.06905} {GLaM: Efficient Scaling of Language Models with Mixture-of-Experts}.

\bibitem[{Dua et~al.(2021)Dua, Bhosale, Goswami, Cross, Lewis, and Fan}]{dua2021trickstrainingsparsetranslation}
Dheeru Dua, Shruti Bhosale, Vedanuj Goswami, James Cross, Mike Lewis, and Angela Fan. 2021.
\newblock \href {http://arxiv.org/abs/2110.08246} {Tricks for Training Sparse Translation Models}.

\bibitem[{Dubey et~al.(2024)Dubey, Jauhri, Pandey, Kadian, Al-Dahle, Letman, Mathur, Schelten, Yang, Fan, Goyal, Hartshorn, Yang, Mitra, Sravankumar, Korenev, Hinsvark, Rao, Zhang, Rodriguez, Gregerson et~al.}]{dubey2024llama3herdmodels}
Abhimanyu Dubey, Abhinav Jauhri, Abhinav Pandey, Abhishek Kadian, Ahmad Al-Dahle, Aiesha Letman, Akhil Mathur, Alan Schelten, Amy Yang, Angela Fan, Anirudh Goyal, Anthony Hartshorn, Aobo Yang, Archi Mitra, Archie Sravankumar, Artem Korenev, Arthur Hinsvark, Arun Rao, Aston Zhang, Aurelien Rodriguez, Austen Gregerson, et~al. 2024.
\newblock \href {http://arxiv.org/abs/2407.21783} {The Llama 3 Herd of Models}.

\bibitem[{Dubois et~al.(2024)Dubois, Galambosi, Liang, and Hashimoto}]{dubois2024lengthcontrolledalpacaevalsimpleway}
Yann Dubois, Balázs Galambosi, Percy Liang, and Tatsunori~B. Hashimoto. 2024.
\newblock \href {http://arxiv.org/abs/2404.04475} {Length-Controlled AlpacaEval: A Simple Way to Debias Automatic Evaluators}.

\bibitem[{Eigen et~al.(2014)Eigen, Ranzato, and Sutskever}]{eigen2014learningfactoredrepresentationsdeep}
David Eigen, Marc'Aurelio Ranzato, and Ilya Sutskever. 2014.
\newblock \href {http://arxiv.org/abs/1312.4314} {Learning Factored Representations in a Deep Mixture of Experts}.

\bibitem[{Enevoldsen et~al.(2024)Enevoldsen, Kardos, Muennighoff, and Nielbo}]{enevoldsen2024scandinavianembeddingbenchmarkscomprehensive}
Kenneth Enevoldsen, Márton Kardos, Niklas Muennighoff, and Kristoffer~Laigaard Nielbo. 2024.
\newblock \href {http://arxiv.org/abs/2406.02396} {The Scandinavian Embedding Benchmarks: Comprehensive Assessment of Multilingual and Monolingual Text Embedding}.

\bibitem[{Ethayarajh et~al.(2024)Ethayarajh, Xu, Muennighoff, Jurafsky, and Kiela}]{ethayarajh2024kto}
Kawin Ethayarajh, Winnie Xu, Niklas Muennighoff, Dan Jurafsky, and Douwe Kiela. 2024.
\newblock \href {http://arxiv.org/abs/2402.01306} {KTO: Model Alignment as Prospect Theoretic Optimization}.

\bibitem[{Faysse et~al.(2024)Faysse, Fernandes, Guerreiro, Loison, Alves, Corro, Boizard, Alves, Rei, Martins, Casademunt, Yvon, Martins, Viaud, Hudelot, and Colombo}]{faysse2024croissantllmtrulybilingualfrenchenglish}
Manuel Faysse, Patrick Fernandes, Nuno~M. Guerreiro, António Loison, Duarte~M. Alves, Caio Corro, Nicolas Boizard, João Alves, Ricardo Rei, Pedro~H. Martins, Antoni~Bigata Casademunt, François Yvon, André F.~T. Martins, Gautier Viaud, Céline Hudelot, and Pierre Colombo. 2024.
\newblock \href {http://arxiv.org/abs/2402.00786} {CroissantLLM: A Truly Bilingual French-English Language Model}.

\bibitem[{Fedus et~al.(2022)Fedus, Zoph, and Shazeer}]{fedus2022switch}
William Fedus, Barret Zoph, and Noam Shazeer. 2022.
\newblock \href {http://arxiv.org/abs/2101.03961} {Switch Transformers: Scaling to Trillion Parameter Models with Simple and Efficient Sparsity}.

\bibitem[{Gadre et~al.(2024)Gadre, Smyrnis, Shankar, Gururangan, Wortsman, Shao, Mercat, Fang, Li, Keh, Xin, Nezhurina, Vasiljevic, Jitsev, Soldaini, Dimakis, Ilharco, Koh, Song, Kollar, Carmon, Dave, Heckel, Muennighoff, and Schmidt}]{gadre2024language}
Samir~Yitzhak Gadre, Georgios Smyrnis, Vaishaal Shankar, Suchin Gururangan, Mitchell Wortsman, Rulin Shao, Jean Mercat, Alex Fang, Jeffrey Li, Sedrick Keh, Rui Xin, Marianna Nezhurina, Igor Vasiljevic, Jenia Jitsev, Luca Soldaini, Alexandros~G. Dimakis, Gabriel Ilharco, Pang~Wei Koh, Shuran Song, Thomas Kollar, Yair Carmon, Achal Dave, Reinhard Heckel, Niklas Muennighoff, and Ludwig Schmidt. 2024.
\newblock \href {http://arxiv.org/abs/2403.08540} {Language models scale reliably with over-training and on downstream tasks}.

\bibitem[{Gale et~al.(2022)Gale, Narayanan, Young, and Zaharia}]{gale2022megablocksefficientsparsetraining}
Trevor Gale, Deepak Narayanan, Cliff Young, and Matei Zaharia. 2022.
\newblock \href {http://arxiv.org/abs/2211.15841} {MegaBlocks: Efficient Sparse Training with Mixture-of-Experts}.

\bibitem[{Gao et~al.(2020)Gao, Biderman, Black, Golding, Hoppe, Foster, Phang, He, Thite, Nabeshima, Presser, and Leahy}]{gao2020pile800gbdatasetdiverse}
Leo Gao, Stella Biderman, Sid Black, Laurence Golding, Travis Hoppe, Charles Foster, Jason Phang, Horace He, Anish Thite, Noa Nabeshima, Shawn Presser, and Connor Leahy. 2020.
\newblock \href {http://arxiv.org/abs/2101.00027} {The Pile: An 800GB Dataset of Diverse Text for Language Modeling}.

\bibitem[{Gao et~al.(2021)Gao, Tow, Biderman, Black, DiPofi, Foster, Golding, Hsu, McDonell, Muennighoff, Phang, Reynolds, Tang, Thite, Wang, Wang, and Zou}]{eval-harness}
Leo Gao, Jonathan Tow, Stella Biderman, Sid Black, Anthony DiPofi, Charles Foster, Laurence Golding, Jeffrey Hsu, Kyle McDonell, Niklas Muennighoff, Jason Phang, Laria Reynolds, Eric Tang, Anish Thite, Ben Wang, Kevin Wang, and Andy Zou. 2021.
\newblock \href {https://doi.org/10.5281/zenodo.5371628} {A framework for few-shot language model evaluation}.

\bibitem[{GLM et~al.(2024)GLM, Zeng, Xu, Wang, Zhang, Yin, Rojas, Feng, Zhao, Lai, Yu, Wang, Sun, Zhang, Cheng, Gui, Tang, Zhang, Li, Zhao, Wu, Zhong, Liu, Huang, Zhang, Zheng, Lu, Duan, Zhang, Cao, Yang, Tam, Zhao, Liu, Xia, Zhang, Gu, Lv, Liu, Liu, Yang, Song, Zhang, An, Xu, Niu, Yang, Li, Bai, Dong, Qi, Wang, Yang, Du, Hou, and Wang}]{glm2024chatglm}
Team GLM, Aohan Zeng, Bin Xu, Bowen Wang, Chenhui Zhang, Da~Yin, Diego Rojas, Guanyu Feng, Hanlin Zhao, Hanyu Lai, Hao Yu, Hongning Wang, Jiadai Sun, Jiajie Zhang, Jiale Cheng, Jiayi Gui, Jie Tang, Jing Zhang, Juanzi Li, Lei Zhao, Lindong Wu, Lucen Zhong, Mingdao Liu, Minlie Huang, Peng Zhang, Qinkai Zheng, Rui Lu, Shuaiqi Duan, Shudan Zhang, Shulin Cao, Shuxun Yang, Weng~Lam Tam, Wenyi Zhao, Xiao Liu, Xiao Xia, Xiaohan Zhang, Xiaotao Gu, Xin Lv, Xinghan Liu, Xinyi Liu, Xinyue Yang, Xixuan Song, Xunkai Zhang, Yifan An, Yifan Xu, Yilin Niu, Yuantao Yang, Yueyan Li, Yushi Bai, Yuxiao Dong, Zehan Qi, Zhaoyu Wang, Zhen Yang, Zhengxiao Du, Zhenyu Hou, and Zihan Wang. 2024.
\newblock \href {http://arxiv.org/abs/2406.12793} {ChatGLM: A Family of Large Language Models from GLM-130B to GLM-4 All Tools}.

\bibitem[{Glorioso et~al.(2024)Glorioso, Anthony, Tokpanov, Whittington, Pilault, Ibrahim, and Millidge}]{glorioso2024zambacompact7bssm}
Paolo Glorioso, Quentin Anthony, Yury Tokpanov, James Whittington, Jonathan Pilault, Adam Ibrahim, and Beren Millidge. 2024.
\newblock \href {http://arxiv.org/abs/2405.16712} {Zamba: A Compact 7B SSM Hybrid Model}.

\bibitem[{Gordon et~al.(2012)Gordon, Kozareva, and Roemmele}]{gordon2012semeval}
Andrew Gordon, Zornitsa Kozareva, and Melissa Roemmele. 2012.
\newblock \href {https://aclanthology.org/S12-1052} {{S}em{E}val-2012 Task 7: Choice of Plausible Alternatives: An Evaluation of Commonsense Causal Reasoning}.

\bibitem[{Groeneveld et~al.(2023)Groeneveld, Awadalla, Beltagy, Bhagia, Magnusson, Peng, Tafjord, Walsh, Richardson, and Dodge}]{groeneveld2023catwalkunifiedlanguagemodel}
Dirk Groeneveld, Anas Awadalla, Iz~Beltagy, Akshita Bhagia, Ian Magnusson, Hao Peng, Oyvind Tafjord, Pete Walsh, Kyle Richardson, and Jesse Dodge. 2023.
\newblock \href {http://arxiv.org/abs/2312.10253} {Catwalk: A Unified Language Model Evaluation Framework for Many Datasets}.

\bibitem[{Groeneveld et~al.(2024)Groeneveld, Beltagy, Walsh, Bhagia, Kinney, Tafjord, Jha, Ivison, Magnusson, Wang, Arora, Atkinson, Authur, Chandu, Cohan, Dumas, Elazar, Gu, Hessel, Khot, Merrill, Morrison, Muennighoff, Naik, Nam, Peters, Pyatkin, Ravichander, Schwenk, Shah, Smith, Strubell, Subramani, Wortsman, Dasigi, Lambert, Richardson, Zettlemoyer, Dodge, Lo, Soldaini, Smith, and Hajishirzi}]{groeneveld2024olmo}
Dirk Groeneveld, Iz~Beltagy, Pete Walsh, Akshita Bhagia, Rodney Kinney, Oyvind Tafjord, Ananya~Harsh Jha, Hamish Ivison, Ian Magnusson, Yizhong Wang, Shane Arora, David Atkinson, Russell Authur, Khyathi~Raghavi Chandu, Arman Cohan, Jennifer Dumas, Yanai Elazar, Yuling Gu, Jack Hessel, Tushar Khot, William Merrill, Jacob Morrison, Niklas Muennighoff, Aakanksha Naik, Crystal Nam, Matthew~E. Peters, Valentina Pyatkin, Abhilasha Ravichander, Dustin Schwenk, Saurabh Shah, Will Smith, Emma Strubell, Nishant Subramani, Mitchell Wortsman, Pradeep Dasigi, Nathan Lambert, Kyle Richardson, Luke Zettlemoyer, Jesse Dodge, Kyle Lo, Luca Soldaini, Noah~A. Smith, and Hannaneh Hajishirzi. 2024.
\newblock \href {http://arxiv.org/abs/2402.00838} {OLMo: Accelerating the Science of Language Models}.

\bibitem[{Gross et~al.(2017)Gross, Ranzato, and Szlam}]{gross2017hardmixturesexpertslarge}
Sam Gross, Marc'Aurelio Ranzato, and Arthur Szlam. 2017.
\newblock \href {http://arxiv.org/abs/1704.06363} {Hard Mixtures of Experts for Large Scale Weakly Supervised Vision}.

\bibitem[{Gu et~al.(2024)Gu, Tafjord, Kuehl, Haddad, Dodge, and Hajishirzi}]{gu2024olmesstandardlanguagemodel}
Yuling Gu, Oyvind Tafjord, Bailey Kuehl, Dany Haddad, Jesse Dodge, and Hannaneh Hajishirzi. 2024.
\newblock \href {http://arxiv.org/abs/2406.08446} {OLMES: A Standard for Language Model Evaluations}.

\bibitem[{Gunasekar et~al.(2023)Gunasekar, Zhang, Aneja, Mendes, Giorno, Gopi, Javaheripi, Kauffmann, de~Rosa, Saarikivi, Salim, Shah, Behl, Wang, Bubeck, Eldan, Kalai, Lee, and Li}]{gunasekar2023textbooksneed}
Suriya Gunasekar, Yi~Zhang, Jyoti Aneja, Caio César~Teodoro Mendes, Allie~Del Giorno, Sivakanth Gopi, Mojan Javaheripi, Piero Kauffmann, Gustavo de~Rosa, Olli Saarikivi, Adil Salim, Shital Shah, Harkirat~Singh Behl, Xin Wang, Sébastien Bubeck, Ronen Eldan, Adam~Tauman Kalai, Yin~Tat Lee, and Yuanzhi Li. 2023.
\newblock \href {http://arxiv.org/abs/2306.11644} {Textbooks Are All You Need}.

\bibitem[{He(2024)}]{he2024mixturemillionexperts}
Xu~Owen He. 2024.
\newblock \href {http://arxiv.org/abs/2407.04153} {Mixture of A Million Experts}.

\bibitem[{Hendrycks et~al.(2021{\natexlab{a}})Hendrycks, Burns, Basart, Zou, Mazeika, Song, and Steinhardt}]{hendrycks2021measuringmassivemultitasklanguage}
Dan Hendrycks, Collin Burns, Steven Basart, Andy Zou, Mantas Mazeika, Dawn Song, and Jacob Steinhardt. 2021{\natexlab{a}}.
\newblock \href {http://arxiv.org/abs/2009.03300} {Measuring Massive Multitask Language Understanding}.

\bibitem[{Hendrycks et~al.(2021{\natexlab{b}})Hendrycks, Burns, Kadavath, Arora, Basart, Tang, Song, and Steinhardt}]{hendrycks2021measuringmathematicalproblemsolving}
Dan Hendrycks, Collin Burns, Saurav Kadavath, Akul Arora, Steven Basart, Eric Tang, Dawn Song, and Jacob Steinhardt. 2021{\natexlab{b}}.
\newblock \href {http://arxiv.org/abs/2103.03874} {Measuring Mathematical Problem Solving With the MATH Dataset}.

\bibitem[{Hoffmann et~al.(2022)Hoffmann, Borgeaud, Mensch, Buchatskaya, Cai, Rutherford, de~Las~Casas, Hendricks, Welbl, Clark, Hennigan, Noland, Millican, van~den Driessche, Damoc, Guy, Osindero, Simonyan, Elsen, Rae, Vinyals, and Sifre}]{hoffmann2022trainingcomputeoptimallargelanguage}
Jordan Hoffmann, Sebastian Borgeaud, Arthur Mensch, Elena Buchatskaya, Trevor Cai, Eliza Rutherford, Diego de~Las~Casas, Lisa~Anne Hendricks, Johannes Welbl, Aidan Clark, Tom Hennigan, Eric Noland, Katie Millican, George van~den Driessche, Bogdan Damoc, Aurelia Guy, Simon Osindero, Karen Simonyan, Erich Elsen, Jack~W. Rae, Oriol Vinyals, and Laurent Sifre. 2022.
\newblock \href {http://arxiv.org/abs/2203.15556} {Training Compute-Optimal Large Language Models}.

\bibitem[{Hong et~al.(2024)Hong, Lee, and Thorne}]{hong2024orpomonolithicpreferenceoptimization}
Jiwoo Hong, Noah Lee, and James Thorne. 2024.
\newblock \href {http://arxiv.org/abs/2403.07691} {ORPO: Monolithic Preference Optimization without Reference Model}.

\bibitem[{Hu et~al.(2024)Hu, Tu, Han, He, Cui, Long, Zheng, Fang, Huang, Zhao, Zhang, Thai, Zhang, Wang, Yao, Zhao, Zhou, Cai, Zhai, Ding, Jia, Zeng, Li, Liu, and Sun}]{hu2024minicpm}
Shengding Hu, Yuge Tu, Xu~Han, Chaoqun He, Ganqu Cui, Xiang Long, Zhi Zheng, Yewei Fang, Yuxiang Huang, Weilin Zhao, Xinrong Zhang, Zheng~Leng Thai, Kaihuo Zhang, Chongyi Wang, Yuan Yao, Chenyang Zhao, Jie Zhou, Jie Cai, Zhongwu Zhai, Ning Ding, Chao Jia, Guoyang Zeng, Dahai Li, Zhiyuan Liu, and Maosong Sun. 2024.
\newblock \href {http://arxiv.org/abs/2404.06395} {MiniCPM: Unveiling the Potential of Small Language Models with Scalable Training Strategies}.

\bibitem[{Huang et~al.(2018)Huang, Vaswani, Uszkoreit, Shazeer, Simon, Hawthorne, Dai, Hoffman, Dinculescu, and Eck}]{huang2018musictransformer}
Cheng-Zhi~Anna Huang, Ashish Vaswani, Jakob Uszkoreit, Noam Shazeer, Ian Simon, Curtis Hawthorne, Andrew~M. Dai, Matthew~D. Hoffman, Monica Dinculescu, and Douglas Eck. 2018.
\newblock \href {http://arxiv.org/abs/1809.04281} {Music Transformer}.

\bibitem[{Ivison et~al.(2023)Ivison, Wang, Pyatkin, Lambert, Peters, Dasigi, Jang, Wadden, Smith, Beltagy, and Hajishirzi}]{ivison2023camels}
Hamish Ivison, Yizhong Wang, Valentina Pyatkin, Nathan Lambert, Matthew Peters, Pradeep Dasigi, Joel Jang, David Wadden, Noah~A. Smith, Iz~Beltagy, and Hannaneh Hajishirzi. 2023.
\newblock \href {http://arxiv.org/abs/2311.10702} {Camels in a Changing Climate: Enhancing LM Adaptation with Tulu 2}.

\bibitem[{Jaszczur et~al.(2021)Jaszczur, Chowdhery, Mohiuddin, Łukasz Kaiser, Gajewski, Michalewski, and Kanerva}]{jaszczur2021sparsescalingtransformers}
Sebastian Jaszczur, Aakanksha Chowdhery, Afroz Mohiuddin, Łukasz Kaiser, Wojciech Gajewski, Henryk Michalewski, and Jonni Kanerva. 2021.
\newblock \href {http://arxiv.org/abs/2111.12763} {Sparse is Enough in Scaling Transformers}.

\bibitem[{Jiang et~al.(2023)Jiang, Sablayrolles, Mensch, Bamford, Chaplot, de~las Casas, Bressand, Lengyel, Lample, Saulnier, Lavaud, Lachaux, Stock, Scao, Lavril, Wang, Lacroix, and Sayed}]{jiang2023mistral}
Albert~Q. Jiang, Alexandre Sablayrolles, Arthur Mensch, Chris Bamford, Devendra~Singh Chaplot, Diego de~las Casas, Florian Bressand, Gianna Lengyel, Guillaume Lample, Lucile Saulnier, Lélio~Renard Lavaud, Marie-Anne Lachaux, Pierre Stock, Teven~Le Scao, Thibaut Lavril, Thomas Wang, Timothée Lacroix, and William~El Sayed. 2023.
\newblock \href {http://arxiv.org/abs/2310.06825} {Mistral 7B}.

\bibitem[{Jiang et~al.(2024)Jiang, Sablayrolles, Roux, Mensch, Savary, Bamford, Chaplot, de~las Casas, Hanna, Bressand, Lengyel, Bour, Lample, Lavaud, Saulnier, Lachaux, Stock, Subramanian, Yang, Antoniak, Scao, Gervet, Lavril, Wang, Lacroix, and Sayed}]{jiang2024mixtral}
Albert~Q. Jiang, Alexandre Sablayrolles, Antoine Roux, Arthur Mensch, Blanche Savary, Chris Bamford, Devendra~Singh Chaplot, Diego de~las Casas, Emma~Bou Hanna, Florian Bressand, Gianna Lengyel, Guillaume Bour, Guillaume Lample, Lélio~Renard Lavaud, Lucile Saulnier, Marie-Anne Lachaux, Pierre Stock, Sandeep Subramanian, Sophia Yang, Szymon Antoniak, Teven~Le Scao, Théophile Gervet, Thibaut Lavril, Thomas Wang, Timothée Lacroix, and William~El Sayed. 2024.
\newblock \href {http://arxiv.org/abs/2401.04088} {Mixtral of Experts}.

\bibitem[{Kaplan et~al.(2020)Kaplan, McCandlish, Henighan, Brown, Chess, Child, Gray, Radford, Wu, and Amodei}]{kaplan2020scalinglawsneurallanguage}
Jared Kaplan, Sam McCandlish, Tom Henighan, Tom~B. Brown, Benjamin Chess, Rewon Child, Scott Gray, Alec Radford, Jeffrey Wu, and Dario Amodei. 2020.
\newblock \href {http://arxiv.org/abs/2001.08361} {Scaling Laws for Neural Language Models}.

\bibitem[{Karpathy(2024)}]{llmsize}
Andrej Karpathy. 2024.
\newblock \href {https://x.com/karpathy/status/1814038096218083497} {LLM model size competition is intensifying… backwards!}

\bibitem[{Kiela et~al.(2021)Kiela, Firooz, Mohan, Goswami, Singh, Fitzpatrick, Bull, Lipstein, Nelli, Zhu et~al.}]{kiela2021hateful}
Douwe Kiela, Hamed Firooz, Aravind Mohan, Vedanuj Goswami, Amanpreet Singh, Casey~A Fitzpatrick, Peter Bull, Greg Lipstein, Tony Nelli, Ron Zhu, et~al. 2021.
\newblock \href {https://proceedings.mlr.press/v133/kiela21a.html} {The hateful memes challenge: Competition report}.

\bibitem[{Kingma and Ba(2017)}]{kingma2017adam}
Diederik~P. Kingma and Jimmy Ba. 2017.
\newblock \href {http://arxiv.org/abs/1412.6980} {Adam: A Method for Stochastic Optimization}.

\bibitem[{Kocetkov et~al.(2022)Kocetkov, Li, Allal, Li, Mou, Ferrandis, Jernite, Mitchell, Hughes, Wolf, Bahdanau, von Werra, and de~Vries}]{kocetkov2022stack3tbpermissively}
Denis Kocetkov, Raymond Li, Loubna~Ben Allal, Jia Li, Chenghao Mou, Carlos~Muñoz Ferrandis, Yacine Jernite, Margaret Mitchell, Sean Hughes, Thomas Wolf, Dzmitry Bahdanau, Leandro von Werra, and Harm de~Vries. 2022.
\newblock \href {http://arxiv.org/abs/2211.15533} {The Stack: 3 TB of permissively licensed source code}.

\bibitem[{Komatsuzaki et~al.(2023)Komatsuzaki, Puigcerver, Lee-Thorp, Ruiz, Mustafa, Ainslie, Tay, Dehghani, and Houlsby}]{komatsuzaki2023sparse}
Aran Komatsuzaki, Joan Puigcerver, James Lee-Thorp, Carlos~Riquelme Ruiz, Basil Mustafa, Joshua Ainslie, Yi~Tay, Mostafa Dehghani, and Neil Houlsby. 2023.
\newblock \href {http://arxiv.org/abs/2212.05055} {Sparse Upcycling: Training Mixture-of-Experts from Dense Checkpoints}.

\bibitem[{Krajewski et~al.(2024)Krajewski, Ludziejewski, Adamczewski, Pióro, Krutul, Antoniak, Ciebiera, Król, Odrzygóźdź, Sankowski, Cygan, and Jaszczur}]{krajewski2024scaling}
Jakub Krajewski, Jan Ludziejewski, Kamil Adamczewski, Maciej Pióro, Michał Krutul, Szymon Antoniak, Kamil Ciebiera, Krystian Król, Tomasz Odrzygóźdź, Piotr Sankowski, Marek Cygan, and Sebastian Jaszczur. 2024.
\newblock \href {http://arxiv.org/abs/2402.07871} {Scaling Laws for Fine-Grained Mixture of Experts}.

\bibitem[{Lambert et~al.(2025)Lambert, Morrison, Pyatkin, Huang, Ivison, Brahman, Miranda, Liu, Dziri, Lyu, Gu, Malik, Graf, Hwang, Yang, Bras, Tafjord, Wilhelm, Soldaini, Smith, Wang, Dasigi, and Hajishirzi}]{lambert2025tulu3pushingfrontiers}
Nathan Lambert, Jacob Morrison, Valentina Pyatkin, Shengyi Huang, Hamish Ivison, Faeze Brahman, Lester James~V. Miranda, Alisa Liu, Nouha Dziri, Shane Lyu, Yuling Gu, Saumya Malik, Victoria Graf, Jena~D. Hwang, Jiangjiang Yang, Ronan~Le Bras, Oyvind Tafjord, Chris Wilhelm, Luca Soldaini, Noah~A. Smith, Yizhong Wang, Pradeep Dasigi, and Hannaneh Hajishirzi. 2025.
\newblock \href {http://arxiv.org/abs/2411.15124} {{Tulu 3: Pushing Frontiers in Open Language Model Post-Training}}.

\bibitem[{Lepikhin et~al.(2020)Lepikhin, Lee, Xu, Chen, Firat, Huang, Krikun, Shazeer, and Chen}]{lepikhin2020gshardscalinggiantmodels}
Dmitry Lepikhin, HyoukJoong Lee, Yuanzhong Xu, Dehao Chen, Orhan Firat, Yanping Huang, Maxim Krikun, Noam Shazeer, and Zhifeng Chen. 2020.
\newblock \href {http://arxiv.org/abs/2006.16668} {GShard: Scaling Giant Models with Conditional Computation and Automatic Sharding}.

\bibitem[{Lewis et~al.(2021)Lewis, Bhosale, Dettmers, Goyal, and Zettlemoyer}]{lewis2021baselayerssimplifyingtraining}
Mike Lewis, Shruti Bhosale, Tim Dettmers, Naman Goyal, and Luke Zettlemoyer. 2021.
\newblock \href {http://arxiv.org/abs/2103.16716} {BASE Layers: Simplifying Training of Large, Sparse Models}.

\bibitem[{Li et~al.(2024{\natexlab{a}})Li, Fang, Smyrnis, Ivgi, Jordan, Gadre, Bansal, Guha, Keh, Arora, Garg, Xin, Muennighoff, Heckel, Mercat, Chen, Gururangan, Wortsman, Albalak, Bitton, Nezhurina, Abbas, Hsieh, Ghosh, Gardner, Kilian, Zhang, Shao, Pratt, Sanyal, Ilharco, Daras, Marathe, Gokaslan, Zhang, Chandu, Nguyen, Vasiljevic, Kakade, Song, Sanghavi, Faghri, Oh, Zettlemoyer, Lo, El-Nouby, Pouransari, Toshev, Wang, Groeneveld, Soldaini, Koh, Jitsev, Kollar, Dimakis, Carmon, Dave, Schmidt, and Shankar}]{li2024datacomplm}
Jeffrey Li, Alex Fang, Georgios Smyrnis, Maor Ivgi, Matt Jordan, Samir Gadre, Hritik Bansal, Etash Guha, Sedrick Keh, Kushal Arora, Saurabh Garg, Rui Xin, Niklas Muennighoff, Reinhard Heckel, Jean Mercat, Mayee Chen, Suchin Gururangan, Mitchell Wortsman, Alon Albalak, Yonatan Bitton, Marianna Nezhurina, Amro Abbas, Cheng-Yu Hsieh, Dhruba Ghosh, Josh Gardner, Maciej Kilian, Hanlin Zhang, Rulin Shao, Sarah Pratt, Sunny Sanyal, Gabriel Ilharco, Giannis Daras, Kalyani Marathe, Aaron Gokaslan, Jieyu Zhang, Khyathi Chandu, Thao Nguyen, Igor Vasiljevic, Sham Kakade, Shuran Song, Sujay Sanghavi, Fartash Faghri, Sewoong Oh, Luke Zettlemoyer, Kyle Lo, Alaaeldin El-Nouby, Hadi Pouransari, Alexander Toshev, Stephanie Wang, Dirk Groeneveld, Luca Soldaini, Pang~Wei Koh, Jenia Jitsev, Thomas Kollar, Alexandros~G. Dimakis, Yair Carmon, Achal Dave, Ludwig Schmidt, and Vaishaal Shankar. 2024{\natexlab{a}}.
\newblock \href {http://arxiv.org/abs/2406.11794} {DataComp-LM: In search of the next generation of training sets for language models}.

\bibitem[{Li et~al.(2022)Li, Gururangan, Dettmers, Lewis, Althoff, Smith, and Zettlemoyer}]{li2022branchtrainmergeembarrassinglyparalleltraining}
Margaret Li, Suchin Gururangan, Tim Dettmers, Mike Lewis, Tim Althoff, Noah~A. Smith, and Luke Zettlemoyer. 2022.
\newblock \href {http://arxiv.org/abs/2208.03306} {Branch-Train-Merge: Embarrassingly Parallel Training of Expert Language Models}.

\bibitem[{Li et~al.(2023{\natexlab{a}})Li, Allal, Zi, Muennighoff, Kocetkov, Mou, Marone, Akiki, Li, Chim et~al.}]{li2023starcoder}
Raymond Li, Loubna~Ben Allal, Yangtian Zi, Niklas Muennighoff, Denis Kocetkov, Chenghao Mou, Marc Marone, Christopher Akiki, Jia Li, Jenny Chim, et~al. 2023{\natexlab{a}}.
\newblock \href {http://arxiv.org/abs/2305.06161} {StarCoder: may the source be with you!}

\bibitem[{Li et~al.(2023{\natexlab{b}})Li, Zhang, Dubois, Taori, Gulrajani, Guestrin, Liang, and Hashimoto}]{alpaca_eval}
Xuechen Li, Tianyi Zhang, Yann Dubois, Rohan Taori, Ishaan Gulrajani, Carlos Guestrin, Percy Liang, and Tatsunori~B. Hashimoto. 2023{\natexlab{b}}.
\newblock \href {https://github.com/tatsu-lab/alpaca_eval} {AlpacaEval: An Automatic Evaluator of Instruction-following Models}.

\bibitem[{Li et~al.(2023{\natexlab{c}})Li, Bubeck, Eldan, Giorno, Gunasekar, and Lee}]{li2023textbooksneediiphi15}
Yuanzhi Li, Sébastien Bubeck, Ronen Eldan, Allie~Del Giorno, Suriya Gunasekar, and Yin~Tat Lee. 2023{\natexlab{c}}.
\newblock \href {http://arxiv.org/abs/2309.05463} {Textbooks Are All You Need II: phi-1.5 technical report}.

\bibitem[{Li et~al.(2024{\natexlab{b}})Li, Jiang, Hu, Wang, Zhong, Luo, Ma, and Zhang}]{li2024unimoescalingunifiedmultimodal}
Yunxin Li, Shenyuan Jiang, Baotian Hu, Longyue Wang, Wanqi Zhong, Wenhan Luo, Lin Ma, and Min Zhang. 2024{\natexlab{b}}.
\newblock \href {http://arxiv.org/abs/2405.11273} {Uni-MoE: Scaling Unified Multimodal LLMs with Mixture of Experts}.

\bibitem[{Liang et~al.(2023)Liang, Bommasani, Lee, Tsipras, Soylu, Yasunaga, Zhang, Narayanan, Wu, Kumar, Newman, Yuan, Yan, Zhang, Cosgrove, Manning, Ré, Acosta-Navas, Hudson, Zelikman, Durmus, Ladhak, Rong, Ren, Yao, Wang, Santhanam, Orr, Zheng, Yuksekgonul, Suzgun, Kim, Guha, Chatterji, Khattab, Henderson, Huang, Chi, Xie, Santurkar, Ganguli, Hashimoto, Icard, Zhang, Chaudhary, Wang, Li, Mai, Zhang, and Koreeda}]{liang2023holisticevaluationlanguagemodels}
Percy Liang, Rishi Bommasani, Tony Lee, Dimitris Tsipras, Dilara Soylu, Michihiro Yasunaga, Yian Zhang, Deepak Narayanan, Yuhuai Wu, Ananya Kumar, Benjamin Newman, Binhang Yuan, Bobby Yan, Ce~Zhang, Christian Cosgrove, Christopher~D. Manning, Christopher Ré, Diana Acosta-Navas, Drew~A. Hudson, Eric Zelikman, Esin Durmus, Faisal Ladhak, Frieda Rong, Hongyu Ren, Huaxiu Yao, Jue Wang, Keshav Santhanam, Laurel Orr, Lucia Zheng, Mert Yuksekgonul, Mirac Suzgun, Nathan Kim, Neel Guha, Niladri Chatterji, Omar Khattab, Peter Henderson, Qian Huang, Ryan Chi, Sang~Michael Xie, Shibani Santurkar, Surya Ganguli, Tatsunori Hashimoto, Thomas Icard, Tianyi Zhang, Vishrav Chaudhary, William Wang, Xuechen Li, Yifan Mai, Yuhui Zhang, and Yuta Koreeda. 2023.
\newblock \href {http://arxiv.org/abs/2211.09110} {Holistic Evaluation of Language Models}.

\bibitem[{Lieber et~al.(2024)Lieber, Lenz, Bata, Cohen, Osin, Dalmedigos, Safahi, Meirom, Belinkov, Shalev-Shwartz, Abend, Alon, Asida, Bergman, Glozman, Gokhman, Manevich, Ratner, Rozen, Shwartz, Zusman, and Shoham}]{lieber2024jambahybridtransformermambalanguage}
Opher Lieber, Barak Lenz, Hofit Bata, Gal Cohen, Jhonathan Osin, Itay Dalmedigos, Erez Safahi, Shaked Meirom, Yonatan Belinkov, Shai Shalev-Shwartz, Omri Abend, Raz Alon, Tomer Asida, Amir Bergman, Roman Glozman, Michael Gokhman, Avashalom Manevich, Nir Ratner, Noam Rozen, Erez Shwartz, Mor Zusman, and Yoav Shoham. 2024.
\newblock \href {http://arxiv.org/abs/2403.19887} {Jamba: A Hybrid Transformer-Mamba Language Model}.

\bibitem[{Lin et~al.(2024{\natexlab{a}})Lin, Tang, Ye, Cui, Zhu, Jin, Huang, Zhang, Pang, Ning, and Yuan}]{lin2024moellavamixtureexpertslarge}
Bin Lin, Zhenyu Tang, Yang Ye, Jiaxi Cui, Bin Zhu, Peng Jin, Jinfa Huang, Junwu Zhang, Yatian Pang, Munan Ning, and Li~Yuan. 2024{\natexlab{a}}.
\newblock \href {http://arxiv.org/abs/2401.15947} {MoE-LLaVA: Mixture of Experts for Large Vision-Language Models}.

\bibitem[{Lin et~al.(2022)Lin, Hilton, and Evans}]{lin2022truthfulqa}
Stephanie Lin, Jacob Hilton, and Owain Evans. 2022.
\newblock \href {http://arxiv.org/abs/2109.07958} {TruthfulQA: Measuring How Models Mimic Human Falsehoods}.

\bibitem[{Lin et~al.(2024{\natexlab{b}})Lin, Shrivastava, Luo, Iyer, Lewis, Gosh, Zettlemoyer, and Aghajanyan}]{lin2024momaefficientearlyfusionpretraining}
Xi~Victoria Lin, Akshat Shrivastava, Liang Luo, Srinivasan Iyer, Mike Lewis, Gargi Gosh, Luke Zettlemoyer, and Armen Aghajanyan. 2024{\natexlab{b}}.
\newblock \href {http://arxiv.org/abs/2407.21770} {MoMa: Efficient Early-Fusion Pre-training with Mixture of Modality-Aware Experts}.

\bibitem[{Liu et~al.(2024{\natexlab{a}})Liu, Zheng, Muennighoff, Zeng, Dou, Pang, Jiang, and Lin}]{liu2024regmixdatamixtureregression}
Qian Liu, Xiaosen Zheng, Niklas Muennighoff, Guangtao Zeng, Longxu Dou, Tianyu Pang, Jing Jiang, and Min Lin. 2024{\natexlab{a}}.
\newblock \href {http://arxiv.org/abs/2407.01492} {RegMix: Data Mixture as Regression for Language Model Pre-training}.

\bibitem[{Liu et~al.(2024{\natexlab{b}})Liu, Blondel, Riquelme, and Puigcerver}]{liu2024routersvisionmixtureexperts}
Tianlin Liu, Mathieu Blondel, Carlos Riquelme, and Joan Puigcerver. 2024{\natexlab{b}}.
\newblock \href {http://arxiv.org/abs/2401.15969} {Routers in Vision Mixture of Experts: An Empirical Study}.

\bibitem[{Liu et~al.(2023)Liu, Qiao, Neiswanger, Wang, Tan, Tao, Li, Wang, Sun, Pangarkar, Fan, Gu, Miller, Zhuang, He, Li, Koto, Tang, Ranjan, Shen, Ren, Iriondo, Mu, Hu, Schulze, Nakov, Baldwin, and Xing}]{liu2023llm360}
Zhengzhong Liu, Aurick Qiao, Willie Neiswanger, Hongyi Wang, Bowen Tan, Tianhua Tao, Junbo Li, Yuqi Wang, Suqi Sun, Omkar Pangarkar, Richard Fan, Yi~Gu, Victor Miller, Yonghao Zhuang, Guowei He, Haonan Li, Fajri Koto, Liping Tang, Nikhil Ranjan, Zhiqiang Shen, Xuguang Ren, Roberto Iriondo, Cun Mu, Zhiting Hu, Mark Schulze, Preslav Nakov, Tim Baldwin, and Eric~P. Xing. 2023.
\newblock \href {http://arxiv.org/abs/2312.06550} {LLM360: Towards Fully Transparent Open-Source LLMs}.

\bibitem[{Longpre et~al.(2023{\natexlab{a}})Longpre, Hou, Vu, Webson, Chung, Tay, Zhou, Le, Zoph, Wei, and Roberts}]{longpre2023flancollectiondesigningdata}
Shayne Longpre, Le~Hou, Tu~Vu, Albert Webson, Hyung~Won Chung, Yi~Tay, Denny Zhou, Quoc~V. Le, Barret Zoph, Jason Wei, and Adam Roberts. 2023{\natexlab{a}}.
\newblock \href {http://arxiv.org/abs/2301.13688} {The Flan Collection: Designing Data and Methods for Effective Instruction Tuning}.

\bibitem[{Longpre et~al.(2023{\natexlab{b}})Longpre, Mahari, Chen, Obeng-Marnu, Sileo, Brannon, Muennighoff, Khazam, Kabbara, Perisetla, Wu, Shippole, Bollacker, Wu, Villa, Pentland, and Hooker}]{longpre2023dataprovenanceinitiativelarge}
Shayne Longpre, Robert Mahari, Anthony Chen, Naana Obeng-Marnu, Damien Sileo, William Brannon, Niklas Muennighoff, Nathan Khazam, Jad Kabbara, Kartik Perisetla, Xinyi Wu, Enrico Shippole, Kurt Bollacker, Tongshuang Wu, Luis Villa, Sandy Pentland, and Sara Hooker. 2023{\natexlab{b}}.
\newblock \href {http://arxiv.org/abs/2310.16787} {The Data Provenance Initiative: A Large Scale Audit of Dataset Licensing \& Attribution in AI}.

\bibitem[{Longpre et~al.(2024)Longpre, Mahari, Lee, Lund, Oderinwale, Brannon, Saxena, Obeng-Marnu, South, Hunter, Klyman, Klamm, Schoelkopf, Singh, Cherep, Anis, Dinh, Chitongo, Yin, Sileo, Mataciunas, Misra, Alghamdi, Shippole, Zhang, Materzynska, Qian, Tiwary, Miranda, Dey, Liang, Hamdy, Muennighoff, Ye, Kim, Mohanty, Gupta, Sharma, Chien, Zhou, Li, Xiong, Villa, Biderman, Li, Ippolito, Hooker, Kabbara, and Pentland}]{longpre2024consentcrisisrapiddecline}
Shayne Longpre, Robert Mahari, Ariel Lee, Campbell Lund, Hamidah Oderinwale, William Brannon, Nayan Saxena, Naana Obeng-Marnu, Tobin South, Cole Hunter, Kevin Klyman, Christopher Klamm, Hailey Schoelkopf, Nikhil Singh, Manuel Cherep, Ahmad Anis, An~Dinh, Caroline Chitongo, Da~Yin, Damien Sileo, Deividas Mataciunas, Diganta Misra, Emad Alghamdi, Enrico Shippole, Jianguo Zhang, Joanna Materzynska, Kun Qian, Kush Tiwary, Lester Miranda, Manan Dey, Minnie Liang, Mohammed Hamdy, Niklas Muennighoff, Seonghyeon Ye, Seungone Kim, Shrestha Mohanty, Vipul Gupta, Vivek Sharma, Vu~Minh Chien, Xuhui Zhou, Yizhi Li, Caiming Xiong, Luis Villa, Stella Biderman, Hanlin Li, Daphne Ippolito, Sara Hooker, Jad Kabbara, and Sandy Pentland. 2024.
\newblock \href {http://arxiv.org/abs/2407.14933} {Consent in Crisis: The Rapid Decline of the AI Data Commons}.

\bibitem[{Loshchilov and Hutter(2019)}]{loshchilov2019decoupled}
Ilya Loshchilov and Frank Hutter. 2019.
\newblock \href {http://arxiv.org/abs/1711.05101} {Decoupled Weight Decay Regularization}.

\bibitem[{Lovenia et~al.(2024)Lovenia, Mahendra, Akbar, Miranda, Santoso, Aco, Fadhilah, Mansurov, Imperial, Kampman, Moniz, Habibi, Hudi, Montalan, Ignatius, Lopo, Nixon, Karlsson, Jaya, Diandaru, Gao, Amadeus, Wang, Cruz, Whitehouse, Parmonangan, Khelli, Zhang, Susanto, Ryanda, Hermawan, Velasco, Kautsar, Hendria, Moslem, Flynn, Adilazuarda, Li, Lee, Damanhuri, Sun, Qorib, Djanibekov, Leong, Do, Muennighoff, Pansuwan, Putra, Xu, Tai, Purwarianti, Ruder, Tjhi, Limkonchotiwat, Aji, Keh, Winata, Zhang, Koto, Yong, and Cahyawijaya}]{lovenia2024seacrowdmultilingualmultimodaldata}
Holy Lovenia, Rahmad Mahendra, Salsabil~Maulana Akbar, Lester James~V. Miranda, Jennifer Santoso, Elyanah Aco, Akhdan Fadhilah, Jonibek Mansurov, Joseph~Marvin Imperial, Onno~P. Kampman, Joel Ruben~Antony Moniz, Muhammad Ravi~Shulthan Habibi, Frederikus Hudi, Railey Montalan, Ryan Ignatius, Joanito~Agili Lopo, William Nixon, Börje~F. Karlsson, James Jaya, Ryandito Diandaru, Yuze Gao, Patrick Amadeus, Bin Wang, Jan Christian~Blaise Cruz, Chenxi Whitehouse, Ivan~Halim Parmonangan, Maria Khelli, Wenyu Zhang, Lucky Susanto, Reynard~Adha Ryanda, Sonny~Lazuardi Hermawan, Dan~John Velasco, Muhammad Dehan~Al Kautsar, Willy~Fitra Hendria, Yasmin Moslem, Noah Flynn, Muhammad~Farid Adilazuarda, Haochen Li, Johanes Lee, R.~Damanhuri, Shuo Sun, Muhammad~Reza Qorib, Amirbek Djanibekov, Wei~Qi Leong, Quyet~V. Do, Niklas Muennighoff, Tanrada Pansuwan, Ilham~Firdausi Putra, Yan Xu, Ngee~Chia Tai, Ayu Purwarianti, Sebastian Ruder, William Tjhi, Peerat Limkonchotiwat, Alham~Fikri Aji, Sedrick Keh, Genta~Indra Winata, Ruochen
  Zhang, Fajri Koto, Zheng-Xin Yong, and Samuel Cahyawijaya. 2024.
\newblock \href {http://arxiv.org/abs/2406.10118} {SEACrowd: A Multilingual Multimodal Data Hub and Benchmark Suite for Southeast Asian Languages}.

\bibitem[{Lozhkov et~al.(2024)Lozhkov, Li, Allal, Cassano, Lamy-Poirier, Tazi, Tang, Pykhtar, Liu, Wei, Liu, Tian, Kocetkov, Zucker, Belkada, Wang, Liu, Abulkhanov, Paul, Li, Li, Risdal, Li, Zhu, Zhuo, Zheltonozhskii, Dade, Yu, Krauß, Jain, Su, He, Dey, Abati, Chai, Muennighoff, Tang, Oblokulov, Akiki, Marone, Mou, Mishra, Gu, Hui, Dao, Zebaze, Dehaene, Patry, Xu, McAuley, Hu, Scholak, Paquet, Robinson, Anderson, Chapados, Patwary, Tajbakhsh, Jernite, Ferrandis, Zhang, Hughes, Wolf, Guha, von Werra, and de~Vries}]{lozhkov2024starcoder2stackv2}
Anton Lozhkov, Raymond Li, Loubna~Ben Allal, Federico Cassano, Joel Lamy-Poirier, Nouamane Tazi, Ao~Tang, Dmytro Pykhtar, Jiawei Liu, Yuxiang Wei, Tianyang Liu, Max Tian, Denis Kocetkov, Arthur Zucker, Younes Belkada, Zijian Wang, Qian Liu, Dmitry Abulkhanov, Indraneil Paul, Zhuang Li, Wen-Ding Li, Megan Risdal, Jia Li, Jian Zhu, Terry~Yue Zhuo, Evgenii Zheltonozhskii, Nii Osae~Osae Dade, Wenhao Yu, Lucas Krauß, Naman Jain, Yixuan Su, Xuanli He, Manan Dey, Edoardo Abati, Yekun Chai, Niklas Muennighoff, Xiangru Tang, Muhtasham Oblokulov, Christopher Akiki, Marc Marone, Chenghao Mou, Mayank Mishra, Alex Gu, Binyuan Hui, Tri Dao, Armel Zebaze, Olivier Dehaene, Nicolas Patry, Canwen Xu, Julian McAuley, Han Hu, Torsten Scholak, Sebastien Paquet, Jennifer Robinson, Carolyn~Jane Anderson, Nicolas Chapados, Mostofa Patwary, Nima Tajbakhsh, Yacine Jernite, Carlos~Muñoz Ferrandis, Lingming Zhang, Sean Hughes, Thomas Wolf, Arjun Guha, Leandro von Werra, and Harm de~Vries. 2024.
\newblock \href {http://arxiv.org/abs/2402.19173} {StarCoder 2 and The Stack v2: The Next Generation}.

\bibitem[{Luukkonen et~al.(2023)Luukkonen, Komulainen, Luoma, Eskelinen, Kanerva, Kupari, Ginter, Laippala, Muennighoff, Piktus, Wang, Tazi, Scao, Wolf, Suominen, Sairanen, Merioksa, Heinonen, Vahtola, Antao, and Pyysalo}]{luukkonen2023fingpt}
Risto Luukkonen, Ville Komulainen, Jouni Luoma, Anni Eskelinen, Jenna Kanerva, Hanna-Mari Kupari, Filip Ginter, Veronika Laippala, Niklas Muennighoff, Aleksandra Piktus, Thomas Wang, Nouamane Tazi, Teven~Le Scao, Thomas Wolf, Osma Suominen, Samuli Sairanen, Mikko Merioksa, Jyrki Heinonen, Aija Vahtola, Samuel Antao, and Sampo Pyysalo. 2023.
\newblock \href {http://arxiv.org/abs/2311.05640} {FinGPT: Large Generative Models for a Small Language}.

\bibitem[{Magnusson et~al.(2023)Magnusson, Bhagia, Hofmann, Soldaini, Jha, Tafjord, Schwenk, Walsh, Elazar, Lo, Groeneveld, Beltagy, Hajishirzi, Smith, Richardson, and Dodge}]{magnusson2023palomabenchmarkevaluatinglanguage}
Ian Magnusson, Akshita Bhagia, Valentin Hofmann, Luca Soldaini, Ananya~Harsh Jha, Oyvind Tafjord, Dustin Schwenk, Evan~Pete Walsh, Yanai Elazar, Kyle Lo, Dirk Groeneveld, Iz~Beltagy, Hannaneh Hajishirzi, Noah~A. Smith, Kyle Richardson, and Jesse Dodge. 2023.
\newblock \href {http://arxiv.org/abs/2312.10523} {Paloma: A Benchmark for Evaluating Language Model Fit}.

\bibitem[{McKinzie et~al.(2024)McKinzie, Gan, Fauconnier, Dodge, Zhang, Dufter, Shah, Du, Peng, Weers, Belyi, Zhang, Singh, Kang, Jain, Hè, Schwarzer, Gunter, Kong, Zhang, Wang, Wang, Du, Lei, Wiseman, Yin, Lee, Wang, Pang, Grasch, Toshev, and Yang}]{mckinzie2024mm1methodsanalysis}
Brandon McKinzie, Zhe Gan, Jean-Philippe Fauconnier, Sam Dodge, Bowen Zhang, Philipp Dufter, Dhruti Shah, Xianzhi Du, Futang Peng, Floris Weers, Anton Belyi, Haotian Zhang, Karanjeet Singh, Doug Kang, Ankur Jain, Hongyu Hè, Max Schwarzer, Tom Gunter, Xiang Kong, Aonan Zhang, Jianyu Wang, Chong Wang, Nan Du, Tao Lei, Sam Wiseman, Guoli Yin, Mark Lee, Zirui Wang, Ruoming Pang, Peter Grasch, Alexander Toshev, and Yinfei Yang. 2024.
\newblock \href {http://arxiv.org/abs/2403.09611} {MM1: Methods, Analysis \& Insights from Multimodal LLM Pre-training}.

\bibitem[{Mehta et~al.(2024)Mehta, Sekhavat, Cao, Horton, Jin, Sun, Mirzadeh, Najibi, Belenko, Zatloukal, and Rastegari}]{mehta2024openelm}
Sachin Mehta, Mohammad~Hossein Sekhavat, Qingqing Cao, Maxwell Horton, Yanzi Jin, Chenfan Sun, Iman Mirzadeh, Mahyar Najibi, Dmitry Belenko, Peter Zatloukal, and Mohammad Rastegari. 2024.
\newblock \href {http://arxiv.org/abs/2404.14619} {OpenELM: An Efficient Language Model Family with Open Training and Inference Framework}.

\bibitem[{Meng et~al.(2024)Meng, Xia, and Chen}]{meng2024simposimplepreferenceoptimization}
Yu~Meng, Mengzhou Xia, and Danqi Chen. 2024.
\newblock \href {http://arxiv.org/abs/2405.14734} {SimPO: Simple Preference Optimization with a Reference-Free Reward}.

\bibitem[{Merity et~al.(2016)Merity, Xiong, Bradbury, and Socher}]{merity2016pointersentinelmixturemodels}
Stephen Merity, Caiming Xiong, James Bradbury, and Richard Socher. 2016.
\newblock \href {http://arxiv.org/abs/1609.07843} {Pointer Sentinel Mixture Models}.

\bibitem[{Micikevicius et~al.(2018)Micikevicius, Narang, Alben, Diamos, Elsen, Garcia, Ginsburg, Houston, Kuchaiev, Venkatesh, and Wu}]{micikevicius2018mixed}
Paulius Micikevicius, Sharan Narang, Jonah Alben, Gregory Diamos, Erich Elsen, David Garcia, Boris Ginsburg, Michael Houston, Oleksii Kuchaiev, Ganesh Venkatesh, and Hao Wu. 2018.
\newblock \href {http://arxiv.org/abs/1710.03740} {Mixed Precision Training}.

\bibitem[{Mihaylov et~al.(2018)Mihaylov, Clark, Khot, and Sabharwal}]{mihaylov2018suitarmorconductelectricity}
Todor Mihaylov, Peter Clark, Tushar Khot, and Ashish Sabharwal. 2018.
\newblock \href {http://arxiv.org/abs/1809.02789} {Can a Suit of Armor Conduct Electricity? A New Dataset for Open Book Question Answering}.

\bibitem[{Mishra et~al.(2022)Mishra, Khashabi, Baral, and Hajishirzi}]{mishra2022crosstask}
Swaroop Mishra, Daniel Khashabi, Chitta Baral, and Hannaneh Hajishirzi. 2022.
\newblock \href {http://arxiv.org/abs/2104.08773} {Cross-Task Generalization via Natural Language Crowdsourcing Instructions}.

\bibitem[{Muennighoff(2020)}]{muennighoff2020vilio}
Niklas Muennighoff. 2020.
\newblock \href {http://arxiv.org/abs/2012.07788} {Vilio: State-of-the-art Visio-Linguistic Models applied to Hateful Memes}.

\bibitem[{Muennighoff et~al.(2023{\natexlab{a}})Muennighoff, Liu, Zebaze, Zheng, Hui, Zhuo, Singh, Tang, von Werra, and Longpre}]{muennighoff2023octopack}
Niklas Muennighoff, Qian Liu, Armel Zebaze, Qinkai Zheng, Binyuan Hui, Terry~Yue Zhuo, Swayam Singh, Xiangru Tang, Leandro von Werra, and Shayne Longpre. 2023{\natexlab{a}}.
\newblock \href {http://arxiv.org/abs/2308.07124} {OctoPack: Instruction Tuning Code Large Language Models}.

\bibitem[{Muennighoff et~al.(2023{\natexlab{b}})Muennighoff, Rush, Barak, Scao, Piktus, Tazi, Pyysalo, Wolf, and Raffel}]{muennighoff2023scaling}
Niklas Muennighoff, Alexander~M. Rush, Boaz Barak, Teven~Le Scao, Aleksandra Piktus, Nouamane Tazi, Sampo Pyysalo, Thomas Wolf, and Colin Raffel. 2023{\natexlab{b}}.
\newblock \href {http://arxiv.org/abs/2305.16264} {Scaling Data-Constrained Language Models}.

\bibitem[{Muennighoff et~al.(2024)Muennighoff, Su, Wang, Yang, Wei, Yu, Singh, and Kiela}]{muennighoff2024generativerepresentationalinstructiontuning}
Niklas Muennighoff, Hongjin Su, Liang Wang, Nan Yang, Furu Wei, Tao Yu, Amanpreet Singh, and Douwe Kiela. 2024.
\newblock \href {http://arxiv.org/abs/2402.09906} {Generative Representational Instruction Tuning}.

\bibitem[{Muennighoff et~al.(2023{\natexlab{c}})Muennighoff, Wang, Sutawika, Roberts, Biderman, Scao, Bari, Shen, Yong, Schoelkopf, Tang, Radev, Aji, Almubarak, Albanie, Alyafeai, Webson, Raff, and Raffel}]{muennighoff2023crosslingual}
Niklas Muennighoff, Thomas Wang, Lintang Sutawika, Adam Roberts, Stella Biderman, Teven~Le Scao, M~Saiful Bari, Sheng Shen, Zheng-Xin Yong, Hailey Schoelkopf, Xiangru Tang, Dragomir Radev, Alham~Fikri Aji, Khalid Almubarak, Samuel Albanie, Zaid Alyafeai, Albert Webson, Edward Raff, and Colin Raffel. 2023{\natexlab{c}}.
\newblock \href {http://arxiv.org/abs/2211.01786} {Crosslingual Generalization through Multitask Finetuning}.

\bibitem[{Muqeeth et~al.(2024)Muqeeth, Liu, and Raffel}]{muqeeth2024softmergingexpertsadaptive}
Mohammed Muqeeth, Haokun Liu, and Colin Raffel. 2024.
\newblock \href {http://arxiv.org/abs/2306.03745} {Soft Merging of Experts with Adaptive Routing}.

\bibitem[{Mustafa et~al.(2022)Mustafa, Riquelme, Puigcerver, Jenatton, and Houlsby}]{mustafa2022multimodalcontrastivelearninglimoe}
Basil Mustafa, Carlos Riquelme, Joan Puigcerver, Rodolphe Jenatton, and Neil Houlsby. 2022.
\newblock \href {http://arxiv.org/abs/2206.02770} {Multimodal Contrastive Learning with LIMoE: the Language-Image Mixture of Experts}.

\bibitem[{Nvidia et~al.(2024)Nvidia, :, Adler, Agarwal, Aithal, Anh, Bhattacharya, Brundyn, Casper, Catanzaro, Clay, Cohen, Das, Dattagupta, Delalleau, Derczynski, Dong, Egert, Evans, Ficek, Fridman, Ghosh, Ginsburg, Gitman, Grzegorzek, Hero, Huang, Jawa, Jennings, Jhunjhunwala, Kamalu, Khan, Kuchaiev, LeGresley, Li, Liu, Liu, Long, Mahabaleshwarkar, Majumdar, Maki, Martinez, de~Melo, Moshkov, Narayanan, Narenthiran, Navarro, Nguyen, Nitski, Noroozi, Nutheti, Parisien, Parmar, Patwary, Pawelec, Ping, Prabhumoye, Roy, Saar, Sabavat, Satheesh, Scowcroft, Sewall, Shamis, Shen, Shoeybi, Sizer, Smelyanskiy, Soares, Sreedhar, Su, Subramanian, Sun, Toshniwal, Wang, Wang, You, Zeng, Zhang, Zhang, Zhang, Zhang, and Zhu}]{nvidia2024nemotron4340btechnicalreport}
Nvidia, :, Bo~Adler, Niket Agarwal, Ashwath Aithal, Dong~H. Anh, Pallab Bhattacharya, Annika Brundyn, Jared Casper, Bryan Catanzaro, Sharon Clay, Jonathan Cohen, Sirshak Das, Ayush Dattagupta, Olivier Delalleau, Leon Derczynski, Yi~Dong, Daniel Egert, Ellie Evans, Aleksander Ficek, Denys Fridman, Shaona Ghosh, Boris Ginsburg, Igor Gitman, Tomasz Grzegorzek, Robert Hero, Jining Huang, Vibhu Jawa, Joseph Jennings, Aastha Jhunjhunwala, John Kamalu, Sadaf Khan, Oleksii Kuchaiev, Patrick LeGresley, Hui Li, Jiwei Liu, Zihan Liu, Eileen Long, Ameya~Sunil Mahabaleshwarkar, Somshubra Majumdar, James Maki, Miguel Martinez, Maer~Rodrigues de~Melo, Ivan Moshkov, Deepak Narayanan, Sean Narenthiran, Jesus Navarro, Phong Nguyen, Osvald Nitski, Vahid Noroozi, Guruprasad Nutheti, Christopher Parisien, Jupinder Parmar, Mostofa Patwary, Krzysztof Pawelec, Wei Ping, Shrimai Prabhumoye, Rajarshi Roy, Trisha Saar, Vasanth Rao~Naik Sabavat, Sanjeev Satheesh, Jane~Polak Scowcroft, Jason Sewall, Pavel Shamis, Gerald Shen, Mohammad
  Shoeybi, Dave Sizer, Misha Smelyanskiy, Felipe Soares, Makesh~Narsimhan Sreedhar, Dan Su, Sandeep Subramanian, Shengyang Sun, Shubham Toshniwal, Hao Wang, Zhilin Wang, Jiaxuan You, Jiaqi Zeng, Jimmy Zhang, Jing Zhang, Vivienne Zhang, Yian Zhang, and Chen Zhu. 2024.
\newblock \href {http://arxiv.org/abs/2406.11704} {Nemotron-4 340B Technical Report}.

\bibitem[{OLMo et~al.(2024)OLMo, Walsh, Soldaini, Groeneveld, Lo, Arora, Bhagia, Gu, Huang, Jordan, Lambert, Schwenk, Tafjord, Anderson, Atkinson, Brahman, Clark, Dasigi, Dziri, Guerquin, Ivison, Koh, Liu, Malik, Merrill, Miranda, Morrison, Murray, Nam, Pyatkin, Rangapur, Schmitz, Skjonsberg, Wadden, Wilhelm, Wilson, Zettlemoyer, Farhadi, Smith, and Hajishirzi}]{OLMo2}
Team OLMo, Pete Walsh, Luca Soldaini, Dirk Groeneveld, Kyle Lo, Shane Arora, Akshita Bhagia, Yuling Gu, Shengyi Huang, Matt Jordan, Nathan Lambert, Dustin Schwenk, Oyvind Tafjord, Taira Anderson, David Atkinson, Faeze Brahman, Christopher Clark, Pradeep Dasigi, Nouha Dziri, Michal Guerquin, Hamish Ivison, Pang~Wei Koh, Jiacheng Liu, Saumya Malik, William Merrill, Lester James~Validad Miranda, Jacob~Daniel Morrison, Tyler~C. Murray, Crystal Nam, Valentina Pyatkin, Aman Rangapur, Michael Schmitz, Sam Skjonsberg, David Wadden, Chris Wilhelm, Michael Wilson, Luke~S. Zettlemoyer, Ali Farhadi, Noah~A. Smith, and Hanna Hajishirzi. 2024.
\newblock \href {https://api.semanticscholar.org/CorpusID:275213098} {{2 OLMo 2 Furious}}.
\newblock \emph{arXiv preprint}.

\bibitem[{OpenAI et~al.(2023)OpenAI, Achiam, Adler, Agarwal, Ahmad, Akkaya, Aleman, Almeida, Altenschmidt, Altman et~al.}]{openai2023gpt4}
OpenAI, Josh Achiam, Steven Adler, Sandhini Agarwal, Lama Ahmad, Ilge Akkaya, Florencia~Leoni Aleman, Diogo Almeida, Janko Altenschmidt, Sam Altman, et~al. 2023.
\newblock \href {http://arxiv.org/abs/2303.08774} {GPT-4 Technical Report}.

\bibitem[{Pan et~al.(2024)Pan, Shen, Liu, Mishra, Zhang, Oliva, Raffel, and Panda}]{pan2024densetrainingsparseinference}
Bowen Pan, Yikang Shen, Haokun Liu, Mayank Mishra, Gaoyuan Zhang, Aude Oliva, Colin Raffel, and Rameswar Panda. 2024.
\newblock \href {http://arxiv.org/abs/2404.05567} {Dense Training, Sparse Inference: Rethinking Training of Mixture-of-Experts Language Models}.

\bibitem[{Parmar et~al.(2024)Parmar, Prabhumoye, Jennings, Patwary, Subramanian, Su, Zhu, Narayanan, Jhunjhunwala, Dattagupta, Jawa, Liu, Mahabaleshwarkar, Nitski, Brundyn, Maki, Martinez, You, Kamalu, LeGresley, Fridman, Casper, Aithal, Kuchaiev, Shoeybi, Cohen, and Catanzaro}]{parmar2024nemotron415btechnicalreport}
Jupinder Parmar, Shrimai Prabhumoye, Joseph Jennings, Mostofa Patwary, Sandeep Subramanian, Dan Su, Chen Zhu, Deepak Narayanan, Aastha Jhunjhunwala, Ayush Dattagupta, Vibhu Jawa, Jiwei Liu, Ameya Mahabaleshwarkar, Osvald Nitski, Annika Brundyn, James Maki, Miguel Martinez, Jiaxuan You, John Kamalu, Patrick LeGresley, Denys Fridman, Jared Casper, Ashwath Aithal, Oleksii Kuchaiev, Mohammad Shoeybi, Jonathan Cohen, and Bryan Catanzaro. 2024.
\newblock \href {http://arxiv.org/abs/2402.16819} {Nemotron-4 15B Technical Report}.

\bibitem[{Paster et~al.(2023)Paster, Santos, Azerbayev, and Ba}]{paster2023openwebmath}
Keiran Paster, Marco~Dos Santos, Zhangir Azerbayev, and Jimmy Ba. 2023.
\newblock \href {http://arxiv.org/abs/2310.06786} {OpenWebMath: An Open Dataset of High-Quality Mathematical Web Text}.

\bibitem[{Penedo et~al.(2023)Penedo, Malartic, Hesslow, Cojocaru, Cappelli, Alobeidli, Pannier, Almazrouei, and Launay}]{penedo2023refinedwebdatasetfalconllm}
Guilherme Penedo, Quentin Malartic, Daniel Hesslow, Ruxandra Cojocaru, Alessandro Cappelli, Hamza Alobeidli, Baptiste Pannier, Ebtesam Almazrouei, and Julien Launay. 2023.
\newblock \href {http://arxiv.org/abs/2306.01116} {The RefinedWeb Dataset for Falcon LLM: Outperforming Curated Corpora with Web Data, and Web Data Only}.

\bibitem[{Peng et~al.(2023)Peng, Alcaide, Anthony, Albalak, Arcadinho, Biderman, Cao, Cheng, Chung, Grella, GV, He, Hou, Lin, Kazienko, Kocon, Kong, Koptyra, Lau, Mantri, Mom, Saito, Song, Tang, Wang, Wind, Wozniak, Zhang, Zhang, Zhao, Zhou, Zhou, Zhu, and Zhu}]{peng2023rwkv}
Bo~Peng, Eric Alcaide, Quentin Anthony, Alon Albalak, Samuel Arcadinho, Stella Biderman, Huanqi Cao, Xin Cheng, Michael Chung, Matteo Grella, Kranthi~Kiran GV, Xuzheng He, Haowen Hou, Jiaju Lin, Przemyslaw Kazienko, Jan Kocon, Jiaming Kong, Bartlomiej Koptyra, Hayden Lau, Krishna Sri~Ipsit Mantri, Ferdinand Mom, Atsushi Saito, Guangyu Song, Xiangru Tang, Bolun Wang, Johan~S. Wind, Stanislaw Wozniak, Ruichong Zhang, Zhenyuan Zhang, Qihang Zhao, Peng Zhou, Qinghua Zhou, Jian Zhu, and Rui-Jie Zhu. 2023.
\newblock \href {http://arxiv.org/abs/2305.13048} {RWKV: Reinventing RNNs for the Transformer Era}.

\bibitem[{Peng et~al.(2024)Peng, Goldstein, Anthony, Albalak, Alcaide, Biderman, Cheah, Du, Ferdinan, Hou, Kazienko, GV, Kocoń, Koptyra, Krishna, au2, Muennighoff, Obeid, Saito, Song, Tu, Woźniak, Zhang, Zhao, Zhao, Zhou, Zhu, and Zhu}]{peng2024eagle}
Bo~Peng, Daniel Goldstein, Quentin Anthony, Alon Albalak, Eric Alcaide, Stella Biderman, Eugene Cheah, Xingjian Du, Teddy Ferdinan, Haowen Hou, Przemysław Kazienko, Kranthi~Kiran GV, Jan Kocoń, Bartłomiej Koptyra, Satyapriya Krishna, Ronald McClelland~Jr. au2, Niklas Muennighoff, Fares Obeid, Atsushi Saito, Guangyu Song, Haoqin Tu, Stanisław Woźniak, Ruichong Zhang, Bingchen Zhao, Qihang Zhao, Peng Zhou, Jian Zhu, and Rui-Jie Zhu. 2024.
\newblock \href {http://arxiv.org/abs/2404.05892} {Eagle and Finch: RWKV with Matrix-Valued States and Dynamic Recurrence}.

\bibitem[{Press and Wolf(2017)}]{press2017usingoutputembeddingimprove}
Ofir Press and Lior Wolf. 2017.
\newblock \href {http://arxiv.org/abs/1608.05859} {Using the Output Embedding to Improve Language Models}.

\bibitem[{Radford et~al.(2022)Radford, Kim, Xu, Brockman, McLeavey, and Sutskever}]{radford2022robustspeechrecognitionlargescale}
Alec Radford, Jong~Wook Kim, Tao Xu, Greg Brockman, Christine McLeavey, and Ilya Sutskever. 2022.
\newblock \href {http://arxiv.org/abs/2212.04356} {Robust Speech Recognition via Large-Scale Weak Supervision}.

\bibitem[{Radford et~al.(2019)Radford, Wu, Child, Luan, Amodei, Sutskever et~al.}]{radford2019language}
Alec Radford, Jeffrey Wu, Rewon Child, David Luan, Dario Amodei, Ilya Sutskever, et~al. 2019.
\newblock \href {https://d4mucfpksywv.cloudfront.net/better-language-models/language_models_are_unsupervised_multitask_learners.pdf} {Language models are unsupervised multitask learners}.

\bibitem[{Rafailov et~al.(2023)Rafailov, Sharma, Mitchell, Ermon, Manning, and Finn}]{rafailov2023direct}
Rafael Rafailov, Archit Sharma, Eric Mitchell, Stefano Ermon, Christopher~D. Manning, and Chelsea Finn. 2023.
\newblock \href {http://arxiv.org/abs/2305.18290} {Direct Preference Optimization: Your Language Model is Secretly a Reward Model}.

\bibitem[{Raffel et~al.(2023)Raffel, Shazeer, Roberts, Lee, Narang, Matena, Zhou, Li, and Liu}]{raffel2023exploring}
Colin Raffel, Noam Shazeer, Adam Roberts, Katherine Lee, Sharan Narang, Michael Matena, Yanqi Zhou, Wei Li, and Peter~J. Liu. 2023.
\newblock \href {http://arxiv.org/abs/1910.10683} {Exploring the Limits of Transfer Learning with a Unified Text-to-Text Transformer}.

\bibitem[{Rajani et~al.(2023)Rajani, Tunstall, Beeching, Lambert, Rush, and Wolf}]{no_robots}
Nazneen Rajani, Lewis Tunstall, Edward Beeching, Nathan Lambert, Alexander~M. Rush, and Thomas Wolf. 2023.
\newblock \href {https://huggingface.co/datasets/HuggingFaceH4/no_robots} {No Robots}.

\bibitem[{Rajbhandari et~al.(2022)Rajbhandari, Li, Yao, Zhang, Aminabadi, Awan, Rasley, and He}]{rajbhandari2022deepspeedmoeadvancingmixtureofexpertsinference}
Samyam Rajbhandari, Conglong Li, Zhewei Yao, Minjia Zhang, Reza~Yazdani Aminabadi, Ammar~Ahmad Awan, Jeff Rasley, and Yuxiong He. 2022.
\newblock \href {http://arxiv.org/abs/2201.05596} {DeepSpeed-MoE: Advancing Mixture-of-Experts Inference and Training to Power Next-Generation AI Scale}.

\bibitem[{Rajbhandari et~al.(2020)Rajbhandari, Rasley, Ruwase, and He}]{rajbhandari2020zero}
Samyam Rajbhandari, Jeff Rasley, Olatunji Ruwase, and Yuxiong He. 2020.
\newblock \href {http://arxiv.org/abs/1910.02054} {ZeRO: Memory Optimizations Toward Training Trillion Parameter Models}.

\bibitem[{Raposo et~al.(2024)Raposo, Ritter, Richards, Lillicrap, Humphreys, and Santoro}]{raposo2024mixtureofdepthsdynamicallyallocatingcompute}
David Raposo, Sam Ritter, Blake Richards, Timothy Lillicrap, Peter~Conway Humphreys, and Adam Santoro. 2024.
\newblock \href {http://arxiv.org/abs/2404.02258} {Mixture-of-Depths: Dynamically allocating compute in transformer-based language models}.

\bibitem[{Reid et~al.(2022)Reid, Zhong, Gururangan, and Zettlemoyer}]{reid2022m2d2massivelymultidomainlanguage}
Machel Reid, Victor Zhong, Suchin Gururangan, and Luke Zettlemoyer. 2022.
\newblock \href {http://arxiv.org/abs/2210.07370} {M2D2: A Massively Multi-domain Language Modeling Dataset}.

\bibitem[{Ren et~al.(2023)Ren, Zhou, Meng, Huang, Wang, Wang, Li, Zhang, Podolskiy, Arshinov, Bout, Piontkovskaya, Wei, Jiang, Su, Liu, and Yao}]{ren2023pangusigmatrillionparameterlanguage}
Xiaozhe Ren, Pingyi Zhou, Xinfan Meng, Xinjing Huang, Yadao Wang, Weichao Wang, Pengfei Li, Xiaoda Zhang, Alexander Podolskiy, Grigory Arshinov, Andrey Bout, Irina Piontkovskaya, Jiansheng Wei, Xin Jiang, Teng Su, Qun Liu, and Jun Yao. 2023.
\newblock \href {http://arxiv.org/abs/2303.10845} {PanGu-Sigma: Towards Trillion Parameter Language Model with Sparse Heterogeneous Computing}.

\bibitem[{Roller et~al.(2021)Roller, Sukhbaatar, Szlam, and Weston}]{roller2021hashlayerslargesparse}
Stephen Roller, Sainbayar Sukhbaatar, Arthur Szlam, and Jason Weston. 2021.
\newblock \href {http://arxiv.org/abs/2106.04426} {Hash Layers For Large Sparse Models}.

\bibitem[{Röttger et~al.(2024)Röttger, Kirk, Vidgen, Attanasio, Bianchi, and Hovy}]{röttger2024xstest}
Paul Röttger, Hannah~Rose Kirk, Bertie Vidgen, Giuseppe Attanasio, Federico Bianchi, and Dirk Hovy. 2024.
\newblock \href {http://arxiv.org/abs/2308.01263} {XSTest: A Test Suite for Identifying Exaggerated Safety Behaviours in Large Language Models}.

\bibitem[{Sakaguchi et~al.(2019)Sakaguchi, Bras, Bhagavatula, and Choi}]{sakaguchi2019winograndeadversarialwinogradschema}
Keisuke Sakaguchi, Ronan~Le Bras, Chandra Bhagavatula, and Yejin Choi. 2019.
\newblock \href {http://arxiv.org/abs/1907.10641} {WinoGrande: An Adversarial Winograd Schema Challenge at Scale}.

\bibitem[{Sanh et~al.(2022)Sanh, Webson, Raffel, Bach, Sutawika, Alyafeai, Chaffin, Stiegler, Scao, Raja et~al.}]{sanh2022multitask}
Victor Sanh, Albert Webson, Colin Raffel, Stephen~H. Bach, Lintang Sutawika, Zaid Alyafeai, Antoine Chaffin, Arnaud Stiegler, Teven~Le Scao, Arun Raja, et~al. 2022.
\newblock \href {http://arxiv.org/abs/2110.08207} {Multitask Prompted Training Enables Zero-Shot Task Generalization}.

\bibitem[{Sap et~al.(2019)Sap, Rashkin, Chen, LeBras, and Choi}]{sap2019socialiqacommonsensereasoningsocial}
Maarten Sap, Hannah Rashkin, Derek Chen, Ronan LeBras, and Yejin Choi. 2019.
\newblock \href {http://arxiv.org/abs/1904.09728} {SocialIQA: Commonsense Reasoning about Social Interactions}.

\bibitem[{Scao et~al.(2022)Scao, Wang, Hesslow, Saulnier, Bekman, Bari, Biderman, Elsahar, Muennighoff, Phang, Press, Raffel, Sanh, Shen, Sutawika, Tae, Yong, Launay, and Beltagy}]{scao2022language}
Teven~Le Scao, Thomas Wang, Daniel Hesslow, Lucile Saulnier, Stas Bekman, M~Saiful Bari, Stella Biderman, Hady Elsahar, Niklas Muennighoff, Jason Phang, Ofir Press, Colin Raffel, Victor Sanh, Sheng Shen, Lintang Sutawika, Jaesung Tae, Zheng~Xin Yong, Julien Launay, and Iz~Beltagy. 2022.
\newblock \href {http://arxiv.org/abs/2210.15424} {What Language Model to Train if You Have One Million GPU Hours?}

\bibitem[{Shazeer(2019)}]{shazeer2019fasttransformerdecodingwritehead}
Noam Shazeer. 2019.
\newblock \href {http://arxiv.org/abs/1911.02150} {Fast Transformer Decoding: One Write-Head is All You Need}.

\bibitem[{Shazeer(2020)}]{shazeer2020glu}
Noam Shazeer. 2020.
\newblock \href {http://arxiv.org/abs/2002.05202} {GLU Variants Improve Transformer}.

\bibitem[{Shazeer et~al.(2017)Shazeer, Mirhoseini, Maziarz, Davis, Le, Hinton, and Dean}]{shazeer2017outrageously}
Noam Shazeer, Azalia Mirhoseini, Krzysztof Maziarz, Andy Davis, Quoc Le, Geoffrey Hinton, and Jeff Dean. 2017.
\newblock \href {http://arxiv.org/abs/1701.06538} {Outrageously Large Neural Networks: The Sparsely-Gated Mixture-of-Experts Layer}.

\bibitem[{Shazeer and Stern(2018)}]{shazeer2018adafactoradaptivelearningrates}
Noam Shazeer and Mitchell Stern. 2018.
\newblock \href {http://arxiv.org/abs/1804.04235} {Adafactor: Adaptive Learning Rates with Sublinear Memory Cost}.

\bibitem[{Shen et~al.(2023{\natexlab{a}})Shen, Hou, Zhou, Du, Longpre, Wei, Chung, Zoph, Fedus, Chen, Vu, Wu, Chen, Webson, Li, Zhao, Yu, Keutzer, Darrell, and Zhou}]{shen2023mixtureofexperts}
Sheng Shen, Le~Hou, Yanqi Zhou, Nan Du, Shayne Longpre, Jason Wei, Hyung~Won Chung, Barret Zoph, William Fedus, Xinyun Chen, Tu~Vu, Yuexin Wu, Wuyang Chen, Albert Webson, Yunxuan Li, Vincent Zhao, Hongkun Yu, Kurt Keutzer, Trevor Darrell, and Denny Zhou. 2023{\natexlab{a}}.
\newblock \href {http://arxiv.org/abs/2305.14705} {Mixture-of-Experts Meets Instruction Tuning:A Winning Combination for Large Language Models}.

\bibitem[{Shen et~al.(2023{\natexlab{b}})Shen, Yao, Li, Darrell, Keutzer, and He}]{shen2023scalingvisionlanguagemodelssparse}
Sheng Shen, Zhewei Yao, Chunyuan Li, Trevor Darrell, Kurt Keutzer, and Yuxiong He. 2023{\natexlab{b}}.
\newblock \href {http://arxiv.org/abs/2303.07226} {Scaling Vision-Language Models with Sparse Mixture of Experts}.

\bibitem[{Shen et~al.(2024)Shen, Guo, Cai, and Qin}]{shen2024jetmoe}
Yikang Shen, Zhen Guo, Tianle Cai, and Zengyi Qin. 2024.
\newblock \href {http://arxiv.org/abs/2404.07413} {JetMoE: Reaching Llama2 Performance with 0.1M Dollars}.

\bibitem[{Shoeybi et~al.(2020)Shoeybi, Patwary, Puri, LeGresley, Casper, and Catanzaro}]{shoeybi2020megatronlmtrainingmultibillionparameter}
Mohammad Shoeybi, Mostofa Patwary, Raul Puri, Patrick LeGresley, Jared Casper, and Bryan Catanzaro. 2020.
\newblock \href {http://arxiv.org/abs/1909.08053} {Megatron-LM: Training Multi-Billion Parameter Language Models Using Model Parallelism}.

\bibitem[{Singh et~al.(2024)Singh, Vargus, Dsouza, Karlsson, Mahendiran, Ko, Shandilya, Patel, Mataciunas, OMahony, Zhang, Hettiarachchi, Wilson, Machado, Moura, Krzemiński, Fadaei, Ergün, Okoh, Alaagib, Mudannayake, Alyafeai, Chien, Ruder, Guthikonda, Alghamdi, Gehrmann, Muennighoff, Bartolo, Kreutzer, Üstün, Fadaee, and Hooker}]{singh2024aya}
Shivalika Singh, Freddie Vargus, Daniel Dsouza, Börje~F. Karlsson, Abinaya Mahendiran, Wei-Yin Ko, Herumb Shandilya, Jay Patel, Deividas Mataciunas, Laura OMahony, Mike Zhang, Ramith Hettiarachchi, Joseph Wilson, Marina Machado, Luisa~Souza Moura, Dominik Krzemiński, Hakimeh Fadaei, Irem Ergün, Ifeoma Okoh, Aisha Alaagib, Oshan Mudannayake, Zaid Alyafeai, Vu~Minh Chien, Sebastian Ruder, Surya Guthikonda, Emad~A. Alghamdi, Sebastian Gehrmann, Niklas Muennighoff, Max Bartolo, Julia Kreutzer, Ahmet Üstün, Marzieh Fadaee, and Sara Hooker. 2024.
\newblock \href {http://arxiv.org/abs/2402.06619} {Aya Dataset: An Open-Access Collection for Multilingual Instruction Tuning}.

\bibitem[{Snowflake(2024{\natexlab{a}})}]{arcticcookbook}
Snowflake. 2024{\natexlab{a}}.
\newblock \href {https://medium.com/snowflake/snowflake-arctic-cookbook-series-exploring-mixture-of-experts-moe-c7d6b8f14d16} {Snowflake Arctic Cookbook Series: Exploring Mixture of Experts (MoE)}.

\bibitem[{Snowflake(2024{\natexlab{b}})}]{arctic}
Snowflake. 2024{\natexlab{b}}.
\newblock \href {https://www.snowflake.com/blog/arctic-open-efficient-foundation-language-models-snowflake/} {Snowflake Arctic: The Best LLM for Enterprise AI — Efficiently Intelligent, Truly Open}.

\bibitem[{Soldaini et~al.(2024)Soldaini, Kinney, Bhagia, Schwenk, Atkinson, Authur, Bogin, Chandu, Dumas, Elazar, Hofmann, Jha, Kumar, Lucy, Lyu, Lambert, Magnusson, Morrison, Muennighoff, Naik, Nam, Peters, Ravichander, Richardson, Shen, Strubell, Subramani, Tafjord, Walsh, Zettlemoyer, Smith, Hajishirzi, Beltagy, Groeneveld, Dodge, and Lo}]{soldaini2024dolma}
Luca Soldaini, Rodney Kinney, Akshita Bhagia, Dustin Schwenk, David Atkinson, Russell Authur, Ben Bogin, Khyathi Chandu, Jennifer Dumas, Yanai Elazar, Valentin Hofmann, Ananya~Harsh Jha, Sachin Kumar, Li~Lucy, Xinxi Lyu, Nathan Lambert, Ian Magnusson, Jacob Morrison, Niklas Muennighoff, Aakanksha Naik, Crystal Nam, Matthew~E. Peters, Abhilasha Ravichander, Kyle Richardson, Zejiang Shen, Emma Strubell, Nishant Subramani, Oyvind Tafjord, Pete Walsh, Luke Zettlemoyer, Noah~A. Smith, Hannaneh Hajishirzi, Iz~Beltagy, Dirk Groeneveld, Jesse Dodge, and Kyle Lo. 2024.
\newblock \href {http://arxiv.org/abs/2402.00159} {Dolma: an Open Corpus of Three Trillion Tokens for Language Model Pretraining Research}.

\bibitem[{Soldaini and Lo(2023)}]{peS2o}
Luca Soldaini and Kyle Lo. 2023.
\newblock \href {https://github.com/allenai/pes2o} {{peS2o (Pretraining Efficiently on S2ORC) Dataset}}.

\bibitem[{Son et~al.(2024)Son, Lee, Kim, Kim, Muennighoff, Choi, Park, Yoo, and Biderman}]{son2024kmmlumeasuringmassivemultitask}
Guijin Son, Hanwool Lee, Sungdong Kim, Seungone Kim, Niklas Muennighoff, Taekyoon Choi, Cheonbok Park, Kang~Min Yoo, and Stella Biderman. 2024.
\newblock \href {http://arxiv.org/abs/2402.11548} {KMMLU: Measuring Massive Multitask Language Understanding in Korean}.

\bibitem[{Su et~al.(2023)Su, Lu, Pan, Murtadha, Wen, and Liu}]{su2023roformerenhancedtransformerrotary}
Jianlin Su, Yu~Lu, Shengfeng Pan, Ahmed Murtadha, Bo~Wen, and Yunfeng Liu. 2023.
\newblock \href {http://arxiv.org/abs/2104.09864} {RoFormer: Enhanced Transformer with Rotary Position Embedding}.

\bibitem[{Su et~al.(2020)Su, Zhu, Cao, Li, Lu, Wei, and Dai}]{su2020vlbertpretraininggenericvisuallinguistic}
Weijie Su, Xizhou Zhu, Yue Cao, Bin Li, Lewei Lu, Furu Wei, and Jifeng Dai. 2020.
\newblock \href {http://arxiv.org/abs/1908.08530} {VL-BERT: Pre-training of Generic Visual-Linguistic Representations}.

\bibitem[{Sukhbaatar et~al.(2024)Sukhbaatar, Golovneva, Sharma, Xu, Lin, Rozière, Kahn, Li, tau Yih, Weston, and Li}]{sukhbaatar2024branchtrainmixmixingexpertllms}
Sainbayar Sukhbaatar, Olga Golovneva, Vasu Sharma, Hu~Xu, Xi~Victoria Lin, Baptiste Rozière, Jacob Kahn, Daniel Li, Wen tau Yih, Jason Weston, and Xian Li. 2024.
\newblock \href {http://arxiv.org/abs/2403.07816} {Branch-Train-MiX: Mixing Expert LLMs into a Mixture-of-Experts LLM}.

\bibitem[{Suzgun et~al.(2022)Suzgun, Scales, Schärli, Gehrmann, Tay, Chung, Chowdhery, Le, Chi, Zhou, and Wei}]{suzgun2022challenging}
Mirac Suzgun, Nathan Scales, Nathanael Schärli, Sebastian Gehrmann, Yi~Tay, Hyung~Won Chung, Aakanksha Chowdhery, Quoc~V. Le, Ed~H. Chi, Denny Zhou, and Jason Wei. 2022.
\newblock \href {http://arxiv.org/abs/2210.09261} {Challenging BIG-Bench Tasks and Whether Chain-of-Thought Can Solve Them}.

\bibitem[{Talmor et~al.(2019)Talmor, Herzig, Lourie, and Berant}]{talmor2019commonsenseqaquestionansweringchallenge}
Alon Talmor, Jonathan Herzig, Nicholas Lourie, and Jonathan Berant. 2019.
\newblock \href {http://arxiv.org/abs/1811.00937} {CommonsenseQA: A Question Answering Challenge Targeting Commonsense Knowledge}.

\bibitem[{Tan et~al.(2023)Tan, Shen, Chen, Courville, and Gan}]{tan2023sparse}
Shawn Tan, Yikang Shen, Zhenfang Chen, Aaron Courville, and Chuang Gan. 2023.
\newblock \href {http://arxiv.org/abs/2310.07096} {Sparse Universal Transformer}.

\bibitem[{Tao et~al.(2024)Tao, Liu, Dou, Muennighoff, Wan, Luo, Lin, and Wong}]{tao2024scalinglawsvocabularylarger}
Chaofan Tao, Qian Liu, Longxu Dou, Niklas Muennighoff, Zhongwei Wan, Ping Luo, Min Lin, and Ngai Wong. 2024.
\newblock \href {http://arxiv.org/abs/2407.13623} {Scaling Laws with Vocabulary: Larger Models Deserve Larger Vocabularies}.

\bibitem[{Team(2024{\natexlab{a}})}]{chameleonteam2024chameleon}
Chameleon Team. 2024{\natexlab{a}}.
\newblock \href {http://arxiv.org/abs/2405.09818} {Chameleon: Mixed-Modal Early-Fusion Foundation Models}.

\bibitem[{Team et~al.(2023)Team, Anil, Borgeaud, Wu, Alayrac, Yu, Soricut, Schalkwyk, Dai, Hauth et~al.}]{geminiteam2023gemini}
Gemini Team, Rohan Anil, Sebastian Borgeaud, Yonghui Wu, Jean-Baptiste Alayrac, Jiahui Yu, Radu Soricut, Johan Schalkwyk, Andrew~M. Dai, Anja Hauth, et~al. 2023.
\newblock \href {http://arxiv.org/abs/2312.11805} {Gemini: A Family of Highly Capable Multimodal Models}.

\bibitem[{Team et~al.(2024{\natexlab{a}})Team, Georgiev, Lei, Burnell, Bai, Gulati, Tanzer, Vincent, Pan, Wang, Mariooryad, Ding, Geng, Alcober, Frostig, Omernick, Walker, Paduraru, Sorokin, Tacchetti, Gaffney, Daruki, Sercinoglu, Gleicher, Love, Voigtlaender, Jain et~al.}]{geminiteam2024gemini}
Gemini Team, Petko Georgiev, Ving~Ian Lei, Ryan Burnell, Libin Bai, Anmol Gulati, Garrett Tanzer, Damien Vincent, Zhufeng Pan, Shibo Wang, Soroosh Mariooryad, Yifan Ding, Xinyang Geng, Fred Alcober, Roy Frostig, Mark Omernick, Lexi Walker, Cosmin Paduraru, Christina Sorokin, Andrea Tacchetti, Colin Gaffney, Samira Daruki, Olcan Sercinoglu, Zach Gleicher, Juliette Love, Paul Voigtlaender, Rohan Jain, et~al. 2024{\natexlab{a}}.
\newblock \href {http://arxiv.org/abs/2403.05530} {Gemini 1.5: Unlocking multimodal understanding across millions of tokens of context}.

\bibitem[{Team et~al.(2024{\natexlab{b}})Team, Mesnard, Hardin, Dadashi, Bhupatiraju, Pathak, Sifre, Rivière, Kale, Love, Tafti, Hussenot, Sessa, Chowdhery, Roberts, Barua, Botev, Castro-Ros, Slone, Héliou, Tacchetti, Bulanova, Paterson, Tsai, Shahriari, Lan, Choquette-Choo, Crepy, Cer, Ippolito, Reid, Buchatskaya, Ni, Noland, Yan, Tucker, Muraru, Rozhdestvenskiy, Michalewski, Tenney, Grishchenko, Austin, Keeling, Labanowski, Lespiau, Stanway, Brennan, Chen, Ferret, Chiu, Mao-Jones, Lee, Yu, Millican, Sjoesund, Lee, Dixon, Reid, Mikuła, Wirth, Sharman, Chinaev, Thain, Bachem, Chang, Wahltinez, Bailey, Michel, Yotov, Chaabouni, Comanescu, Jana, Anil, McIlroy, Liu, Mullins, Smith, Borgeaud, Girgin, Douglas, Pandya, Shakeri, De, Klimenko, Hennigan, Feinberg, Stokowiec, hui Chen, Ahmed, Gong, Warkentin, Peran, Giang, Farabet, Vinyals, Dean, Kavukcuoglu, Hassabis, Ghahramani, Eck, Barral, Pereira, Collins, Joulin, Fiedel, Senter, Andreev, and Kenealy}]{gemmateam2024gemmaopenmodelsbased}
Gemma Team, Thomas Mesnard, Cassidy Hardin, Robert Dadashi, Surya Bhupatiraju, Shreya Pathak, Laurent Sifre, Morgane Rivière, Mihir~Sanjay Kale, Juliette Love, Pouya Tafti, Léonard Hussenot, Pier~Giuseppe Sessa, Aakanksha Chowdhery, Adam Roberts, Aditya Barua, Alex Botev, Alex Castro-Ros, Ambrose Slone, Amélie Héliou, Andrea Tacchetti, Anna Bulanova, Antonia Paterson, Beth Tsai, Bobak Shahriari, Charline~Le Lan, Christopher~A. Choquette-Choo, Clément Crepy, Daniel Cer, Daphne Ippolito, David Reid, Elena Buchatskaya, Eric Ni, Eric Noland, Geng Yan, George Tucker, George-Christian Muraru, Grigory Rozhdestvenskiy, Henryk Michalewski, Ian Tenney, Ivan Grishchenko, Jacob Austin, James Keeling, Jane Labanowski, Jean-Baptiste Lespiau, Jeff Stanway, Jenny Brennan, Jeremy Chen, Johan Ferret, Justin Chiu, Justin Mao-Jones, Katherine Lee, Kathy Yu, Katie Millican, Lars~Lowe Sjoesund, Lisa Lee, Lucas Dixon, Machel Reid, Maciej Mikuła, Mateo Wirth, Michael Sharman, Nikolai Chinaev, Nithum Thain, Olivier Bachem,
  Oscar Chang, Oscar Wahltinez, Paige Bailey, Paul Michel, Petko Yotov, Rahma Chaabouni, Ramona Comanescu, Reena Jana, Rohan Anil, Ross McIlroy, Ruibo Liu, Ryan Mullins, Samuel~L Smith, Sebastian Borgeaud, Sertan Girgin, Sholto Douglas, Shree Pandya, Siamak Shakeri, Soham De, Ted Klimenko, Tom Hennigan, Vlad Feinberg, Wojciech Stokowiec, Yu~hui Chen, Zafarali Ahmed, Zhitao Gong, Tris Warkentin, Ludovic Peran, Minh Giang, Clément Farabet, Oriol Vinyals, Jeff Dean, Koray Kavukcuoglu, Demis Hassabis, Zoubin Ghahramani, Douglas Eck, Joelle Barral, Fernando Pereira, Eli Collins, Armand Joulin, Noah Fiedel, Evan Senter, Alek Andreev, and Kathleen Kenealy. 2024{\natexlab{b}}.
\newblock \href {http://arxiv.org/abs/2403.08295} {Gemma: Open Models Based on Gemini Research and Technology}.

\bibitem[{Team et~al.(2024{\natexlab{c}})Team, Riviere, Pathak, Sessa, Hardin, Bhupatiraju, Hussenot, Mesnard, Shahriari, Ramé, Ferret, Liu, Tafti, Friesen et~al.}]{gemmateam2024gemma2improvingopen}
Gemma Team, Morgane Riviere, Shreya Pathak, Pier~Giuseppe Sessa, Cassidy Hardin, Surya Bhupatiraju, Léonard Hussenot, Thomas Mesnard, Bobak Shahriari, Alexandre Ramé, Johan Ferret, Peter Liu, Pouya Tafti, Abe Friesen, et~al. 2024{\natexlab{c}}.
\newblock \href {http://arxiv.org/abs/2408.00118} {Gemma 2: Improving Open Language Models at a Practical Size}.

\bibitem[{Team et~al.(2024{\natexlab{d}})Team, Lenz, Arazi, Bergman, Manevich, Peleg, Aviram, Almagor, Fridman, Padnos, Gissin, Jannai, Muhlgay, Zimberg, Gerber, Dolev, Krakovsky, Safahi, Schwartz, Cohen, Shachaf, Rozenblum, Bata, Blass, Magar, Dalmedigos, Osin, Fadlon, Rozman, Danos, Gokhman, Zusman, Gidron, Ratner, Gat, Rozen, Fried, Leshno, Antverg, Abend, Lieber, Dagan, Cohavi, Alon, Belson, Cohen, Gilad, Glozman, Lev, Meirom, Delbari, Ness, Asida, Gal, Braude, Pumerantz, Cohen, Belinkov, Globerson, Levy, and Shoham}]{jambateam2024jamba15hybridtransformermambamodels}
Jamba Team, Barak Lenz, Alan Arazi, Amir Bergman, Avshalom Manevich, Barak Peleg, Ben Aviram, Chen Almagor, Clara Fridman, Dan Padnos, Daniel Gissin, Daniel Jannai, Dor Muhlgay, Dor Zimberg, Edden~M Gerber, Elad Dolev, Eran Krakovsky, Erez Safahi, Erez Schwartz, Gal Cohen, Gal Shachaf, Haim Rozenblum, Hofit Bata, Ido Blass, Inbal Magar, Itay Dalmedigos, Jhonathan Osin, Julie Fadlon, Maria Rozman, Matan Danos, Michael Gokhman, Mor Zusman, Naama Gidron, Nir Ratner, Noam Gat, Noam Rozen, Oded Fried, Ohad Leshno, Omer Antverg, Omri Abend, Opher Lieber, Or~Dagan, Orit Cohavi, Raz Alon, Ro'i Belson, Roi Cohen, Rom Gilad, Roman Glozman, Shahar Lev, Shaked Meirom, Tal Delbari, Tal Ness, Tomer Asida, Tom~Ben Gal, Tom Braude, Uriya Pumerantz, Yehoshua Cohen, Yonatan Belinkov, Yuval Globerson, Yuval~Peleg Levy, and Yoav Shoham. 2024{\natexlab{d}}.
\newblock \href {http://arxiv.org/abs/2408.12570} {Jamba-1.5: Hybrid Transformer-Mamba Models at Scale}.

\bibitem[{Team(2023)}]{MosaicML2023Introducing}
MosaicML~NLP Team. 2023.
\newblock \href {https://mosaicml.com/blog/mpt-7b} {Introducing MPT-7B: A New Standard for Open-Source, Commercially Usable LLMs}.

\bibitem[{Team(2024{\natexlab{b}})}]{qwen_moe}
Qwen Team. 2024{\natexlab{b}}.
\newblock \href {https://qwenlm.github.io/blog/qwen-moe/} {Qwen1.5-MoE: Matching 7B Model Performance with 1/3 Activated Parameters"}.

\bibitem[{Team et~al.(2024{\natexlab{e}})Team, Ormazabal, Zheng, de~Masson~d'Autume, Yogatama, Fu, Ong, Chen, Lamprecht, Pham, Ong, Aleksiev, Li, Henderson, Bain, Artetxe, Relan, Padlewski, Liu, Chen, Phua, Yang, Tay, Wang, Zhu, and Xie}]{rekateam2024rekacoreflashedge}
Reka Team, Aitor Ormazabal, Che Zheng, Cyprien de~Masson~d'Autume, Dani Yogatama, Deyu Fu, Donovan Ong, Eric Chen, Eugenie Lamprecht, Hai Pham, Isaac Ong, Kaloyan Aleksiev, Lei Li, Matthew Henderson, Max Bain, Mikel Artetxe, Nishant Relan, Piotr Padlewski, Qi~Liu, Ren Chen, Samuel Phua, Yazheng Yang, Yi~Tay, Yuqi Wang, Zhongkai Zhu, and Zhihui Xie. 2024{\natexlab{e}}.
\newblock \href {http://arxiv.org/abs/2404.12387} {Reka Core, Flash, and Edge: A Series of Powerful Multimodal Language Models}.

\bibitem[{Touvron et~al.(2023{\natexlab{a}})Touvron, Lavril, Izacard, Martinet, Lachaux, Lacroix, Rozière, Goyal, Hambro, Azhar, Rodriguez, Joulin, Grave, and Lample}]{touvron2023llama}
Hugo Touvron, Thibaut Lavril, Gautier Izacard, Xavier Martinet, Marie-Anne Lachaux, Timothée Lacroix, Baptiste Rozière, Naman Goyal, Eric Hambro, Faisal Azhar, Aurelien Rodriguez, Armand Joulin, Edouard Grave, and Guillaume Lample. 2023{\natexlab{a}}.
\newblock \href {http://arxiv.org/abs/2302.13971} {LLaMA: Open and Efficient Foundation Language Models}.

\bibitem[{Touvron et~al.(2023{\natexlab{b}})Touvron, Martin, Stone, Albert, Almahairi, Babaei, Bashlykov, Batra, Bhargava, Bhosale, Bikel, Blecher, Ferrer, Chen, Cucurull, Esiobu, Fernandes, Fu, Fu, Fuller, Gao, Goswami, Goyal, Hartshorn, Hosseini, Hou, Inan, Kardas, Kerkez, Khabsa, Kloumann, Korenev, Koura, Lachaux, Lavril, Lee, Liskovich, Lu, Mao, Martinet, Mihaylov, Mishra, Molybog, Nie, Poulton, Reizenstein, Rungta, Saladi, Schelten, Silva, Smith, Subramanian, Tan, Tang, Taylor, Williams, Kuan, Xu, Yan, Zarov, Zhang, Fan, Kambadur, Narang, Rodriguez, Stojnic, Edunov, and Scialom}]{touvron2023llama2openfoundation}
Hugo Touvron, Louis Martin, Kevin Stone, Peter Albert, Amjad Almahairi, Yasmine Babaei, Nikolay Bashlykov, Soumya Batra, Prajjwal Bhargava, Shruti Bhosale, Dan Bikel, Lukas Blecher, Cristian~Canton Ferrer, Moya Chen, Guillem Cucurull, David Esiobu, Jude Fernandes, Jeremy Fu, Wenyin Fu, Brian Fuller, Cynthia Gao, Vedanuj Goswami, Naman Goyal, Anthony Hartshorn, Saghar Hosseini, Rui Hou, Hakan Inan, Marcin Kardas, Viktor Kerkez, Madian Khabsa, Isabel Kloumann, Artem Korenev, Punit~Singh Koura, Marie-Anne Lachaux, Thibaut Lavril, Jenya Lee, Diana Liskovich, Yinghai Lu, Yuning Mao, Xavier Martinet, Todor Mihaylov, Pushkar Mishra, Igor Molybog, Yixin Nie, Andrew Poulton, Jeremy Reizenstein, Rashi Rungta, Kalyan Saladi, Alan Schelten, Ruan Silva, Eric~Michael Smith, Ranjan Subramanian, Xiaoqing~Ellen Tan, Binh Tang, Ross Taylor, Adina Williams, Jian~Xiang Kuan, Puxin Xu, Zheng Yan, Iliyan Zarov, Yuchen Zhang, Angela Fan, Melanie Kambadur, Sharan Narang, Aurelien Rodriguez, Robert Stojnic, Sergey Edunov, and Thomas
  Scialom. 2023{\natexlab{b}}.
\newblock \href {http://arxiv.org/abs/2307.09288} {Llama 2: Open Foundation and Fine-Tuned Chat Models}.

\bibitem[{Tunstall et~al.(2023)Tunstall, Beeching, Lambert, Rajani, Rasul, Belkada, Huang, von Werra, Fourrier, Habib, Sarrazin, Sanseviero, Rush, and Wolf}]{tunstall2023zephyr}
Lewis Tunstall, Edward Beeching, Nathan Lambert, Nazneen Rajani, Kashif Rasul, Younes Belkada, Shengyi Huang, Leandro von Werra, Clémentine Fourrier, Nathan Habib, Nathan Sarrazin, Omar Sanseviero, Alexander~M. Rush, and Thomas Wolf. 2023.
\newblock \href {http://arxiv.org/abs/2310.16944} {Zephyr: Direct Distillation of LM Alignment}.

\bibitem[{Vaswani et~al.(2023)Vaswani, Shazeer, Parmar, Uszkoreit, Jones, Gomez, Kaiser, and Polosukhin}]{vaswani2023attention}
Ashish Vaswani, Noam Shazeer, Niki Parmar, Jakob Uszkoreit, Llion Jones, Aidan~N. Gomez, Lukasz Kaiser, and Illia Polosukhin. 2023.
\newblock \href {http://arxiv.org/abs/1706.03762} {Attention Is All You Need}.

\bibitem[{Wang and Komatsuzaki(2021)}]{wang2021gpt}
Ben Wang and Aran Komatsuzaki. 2021.
\newblock \href {https://github.com/kingoflolz/mesh-transformer-jax} {GPT-J-6B: A 6 Billion Parameter Autoregressive Language Model}.

\bibitem[{Wang et~al.(2024{\natexlab{a}})Wang, Li, Song, Xu, Tang, Zhuge, Pan, Song, Li, Singh, Tran, Li, Ma, Zheng, Qian, Shao, Muennighoff, Zhang, Hui, Lin, Brennan, Peng, Ji, and Neubig}]{wang2024opendevinopenplatformai}
Xingyao Wang, Boxuan Li, Yufan Song, Frank~F. Xu, Xiangru Tang, Mingchen Zhuge, Jiayi Pan, Yueqi Song, Bowen Li, Jaskirat Singh, Hoang~H. Tran, Fuqiang Li, Ren Ma, Mingzhang Zheng, Bill Qian, Yanjun Shao, Niklas Muennighoff, Yizhe Zhang, Binyuan Hui, Junyang Lin, Robert Brennan, Hao Peng, Heng Ji, and Graham Neubig. 2024{\natexlab{a}}.
\newblock \href {http://arxiv.org/abs/2407.16741} {OpenDevin: An Open Platform for AI Software Developers as Generalist Agents}.

\bibitem[{Wang et~al.(2023)Wang, Ivison, Dasigi, Hessel, Khot, Chandu, Wadden, MacMillan, Smith, Beltagy, and Hajishirzi}]{wang2023far}
Yizhong Wang, Hamish Ivison, Pradeep Dasigi, Jack Hessel, Tushar Khot, Khyathi~Raghavi Chandu, David Wadden, Kelsey MacMillan, Noah~A. Smith, Iz~Beltagy, and Hannaneh Hajishirzi. 2023.
\newblock \href {http://arxiv.org/abs/2306.04751} {How Far Can Camels Go? Exploring the State of Instruction Tuning on Open Resources}.

\bibitem[{Wang et~al.(2024{\natexlab{b}})Wang, Dong, Delalleau, Zeng, Shen, Egert, Zhang, Sreedhar, and Kuchaiev}]{wang2024helpsteer2opensourcedatasettraining}
Zhilin Wang, Yi~Dong, Olivier Delalleau, Jiaqi Zeng, Gerald Shen, Daniel Egert, Jimmy~J. Zhang, Makesh~Narsimhan Sreedhar, and Oleksii Kuchaiev. 2024{\natexlab{b}}.
\newblock \href {http://arxiv.org/abs/2406.08673} {HelpSteer2: Open-source dataset for training top-performing reward models}.

\bibitem[{Wei et~al.(2022)Wei, Bosma, Zhao, Guu, Yu, Lester, Du, Dai, and Le}]{wei2022finetuned}
Jason Wei, Maarten Bosma, Vincent~Y. Zhao, Kelvin Guu, Adams~Wei Yu, Brian Lester, Nan Du, Andrew~M. Dai, and Quoc~V. Le. 2022.
\newblock \href {http://arxiv.org/abs/2109.01652} {Finetuned Language Models Are Zero-Shot Learners}.

\bibitem[{Wei et~al.(2024)Wei, Zhu, Zhao, Cheng, Li, Lü, Cheng, Zhang, Zhang, Zeng, Wang, Ma, Hu, Yan, Fang, and Zhou}]{wei2024skyworkmoe}
Tianwen Wei, Bo~Zhu, Liang Zhao, Cheng Cheng, Biye Li, Weiwei Lü, Peng Cheng, Jianhao Zhang, Xiaoyu Zhang, Liang Zeng, Xiaokun Wang, Yutuan Ma, Rui Hu, Shuicheng Yan, Han Fang, and Yahui Zhou. 2024.
\newblock \href {http://arxiv.org/abs/2406.06563} {Skywork-MoE: A Deep Dive into Training Techniques for Mixture-of-Experts Language Models}.

\bibitem[{Welbl et~al.(2017)Welbl, Liu, and Gardner}]{welbl2017crowdsourcingmultiplechoicescience}
Johannes Welbl, Nelson~F. Liu, and Matt Gardner. 2017.
\newblock \href {http://arxiv.org/abs/1707.06209} {Crowdsourcing Multiple Choice Science Questions}.

\bibitem[{Workshop et~al.(2023)Workshop, Scao, Fan, Akiki, Pavlick, Ilić, Hesslow, Castagné, Luccioni, Yvon, Gallé, Tow, Rush, Biderman, Webson, Ammanamanchi, Wang, Sagot, Muennighoff et~al.}]{workshop2023bloom}
BigScience Workshop, Teven~Le Scao, Angela Fan, Christopher Akiki, Ellie Pavlick, Suzana Ilić, Daniel Hesslow, Roman Castagné, Alexandra~Sasha Luccioni, François Yvon, Matthias Gallé, Jonathan Tow, Alexander~M. Rush, Stella Biderman, Albert Webson, Pawan~Sasanka Ammanamanchi, Thomas Wang, Benoît Sagot, Niklas Muennighoff, et~al. 2023.
\newblock \href {http://arxiv.org/abs/2211.05100} {BLOOM: A 176B-Parameter Open-Access Multilingual Language Model}.

\bibitem[{Wu et~al.(2024{\natexlab{a}})Wu, Hu, Wang, Pang, and Soricut}]{wu2024omnismolaboostinggeneralistmultimodal}
Jialin Wu, Xia Hu, Yaqing Wang, Bo~Pang, and Radu Soricut. 2024{\natexlab{a}}.
\newblock \href {http://arxiv.org/abs/2312.00968} {Omni-SMoLA: Boosting Generalist Multimodal Models with Soft Mixture of Low-rank Experts}.

\bibitem[{Wu et~al.(2024{\natexlab{b}})Wu, Luo, Chen, Li, Zhao, Yu, Wang, Wang, Wang, Qiao, He, Zhang, Sun, Mao, and Shen}]{wu2024yuan20m32mixtureexperts}
Shaohua Wu, Jiangang Luo, Xi~Chen, Lingjun Li, Xudong Zhao, Tong Yu, Chao Wang, Yue Wang, Fei Wang, Weixu Qiao, Houbo He, Zeru Zhang, Zeyu Sun, Junxiong Mao, and Chong Shen. 2024{\natexlab{b}}.
\newblock \href {http://arxiv.org/abs/2405.17976} {Yuan 2.0-M32: Mixture of Experts with Attention Router}.

\bibitem[{xAI(2024)}]{grok}
xAI. 2024.
\newblock \href {https://x.ai/blog/grok-os} {Open Release of Grok-1}.

\bibitem[{Xiao et~al.(2023)Xiao, Liu, Zhang, and Muennighoff}]{xiao2023cpack}
Shitao Xiao, Zheng Liu, Peitian Zhang, and Niklas Muennighoff. 2023.
\newblock \href {http://arxiv.org/abs/2309.07597} {C-Pack: Packaged Resources To Advance General Chinese Embedding}.

\bibitem[{Xu et~al.(2024)Xu, Guan, Greene, and Kechadi}]{xu2024benchmarkdatacontaminationlarge}
Cheng Xu, Shuhao Guan, Derek Greene, and M-Tahar Kechadi. 2024.
\newblock \href {http://arxiv.org/abs/2406.04244} {Benchmark Data Contamination of Large Language Models: A Survey}.

\bibitem[{Xue et~al.(2024)Xue, Zheng, Fu, Ni, Zheng, Zhou, and You}]{xue2024openmoe}
Fuzhao Xue, Zian Zheng, Yao Fu, Jinjie Ni, Zangwei Zheng, Wangchunshu Zhou, and Yang You. 2024.
\newblock \href {http://arxiv.org/abs/2402.01739} {OpenMoE: An Early Effort on Open Mixture-of-Experts Language Models}.

\bibitem[{Yang et~al.(2023)Yang, Xiao, Wang, Zhang, Bian, Yin, Lv, Pan, Wang, Yan, Yang, Deng, Wang, Liu, Ai, Dong, Zhao, Xu, Sun, Zhang, Liu, Ji, Xie, Dai, Fang, Su, Song, Liu, Ru, Ma, Wang, Liu, Lin, Nie, Guo, Sun, Zhang, Li, Li, Cheng, Chen, Zeng, Wang, Chen, Men, Yu, Pan, Shen, Wang, Li, Jiang, Gao, Zhang, Zhou, and Wu}]{yang2023baichuan2openlargescale}
Aiyuan Yang, Bin Xiao, Bingning Wang, Borong Zhang, Ce~Bian, Chao Yin, Chenxu Lv, Da~Pan, Dian Wang, Dong Yan, Fan Yang, Fei Deng, Feng Wang, Feng Liu, Guangwei Ai, Guosheng Dong, Haizhou Zhao, Hang Xu, Haoze Sun, Hongda Zhang, Hui Liu, Jiaming Ji, Jian Xie, JunTao Dai, Kun Fang, Lei Su, Liang Song, Lifeng Liu, Liyun Ru, Luyao Ma, Mang Wang, Mickel Liu, MingAn Lin, Nuolan Nie, Peidong Guo, Ruiyang Sun, Tao Zhang, Tianpeng Li, Tianyu Li, Wei Cheng, Weipeng Chen, Xiangrong Zeng, Xiaochuan Wang, Xiaoxi Chen, Xin Men, Xin Yu, Xuehai Pan, Yanjun Shen, Yiding Wang, Yiyu Li, Youxin Jiang, Yuchen Gao, Yupeng Zhang, Zenan Zhou, and Zhiying Wu. 2023.
\newblock \href {http://arxiv.org/abs/2309.10305} {Baichuan 2: Open Large-scale Language Models}.

\bibitem[{Yang et~al.(2024{\natexlab{a}})Yang, Yang, Hui, Zheng, Yu, Zhou, Li, Li, Liu, Huang, Dong, Wei, Lin, Tang, Wang, Yang, Tu, Zhang, Ma, Yang, Xu, Zhou, Bai, He, Lin, Dang, Lu, Chen, Yang, Li, Xue, Ni, Zhang, Wang, Peng, Men, Gao, Lin, Wang, Bai, Tan, Zhu, Li, Liu, Ge, Deng, Zhou, Ren, Zhang, Wei, Ren, Liu, Fan, Yao, Zhang, Wan, Chu, Liu, Cui, Zhang, Guo, and Fan}]{yang2024qwen2technicalreport}
An~Yang, Baosong Yang, Binyuan Hui, Bo~Zheng, Bowen Yu, Chang Zhou, Chengpeng Li, Chengyuan Li, Dayiheng Liu, Fei Huang, Guanting Dong, Haoran Wei, Huan Lin, Jialong Tang, Jialin Wang, Jian Yang, Jianhong Tu, Jianwei Zhang, Jianxin Ma, Jianxin Yang, Jin Xu, Jingren Zhou, Jinze Bai, Jinzheng He, Junyang Lin, Kai Dang, Keming Lu, Keqin Chen, Kexin Yang, Mei Li, Mingfeng Xue, Na~Ni, Pei Zhang, Peng Wang, Ru~Peng, Rui Men, Ruize Gao, Runji Lin, Shijie Wang, Shuai Bai, Sinan Tan, Tianhang Zhu, Tianhao Li, Tianyu Liu, Wenbin Ge, Xiaodong Deng, Xiaohuan Zhou, Xingzhang Ren, Xinyu Zhang, Xipin Wei, Xuancheng Ren, Xuejing Liu, Yang Fan, Yang Yao, Yichang Zhang, Yu~Wan, Yunfei Chu, Yuqiong Liu, Zeyu Cui, Zhenru Zhang, Zhifang Guo, and Zhihao Fan. 2024{\natexlab{a}}.
\newblock \href {http://arxiv.org/abs/2407.10671} {Qwen2 Technical Report}.

\bibitem[{Yang et~al.(2024{\natexlab{b}})Yang, Jimenez, Wettig, Lieret, Yao, Narasimhan, and Press}]{yang2024sweagentagentcomputerinterfacesenable}
John Yang, Carlos~E. Jimenez, Alexander Wettig, Kilian Lieret, Shunyu Yao, Karthik Narasimhan, and Ofir Press. 2024{\natexlab{b}}.
\newblock \href {http://arxiv.org/abs/2405.15793} {SWE-agent: Agent-Computer Interfaces Enable Automated Software Engineering}.

\bibitem[{Yong et~al.(2023)Yong, Schoelkopf, Muennighoff, Aji, Adelani, Almubarak, Bari, Sutawika, Kasai, Baruwa, Winata, Biderman, Raff, Radev, and Nikoulina}]{yong2023bloom1addinglanguagesupport}
Zheng-Xin Yong, Hailey Schoelkopf, Niklas Muennighoff, Alham~Fikri Aji, David~Ifeoluwa Adelani, Khalid Almubarak, M~Saiful Bari, Lintang Sutawika, Jungo Kasai, Ahmed Baruwa, Genta~Indra Winata, Stella Biderman, Edward Raff, Dragomir Radev, and Vassilina Nikoulina. 2023.
\newblock \href {http://arxiv.org/abs/2212.09535} {BLOOM+1: Adding Language Support to BLOOM for Zero-Shot Prompting}.

\bibitem[{Yu et~al.(2024)Yu, Jiang, Shi, Yu, Liu, Zhang, Kwok, Li, Weller, and Liu}]{yu2024metamathbootstrapmathematicalquestions}
Longhui Yu, Weisen Jiang, Han Shi, Jincheng Yu, Zhengying Liu, Yu~Zhang, James~T. Kwok, Zhenguo Li, Adrian Weller, and Weiyang Liu. 2024.
\newblock \href {http://arxiv.org/abs/2309.12284} {MetaMath: Bootstrap Your Own Mathematical Questions for Large Language Models}.

\bibitem[{Yun et~al.(2024)Yun, Zhuang, Fu, Xing, and Zhang}]{yun2024inferenceoptimalmixtureofexpertlargelanguage}
Longfei Yun, Yonghao Zhuang, Yao Fu, Eric~P Xing, and Hao Zhang. 2024.
\newblock \href {http://arxiv.org/abs/2404.02852} {Toward Inference-optimal Mixture-of-Expert Large Language Models}.

\bibitem[{Zadouri et~al.(2023)Zadouri, Üstün, Ahmadian, Ermiş, Locatelli, and Hooker}]{zadouri2023pushingmixtureexpertslimit}
Ted Zadouri, Ahmet Üstün, Arash Ahmadian, Beyza Ermiş, Acyr Locatelli, and Sara Hooker. 2023.
\newblock \href {http://arxiv.org/abs/2309.05444} {Pushing Mixture of Experts to the Limit: Extremely Parameter Efficient MoE for Instruction Tuning}.

\bibitem[{Zellers et~al.(2019)Zellers, Holtzman, Bisk, Farhadi, and Choi}]{zellers2019hellaswagmachinereallyfinish}
Rowan Zellers, Ari Holtzman, Yonatan Bisk, Ali Farhadi, and Yejin Choi. 2019.
\newblock \href {http://arxiv.org/abs/1905.07830} {HellaSwag: Can a Machine Really Finish Your Sentence?}

\bibitem[{Zhang and Sennrich(2019)}]{zhang2019rootmeansquarelayer}
Biao Zhang and Rico Sennrich. 2019.
\newblock \href {http://arxiv.org/abs/1910.07467} {Root Mean Square Layer Normalization}.

\bibitem[{Zhang et~al.(2024{\natexlab{a}})Zhang, Qu, Liu, Zhang, Lin, Yu, Pan, Cheng, Liu, Lin, Yuan, Zheng, Pang, Du, Liang, Ma, Li, Ma, Lin, Benetos, Yang, Zhou, Ma, Liu, Niu, Wang, Que, Liu, Liu, Guo, Gao, Zhou, Zhang, Zhou, Wang, Bai, Zhang, Zhang, Wang, Yang, Zhao, Zhang, Ouyang, Huang, and Chen}]{zhang2024mapneohighlycapabletransparent}
Ge~Zhang, Scott Qu, Jiaheng Liu, Chenchen Zhang, Chenghua Lin, Chou~Leuang Yu, Danny Pan, Esther Cheng, Jie Liu, Qunshu Lin, Raven Yuan, Tuney Zheng, Wei Pang, Xinrun Du, Yiming Liang, Yinghao Ma, Yizhi Li, Ziyang Ma, Bill Lin, Emmanouil Benetos, Huan Yang, Junting Zhou, Kaijing Ma, Minghao Liu, Morry Niu, Noah Wang, Quehry Que, Ruibo Liu, Sine Liu, Shawn Guo, Soren Gao, Wangchunshu Zhou, Xinyue Zhang, Yizhi Zhou, Yubo Wang, Yuelin Bai, Yuhan Zhang, Yuxiang Zhang, Zenith Wang, Zhenzhu Yang, Zijian Zhao, Jiajun Zhang, Wanli Ouyang, Wenhao Huang, and Wenhu Chen. 2024{\natexlab{a}}.
\newblock \href {http://arxiv.org/abs/2405.19327} {MAP-Neo: Highly Capable and Transparent Bilingual Large Language Model Series}.

\bibitem[{Zhang et~al.(2024{\natexlab{b}})Zhang, Zeng, Wang, and Lu}]{zhang2024tinyllamaopensourcesmalllanguage}
Peiyuan Zhang, Guangtao Zeng, Tianduo Wang, and Wei Lu. 2024{\natexlab{b}}.
\newblock \href {http://arxiv.org/abs/2401.02385} {TinyLlama: An Open-Source Small Language Model}.

\bibitem[{Zhang et~al.(2024{\natexlab{c}})Zhang, Gritsch, Gnaneshwar, Guo, Cairuz, Venkitesh, Foerster, Blunsom, Ruder, Ustun, and Locatelli}]{zhang2024bamjustlikethat}
Qizhen Zhang, Nikolas Gritsch, Dwaraknath Gnaneshwar, Simon Guo, David Cairuz, Bharat Venkitesh, Jakob Foerster, Phil Blunsom, Sebastian Ruder, Ahmet Ustun, and Acyr Locatelli. 2024{\natexlab{c}}.
\newblock \href {http://arxiv.org/abs/2408.08274} {BAM! Just Like That: Simple and Efficient Parameter Upcycling for Mixture of Experts}.

\bibitem[{Zhang et~al.(2022)Zhang, Roller, Goyal, Artetxe, Chen, Chen, Dewan, Diab, Li, Lin, Mihaylov, Ott, Shleifer, Shuster, Simig, Koura, Sridhar, Wang, and Zettlemoyer}]{zhang2022optopenpretrainedtransformer}
Susan Zhang, Stephen Roller, Naman Goyal, Mikel Artetxe, Moya Chen, Shuohui Chen, Christopher Dewan, Mona Diab, Xian Li, Xi~Victoria Lin, Todor Mihaylov, Myle Ott, Sam Shleifer, Kurt Shuster, Daniel Simig, Punit~Singh Koura, Anjali Sridhar, Tianlu Wang, and Luke Zettlemoyer. 2022.
\newblock \href {http://arxiv.org/abs/2205.01068} {OPT: Open Pre-trained Transformer Language Models}.

\bibitem[{Zhao et~al.(2023)Zhao, Gu, Varma, Luo, Huang, Xu, Wright, Shojanazeri, Ott, Shleifer, Desmaison, Balioglu, Damania, Nguyen, Chauhan, Hao, Mathews, and Li}]{zhao2023pytorch}
Yanli Zhao, Andrew Gu, Rohan Varma, Liang Luo, Chien-Chin Huang, Min Xu, Less Wright, Hamid Shojanazeri, Myle Ott, Sam Shleifer, Alban Desmaison, Can Balioglu, Pritam Damania, Bernard Nguyen, Geeta Chauhan, Yuchen Hao, Ajit Mathews, and Shen Li. 2023.
\newblock \href {http://arxiv.org/abs/2304.11277} {PyTorch FSDP: Experiences on Scaling Fully Sharded Data Parallel}.

\bibitem[{Zheng et~al.(2024)Zheng, Zhang, Shen, Liu, Lin, Fu, Chen, and Yue}]{zheng2024opencodeinterpreter}
Tianyu Zheng, Ge~Zhang, Tianhao Shen, Xueling Liu, Bill~Yuchen Lin, Jie Fu, Wenhu Chen, and Xiang Yue. 2024.
\newblock Opencodeinterpreter: Integrating code generation with execution and refinement.
\newblock \emph{arXiv preprint arXiv:2402.14658}.

\bibitem[{Zhong et~al.(2024)Zhong, Xia, Chen, and Lewis}]{zhong2024lory}
Zexuan Zhong, Mengzhou Xia, Danqi Chen, and Mike Lewis. 2024.
\newblock \href {http://arxiv.org/abs/2405.03133} {Lory: Fully Differentiable Mixture-of-Experts for Autoregressive Language Model Pre-training}.

\bibitem[{Zhou et~al.(2023{\natexlab{a}})Zhou, Liu, Xu, Iyer, Sun, Mao, Ma, Efrat, Yu, Yu, Zhang, Ghosh, Lewis, Zettlemoyer, and Levy}]{zhou2023lima}
Chunting Zhou, Pengfei Liu, Puxin Xu, Srini Iyer, Jiao Sun, Yuning Mao, Xuezhe Ma, Avia Efrat, Ping Yu, Lili Yu, Susan Zhang, Gargi Ghosh, Mike Lewis, Luke Zettlemoyer, and Omer Levy. 2023{\natexlab{a}}.
\newblock \href {http://arxiv.org/abs/2305.11206} {LIMA: Less Is More for Alignment}.

\bibitem[{Zhou et~al.(2023{\natexlab{b}})Zhou, Lu, Mishra, Brahma, Basu, Luan, Zhou, and Hou}]{zhou2023instructionfollowingevaluationlargelanguage}
Jeffrey Zhou, Tianjian Lu, Swaroop Mishra, Siddhartha Brahma, Sujoy Basu, Yi~Luan, Denny Zhou, and Le~Hou. 2023{\natexlab{b}}.
\newblock \href {http://arxiv.org/abs/2311.07911} {Instruction-Following Evaluation for Large Language Models}.

\bibitem[{Zhou et~al.(2024)Zhou, Du, Huang, Peng, Lan, Huang, Shakeri, So, Dai, Lu, Chen, Le, Cui, Laudon, and Dean}]{zhou2024brainformerstradingsimplicityefficiency}
Yanqi Zhou, Nan Du, Yanping Huang, Daiyi Peng, Chang Lan, Da~Huang, Siamak Shakeri, David So, Andrew Dai, Yifeng Lu, Zhifeng Chen, Quoc Le, Claire Cui, James Laudon, and Jeff Dean. 2024.
\newblock \href {http://arxiv.org/abs/2306.00008} {Brainformers: Trading Simplicity for Efficiency}.

\bibitem[{Zhou et~al.(2022)Zhou, Lei, Liu, Du, Huang, Zhao, Dai, Chen, Le, and Laudon}]{zhou2022mixtureofexperts}
Yanqi Zhou, Tao Lei, Hanxiao Liu, Nan Du, Yanping Huang, Vincent Zhao, Andrew Dai, Zhifeng Chen, Quoc Le, and James Laudon. 2022.
\newblock \href {http://arxiv.org/abs/2202.09368} {Mixture-of-Experts with Expert Choice Routing}.

\bibitem[{Zhuo et~al.(2024)Zhuo, Zebaze, Suppattarachai, von Werra, de~Vries, Liu, and Muennighoff}]{zhuo2024astraios}
Terry~Yue Zhuo, Armel Zebaze, Nitchakarn Suppattarachai, Leandro von Werra, Harm de~Vries, Qian Liu, and Niklas Muennighoff. 2024.
\newblock \href {http://arxiv.org/abs/2401.00788} {Astraios: Parameter-Efficient Instruction Tuning Code Large Language Models}.

\bibitem[{Zoph et~al.(2022)Zoph, Bello, Kumar, Du, Huang, Dean, Shazeer, and Fedus}]{zoph2022stmoe}
Barret Zoph, Irwan Bello, Sameer Kumar, Nan Du, Yanping Huang, Jeff Dean, Noam Shazeer, and William Fedus. 2022.
\newblock \href {http://arxiv.org/abs/2202.08906} {ST-MoE: Designing Stable and Transferable Sparse Expert Models}.

\bibitem[{Zuo et~al.(2022)Zuo, Liu, Jiao, Kim, Hassan, Zhang, Zhao, and Gao}]{zuo2022tamingsparselyactivatedtransformer}
Simiao Zuo, Xiaodong Liu, Jian Jiao, Young~Jin Kim, Hany Hassan, Ruofei Zhang, Tuo Zhao, and Jianfeng Gao. 2022.
\newblock \href {http://arxiv.org/abs/2110.04260} {Taming Sparsely Activated Transformer with Stochastic Experts}.

\bibitem[{Üstün et~al.(2024)Üstün, Aryabumi, Yong, Ko, D'souza, Onilude, Bhandari, Singh, Ooi, Kayid, Vargus, Blunsom, Longpre, Muennighoff, Fadaee, Kreutzer, and Hooker}]{üstün2024aya}
Ahmet Üstün, Viraat Aryabumi, Zheng-Xin Yong, Wei-Yin Ko, Daniel D'souza, Gbemileke Onilude, Neel Bhandari, Shivalika Singh, Hui-Lee Ooi, Amr Kayid, Freddie Vargus, Phil Blunsom, Shayne Longpre, Niklas Muennighoff, Marzieh Fadaee, Julia Kreutzer, and Sara Hooker. 2024.
\newblock \href {http://arxiv.org/abs/2402.07827} {Aya Model: An Instruction Finetuned Open-Access Multilingual Language Model}.

\end{thebibliography}
\bibliographystyle{acl_natbib}

\newpage
\appendix

\section{Artifacts}
\label{sec:artifacts}

\begin{table}[htbp]
\centering
\begin{tabular}{l|l}
\toprule
\textbf{Artifact} & \textbf{Public link} \\
\midrule
\modelsmall{} & \url{https://hf.co/allenai/OLMoE-1B-7B-0924} \\
\modelsmalldpo{} & \url{https://hf.co/allenai/OLMoE-1B-7B-0924-Instruct} \\
\modelsmallsft{} & \url{https://hf.co/allenai/OLMoE-1B-7B-0924-SFT} \\
\data{} & \url{https://hf.co/datasets/allenai/OLMoE-mix-0924} \\
\multirow{2}{*}{\textbf{SFT data}} & \href{https://hf.co/datasets/allenai/tulu-v3.1-mix-preview-4096-OLMoE}{\texttt{https://hf.co/datasets/allenai/}} \\
& \href{https://hf.co/datasets/allenai/tulu-v3.1-mix-preview-4096-OLMoE}{\texttt{tulu-v3.1-mix-preview-4096-OLMoE}} \\
\multirow{2}{*}{\textbf{KTO/DPO data}} & \href{https://hf.co/datasets/allenai/ultrafeedback_binarized_cleaned}{\texttt{https://hf.co/datasets/allenai/}} \\
& \href{https://hf.co/datasets/allenai/ultrafeedback_binarized_cleaned}{\texttt{ultrafeedback\_binarized\_cleaned}} \\
\textbf{Code} & \url{https://github.com/allenai/OLMoE} \\
\multirow{2}{*}{\textbf{Logs}} & \href{https://wandb.ai/ai2-llm/olmoe/reports/OLMoE-1B-7B-0924--Vmlldzo4OTcyMjU3}{\texttt{https://wandb.ai/ai2-llm/olmoe/reports/}} \\
& \href{https://wandb.ai/ai2-llm/olmoe/reports/OLMoE-1B-7B-0924--Vmlldzo4OTcyMjU3}{\texttt{OLMoE-1B-7B-0924--Vmlldzo4OTcyMjU3}} \\
\midrule
BLOOM-7B & \url{https://hf.co/bigscience/bloom-7b1} \\
DeepSeekMoE-3B-16B & \url{https://hf.co/deepseek-ai/deepseek-moe-16b-base} \\
DeepSeekMoE-3B-16B+chat & \url{https://hf.co/deepseek-ai/deepseek-moe-16b-chat} \\
DeepSeekV2-2B-16B & \url{https://hf.co/deepseek-ai/DeepSeek-V2-Lite} \\
DCLM-1B & \url{https://hf.co/TRI-ML/DCLM-1B} \\
DCLM-7B & \url{https://hf.co/TRI-ML/DCLM-7B} \\
Falcon-7B & \url{https://hf.co/tiiuae/falcon-7b} \\
Gemma2-3B & \url{https://hf.co/google/gemma-2-2b} \\
Gemma2-9B & \url{https://hf.co/google/gemma-2-9b} \\
JetMoE-2B-9B & \url{https://hf.co/jetmoe/jetmoe-8b} \\
JetMoE-2B-9B+SFT & \url{https://hf.co/jetmoe/jetmoe-8b-sft} \\
JetMoE-2B-9B+Chat & \url{https://hf.co/jetmoe/jetmoe-8b-chat} \\
Llama-7B & \url{https://hf.co/huggyllama/llama-7b} \\
Llama2-7B & \url{https://hf.co/meta-llama/Llama-2-7b-hf} \\
Llama3.1-8B & \url{https://hf.co/meta-llama/Meta-Llama-3.1-8B} \\
MPT-7B & \url{https://hf.co/mosaicml/mpt-7b} \\
Mistral-7B & \url{https://hf.co/mistralai/Mistral-7B-v0.1} \\
Mixtral-8x7B & \url{https://hf.co/mistralai/Mixtral-8x7B-v0.1} \\
OLMo-1B (0724) & \url{https://hf.co/allenai/OLMo-1B-0724-hf} \\
OLMo-7B (0724) & \url{https://hf.co/allenai/OLMo-7B-0724-hf} \\
OpenMoE-3B-9B & \url{https://hf.co/OrionZheng/openmoe-8b} \\
Pythia-7B & \url{https://hf.co/EleutherAI/pythia-6.9b} \\
Qwen1.5-3B-14B & \url{https://hf.co/Qwen/Qwen1.5-MoE-A2.7B} \\
Qwen1.5-3B-14B+Chat & \url{https://hf.co/Qwen/Qwen1.5-MoE-A2.7B-Chat} \\
StableLM2-2B & \url{https://hf.co/stabilityai/stablelm-2-1_6b} \\
TinyLlama-1B & 
\url{https://hf.co/TinyLlama/TinyLlama_v1.1} \\
\bottomrule
\end{tabular}
\vspace{1em}
\caption{\textbf{All artifacts released and used in this work.} We point from the name used for a given artifact in this work (e.g. \autoref{fig:overview}) to the URL where it can be obtained.}
\label{tab:artifacts}
\end{table}

\section{Training Configuration}
\label{sec:config}

\paragraph{Pretraining} We display the pretraining hyperparameter configuration of \modelsmall{} in \autoref{sec:config} comparing with other relevant models. We follow \citet{groeneveld2024olmo} using the AdamW optimizer~\citep{loshchilov2019decoupled} with ZeRO~\citep{rajbhandari2020zero} via PyTorch FSDP~\citep{zhao2023pytorch} and mixed-precision training~\citep{micikevicius2018mixed}. Our main model settings differing from \citet{groeneveld2024olmo} are: \textbf{(1) MoE-related changes:} \modelsmall{} is a sparsely activated decoder-only transformer~\citep{vaswani2023attention} using dropless Mixture-of-Experts~\citep{gale2022megablocksefficientsparsetraining}. Unlike most prior MoEs, we use a high granularity~\citep{dai2024deepseekmoeultimateexpertspecialization,krajewski2024scaling} with 64 small experts with an FFN dimension of just 1,024 rather than a few large experts. We further use two auxiliary losses: router z-loss~\citep{zoph2022stmoe} and load balancing loss~\citep{shazeer2017outrageously}. \textbf{(2) Stability improvements:} (a) We use a truncated normal initialization with a standard deviation of 0.02 and a minimum (maximum) cut-off of -0.06 (0.06) corresponding to three standard deviations. (b) We use QK normalization~\citep{chameleonteam2024chameleon,mehta2024openelm,dehghani2023scaling}. (c) We use RMSNorm~\citep{zhang2019rootmeansquarelayer} instead of the non-parametric LayerNorm used in \citet{groeneveld2024olmo}. \textbf{(3) Performance improvements:} Besides some of the stability improvements which also impact performance, we also reduce the AdamW epsilon to 1.0E-08 from the 1.0E-05 used in \citet{groeneveld2024olmo} to speed up convergence. Finally, we train \modelsmall{} for significantly longer than all prior OLMo models amounting to 5T tokens and thus more than one epoch (1.3) following \citet{muennighoff2023scaling}. We shuffle the pretraining dataset before starting the second epoch. For the final 100B tokens, we decay the learning rate linearly from 5.0E-04 to 0. We experiment with many of these settings in \autoref{sec:ablations}.

\paragraph{Adaptation} For finetuning we use Open Instruct~\citep{wang2023far, ivison2023camels}.\footnote{Code: \url{https://github.com/allenai/open-instruct}} We filter all SFT samples to a length of fewer than 4096 tokens to match the sequence length of the model. Following \citet{muennighoff2024generativerepresentationalinstructiontuning}, we aggregate loss at the token level during SFT to improve performance on long generative tasks, such as AlpacaEval. We finetune in BF16 with a global batch size of 128 (4 H100 nodes with 8 GPUs each, a per device batch size of 2, and 2 gradient accumulation steps). We train for 2 epochs with a constant learning rate of 2.0E-5. For DPO~\citep{rafailov2023direct}, we reduce the global batch size to 32 (4 H100 nodes with 8 GPUs each and a per device batch size of 1). We train for 3 epochs with a learning rate of 5.0E-7 and a DPO beta of 0.1. Our adapted models are built on top of our annealed checkpoint, and we include the load balancing loss during both SFT and DPO based on our experiments in \autoref{sec:adapt}. Our preference tuning recipe is heavily optimized for DPO based on extensive experiments by \citet{ivison2023camels}, thus for KTO~\citep{ethayarajh2024kto} we experiment with a few settings in \autoref{sec:addablations}. Our final KTO adaptation uses the same hyperparameters as DPO, except that we use the RMSProp optimizer instead of Adam, which we use for SFT and DPO, and that we reduce the training duration to 1.3 epochs (5,000 steps) for KTO instead of the 3 epochs used for DPO.

\paragraph{Hardware} We pretrain \modelsmall{} on 256 H100 GPUs for approximately 10 days with NV-link interconnect across GPUs and InfiniBand interconnect across nodes. We also use H100 GPUs for all our experiments but some use a cluster with GCP TCPx interconnect across nodes instead. For adaptation, we use 32 H100 GPUs for 33 hours to instruction tune and for another 14 hours to preference tune via DPO. For KTO adaptation we use 8 H100 GPUs for 30 hours instead.

\begin{table}[htbp]
\centering
\begin{tabular}{l|lll|l}
\toprule
& \modelsmall{} & \textbf{JetMoE} & \textbf{OpenMoE} & \textbf{OLMo-1B (0724)} \\
\midrule
Dimension & 2,048 & 2,048 & 2,048 & 2,048 \\
Activation & SwiGLU & SwiGLU & SwiGLU & SwiGLU \\
\rowcolor{lightOlmoeYellow}
FFN dimension & 1,024 & 5,632 & 8,192 & 8,192 \\
Vocab size & 50,304 & 32,000 & 256,384 & 50,304 \\
Attn heads & 16 & 16 & 24 & 16 \\
Num layers & 16 & 24 & 32 & 16 \\
\rowcolor{lightOlmoeYellow}
Layer norm type & RMSNorm & RMSNorm & RMSNorm & non-parametric \\
Layer norm eps & 1.0E-05 & 1.0E-05 & 1.0E-06 & 1.0E-05 \\
\rowcolor{lightOlmoeYellow}
QK-Norm & yes & no & no & no \\
Pos emb. & RoPE & RoPE & RoPE & RoPE \\
RoPE $\theta$ & 10,000 & 10,000 & 10,000 & 10,000 \\
Attention variant & full & MoA & full & full \\
Biases & - & MLP \& Attn & - & - \\
Weight tying\nocite{press2017usingoutputembeddingimprove} & no & yes & no & no \\
\rowcolor{lightOlmoeYellow}
Init dist & trunc normal & ? & ? & normal \\
\rowcolor{lightOlmoeYellow}
Init std & 0.02 & 0.02 & varies & varies \\
\rowcolor{lightOlmoeYellow}
Init trunc & 3$\times$std & - & - & - \\
\rowcolor{lightOlmoeYellow}
MoE layers & Every & Every & Every 6th & - \\
\rowcolor{lightOlmoeYellow}
MoE layer type & dMoE & dMoE & ST-MoE & - \\
\rowcolor{lightOlmoeYellow}
\# Experts & 64 & 8 & 32 & 1 \\
\rowcolor{lightOlmoeYellow}
\# Activated & 8 & 2 & 2 & 1 \\
\midrule
\# Vocab params & 103M & 66M & 525M & 103M \\
\# Active params & 1.3B & 2.2B & 2.6B & 1.3B \\
\rowcolor{lightOlmoeYellow}
\# Total params & 6.9B & 8.5B & 8.7B & 1.3B \\
\midrule
Sequence length & 4,096 & 4,096 & 2,048 & 4,096 \\
\rowcolor{lightOlmoeYellow}
Batch size (samples) & 1,024 & 1,024 & 2,048 & 512 \\
\rowcolor{lightOlmoeYellow}
Batch size (tokens) & $\sim$4M & $\sim$4M & $\sim$4M & $\sim$2M \\
warmup steps & 2,500 & 2,500 & 10,000 & 2,000 \\
peak LR & 4.0E-04 & 5.0E-04 & 0.01 & 4.0E-04 \\
minimum LR & 4.0E-05 & 5.0E-05 & - & 4.0E-05 \\
optimizer & AdamW & AdamW & Adafactor & AdamW \\
weight decay & 0.1 & 0.1 & 0.0 & 0.1 \\
beta1 & 0.9 & ? & 0.9 & 0.9 \\
beta2 & 0.95 & ? & - & 0.95 \\
\rowcolor{lightOlmoeYellow}
AdamW epsilon & 1.0E-08 & ? & - & 1.0E-05 \\
LR schedule & cosine & WSD & Inv Sq Root & cosine \\
gradient clipping & global 1.0 & global 1.0 & global 1.0 & global 1.0 \\
gradient reduce dtype & FP32 & ? & ? & FP32 \\
optimizer state dtype & FP32 & ? & ? & FP32 \\
\rowcolor{lightOlmoeYellow}
LBL weight & 0.01 & 0.01 & 0.01 & - \\
\rowcolor{lightOlmoeYellow}
Router z-loss weight & 0.001 & 0.001 & 0.0001 & - \\
\rowcolor{lightOlmoeYellow}
Pretraining tokens & 5,033B & 1,000B & 1,100B & 2,000B \\
\rowcolor{lightOlmoeYellow}
Annealing tokens & 100B & 250B & - & 50B \\
Annealing schedule & linear & - & - & linear \\
Annealing min LR & 0 & - & - & 0 \\   
\bottomrule
\end{tabular}
\vspace{1em}
\caption{\textbf{Pretraining hyperparameters of \modelsmall{} and comparable models trained from scratch.} We highlight rows where \modelsmall{} differs from OLMo-1B. Active params include vocab params. ``?'' = undisclosed settings, FFN = feed-forward network, Attn = Attention, LR = learning rate, WSD = Weight-Stable-Decay \citep{hu2024minicpm}, LBL = load balancing loss, Inv Sq Root = Inverse Square Root decay \citep{shazeer2018adafactoradaptivelearningrates}, trunc = truncation, std = standard deviation, ``varies'' = stds that are layer or weight-dependent.}
\label{tab:hp}
\end{table}

\FloatBarrier

\section{Evaluation Setup}
\label{sec:evalsetup}

\begin{table}[htbp]
\centering
\setlength{\tabcolsep}{5.3pt}
\begin{tabular}{l|cccc|cccc}
\toprule
\multirow{3}{*}{Dataset ($\downarrow$)} & \multicolumn{4}{c|}{During pretraining} & \multicolumn{4}{c}{After pretraining (OLMES~\citep{gu2024olmesstandardlanguagemodel})} \\ 
& Format & Shot & Norm & Split & Format & Shot & CF Norm & Split \\ 
\midrule
ARC-C~\citep{clark2018thinksolvedquestionanswering} & CF & 0 & char & val & max(MCF,CF) & 5 & pmi & test \\
ARC-E~\citep{clark2018thinksolvedquestionanswering} & CF & 0 & none & val & max(MCF,CF) & 5 & char & test \\ 
BoolQ~\citep{clark2019boolqexploringsurprisingdifficulty} & CF & 0 & none & val & max(MCF,CF) & 5 & none & val \\
COPA~\citep{gordon2012semeval} & CF & 0 & none & val & - & - & - & - \\
CSQA~\citep{talmor2019commonsenseqaquestionansweringchallenge} & CF & 0 & char & val & max(MCF,CF) & 5 & pmi & val \\
HellaSwag~\citep{zellers2019hellaswagmachinereallyfinish} & CF & 0 & char & val & max(MCF,CF) & 5 & char & val \\
MMLU~\citep{hendrycks2021measuringmassivemultitasklanguage} & MCF & 5 & none & val & max(MCF,CF) & 5 & char & test \\ 
MMLU Var & CF & 0-5 & char & val & - & - & - & - \\
OBQA~\citep{mihaylov2018suitarmorconductelectricity} & CF & 0 & char & val & max(MCF,CF) & 5 & pmi & test \\
PIQA~\citep{bisk2019piqareasoningphysicalcommonsense} & CF & 0 & char & val & max(MCF,CF) & 5 & char & val \\
SciQ~\citep{welbl2017crowdsourcingmultiplechoicescience} & CF & 0 & none & val & - & - & - & - \\ 
SocialIQA~\citep{sap2019socialiqacommonsensereasoningsocial} & CF & 0 & char & val & max(MCF,CF) & 5 & char & val \\
Winogrande~\citep{sakaguchi2019winograndeadversarialwinogradschema} & CF & 0 & none & val & max(MCF,CF) & 5 & none & val \\
\bottomrule
\end{tabular}
\vspace{.5em}
\caption{\textbf{Summary of downstream evaluation during and after pretraining (OLMES).} ARC-C and ARC-E refer to ARC-Challenge and -Easy, CSQA=CommonsenseQA, OBQA=OpenBookQA, CF=Completion/Cloze formulation, MCF=Multiple-choice formulation, pmi=pointwise-mutual-information, char=per-character, Var=variants referring to the use of few-shots varying from 0-5.}
\label{tab:evalsetup}
\end{table}

\begin{figure*}[htbp]
\centering
\begin{center}
\includegraphics[width=\textwidth]{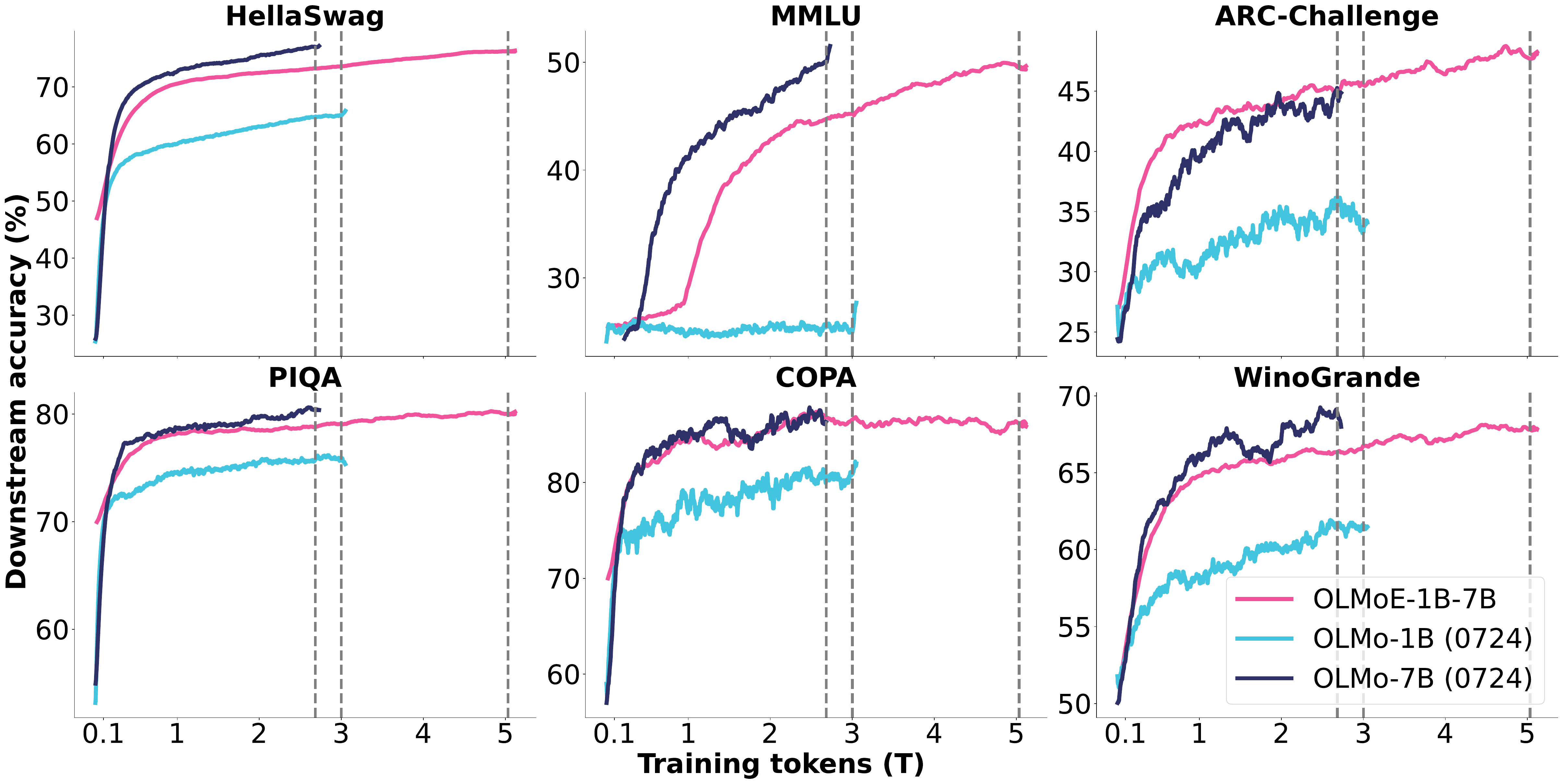}
\caption{\textbf{Evaluation of \modelsmall{} and the current best OLMo models during pretraining.} Grey vertical lines correspond to where the respective run enters annealing with the 1st line being for OLMo-7B, the 2nd for OLMo-1B, and the third for \modelsmall{}. \autoref{fig:trainingevalflops} is a version of this plot with training FLOPs as the x-axis. More results, logs, and configurations: \url{https://wandb.ai/ai2-llm/olmoe/reports/Plot-OLMoE-1B-7B-vs-OLMo-7B-vs-OLMo-1B--Vmlldzo4OTcyMjEz}}
\label{fig:trainingevaltokens}
\end{center}
\end{figure*}

\paragraph{During pretraining} We evaluate using a similar in-loop evaluation setup as \citet{groeneveld2024olmo}, with the addition of more tasks such as CommonsenseQA, PIQA, and different implementations of MMLU. Following \citet{groeneveld2024olmo}, for the majority of the tasks, we perform 0-shot evaluation using the Completion/Cloze Formulation (CF), ranking each answer string using language model probabilities. In terms of probability normalization, there is either no normalization (none) or normalization by the number of characters in the answer (char) when ranking solely based on probability may heavily favor shorter answers~\citep{brown2020language}. For MMLU, the in-loop evaluation also includes a setup where we increase the total number of instances by including a range of 0-shot to 5-shot setups together as we found this provides smoother trends as the training proceeds (``MMLU Var''). We also include the Multiple-Choice Formulation (MCF) version of MMLU, scoring prediction of answer labels like A/B/C/D, which generally starts to rise only later in training as models only gain the multiple-choice capability later (at around 1T tokens for \modelsmall{} in \autoref{fig:trainingevaltokens}). We also evaluate perplexity on selected validation sets from Paloma~\citep{magnusson2023palomabenchmarkevaluatinglanguage,reid2022m2d2massivelymultidomainlanguage,gao2020pile800gbdatasetdiverse,soldaini2024dolma,liang2023holisticevaluationlanguagemodels,merity2016pointersentinelmixturemodels}. All code used for evaluation during pretraining is at \url{https://github.com/allenai/OLMo/tree/61ac104d616ec5435db225796e5c7532c9abd95a/olmo/eval}.

\paragraph{After pretraining - OLMES} We perform evaluations following the OLMES evaluation standard~\citep{gu2024olmesstandardlanguagemodel}, with the suite of tasks in the original paper. OLMES (Open Language Model Evaluation Standard) is a standard for reproducible LM evaluations that is open, practical, and documented, providing recommendations guided by experiments and results from the literature~\citep{biderman2024lessons,eval-harness,groeneveld2023catwalkunifiedlanguagemodel}. It is designed to support comparisons between smaller base models that require the Cloze formulation of multiple-choice questions against larger models that can utilize the Multiple-choice formulation. To make our evaluations reproducible, we follow OLMES in prompt formatting, choice of in-context examples, probability normalization, task formulation, as well as all other details. We summarize this setup in \autoref{tab:eval} and refer to \citet{gu2024olmesstandardlanguagemodel} for more details.

\paragraph{After pretraining - DCLM} For results on the DCLM tasks~\citep{li2024datacomplm} in \autoref{tab:dclm}, we precisely follow their setup using the evaluation code released by the authors at \url{https://github.com/mlfoundations/dclm}. ``Core'' results are the \texttt{low variance} tasks in their evaluation code, while ``Extended'' corresponds to the \texttt{heavy} tasks.

\paragraph{After adaptation} After supervised finetuning and direct preference optimization, we evaluate models using a subset of the evaluations and the same overall setup used in \citet{ivison2023camels} and \citet{wang2023far}. We cover a wide range of model capabilities in our evaluation suite including coding (HumanEval \cite{chen2021evaluating}), general and mathematical reasoning (Big Bench Hard \cite{suzgun2022challenging}, GSM8k \cite{cobbe2021training}), world knowledge (MMLU), general instruction following (AlpacaEval 1.0 \cite{alpaca_eval}, not the length-controlled variant~\citep{dubois2024lengthcontrolledalpacaevalsimpleway}), precise instruction following (IFEval \cite{zhou2023instructionfollowingevaluationlargelanguage}) and safety (XSTest \cite{röttger2024xstest}). We refer to \citet{wang2023far} for more details on each benchmark.

\section{Openness of Models}
\label{sec:openness}

We list the openness of various models summarized in \autoref{fig:overview}. We exclude Switch Transformers~\citep{fedus2022switch}, as it was published over three years ago and is very different from more recent MoE models (MLM objective, Encoder-decoder, etc.).

\paragraph{Grok-86B-314B~\citep{grok}}
\begin{itemize}
\item \textbf{\emoji{olmoe_checkmark} Model:} Their model is licensed under the open-source Apache 2.0 license.
\item \textbf{\emoji{olmoe_cross} Data:} Unavailable.
\item \textbf{\emoji{olmoe_cross} Code:} Unavailable.
\item \textbf{\emoji{olmoe_cross} Logs:} Unavailable.
\end{itemize}

\paragraph{Mixtral-39B-141B and Mixtral-13B-42B~\citep{jiang2024mixtral}}
\begin{itemize}
\item \textbf{\emoji{olmoe_checkmark} Model:} Their model is licensed under the open-source Apache 2.0 license.
\item \textbf{\emoji{olmoe_cross} Data:} Unavailable.
\item \textbf{\emoji{olmoe_cross} Code:} Unavailable.
\item \textbf{\emoji{olmoe_cross} Logs:} Unavailable.
\end{itemize}

\paragraph{DBRX-36B-132B~\citep{dbrx}}
\begin{itemize}
\item \textbf{\emoji{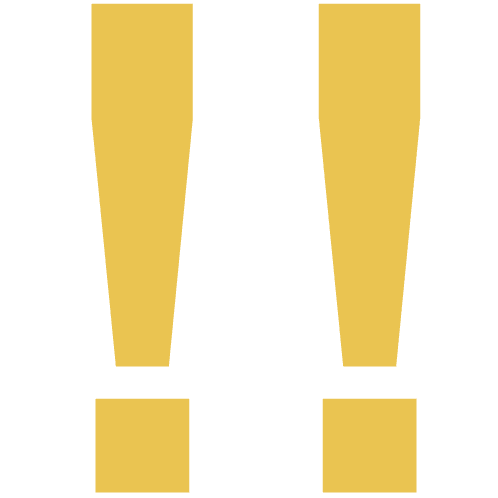} Model:} The model is licensed under a custom non-open-source license\footnote{\url{https://www.databricks.com/legal/open-model-license}} with additional use-case restrictions.\footnote{\url{https://www.databricks.com/legal/acceptable-use-policy-open-model}} 
\item \textbf{\emoji{olmoe_cross} Data:} Unavailable. 
\item \textbf{\emoji{olmoe_cross} Code:} They use closed-source custom adaptations of their public libraries LLM-foundry, composer, and megablocks.\footnote{\url{https://github.com/databricks/dbrx}}
\item \textbf{\emoji{olmoe_cross} Logs:} Unavailable.
\end{itemize}

\paragraph{Skywork-MoE-22B-146B~\citep{wei2024skyworkmoe}}
\begin{itemize}
\item \textbf{\emoji{olmoe_warning} Model:} The model is licensed under a custom non-open-source license.\footnote{\url{https://github.com/SkyworkAI/Skywork/blob/main/Skywork\%20Community\%20License.pdf}}
\item \textbf{\emoji{olmoe_cross} Data:} Unavailable.
\item \textbf{\emoji{olmoe_cross} Code:} Unavailable.
\item \textbf{\emoji{olmoe_cross} Logs:} Unavailable.
\end{itemize}

\paragraph{DeepSeekV2-21B-236B~\citep{deepseekai2024deepseekv2} and DeepSeekMoE-3B-14B~\citep{dai2024deepseekmoeultimateexpertspecialization}}
\begin{itemize}
\item \textbf{\emoji{olmoe_warning} Model:} The models are licensed under custom non-open-source licenses.\footnote{\url{https://github.com/deepseek-ai/DeepSeek-MoE/blob/main/LICENSE-MODEL} and \url{https://github.com/deepseek-ai/DeepSeek-V2/blob/main/LICENSE-MODEL}}
\item \textbf{\emoji{olmoe_cross} Data:} Unavailable.
\item \textbf{\emoji{olmoe_cross} Code:} Unavailable.
\item \textbf{\emoji{olmoe_cross} Logs:} Unavailable.
\end{itemize}

\paragraph{Arctic-17B-480B~\citep{arctic}}
\begin{itemize}
\item \textbf{\emoji{olmoe_checkmark} Model:} The model is licensed under the open-source Apache 2.0 license. 
\item \textbf{\emoji{olmoe_warning} Data:} They describe their mixture but do not release it.\footnote{\url{https://medium.com/snowflake/snowflake-arctic-cookbook-series-arctics-approach-to-data-b81a8a0958bd}}
\item \textbf{\emoji{olmoe_cross} Code:} Unavailable. 
\item \textbf{\emoji{olmoe_cross} Logs:} Unavailable.
\end{itemize}

\paragraph{Qwen2-14B-57B~\citep{qwen_moe}}
\begin{itemize}
\item \textbf{\emoji{olmoe_checkmark} Model:} The model is licensed under the open-source Apache 2.0 license.
\item \textbf{\emoji{olmoe_cross} Data:} Unavailable.
\item \textbf{\emoji{olmoe_cross} Code:} Unavailable.
\item \textbf{\emoji{olmoe_cross} Logs:} Unavailable.
\end{itemize}

\paragraph{Jamba-12B-52B~\citep{lieber2024jambahybridtransformermambalanguage}}
\begin{itemize}
\item \textbf{\emoji{olmoe_checkmark} Model:} The model is licensed under the open-source Apache 2.0 license.
\item \textbf{\emoji{olmoe_cross} Data:} Unavailable.
\item \textbf{\emoji{olmoe_cross} Code:} Unavailable.
\item \textbf{\emoji{olmoe_cross} Logs:} Unavailable.
\end{itemize}

\paragraph{Qwen1.5-3B-14B~\citep{qwen_moe}}
\begin{itemize}
\item \textbf{\emoji{olmoe_warning} Model:} The model is licensed under a custom non-open-source license.\footnote{\url{https://hf.co/Qwen/Qwen1.5-MoE-A2.7B/blob/main/LICENSE}}
\item \textbf{\emoji{olmoe_cross} Data:} Unavailable.
\item \textbf{\emoji{olmoe_cross} Code:} Unavailable.
\item \textbf{\emoji{olmoe_cross} Logs:} Unavailable.
\end{itemize}

\paragraph{JetMoE-2B-9B~\citep{shen2024jetmoe}}
\begin{itemize}
\item \textbf{\emoji{olmoe_checkmark} Model:} The model is licensed under the open-source Apache 2.0 license.
\item \textbf{\emoji{olmoe_warning} Data:} They describe their mixture but do not release it.
\item \textbf{\emoji{olmoe_warning} Code:} They make their fork of megablocks publicly available,\footnote{\url{https://github.com/yikangshen/megablocks}} however, their Megatron-LM training code is not available.\footnote{\url{https://hf.co/jetmoe/jetmoe-8b/discussions/5\#661ee52c03251697a0b155cc}}
\item \textbf{\emoji{olmoe_cross} Logs:} Unavailable.
\end{itemize}

\paragraph{OpenMoE-2B-9B~\citep{xue2024openmoe}}
\begin{itemize}
\item \textbf{\emoji{olmoe_checkmark} Model:} The model is licensed under the open-source Apache 2.0 license.
\item \textbf{\emoji{olmoe_checkmark} Data:} They make scripts for recreating their data available.
\item \textbf{\emoji{olmoe_checkmark} Code:} They make their code available.\footnote{\url{https://github.com/XueFuzhao/OpenMoE/tree/main?tab=readme-ov-file\#training-with-tpugpu}}
\item \textbf{\emoji{olmoe_cross} Logs:} Unavailable.
\end{itemize}

\paragraph{\modelsmall{}}
\begin{itemize}
\item \textbf{\emoji{olmoe_checkmark} Model:} The model is licensed under the open-source Apache 2.0 license.
\item \textbf{\emoji{olmoe_checkmark} Data:} The data is licensed under the open-source ODC-By 1.0 license.
\item \textbf{\emoji{olmoe_checkmark} Code:} The code is licensed under the open-source Apache 2.0 license.
\item \textbf{\emoji{olmoe_checkmark} Logs:} Logs are available with the same open-source license as the code (Apache 2.0).
\end{itemize}

\FloatBarrier

\section{Additional Evaluation}
\label{sec:addeval}

\begin{figure*}[htbp]
\centering
\begin{center}
\includegraphics[width=\textwidth]{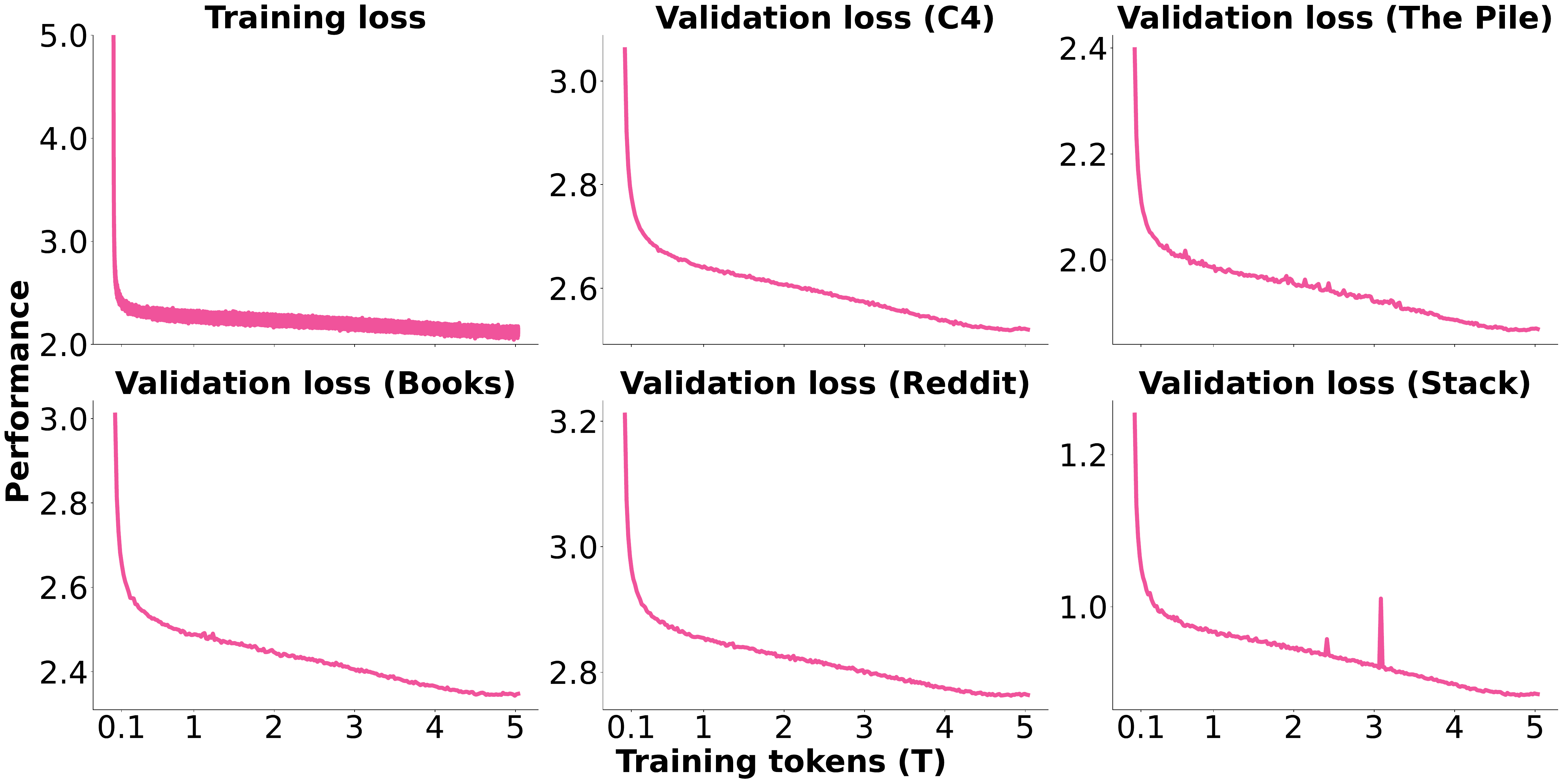}
\caption{\textbf{Losses of \modelsmall{} during training.} The Books, Reddit, and Stack~\citep{kocetkov2022stack3tbpermissively} datasets are from Dolma 1.7~\citep{soldaini2024dolma} via Paloma~\citep{magnusson2023palomabenchmarkevaluatinglanguage}. More results, logs, and configurations: \url{https://wandb.ai/ai2-llm/olmoe/reports/Plot-OLMoE-1B-7B--Vmlldzo4OTcyMjU3}}
\label{fig:loss}
\end{center}
\end{figure*}

\begin{table}
\setlength\tabcolsep{2pt} 
\centering
\begin{tabular}{l|ccccccccccc}
\toprule
\footnotesize{\bf{Model}} & \footnotesize{\bf{ARC\_C}} & \footnotesize{\bf{ARC\_E}} & \footnotesize{\bf{BoolQ}} & \footnotesize{\bf{CSQA}} & \footnotesize{\bf{HSwag}} & \footnotesize{\bf{MMLU}} & \footnotesize{\bf{OBQA}} & \footnotesize{\bf{PIQA}} & \footnotesize{\bf{SIQA}} & \footnotesize{\bf{WinoG}} & \footnotesize{\bf{Avg}}\\
\midrule
\multicolumn{12}{c}{LMs with $\sim$7-9B active parameters} \\
\midrule    
Mistral-7B & 78.6$^\dagger$ & 90.8$^\dagger$ & 89.3 & 72.4$^\dagger$ & 83.0 & 64.0$^\dagger$ & 80.6$^\dagger$ & 82.8 & 71.3$^\dagger$ & 77.9 & 79.1 \\
OLMo-7B (0724) & 68.0$^\dagger$ & 85.7$^\dagger$ & 85.3 & 85.4$^\dagger$ & 80.5 & 54.9$^\dagger$ & 67.6$^\dagger$ & 79.3 & 76.1$^\dagger$ & 73.2 & 75.6 \\
DCLM-7B & 79.8$^\dagger$ & 92.3$^\dagger$ & 87.0 & 77.0 & 82.3 & 64.4$^\dagger$ & 79.6$^\dagger$ & 80.1 & $71.2^\dagger$ & 77.3 & 79.1 \\
Llama2-7B & 54.2 & 84.0 & 86.1 & 74.2 & 78.9 & 46.2$^\dagger$ & 57.8 & 77.5 & 59.6 & 71.7 & 69.0 \\
Llama3.1-8B & 79.5$^\dagger$ & 91.7$^\dagger$ & 88.5 & 74.3$^\dagger$ & 81.6 & 66.9$^\dagger$ & 78.6$^\dagger$ & 81.1 & 71.4$^\dagger$ & 76.6 & 79.0\\
Gemma2-9B & 89.5$^\dagger$ & 95.5$^\dagger$ & 89.4 & 78.8$^\dagger$ & 87.3$^\dagger$ & 70.6$^\dagger$ & 88.4$^\dagger$ & 86.1$^\dagger$ & 76.0$^\dagger$ & 78.8 & 84.0 \\
\midrule
\multicolumn{12}{c}{LMs with $\sim$2-3B active parameters} \\
\midrule
StableLM-2B & 50.6$^\dagger$ & 75.3 & 82.3 & 70.4$^\dagger$ & 70.3 & 40.4$^\dagger$ & 56.6$^\dagger$ & 75.6 & 64.3$^\dagger$ & 65.8 & 65.1 \\
Gemma2-3B & 67.5$^\dagger$ & 84.3$^\dagger$ & 83.6 & 66.4$^\dagger$ & 74.6 & 53.3$^\dagger$ & 68.8$^\dagger$ & 78.5 & 64.7$^\dagger$ & 71.8 & 71.4\\
JetMoE-2B-9B & 61.4$^\dagger$ & 81.9$^\dagger$ & 85.7 & 75.3$^\dagger$ & 81.7 & 49.1$^\dagger$ & 68.0$^\dagger$ & 80.3 & 71.3$^\dagger$ & 70.7 & 72.5 \\
OpenMoE-3B-9B & 29.3 & 50.6 & 63.2 & 21.5 & 44.4 & 27.4 & 34.6 & 63.3 & 42.9 & 51.9$^\dagger$ & 42.9 \\
DeepSeek-3B-16B & 53.4 & 82.7 & 81.9 & 72.7 & 80.4 & 45.5$^\dagger$ & 58.4 & 80.1 & 59.9 & 73.2 & 68.8 \\
DeepSeekV2-2B-16B & 74.0$^\dagger$ & 88.9$^\dagger$ & 84.7 & 73.8 & 81.9 & 58.8$^\dagger$ & 72.4$^\dagger$ & 80.2 & 69.1$^\dagger$ & 74.0 & 75.8 \\
Llama3.2-3B & 69.6$^\dagger$ & 85.1$^\dagger$ & 78.3 & 69.0 & 77.0 & 57.8$^\dagger$ & 67.2$^\dagger$ & 77.4 & 64.9$^\dagger$ & 69.9 & 71.6 \\
Qwen1.5-3B-14B & 77.4$^\dagger$ & 91.6$^\dagger$ & 85.0 & 81.4$^\dagger$ & 80.0 & $62.4^\dagger$ & 80.6$^\dagger$ & 81.0 & 74.1$^\dagger$ & 72.3 & 78.6 \\
\midrule
\multicolumn{12}{c}{LMs with $\sim$1B active parameters} \\
\midrule
OLMo-1B (0724) & 36.4 & 53.5 & 66.8 & 42.4 & 67.5 & 32.1 & 44.2 & 74.0 & 45.2 & 62.9 & 52.5 \\
TinyLlama-1B & 38.1 & 69.5 & 63.6 & 61.1 & 60.8 & 33.6 & 45.0 & 71.7 & 50.4 & 60.1 & 55.4 \\
Pythia-1B & 31.4 & 63.4 & 56.8$^\dagger$ & 50.9 & 48.0 & 31.1 & 40.4 & 68.9 & 46.4 & 52.7 & 49.0 \\
Llama3.2-1B & 43.5 & 71.6 & 69.4 & 59.6 & 67.3 & 38.2 & 42.0 & 73.7 & 52.0 & 62.5 & 58.0 \\
Zamba2-1B & 55.0$^\dagger$ & 85.4 & 76.1 & 70.1 & 73.4 & 44.73$^\dagger$ & 59.8$^\dagger$ & 76.6 & 58.4 & 67.2 & 66.7 \\
DCLM-1B & 57.6$^\dagger$ & 79.5 & 80.9 & 71.3 & 75.1 & 48.5$^\dagger$ & 60.0$^\dagger$ & 76.6 & 60.5$^\dagger$ & 68.1 & 67.8 \\
\rowcolor{lightOlmoeYellow}
\modelsmall{} & 62.1$^\dagger$ & 84.2 & 79.2 & 72.9 & 80.0 & 54.1$^\dagger$ & 65.4$^\dagger$ & 79.8 & 63.0$^\dagger$ & 70.2 & 71.1 \\
\bottomrule
\end{tabular}
\caption{More results on OLMES. $^\dagger$ indicates use of the MCF score, see \autoref{sec:evalsetup}. See \autoref{tab:eval} for details on naming and a summary of these results.}
\label{tab:addolmes}
\end{table}

\begin{table}[htbp]
\centering
\setlength\tabcolsep{4pt}
\footnotesize
\begin{tabular}{l|rrr|rr}
\toprule
\modelsmall{} checkpoint ($\rightarrow$) & step 1,200,000 & step 1,220,000 & annealed & OLMo-1B & OLMo-7B \\
\midrule
AGI Eval LSAT-AR$^*$ & 24.3 & 26.5 & 28.7 & 28.3 & 28.3 \\
AGI Eval LSAT-LR & 40.2 & 38.6 & 37.3 & 30.2 & 42.9 \\
AGI Eval LSAT-RC & 47.4 & 43.7 & 46.6 & 23.5 & 61.6 \\
AGI Eval SAT-En & 55.3 & 54.9 & 52.9 & 28.2 & 73.8 \\
AGI Eval SAT-Math CoT & 5.5 & 4.1 & 6.4 & 1.8 & 6.8 \\
AQuA CoT & 2.4 & 2.9 & 2.0 & 2.9 & 6.1 \\
ARC Challenge$^*$ & 53.3 & 53.4 & 53.8 & 34.6 & 48.1 \\
ARC Easy$^*$ & 77.1 & 78.5 & 77.7 & 64.4 & 75.9 \\
BBQ & 49.8 & 48.3 & 50.6 & 45.8 & 67.2 \\
BigBench CS Algorithms$^*$ & 47.1 & 50.2 & 47.2 & 47.5 & 53.6 \\
BigBench Conceptual Combinations & 51.5 & 50.5 & 56.3 & 31.1 & 68.0 \\
BigBench Conlang Translation & 3.7 & 6.1 & 7.3 & 4.3 & 7.3 \\
BigBench Dyck Languages$^*$ & 19.3 & 15.9 & 21.5 & 26.6 & 22.2 \\
BigBench Elementary Math QA & 26.2 & 27.0 & 26.9 & 26.2 & 30.4 \\
BigBench Language Identification$^*$ & 31.9 & 34.0 & 31.0 & 27.0 & 39.1 \\
BigBench Logical Deduction & 26.6 & 25.3 & 24.6 & 23.6 & 27.3 \\
BigBench Misconceptions & 59.8 & 55.3 & 62.6 & 55.7 & 58.0 \\
BigBench Novel Concepts & 62.5 & 62.5 & 65.6 & 43.8 & 53.1 \\
BigBench Operators$^*$ & 36.2 & 34.3 & 33.8 & 23.8 & 45.2 \\
BigBench QA Wikidata$^*$ & 68.2 & 68.8 & 69.2 & 67.0 & 69.9 \\
BigBench Repeat Copy Logic$^*$ & 15.6 & 15.6 & 18.8 & 3.1 & 9.4 \\
BigBench Strange Stories & 66.7 & 68.4 & 69.5 & 53.4 & 66.1 \\
BigBench Strategy QA & 56.2 & 58.1 & 57.0 & 51.5 & 68.6 \\
BigBench Understanding Fables & 47.1 & 44.4 & 47.6 & 28.0 & 61.4 \\
BoolQ$^*$ & 73.3 & 72.8 & 73.2 & 63.7 & 83.9 \\
COPA$^*$ & 81.0 & 80.0 & 78.0 & 75.0 & 77.0 \\
CoQA$^*$ & 43.7 & 44.4 & 43.7 & 3.4 & 45.4 \\
CommonsenseQA$^*$ & 67.2 & 67.0 & 69.3 & 19.6 & 86.0 \\
Enterprise PII Classification & 52.3 & 53.7 & 52.2 & 57.3 & 50.6 \\
GPQA Diamond & 22.2 & 21.2 & 19.7 & 19.7 & 20.2 \\
GPQA Main & 24.8 & 22.3 & 22.5 & 20.3 & 23.0 \\
GSM8K CoT & 6.4 & 7.4 & 7.4 & 4.9 & 30.6 \\
HellaSwag 0-shot$^*$ & 76.0 & 76.0 & 77.0 & 65.8 & 76.7 \\
HellaSwag 10-shot$^*$ & 77.6 & 77.5 & 78.6 & 66.3 & 78.9 \\
Jeopardy$^*$ & 48.8 & 48.7 & 50.3 & 22.6 & 46.5 \\
LAMBADA$^*$ & 72.7 & 72.2 & 73.3 & 61.1 & 71.8 \\
LogiQA & 34.9 & 34.3 & 34.6 & 28.7 & 31.0 \\
MMLU Few-shot & 52.2 & 51.9 & 53.3 & 28.4 & 55.1 \\
MMLU Zero-shot & 41.6 & 42.7 & 43.3 & 26.2 & 50.0 \\
Math QA & 26.4 & 27.1 & 27.5 & 24.1 & 29.8 \\
OpenBookQA$^*$ & 41.4 & 44.0 & 44.8 & 36.6 & 43.4 \\
PIQA$^*$ & 81.3 & 81.2 & 82.0 & 76.4 & 81.7 \\
PubMedQA & 56.1 & 46.6 & 57.9 & 0.2 & 57.9 \\
SQuAD$^*$ & 52.9 & 52.4 & 52.4 & 0.0 & 65.5 \\
SVAMP CoT & 30.0 & 28.0 & 33.0 & 14.3 & 44.7 \\
Simple Arithmetic, no spaces & 17.6 & 18.1 & 20.1 & 1.2 & 15.3 \\
Simple Arithmetic, with spaces & 19.5 & 20.6 & 22.1 & 1.8 & 16.0 \\
Social IQA & 71.5 & 70.7 & 69.3 & 69.5 & 84.4 \\
Trivia QA & 54.2 & 53.0 & 55.9 & 25.1 & 51.8 \\
Winogender Female & 50.0 & 46.7 & 50.0 & 41.7 & 58.3 \\
Winogender Male & 55.0 & 58.3 & 60.0 & 63.3 & 58.3 \\
Winograd$^*$ & 82.8 & 83.2 & 84.6 & 79.9 & 83.2 \\
Winogrande$^*$ & 68.0 & 68.5 & 69.0 & 61.8 & 67.6 \\
\midrule
Core & 46.3 & 46.5 & 47.2 & 30.2 & 49.8 \\
Extended & 31.3 & 30.9 & 32.5 & 16.9 & 37.0 \\
\bottomrule
\end{tabular}
\caption{\textbf{DCLM evaluation metrics on the Core and Extended task subsets~\citep{li2024datacomplm}.} $^*$=Core tasks. ``annealed'' is the final pretraining checkpoint we use for \modelsmall{} and was annealed from the checkpoint at step 1,200,000. We left the non-annealing pretraining run train a little longer resulting in the 1,220,000 checkpoint.}
\label{tab:dclm}
\end{table}

\FloatBarrier

\section{Additional Experiments}
\label{sec:addablations}

\begin{figure*}[htbp]
\centering
\begin{center}
\includegraphics[width=\textwidth]{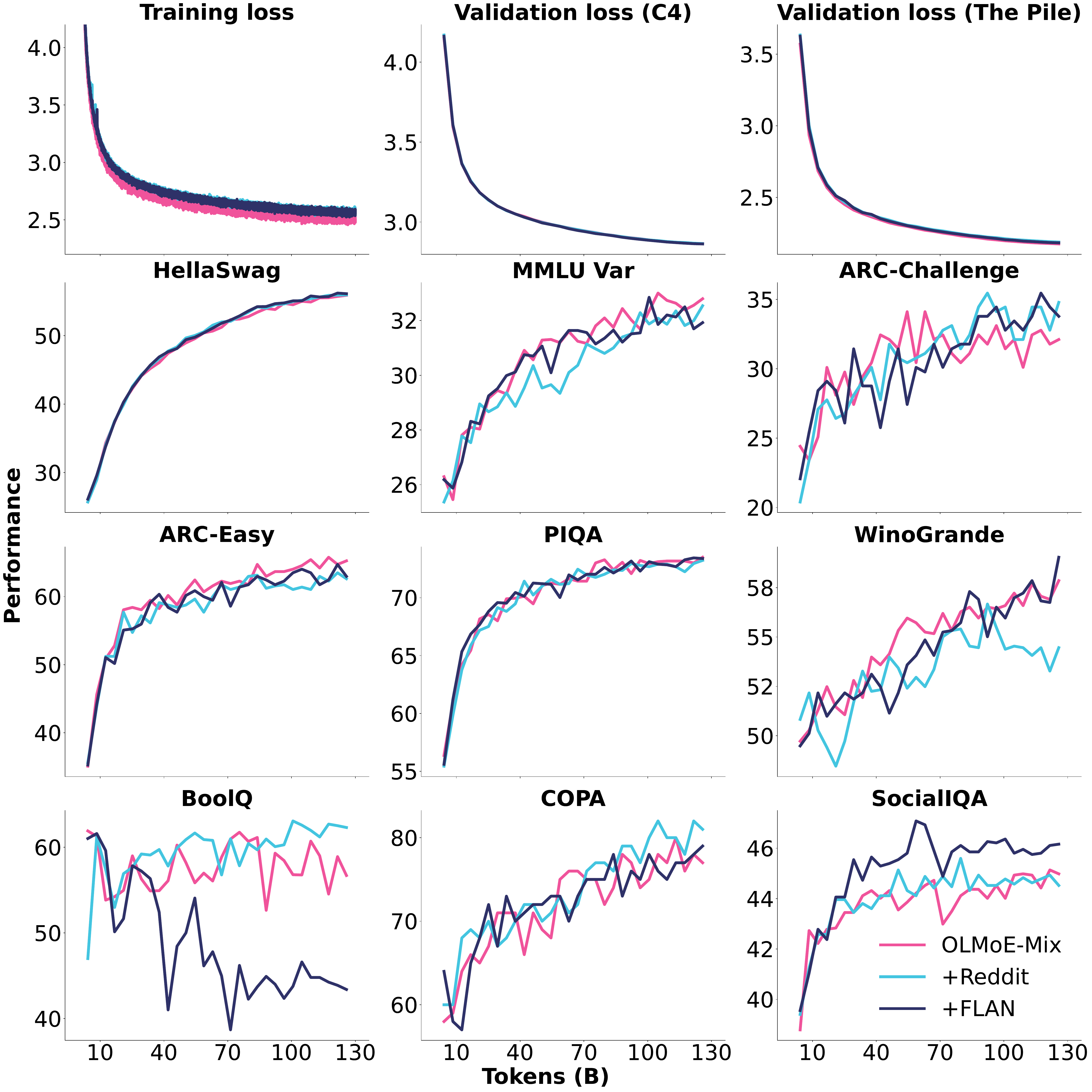}
\caption{\textbf{Adding Reddit or FLAN to \data{}.} More results, logs, and configurations: \url{https://wandb.ai/ai2-llm/olmoe/reports/Plot-Adding-Reddit-FLAN--Vmlldzo4OTg1NTg4}}
\label{fig:datasetredditflan}
\end{center}
\end{figure*}

\paragraph{Adding Reddit or FLAN to \data{}} In \autoref{fig:datasetredditflan} we benchmark adding the Reddit or FLAN~\citep{wei2022finetuned} subsets of Dolma 1.7~\citep{soldaini2024dolma} to our pretraining data mix (\autoref{sec:pretraining}). Overall, we do not find either one to lead to consistent gains, thus we do not use them in our final data mix.

\begin{figure*}[htbp]
\centering
\begin{center}
\includegraphics[width=\textwidth]{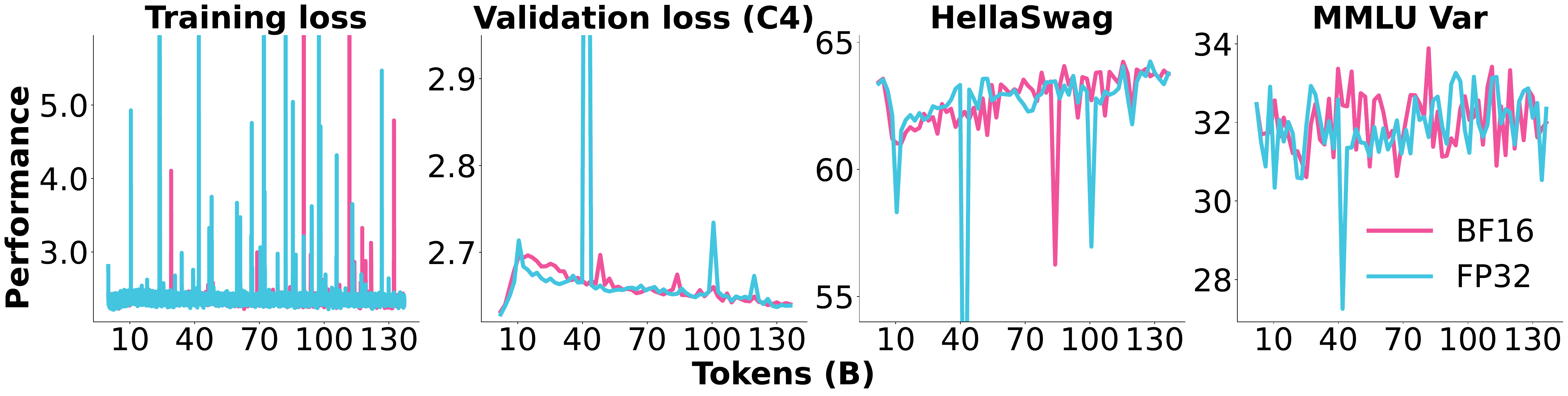}
\caption{\textbf{Load balancing precision.} More results, logs, and configurations: \url{https://wandb.ai/ai2-llm/olmoe/reports/Plot-FP32-LBL--Vmlldzo4NDMxNDA4}}
\label{fig:lbl}
\end{center}
\end{figure*}

\paragraph{Load balancing precision} \citet{fedus2022switch} selectively perform operations related to routing in full precision (FP32) to improve stability. In \autoref{fig:lbl}, we test whether computing the load balancing loss in full precision improves stability, but do not find it to reduce spikes. Thus, we stick with bfloat16 (BF16).

\begin{figure*}[htbp]
\centering
\begin{center}
\includegraphics[width=\textwidth]{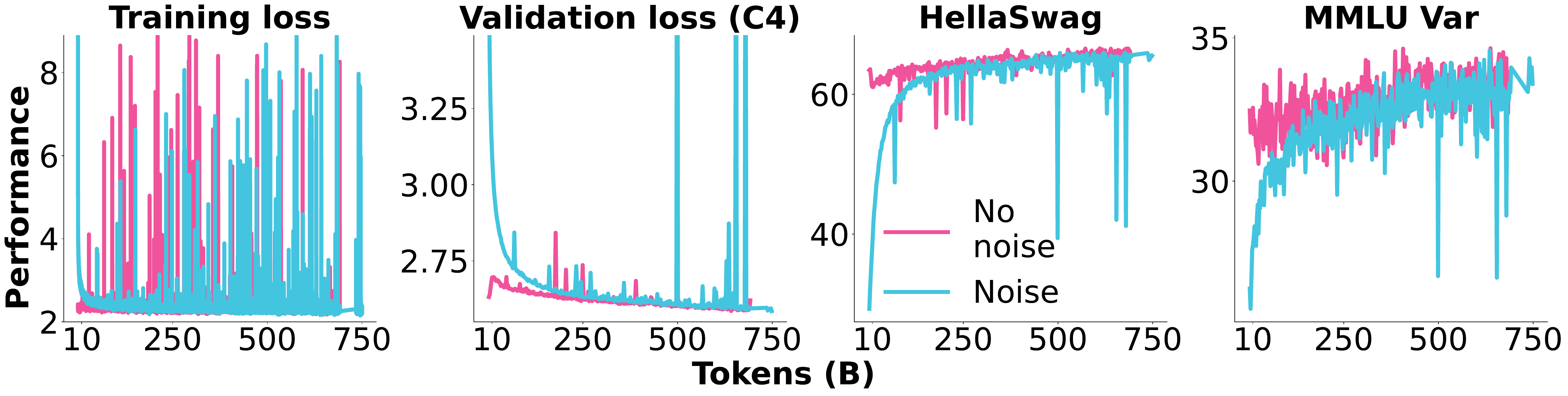}
\caption{\textbf{Adding noise to the upcycled checkpoint.} More results, logs, and configurations: \url{https://wandb.ai/ai2-llm/olmoe/reports/Plot-Noise-upcycle---Vmlldzo4NDA3MzI2}}
\label{fig:noise}
\end{center}
\end{figure*}

\paragraph{Noise upcycling} For the creation of Qwen2-MoE~\citep{yang2024qwen2technicalreport,qwen_moe,bai2023qwen}, the authors add 50\% of gaussian noise to feedforward networks before continuing training in an upcycled setup~\citep{komatsuzaki2023sparse}. \citet{komatsuzaki2023sparse} also report that they experimented with adding noise but did not find it beneficial. In \autoref{fig:noise}, we experiment with regular upcycling versus adding noise by randomly replacing 50\% of each MLP with numbers drawn from a normal distribution with a standard deviation of 0.02 following. We find that after 700 billion tokens, the no noise variant still performs slightly better but both appear to converge to the same performance. If training further, it is possible that the noise variant eventually outperforms the no noise variant, but at that point, it may make more sense to just train the MoE from scratch (\autoref{sec:upcycling}).

\begin{figure*}[htbp]
\centering
\begin{center}
\includegraphics[width=\textwidth]{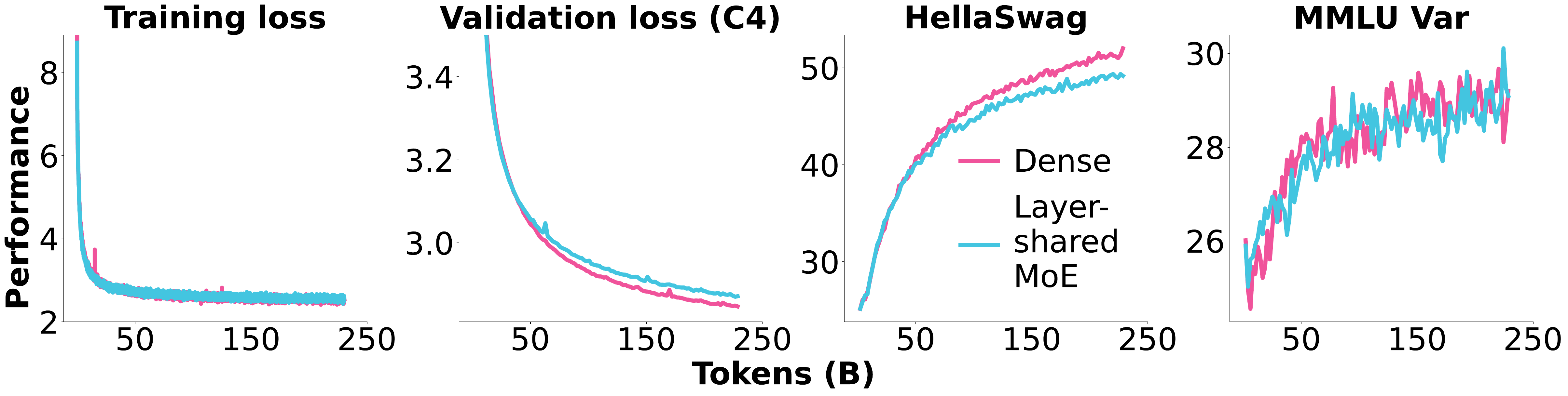}
\caption{\textbf{Sharing the same MoE across layers versus a regular dense LM.} The number of experts in the MoE is equivalent to its number of layers. Thus, because the MoE is shared across layers, it has the same number of total and active parameters as the dense model. More results, logs, and configurations: \url{https://wandb.ai/ai2-llm/olmoe/reports/Plot-Shared-vs-Dense--Vmlldzo4NDI0MTc5}}
\label{fig:layershared}
\end{center}
\end{figure*}

\paragraph{Shared Layer} Some work has investigated Mixture-of-Experts with weights shared across layers in the context of Universal Transformers~\citep{tan2023sparse,csordás2024moeut,dehghani2019universaltransformers}. We test whether layer-shared Mixture-of-Experts can beat non-shared dense models in \autoref{fig:layershared}. The layer-shared MoE uses a load balancing loss that is applied at the model level rather than at the layer level. This gives the model more flexibility by allowing it to completely deactivate certain experts for some layers and even emulate a dense model by always activating one separate expert for each layer. This makes it a generalization of the dense model which motivated our hypothesis that it may perform better than the dense model. However, in practice, we find that both perform similarly with the regular dense models even maintaining a small advantage on validation loss and HellaSwag. One possible advantage of layer-shared MoEs is that they can allow for better load balancing at inference. If prompts come in continuously, then newly incoming prompts can be batched with previous prompts that have already passed through several layers and sent through the MoE module together, as the MoE module is the same regardless of whether it is the first or last layer. Sharing also reduces throughput by around 20\% during training, which further motivates our decision not to use it for \modelsmall{}.

\paragraph{KTO experiments} In \autoref{tab:addadaptationablation} we experiment with the number of steps (5,000 vs. 10,000) and the optimizer (Adam~\citep{kingma2017adam} vs. RMS) used for KTO~\citep{ethayarajh2024kto}. Based on these experiments we use the RMS optimizer and the checkpoint at 5,000 steps in \autoref{sec:adapt}.

\begin{table*}[htbp]
\setlength{\tabcolsep}{3.9pt}
\begin{tabular}{l|ccccccc|c}
\toprule
&  &  &  & \textbf{Human-} & \textbf{Alpaca-} & & \\
\textbf{Task ($\rightarrow$)} & \textbf{MMLU} & \textbf{GSM8k} & \textbf{BBH} & \textbf{Eval} & \textbf{Eval 1.0} & \textbf{XSTest} & \textbf{IFEval} & \textbf{Avg} \\
\textbf{Setup ($\rightarrow$)} & \scriptsize{0-shot} & \scriptsize{8-shot CoT} & \scriptsize{0-shot} & \scriptsize{0-shot} & \scriptsize{0-shot} &\scriptsize{0-shot} & \scriptsize{0-shot} & \scriptsize{0-shot} \\
\textbf{Metric ($\rightarrow$)} & \scriptsize{EM} & \scriptsize{EM} & \scriptsize{EM} & \scriptsize{Pass@10} & \scriptsize{\%win} & \scriptsize{F1} & \scriptsize{Loose Acc} & \\
\midrule
KTO, 5,000 steps, RMS & \textbf{51.2} & \textbf{45.5} & 34.1 & \textbf{57.1} & \textbf{81.6} & \textbf{86.6} & \textbf{47.5} & \textbf{57.7} \\
KTO, 10,000 steps, RMS & 51.0 & 41.0 & 34.7 & 53.8 & 81.0 & 62.3 & \textbf{47.5} & 54.2 \\
\midrule
KTO, 5,000 steps, Adam & \textbf{51.2} & 42.0 & \textbf{35.3} & 55.6 & 81.0 & 84.5 & 46.6 & 56.0 \\
KTO, 10,000 steps, Adam & 51.0 & 43.0 & 34.1 & 54.9 & 79.7 & 62.7 & \textbf{47.5} & 53.3 \\
\bottomrule
\end{tabular}
\caption{\textbf{KTO adaptation experiments.} 5,000 and 10,000 steps correspond to 1.3 and 2.6 epochs on our adaptation dataset (\autoref{sec:pretraining}), respectively.}
\label{tab:addadaptationablation}
\end{table*}

\FloatBarrier

\section{Additional Analysis} 
\label{sec:addanalysis}

\begin{figure*}[htbp]
\centering
\includegraphics[width=\textwidth]{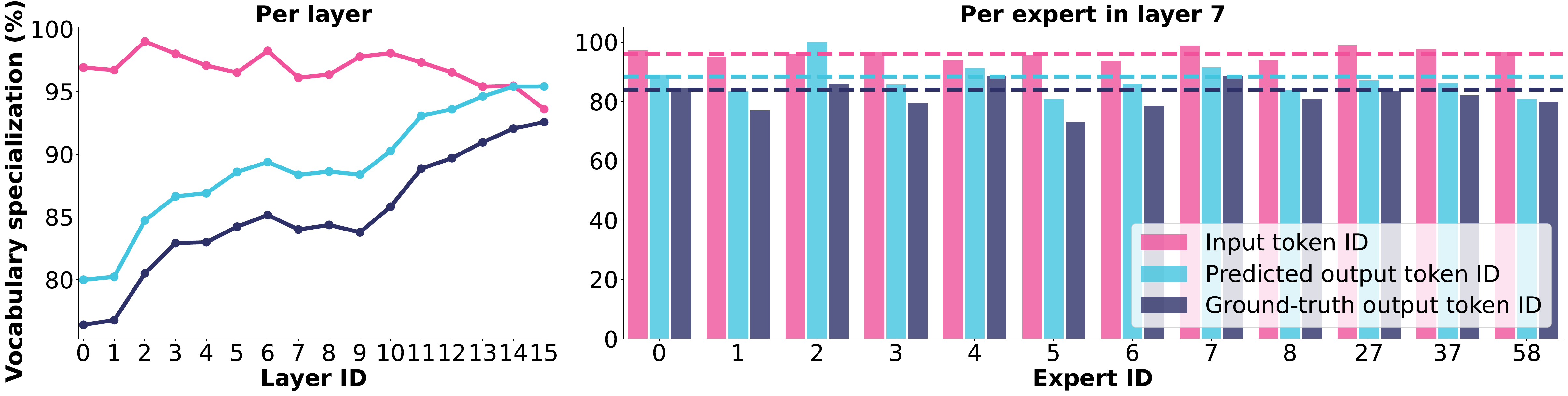}
\caption{\textbf{Vocabulary specialization for \modelsmall{} when considering all 8 activated experts.} Equivalent to $k=8$ in \autoref{eq:token}.}
\label{fig:tokenspecolmoe8}
\end{figure*}

\begin{figure*}[htbp]
\centering
\includegraphics[width=\textwidth]{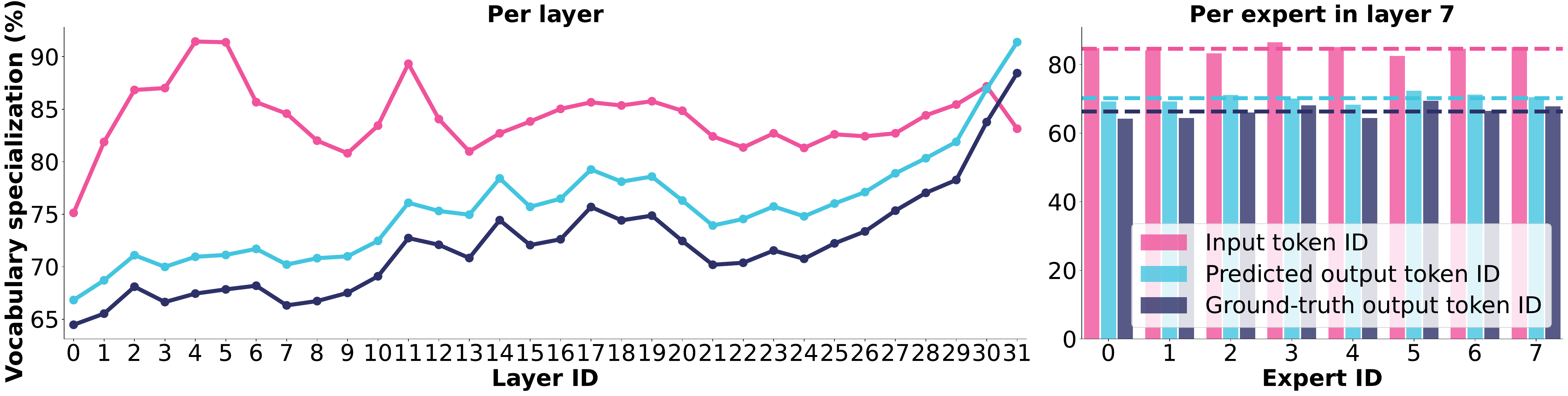}
\caption{\textbf{Vocabulary specialization for Mixtral-8x7B when considering all 2 activated experts.} Equivalent to $k=2$ in \autoref{eq:token}.}
\label{fig:tokenspecmixtral2}
\end{figure*}

\begin{figure*}[htbp]
\centering
\includegraphics[width=\textwidth]{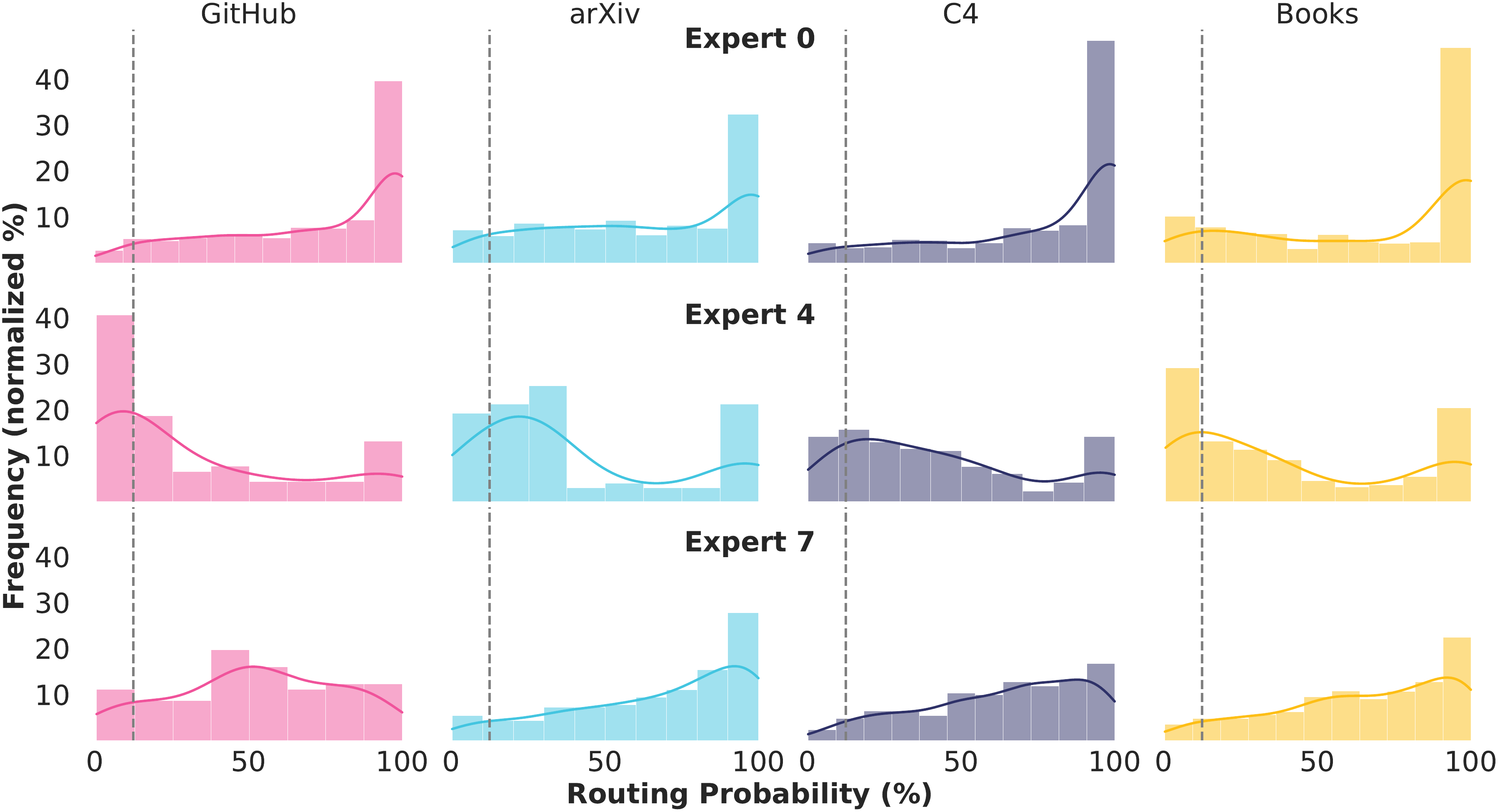}
\vskip 20pt
\includegraphics[width=\textwidth]{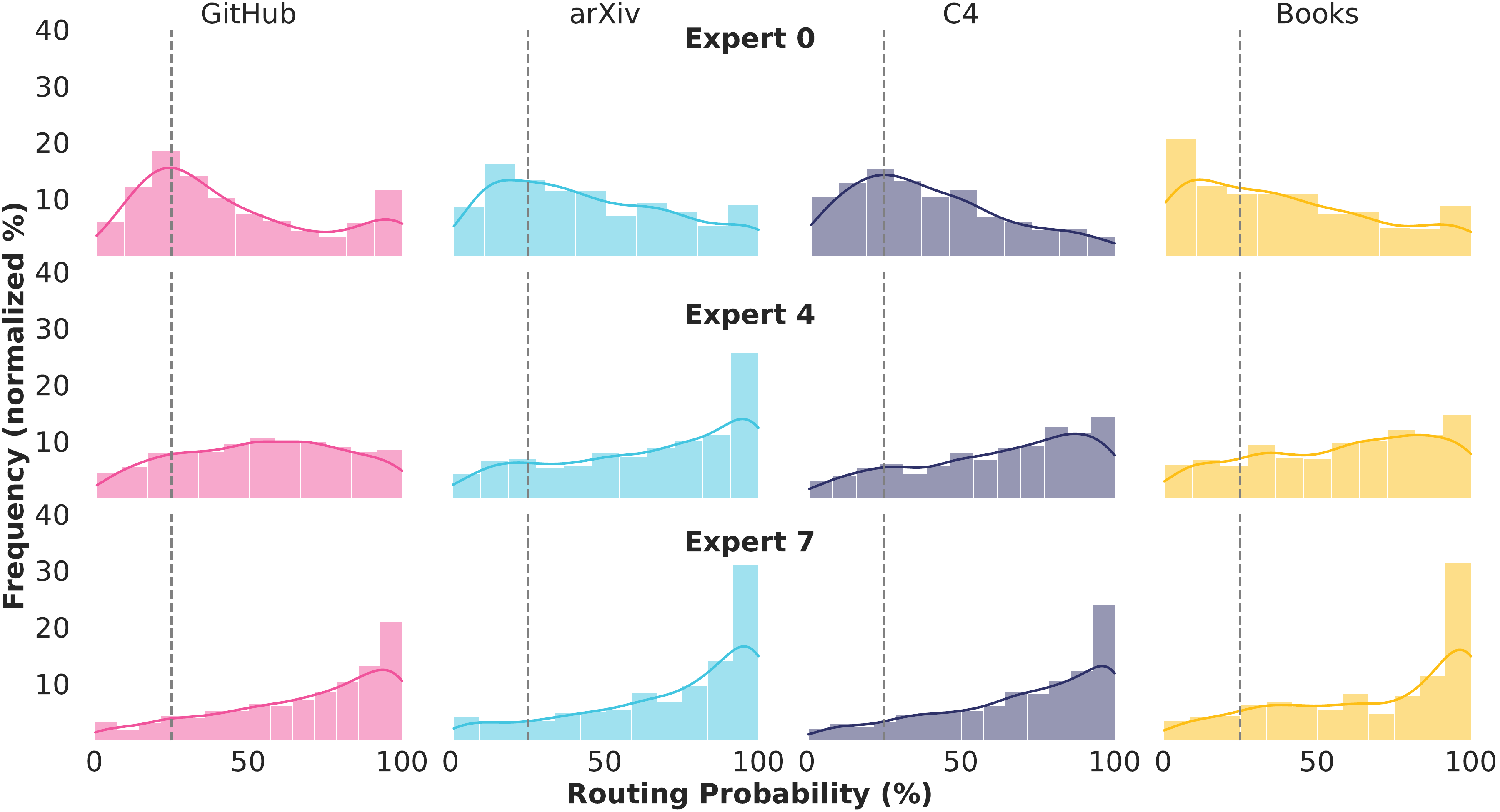}
\caption{\textbf{Vocabulary specialization across domains of \modelsmall{} (top) and Mixtral-8x7B (bottom).} We visualize how often token IDs get routed to specific experts. We only include IDs that appear at least 8 times in the various corpora. Vertical gray lines correspond to uniform routing (8/64=12.5\% for \modelsmall{} as it has 64 experts, 8 of which are activated; 2/8=25\% for Mixtral as it has 8 experts, 2 of which are activated). For example, among all token IDs in GitHub that get routed to Expert 0 at least 8 times for \modelsmall{}, $\sim$40\% of them get routed to Expert 0 with a probability of $\sim$100\% (upper left) indicating that Expert 0 is specialized on those token IDs. For \modelsmall{} there is much frequency at the routing probability extremes (0\% or 100\%) indicating that these experts exclusively focus on certain token IDs, especially for \textit{specific domains} (\autoref{sec:domain}) like GitHub and arXiv.}
\label{fig:tokendomainspec}
\end{figure*}

\begin{figure*}[htbp]
\centering
\includegraphics[width=\textwidth]{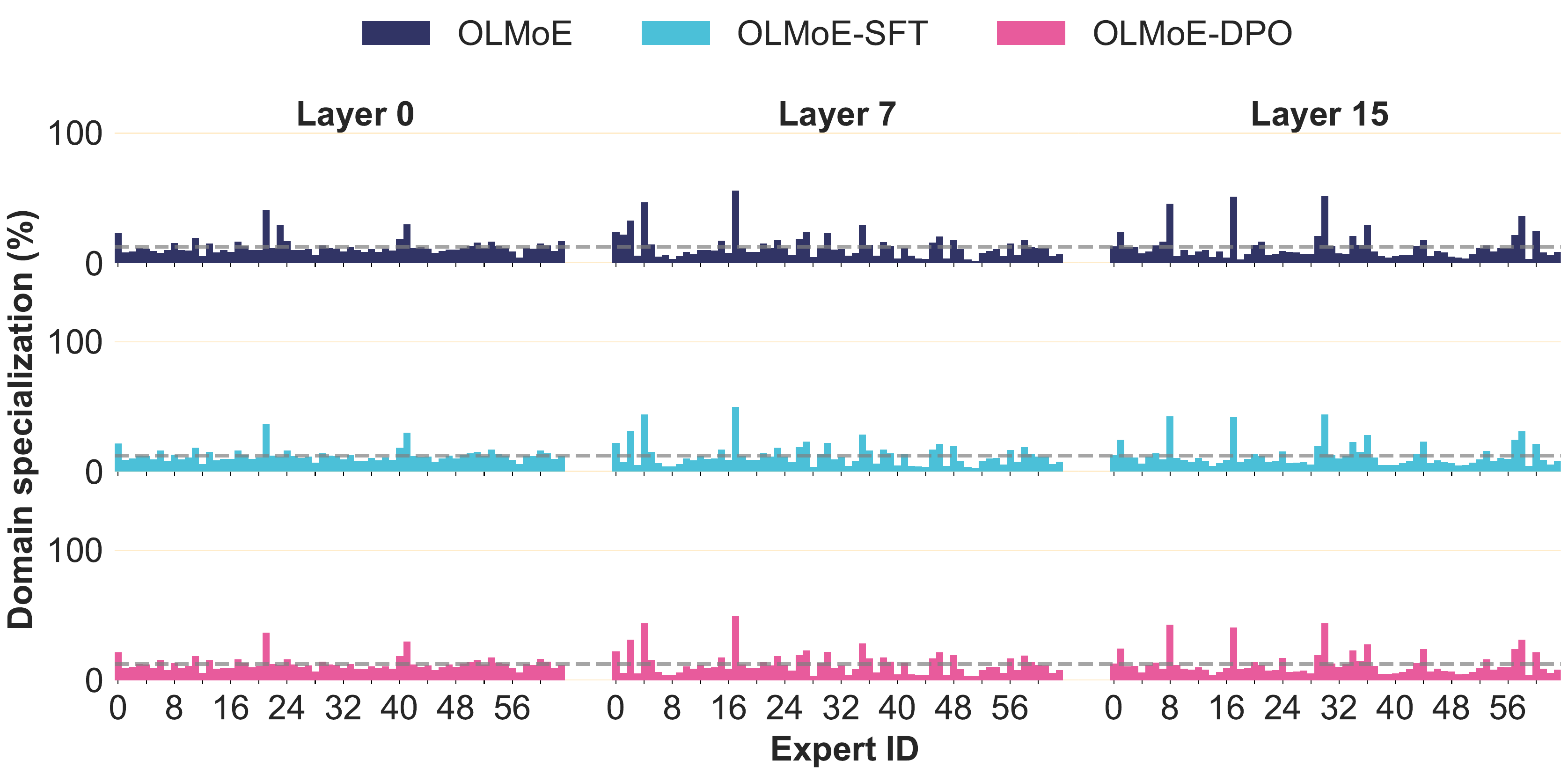}
\caption{\textbf{Load imbalances in selective layers after adaptation.} We visualize how often tokens from our instruction tuning dataset (\autoref{sec:pretraining}) get routed to the 8 active experts out of the 64 total experts ($k=1$ in \autoref{eq:domainspec}). Horizontal gray lines correspond to uniform routing (8/64=12.5\% per expert). Although we run SFT and DPO without loss balancing loss (\autoref{sec:adapt}), we observe that the load distribution does not change substantially.}
\label{fig:routing_tulu}
\end{figure*}

\begin{figure*}[htbp]
\centering
\includegraphics[width=\textwidth]{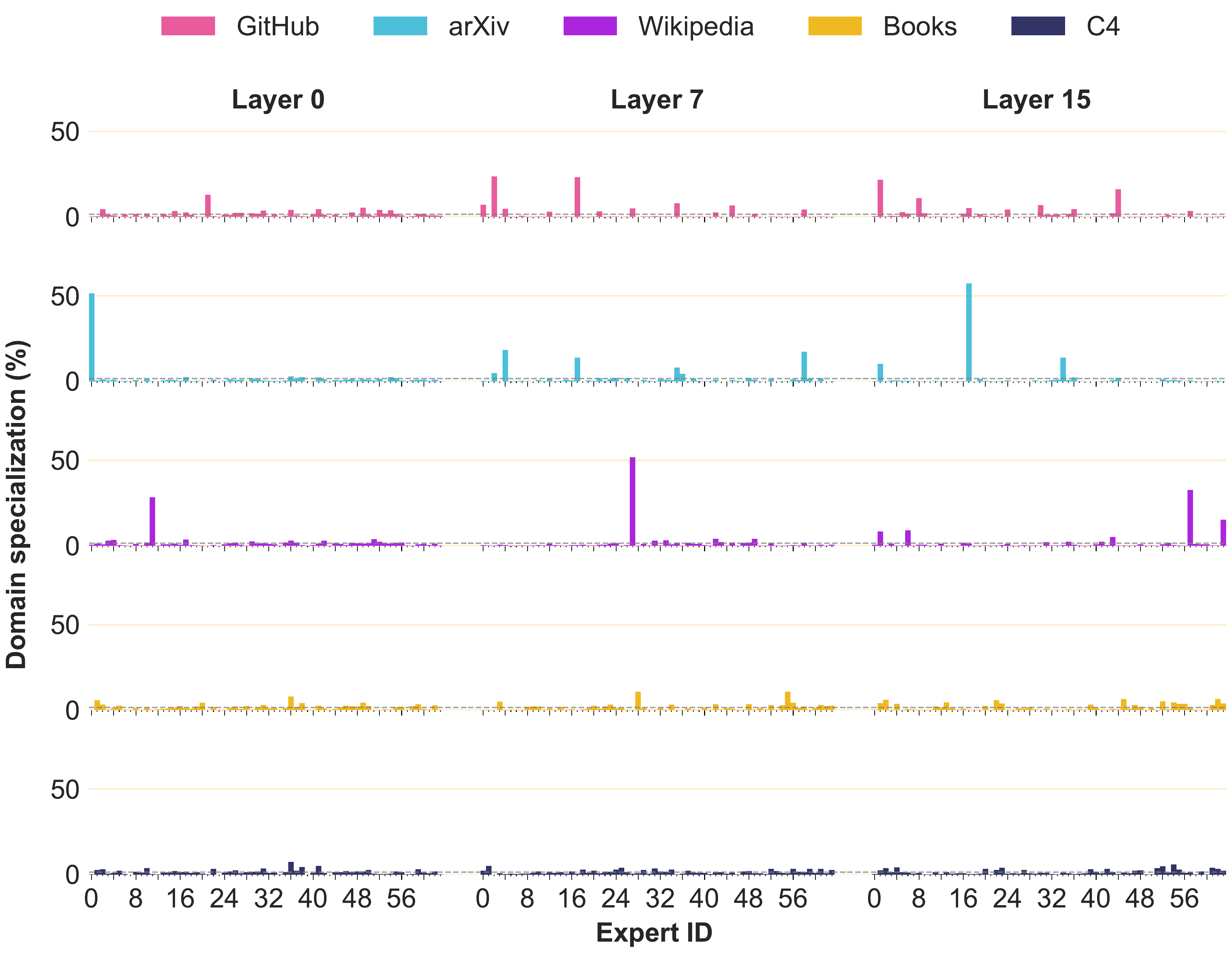}
\vskip 20pt
\includegraphics[width=\textwidth]{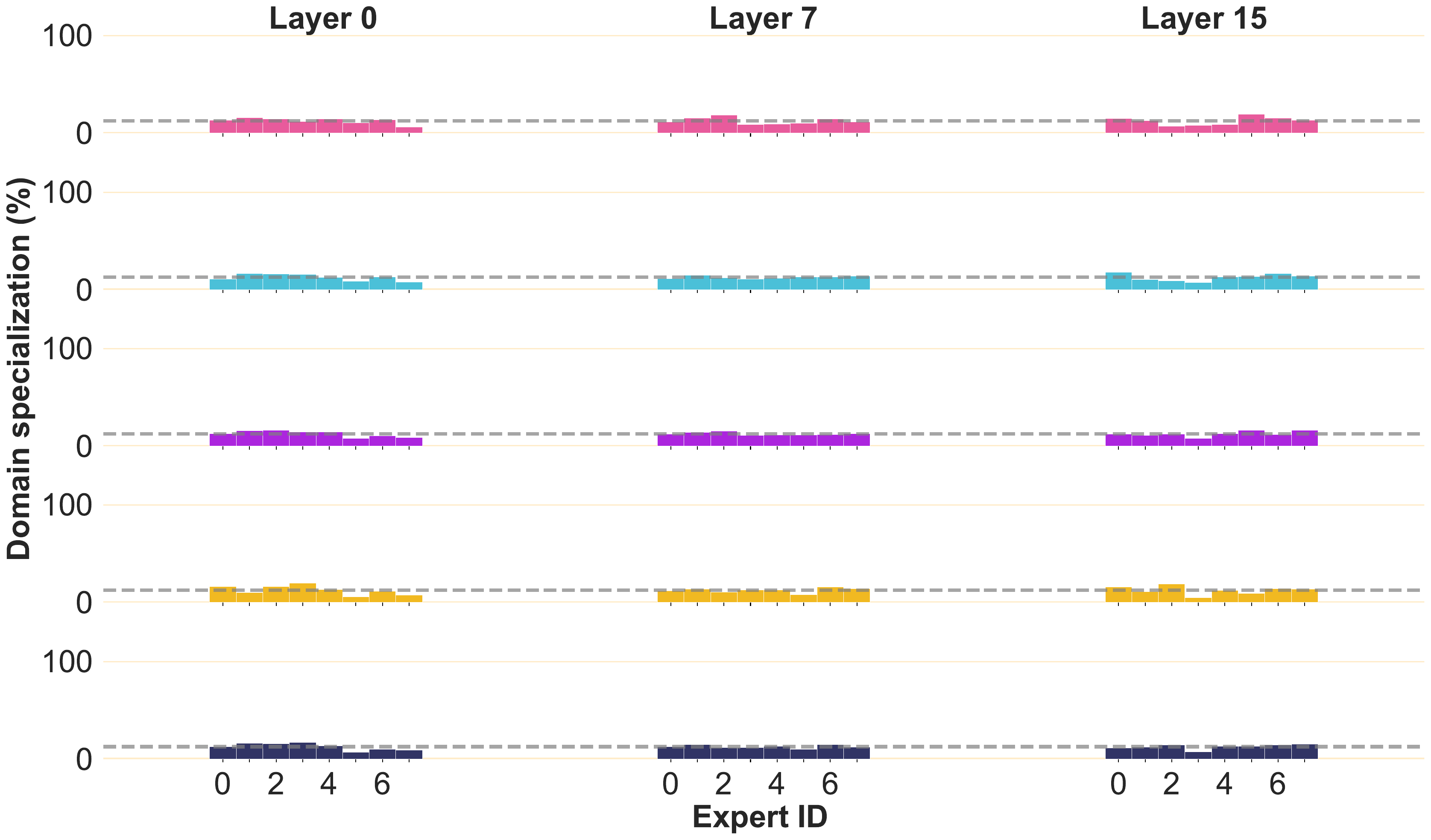}
\caption{\textbf{Domain specialization of \modelsmall{} (top) vs. Mixtral-8x7B (bottom) of the top-$1$ routed expert.} We visualize how often tokens from different domains get routed to the 64 (\model{}) or 8 (Mixtral) experts at the end of pretraining. Unlike in \autoref{fig:domainspec}, here we only consider tokens routed to the top-1 expert ($k=1$ in \autoref{eq:domainspec}). Horizontal gray lines correspond to uniform routing (1/64=1.56\% per expert for \modelsmall{} and 1/8=12.5\% for Mixtral).}
\label{fig:domainspec_top1}
\end{figure*}

\begin{figure*}[htbp]
\centering
\includegraphics[height=0.98\textwidth,angle=90]{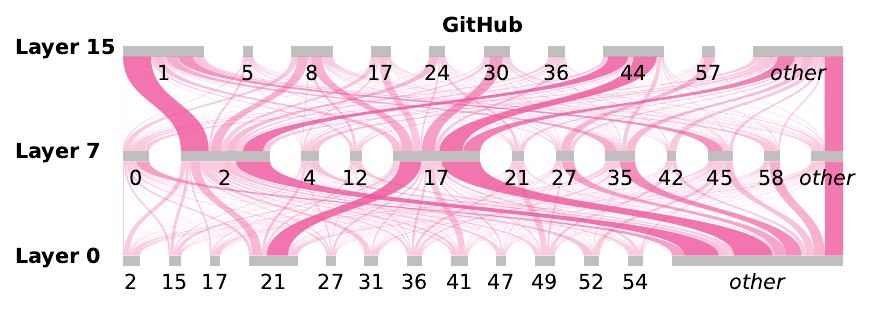}
\includegraphics[height=0.98\textwidth,angle=90]{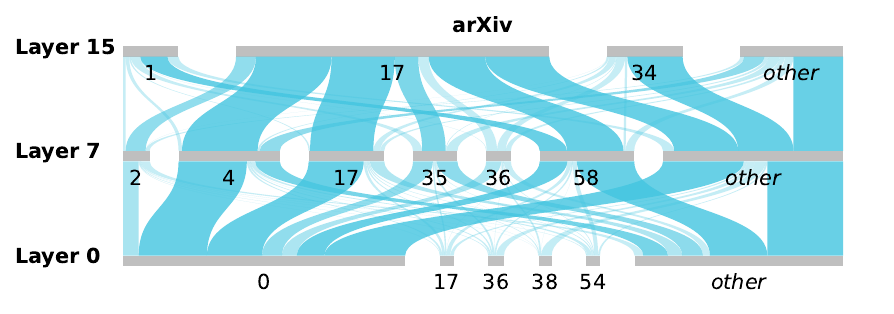}
\includegraphics[height=0.98\textwidth,angle=90]{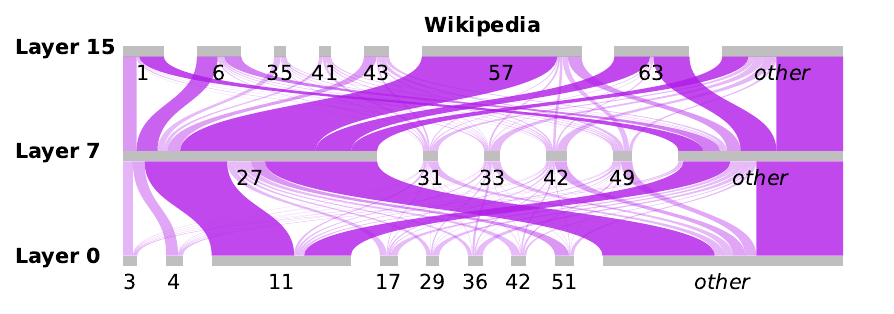}
\includegraphics[height=0.98\textwidth,angle=90]{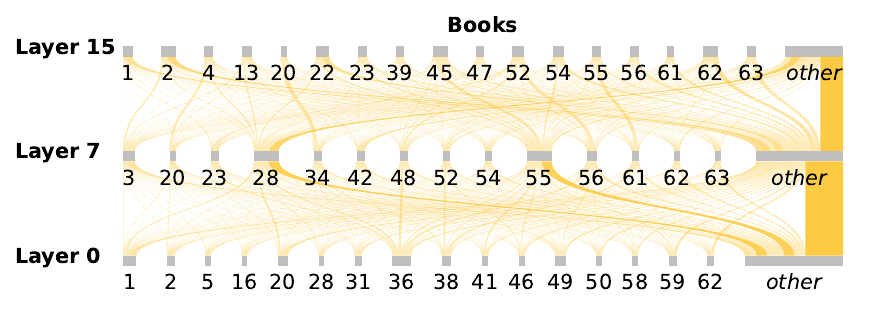}
\caption{\textbf{\modelsmall{} token routing across layers.} We visualize how often tokens from different domains get routed to a pair of experts across layers under top-1 routing, corresponding to \autoref{fig:domainspec_top1}. The size of each rectangle is proportional to the total number of tokens an expert receives, while the flow between two experts shows the proportion of tokens routed to both experts. We only show experts that receive tokens 50\% above random chance and use stronger coloring for larger flows. We observe some instances of cross-layer coordination between pairs of experts, e.g., expert 27 in layer 7 and expert 57 in layer 15 process a substantial fraction of Wikipedia tokens together. The flows between layers 0 $\to$ 7 and 7 $\to$ 15 are independent in this visualization.}
\label{fig:cross_layer_sankey}
\end{figure*}

\begin{figure*}[htbp]
\centering
\includegraphics[height=0.98\textwidth,angle=90]{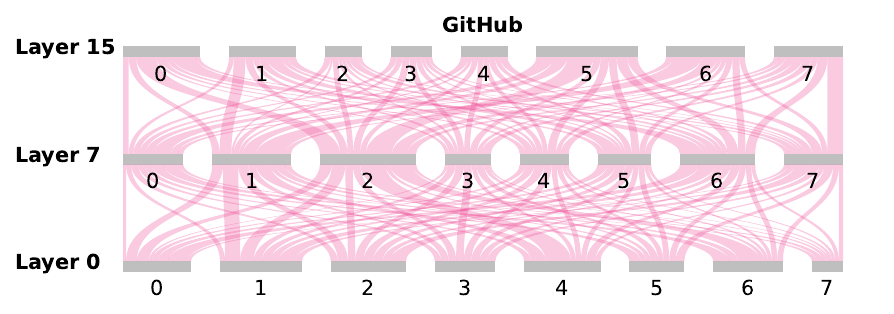}
\includegraphics[height=0.98\textwidth,angle=90]{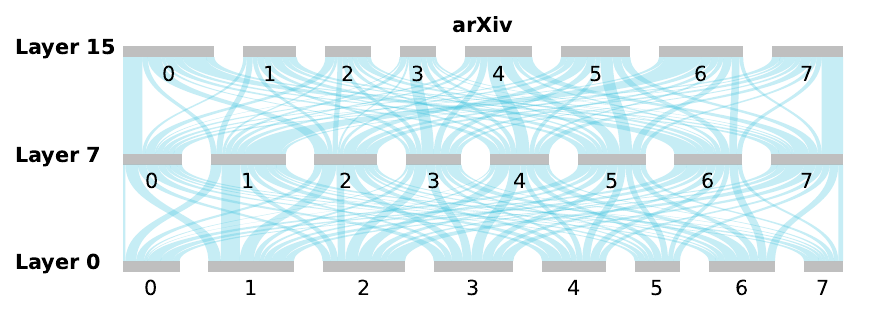}
\includegraphics[height=0.98\textwidth,angle=90]{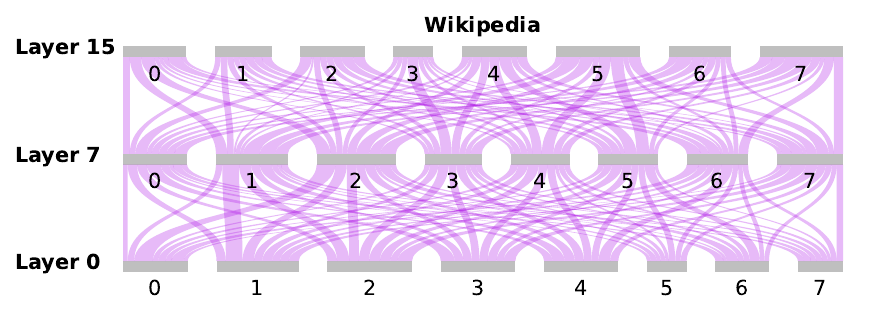}
\includegraphics[height=0.98\textwidth,angle=90]{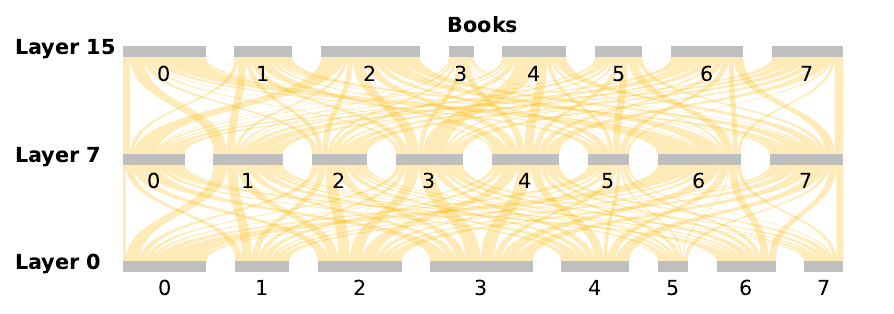}
\caption{\textbf{Mixtral-8x7B token routing across layers.} We visualize how often tokens from different domains get routed to a pair of experts across layers under top-1 routing, corresponding to \autoref{fig:domainspec_top1}. The size of each rectangle is proportional to the total number of tokens an expert receives, while the flow between two experts shows the proportion of tokens routed to both experts. The flows between layers 0 $\to$ 7 and 7 $\to$ 15 are independent in this visualization.}
\label{fig:cross_layer_sankey_mixtral}
\end{figure*}

\FloatBarrier

\section{Limitations and Future Work} 
\label{sec:limitations}

We highlight four key limitations with this release of \modelsmall{}. We look forward to addressing these issues in future iterations of \model{}.

\paragraph{More parameters} \modelsmall{} has 7B total parameters out of which 1B are activated for each input token. This small size makes \modelsmall{} very cheap to use, yet we demonstrate in this work that it outperforms much more expensive models (\autoref{fig:overview}). However, using only 1B parameters for each input token also limits the capabilities of \modelsmall{} as seen by its performance compared to models that use $>$7$\times$ more parameters, such as Llama3.1-8B in \autoref{sec:results}. While it may be possible that more parameters are not needed to match 8B models and beyond~\citep{llmsize}, in the short-term adding parameters is an easy way to improve the performance of \model{}, at least allowing the model to utilize more than 1B parameters per input, possibly via recursion~\citep{dehghani2019universaltransformers} or agentic workflows~\citep{wang2024opendevinopenplatformai,yang2024sweagentagentcomputerinterfacesenable}. Relatedly, changing the allocation of parameters to e.g. vocabulary versus non-vocabulary parameters is another avenue for improvement~\citep{tao2024scalinglawsvocabularylarger}.

\paragraph{More data} We train \modelsmall{} for 5 trillion tokens, however, some recent dense models train significantly longer, such as Llama 3 with 15 trillion tokens~\citep{dubey2024llama3herdmodels}. To the best of our knowledge, there has been no large MoE that has been overtrained~\citep{gadre2024language} as much as \modelsmall{}. Specifically, taking the active parameters of \modelsmall{}, our token multiplier~\citep{gadre2024language} is around 5,000 (5T / 1B). There are likely benefits to training even longer, but to what degree overtraining is effective for MoEs and how it differs from dense models still requires more research~\citep{allenzhu2024physics}.

\paragraph{Multimodal} \modelsmall{} is a text-only large language model, thus it cannot take inputs or produce outputs in other modalities like images or audio. This limits its utility for the large variety of multimodal use cases of such models~\citep{huang2018musictransformer,su2020vlbertpretraininggenericvisuallinguistic,chen2020generative,kiela2021hateful,muennighoff2020vilio,radford2022robustspeechrecognitionlargescale,bai2023qwenvlversatilevisionlanguagemodel,driess2023palmeembodiedmultimodallanguage,dubey2024llama3herdmodels}. There has been early work on open multimodal MoEs~\citep{mustafa2022multimodalcontrastivelearninglimoe,lin2024moellavamixtureexpertslarge,li2024unimoescalingunifiedmultimodal,shen2023scalingvisionlanguagemodelssparse,mckinzie2024mm1methodsanalysis,wu2024omnismolaboostinggeneralistmultimodal} and we look forward to making future versions of \model{} a part of that.

\paragraph{Multilingual} We pretrain \modelsmall{} on a predominantly English corpus and exclusively evaluate on English tasks. This may severely limit the usefulness of our model for research on non-English language models~\citep{lovenia2024seacrowdmultilingualmultimodaldata,singh2024aya,üstün2024aya,enevoldsen2024scandinavianembeddingbenchmarkscomprehensive,son2024kmmlumeasuringmassivemultitask,xiao2023cpack}. While there has been work on training language-specific LMs~\citep{luukkonen2023fingpt,faysse2024croissantllmtrulybilingualfrenchenglish}, it is more likely that as we add more data to build better future iterations of \model{} we will mix in more non-English data due to data constraints~\citep{muennighoff2023scaling}. This may make future \model{} models perform better in non-English languages.

\section{\modelsmallnew{}}
\label{sec:olmoe0125}

We introduced \modelsmall{} in September 2024. In January 2025, we released a better model, \modelsmallnew{}, which we discuss here.

\begin{table}[h]
\centering
\begin{tabular}{lccc}
\toprule
\textbf{Source} & \textbf{Total tokens} & \textbf{Source \%} & \textbf{Mix \%}\\
\midrule
Filtered DCLM & 752B & 6.85 & 50.2 \\
Decontaminated FLAN & 17.0B & 100 & 16.7 \\
StackExchange Q\&A & 1.26B & 200 & 2.47  \\
peS2o & 58.6B  & 16.7 & 9.52  \\
Wikipedia/Wikibooks & 3.70B & 100 & 3.57 \\
Dolmino Math & 10.7B & 200 & 17.5 \\
\bottomrule
\end{tabular}
\vspace{3pt}
\caption{\textsc{Dolmino} composition and sampling distribution used for \modelsmallnew{}.}
\label{tab:dolmino-mix}
\end{table}

For pretraining, \modelsmallnew{} uses the same data mix for the first stage of training. Following OLMo 2~\citep{OLMo2}, we anneal this new model on a curated mix of high-quality sources. 
We sample this mix from the \textsc{Dolmino} dataset,\footnote{\href{https://huggingface.co/datasets/allenai/dolmino-mix-1124}{\texttt{huggingface.co/datasets/allenai/dolmino-mix-1124}}} a collection of high-quality web pages, academic content, question answering pairs, instruction data, and math problems. We use the same 100B tokens sample of \textsc{Dolmino} used to anneal OLMo 2 13B; a summary of this dataset is in \autoref{tab:dolmino-mix}.

\begin{table}[htbp]
\setlength\tabcolsep{2pt} 
\centering
\begin{tabular}{l|ccccccccccc}
\toprule
\footnotesize{\bf{\textsc{OLMoE} release}} & \footnotesize{\bf{ARC\_C}} & \footnotesize{\bf{ARC\_E}} & \footnotesize{\bf{BoolQ}} & \footnotesize{\bf{CSQA}} & \footnotesize{\bf{HSwag}} & \footnotesize{\bf{MMLU}} & \footnotesize{\bf{OBQA}} & \footnotesize{\bf{PIQA}} & \footnotesize{\bf{SIQA}} & \footnotesize{\bf{WinoG}} & \footnotesize{\bf{Avg}}\\
\midrule
\textbf{Sep 2024 (0924)} & 62.1$^\dagger$ & 84.2 & 79.2 & \textbf{72.9} & 80.0 & 54.1$^\dagger$ & 65.4$^\dagger$ & \textbf{79.8} & 63.0$^\dagger$ & 70.2 & 71.1 \\
\textbf{Jan 2025 (0125)} & \textbf{67.5}$^\dagger$ & \textbf{84.4}$^\dagger$ & \textbf{80.6} & 70.8 & \textbf{81.7} & \textbf{56.3}$^\dagger$ & \textbf{69.6}$^\dagger$ & 78.7 & \textbf{66.8}$^\dagger$ & \textbf{70.6} & \textbf{72.7} \\
\bottomrule
\end{tabular}
\vspace{.5em}
\caption{\textbf{\modelsmalldate{} and \modelsmallnew{} on OLMES.}  We bold the best performance. $^\dagger$ indicates use of the MCF score, see \autoref{sec:evalsetup} for evaluation details.}
\label{tab:score-table-main-new}
\end{table}

We compare \modelsmallnew{} with \modelsmall{} In \autoref{tab:score-table-main-new}. Overall, the new model is a notable improvement over the previous iteration being better on average (+1.6) and notable datasets like MMLU (+2.1). 

Following this improved annealing setup, we adapt \modelsmallnew{} using the post-training from T{\"u}lu 3~\citep{lambert2025tulu3pushingfrontiers}. This recipe represents an updated version of the one originally used for \model{}. It features an improved SFT mix, better sampled DPO data, and a PPO step that leverages verifiers as for the model reward.  We compare this new iteration using the evaluation setup from T{\"u}lu (which differs from other evaluations in this paper) in \autoref{tab:adaptresultsnew}. After adaptation, the new model is significantly better, with a 10-point gain on the benchmark average.

The new models and datasets are freely available on the Hugging Face hub.\footnote{\href{https://hf.co/collections/allenai/olmoe-january-2025-67992134f9ebea0a941706ca}{\path{hf.co/collections/allenai/olmoe-january-2025-67992134f9ebea0a941706ca}}} For more information about this release, we refer to its announcement on Ai2's website.\footnote{\href{https://web.archive.org/web/20250212023046/https://allenai.org/blog/olmoe-app}{\path{allenai.org/blog/olmoe-app}}}

\begin{table}[htbp]
\centering
\begin{tabular}{ll|cc|ccc}
\toprule
\multirow{2}{*}{\textbf{Skill}} & \multirow{2}{*}{\textbf{Benchmark}$_\text{(eval)}$} & \multicolumn{2}{c|}{\modelsmalldate} & \multicolumn{3}{c}{\modelsmallnew{}} \\[0.1em]
&&\textbf{+SFT} & \textbf{+DPO} & \textbf{+SFT} & \textbf{+DPO} & \textbf{+RLVR} \\
\midrule
& Avg. & 39.7 & 39.8 & 46.6 & 49.3 & \textbf{49.8} \\
\midrule
Knowledge & MMLU$_\text{(0 shot, CoT)}$ & 54.3 & 54.6 & \textbf{55.3} & 54.9 & 55.1 \\
& PopQA$_\text{(15 shot)}$ & \textbf{21.0} & 20.6 & 20.1 & 19.7 & 19.8 \\
& TruthfulQA$_\text{(6 shot)}$ & 44.7 & 49.1 & 45.5 & 50.0 & \textbf{50.6} \\
\midrule
Reasoning & BigBenchHard$_\text{(3 shot, CoT)}$ & 36.6 & 36.8 & 37.3 & 37.4 & \textbf{38.6} \\
& DROP$_\text{(3 shot)}$ & 34.7 & 34.5 & \textbf{48.6} & 48.4 & 47.9 \\
\midrule
Math & MATH$_\text{(4 shot CoT, Flex)}$ & 8.2 & 8.2 & \textbf{21.4} & 20.4 & \textbf{21.4} \\
& GSM8K$_\text{(8 shot, CoT)}$ & 42.5 & 47.4 & 55.7 & 64.6 & \textbf{72.4} \\
\midrule
Coding & HumanEval$_\text{(pass@10)}$ & \textbf{63.7} & 63.0 & 62.6 & 61.9 & 62.3 \\
& HumanEval+$_\text{(pass@10)}$ & 57.4 & 58.9 & 55.7 & \textbf{57.6} & 54.4 \\
\midrule
IF \& chat & IFEval$_\text{(prompt loose)}$ & 41.2 & 45.3 & 56.6 & 65.6 & \textbf{66.4} \\
& AlpacaEval 2$_\text{(LC \% win)}$ & 6.4 & 7.5 & 5.8 & \textbf{19.5} & 18.0 \\
\midrule
Safety & Safety$_\text{(6 task avg.)}$ & 65.8 & 51.4 & \textbf{94.5} & 91.4 & 90.4 \\
\bottomrule
\end{tabular}
\vspace{.5em}
\caption{\textbf{\modelsmalldate{} and \modelsmallnew{} after adaptation.} We bold the best performance.}
\label{tab:adaptresultsnew}
\end{table}

\FloatBarrier

\section{Change log}
\label{sec:log}

\textbf{V1 → V2 (2025-03):}

\begin{itemize}
    \item Added reference to \modelsmallnew{} in \autoref{sec:olmoe0125}
    \item Corrected OpenMoE active parameters in \autoref{tab:eval} from 2.9B to 2.6B
    \item Corrected our max LR in \autoref{tab:hp} from 5.0E-04 to 4.0E-05
    \item Added Zamba2, Llama3.2, and DeepSeekV2 in \autoref{tab:addolmes}
\end{itemize}

\end{document}